\documentclass[10pt,twocolumn]{article}
\usepackage[a4paper,total={7in,9.77in}]{geometry}
\usepackage{newtxtext}
\usepackage{amsmath,amsfonts,amssymb,amsthm}
\usepackage{algorithmic}
\usepackage{algorithm}
\usepackage{array}
\usepackage[caption=false,font=normalsize,labelfont=sf,textfont=sf]{subfig}
\usepackage{textcomp}
\usepackage{stfloats}
\usepackage{url}
\usepackage{verbatim}
\usepackage{graphicx}
\usepackage{cite}
\usepackage{booktabs}
\usepackage{cases}
\usepackage{diagbox}
\usepackage{contour}
\usepackage{placeins}
\usepackage{indentfirst}
\usepackage{etoolbox}
\usepackage{fancyhdr}
\usepackage{float}
\usepackage{multirow}
\usepackage{threeparttable}
\usepackage{pifont}
\usepackage{makecell}
\usepackage{xcolor}
\usepackage{microtype}
\usepackage{listings}
\usepackage{hyperref}

\AtBeginEnvironment{tabular}{\scriptsize}


\newcommand{\rev}[1]{#1}

\newcommand{\revCell}[1]{#1}
\newcommand{\revEnvBegin}{}
\newcommand{\revEnvEnd}{}
\newcommand{\revTableFootnote}[1]{%
  \par\noindent\begingroup\raggedright\footnotesize #1\par\endgroup}

\providecommand{\checkmark}{\ding{51}}

\newtheorem{theorem}{Theorem}

\newtheorem{suppTheorem}{Theorem}

\newtheorem{suppProposition}{Proposition}

\newcommand{\R}{\mathbb{R}}
\newcommand{\vx}{\mathbf{x}}
\newcommand{\va}{\mathbf{a}}
\newcommand{\vb}{\mathbf{b}}

\newcommand{\vr}{\mathbf{r}}
\newcommand{\vy}{\mathbf{y}}
\newcommand{\vz}{\mathbf{z}}

\newcommand{\mU}{\mathbf{U}}
\newcommand{\mV}{\mathbf{V}}
\newcommand{\mW}{\mathbf{W}}
\newcommand{\mP}{\mathbf{P}}
\newcommand{\mR}{\mathbf{R}}
\newcommand{\mA}{\mathbf{A}}
\newcommand{\mI}{\mathbb{I}}

\newcommand{\mSigma}{\boldsymbol{\Sigma}}

\newcommand{\tablefootnote}[1]{%
  \par\vspace{2pt}\noindent\begin{minipage}{\linewidth}%
  \footnotesize #1\end{minipage}}

\definecolor{codegreen}{HTML}{2E7D5B}
\definecolor{codegray}{HTML}{6B7280}
\definecolor{codepurple}{HTML}{8B5A8C}
\definecolor{codeback}{HTML}{F5F7FA}
\definecolor{codeframe}{HTML}{CBD5E1}
\definecolor{outputback}{HTML}{F8F8F6}
\lstdefinestyle{pythonstyle}{
  language=Python, backgroundcolor=\color{codeback},
  basicstyle=\ttfamily\footnotesize, commentstyle=\color{codegreen},
  keywordstyle=\color{blue}, stringstyle=\color{codepurple},
  numberstyle=\tiny\color{codegray}, numbers=left, numbersep=6pt,
  breaklines=true, breakatwhitespace=false, columns=fullflexible,
  keepspaces=true, showstringspaces=false, frame=single,
  rulecolor=\color{codeframe}, framerule=0.4pt, framesep=5pt,
  framexleftmargin=6pt, framexrightmargin=6pt, framextopmargin=4pt,
  framexbottommargin=4pt, xleftmargin=1.8em, xrightmargin=0.4em,
  aboveskip=0.8em, belowskip=0.8em}
\lstdefinestyle{outputstyle}{
  backgroundcolor=\color{outputback}, basicstyle=\ttfamily\footnotesize,
  columns=fullflexible, keepspaces=true, showstringspaces=false,
  frame=single, rulecolor=\color{codeframe}, framerule=0.4pt,
  framesep=5pt, framexleftmargin=6pt, framexrightmargin=6pt,
  framextopmargin=4pt, framexbottommargin=4pt, xleftmargin=0.4em,
  xrightmargin=0.4em, aboveskip=0.4em, belowskip=0.8em}

\hypersetup{colorlinks=true,linkcolor=blue,citecolor=blue,urlcolor=blue,hypertexnames=false}
\title{Hierarchical Feature-level Reverse Propagation \\ for Post-Training Neural Networks}
\author{
  Ni Ding\textsuperscript{2}, Shuchang Wang\textsuperscript{2},
  Lei He\textsuperscript{1,2}\thanks{Corresponding author: helei2023@tsinghua.edu.cn. This work was supported by the National Key R\&D Program of China, Project ``Development of Large Model Technology and Scenario Library Construction for Autonomous Driving Data Closed-Loop'' (Grant No. 2024YFB2505501).},
  Shengbo Eben Li\textsuperscript{1,2},
  and Keqiang Li\textsuperscript{1,2}\\
  \textsuperscript{1}School of Vehicle and Mobility, Tsinghua University, Beijing 100084, China.\\
  \textsuperscript{2}State Key Laboratory of Intelligent Green Vehicle and Mobility, Tsinghua University, Beijing 100084, China.
}
\date{}

\begin{document}
\maketitle
\revEnvBegin
\begin{abstract}
End-to-end neural networks have become a dominant paradigm in autonomous driving, where reliable deployment requires controllable post-training adaptation and improved transparency of model updates.
In this paper, we propose Feature-level Reverse Propagation for Post-Training (FR-PT), a hierarchical framework that provides explicit intermediate supervision for upstream modules by reconstructing label-conditioned features backward through frozen downstream networks. For the first time, we formulate feature reconstruction via the Computation Consistency Principle (CCP) and Minimum Deviation Principle (MDP), and develop efficient operator-specific reverse computation algorithms with MDP-centered Tikhonov regularization to handle numerically unstable inverse problems. Specifically, FR-PT incorporates circular convolution theorem-based solvers for scalable convolutional reconstruction, nearest embedding for constructing continuous output targets from categorical labels, and iterative reverse propagation for composite residual and Transformer-style blocks.
Extensive experiments on image classification and autonomous driving tasks demonstrate effective and stable adaptation across diverse architectures. Among 85 post-training settings, FR-PT achieves statistically significant improvements over task-level baselines in 63 cases, while the matched backpropagation reference outperforms reconstruction-supervised configurations in only 4 cases. Additional efficiency, conditioning, and favorable-condition analyses characterize the reliability and limitations of reconstructed targets, while feature-response analyses further demonstrate their diagnostic value. Code is available at \url{https://github.com/Dingni2000/FR-PT}.
\end{abstract}\par
\revEnvEnd
\revEnvBegin
\par
\revEnvEnd
\section{Introduction}
\label{sec:introduction}
As a rapidly advancing technology in recent years, artificial intelligence (AI) based on deep neural networks (DNNs) \cite{mathew2021deep} provides promising solutions for autonomous driving (AD).
Featured in optimizing all processing steps simultaneously, \rev{large }end-to-end (E2E) models that directly map raw sensor inputs to driving commands \cite{bojarski2016endendlearningselfdriving, hwang2024emmaendtoendmultimodalmodel, zheng2024genad, 10629039, e2e_2022} have become the dominant paradigm in AD systems for their remarkable performance.
\rev{After deployment, however, AD models continuously encounter new driving scenarios, newly collected data, and evolving task requirements, creating a persistent need for post-training adaptation. Updating the entire model for each iteration can be computationally expensive and may unnecessarily alter modules that have already learned stable and effective representations. This motivates localized post-training, where only selected parts of a pretrained network are optimized.}
\revEnvBegin
\begin{figure}[tp]
    \centering
    \includegraphics[width=\linewidth]{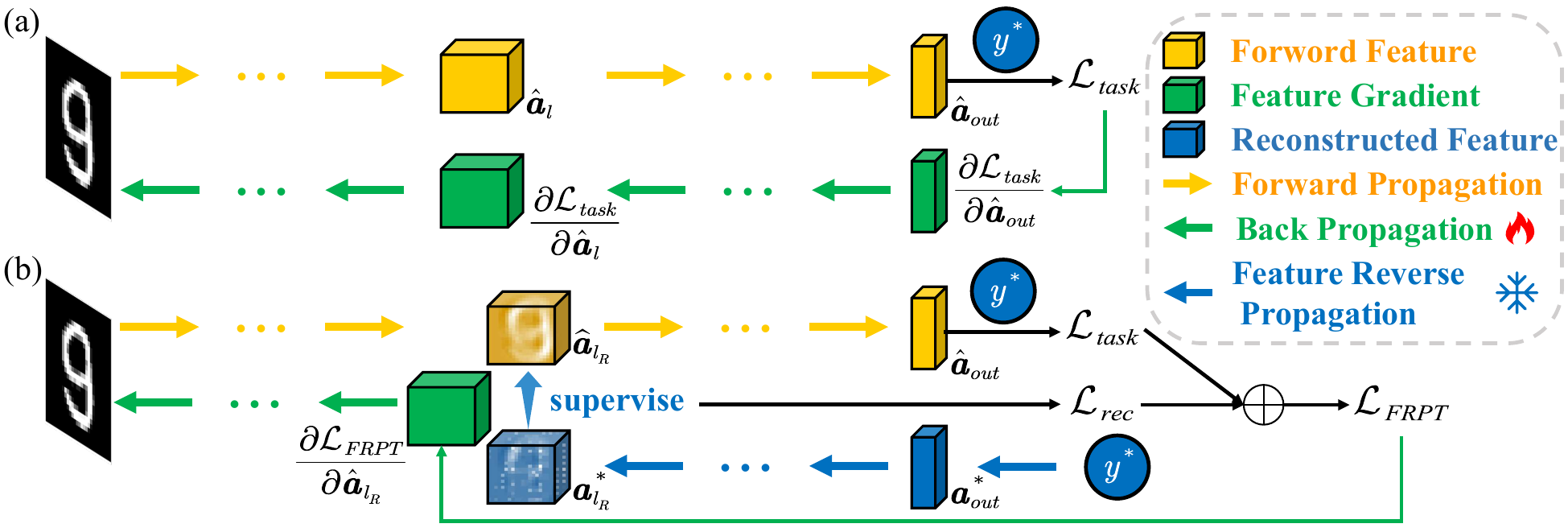}
    \caption{\rev{\textbf{Standard back-propagation and FR-PT.}
    (a) Standard back-propagation propagates feature gradients backward from the final task loss to update the trainable modules.
    (b) FR-PT freezes the downstream modules after reconstruction layer $l_R$ and reverse-propagates reconstructed features $\mathbf{a}^{\ast}$ to supervise the update of the upstream modules.}}
    \label{fig:fig1}
\end{figure}\par
\revEnvEnd
\rev{To facilitate localized optimization, some E2E AD systems introduce modular architectures and impose auxiliary supervision on selected module outputs. Representative examples include BEVFormer V2 \cite{Yang_2023_bevformer}, CLIP-BEVFormer \cite{pan2024clipbevformerenhancingmultiviewimagebased}, DiffStack \cite{pmlr-v205-karkus23a}, ChauffeurNet \cite{Bansal2018ChauffeurNetLT}, and 3D detection models with intermediate depth supervision \cite{9966379}. These methods rely on predefined functional modules whose outputs carry task-specific semantics and can therefore be paired with manually designed annotations. Nevertheless, such targets are not always available, and the resulting training scheme is inherently tied to the model architecture. This limits its applicability to arbitrary hidden representations or seamless E2E networks, where intermediate features generally lack directly accessible ground-truth targets for localized post-training.\par
}
\rev{Conventional back-propagation (BP) propagates gradients from the final task loss to determine how trainable parameters should be updated, but it does not explicitly specify what intermediate representation a selected module should produce. Such indirect supervision also offers limited insight into the network's internal behavior, which is particularly undesirable in complex and safety-critical AD pipelines \cite{explainable_2024, 10.1613/jair.1.12228}. As illustrated in Fig.~\ref{fig:fig1}, feature-level reverse propagation for post-training (FR-PT) follows a different principle: instead of propagating feature gradients through the network, it reconstructs target features $\mathbf{a}^{\ast}$ backward from the final ground-truth supervision and uses them to directly supervise the selected upstream modules. FR-PT thus complements the task loss with explicit feature-level targets and provides a more transparent view of localized optimization through feature reconstruction (FR).
}
\rev{Starting from the final ground-truth label, FR reverses the frozen downstream computation module by module to reconstruct the target feature $\mathbf{a}_{l_R}^{\ast}$ at a selected layer $l_R$ under the proposed Computation Consistency Principle (CCP) and Minimum Deviation Principle (MDP). This target directly supervises post-training of the upstream modules and also serves as a diagnostic reference, since its discrepancy from the original forward feature reflects how the selected representation should change under the fixed downstream mapping. FR-PT is therefore positioned as a localized post-training framework for pretrained networks, complementing conventional BP rather than replacing end-to-end training from scratch.\par
}
\rev{Our contributions are summarized as follows:}
\revEnvBegin
\begin{itemize}
    \item As a localized post-training framework, the proposed FR-PT reconstructs label-conditioned intermediate target features under the CCP/MDP formulation, providing direct supervision for selected modules and feature-deviation diagnostics of internal representation changes.
    \item The FR algorithms for different network operations are systematically developed. In particular, convolutional FR uses FFT-based reconstruction for efficiency, composite modules are handled iteratively, and a parallelizable optimal embedding algorithm maps category labels to reconstructed output targets.
    \item MDP-centered Tikhonov regularization is theoretically shown to stabilize ill-conditioned reconstruction, while a layer-wise analysis characterizes error accumulation across successive reverse operations.
    \item Extensive experiments on image-classification benchmarks and the NAVSIM trajectory-planning task validate FR-PT across different architectures and reconstruction locations.
\end{itemize}\par
\revEnvEnd
\section{Related Work}
\label{sec:related_work}
\subsection{\rev{Decoupled and Target-Based Learning}}
\begin{table*}[ht]
\centering
\caption{\rev{Comparison between FR-PT and representative backpropagation alternatives.}}
\label{tab:relatedwork_FR}
\renewcommand{\arraystretch}{1.25}
\footnotesize
\setlength{\tabcolsep}{2pt}
\begin{tabular}{@{}>{\raggedright\arraybackslash}p{0.075\textwidth}>{\raggedright\arraybackslash}p{0.145\textwidth}>{\raggedright\arraybackslash}p{0.4\textwidth}>{\centering\arraybackslash}p{0.1\textwidth}>{\centering\arraybackslash}p{0.1\textwidth}>{\centering\arraybackslash}p{0.1\textwidth}@{}}
\toprule
\revCell{Method} & \revCell{\makecell[l]{Back-propagated\\information}} & \revCell{Information generation mechanism} & \revCell{\makecell[c]{Additional\\model}} & \revCell{\makecell[c]{Explicit \\feature target}} & \revCell{\makecell[c]{Feature-level\\diagnostic}}\\
\midrule
\revCell{BP} & \revCell{Exact gradients} & \revCell{Chain-rule computation through the downstream graph} & \revCell{$\times$} & \revCell{$\times$} & \revCell{$\times$}\\
\revCell{SG \cite{jaderberg2017decoupledneuralinterfacesusing}} & \revCell{Predicted gradients} & \revCell{Local predictor estimates gradients} & \revCell{\checkmark} & \revCell{$\times$} & \revCell{$\times$}\\
\revCell{TP \cite{bengio2014targetprop}} & \revCell{Target features} & \revCell{Learned approximate inverse of the forward mapping} & \revCell{\checkmark} & \revCell{\checkmark} & \revCell{\checkmark}\\
\revCell{DTP \cite{DTP}} & \revCell{Difference-corrected target features} & \revCell{Approximate inverse mapping with reconstruction-error correction} & \revCell{\checkmark} & \revCell{\checkmark} & \revCell{\checkmark}\\
\revCell{\textbf{FR-PT (Ours)}} & \revCell{Reconstructed target features} & \revCell{Solves the inverse problem formulated by CCP and MDP} & \revCell{$\times$} & \revCell{\checkmark} & \revCell{\checkmark}\\
\bottomrule
\end{tabular}
\end{table*}
\rev{Localized optimization is also related to the broader literature on decoupled learning, which predominantly aims to relax the locking and coordination constraints imposed by conventional BP. Existing approaches mainly restructure module-wise optimization so that different parts of a network can be updated more independently or asynchronously. Jaderberg et al. \cite{jaderberg2017decoupledneuralinterfacesusing} developed decoupled neural interfaces, where synthetic gradients (SG) are used to remove update locking, allowing for asynchronous training of different modules. As a simpler yet more parallelizable alternative, greedy local learning that assigns auxiliary objectives to successive blocks, was later introduced \cite{belilovsky2020decoupled}. Yang et al. \cite{yang2024successive} proposed successive gradient reconciliation to coordinate the local gradients of neighboring modules without breaking gradient isolation, thereby improving deep local learning.}
Similarly, Zhuang et al. \cite{zhuang2021fully} explored delayed gradient updates to achieve a fully decoupled training scheme that can train modules independently, reducing memory overhead while maintaining competitive performance.
Peng et al.\cite{peng2023decoupled} extended decoupled learning by incorporating re-computation and weight prediction strategies to mitigate memory explosion.
\rev{These methods primarily address training-time locking and gradient dependencies, rather than localized post-training with a fixed downstream network, and generally do not provide explicit feature targets for direct local supervision or feature-level diagnosis.}\par
\rev{Target propagation (TP) and difference target propagation (DTP) are the closest conceptual precedents, as they replace layer-wise gradients with target activations \cite{bengio2014targetprop,DTP,NEURIPS2020_e7a425c6}. Classical TP learns feedback mappings that approximately invert successive forward transformations, while DTP introduces a difference correction to compensate for feedback reconstruction error. More recently, Ernoult et al. \cite{ernoult2022scaling} introduced a feedback-training scheme based on learned backpropagation targets, improving the theoretical alignment of DTP with BP while scaling it to more challenging vision benchmarks.Explicit feature targets were partly motivated by concerns over the biological plausibility of exact BP \cite{lillicrap2020backpropbrain}, although biologically motivated alternatives can be difficult to scale in practice \cite{bartunov2018scalability}. FR-PT adopts the feature-target perspective but differs in both objective and construction: it is designed for post-training of a pretrained forward model, requires no separately trained feedback network, and reconstructs each target by solving the inverse problem induced by the actual frozen downstream operator. Table~\ref{tab:relatedwork_FR} summarizes the key differences between FR-PT and representative BP alternatives.\par
}
\rev{Other post-training and trust-aware methods address complementary objectives. Post-training quantization for vision transformers preserves behavior under low-bit weights and activations \cite{liu2021posttrainingquantizationvit}. DeepTrust propagates subjective-logic opinions to quantify predictive trustworthiness \cite{cheng2020hope}, and TrustCNets evaluate or redesign CNN components using trust-related criteria \cite{pmlr-v199-cheng22a}. FR-PT neither compresses the network nor defines a trust score. Its object is the discrepancy between an observed intermediate feature and a label-conditioned target reconstructed through the frozen downstream mapping.}
\subsection{\rev{Explainable AI (XAI)}}
\label{sec:xai}
\rev{Extensive studies have explored diverse XAI techniques for interpreting intermediate representations and model decisions in AD pipelines.}
Abukmeil et al.\cite{9551172} proposed the first explainable semantic segmentation model for AD based on the variational autoencoder, which used multiscale second-order derivatives between the latent space and the encoder layers to capture the curvatures of the neurons' responses.
Gou et al. \cite{9233993} developed a visual analytics system equipped with a disentangled representation learning and semantic adversarial learning, to assess, understand, and improve traffic light detection.
\rev{Moreover, interactive software that allows real-time inspection of neuron activations and feature visualization via regularized optimization can further support intuitive understanding of learned representations \cite{yosinski2015understandingneuralnetworksdeep}. Shao et al. \cite{shao2023interfuser} incorporated interpretability directly into an end-to-end driving pipeline by generating semantically meaningful intermediate features that constrain vehicle actions within safe sets. At the decision level, Nie et al. \cite{nie2024reason2drive} introduced Reason2Drive, which organizes driving interpretation into perception, prediction, and reasoning chains to provide human-readable rationales for downstream planning decisions.}\par
\rev{\rev{Beyond AD-specific approaches, a broad range of XAI methods have been developed for trained visual models. Gradient saliency and integrated gradients attribute a prediction to input variables \cite{Simonyan2013DeepIC,pmlr-v70-sundararajan17a}. CAM-family methods localize class-related spatial responses in convolutional features \cite{Discriminative,cam,layercam}, while perturbation methods such as LIME and RISE estimate local input importance \cite{ribeiro2016why,petsiuk2018rise}. Representation-oriented methods visualize preferred inputs or test sensitivity to human-defined concepts \cite{zeiler2013visualizingunderstandingconvolutionalnetworks,yosinski2015understandingneuralnetworksdeep,kim2018interpretability}. Recently,Wang et al. \cite{wang2024pace} introduced probabilistic concept explainers for vision transformers, which model the distributions of patch embeddings to provide multi-level post-hoc conceptual explanations.In AD, these techniques have been applied to road-scene segmentation, 3D detection, prediction, planning, and runtime introspection \cite{zablocki_explainability_2022,app12115310,9879848,10161132,10260283}. While valuable for diagnosing existing predictions and representations, they do not construct label-conditioned intermediate feature targets for supervising selected modules during post-training.\par }
}
\section{Method}
\label{sec:method}
\revEnvBegin
\begin{figure*}[tp]
    \centering
    \includegraphics[width=\linewidth]{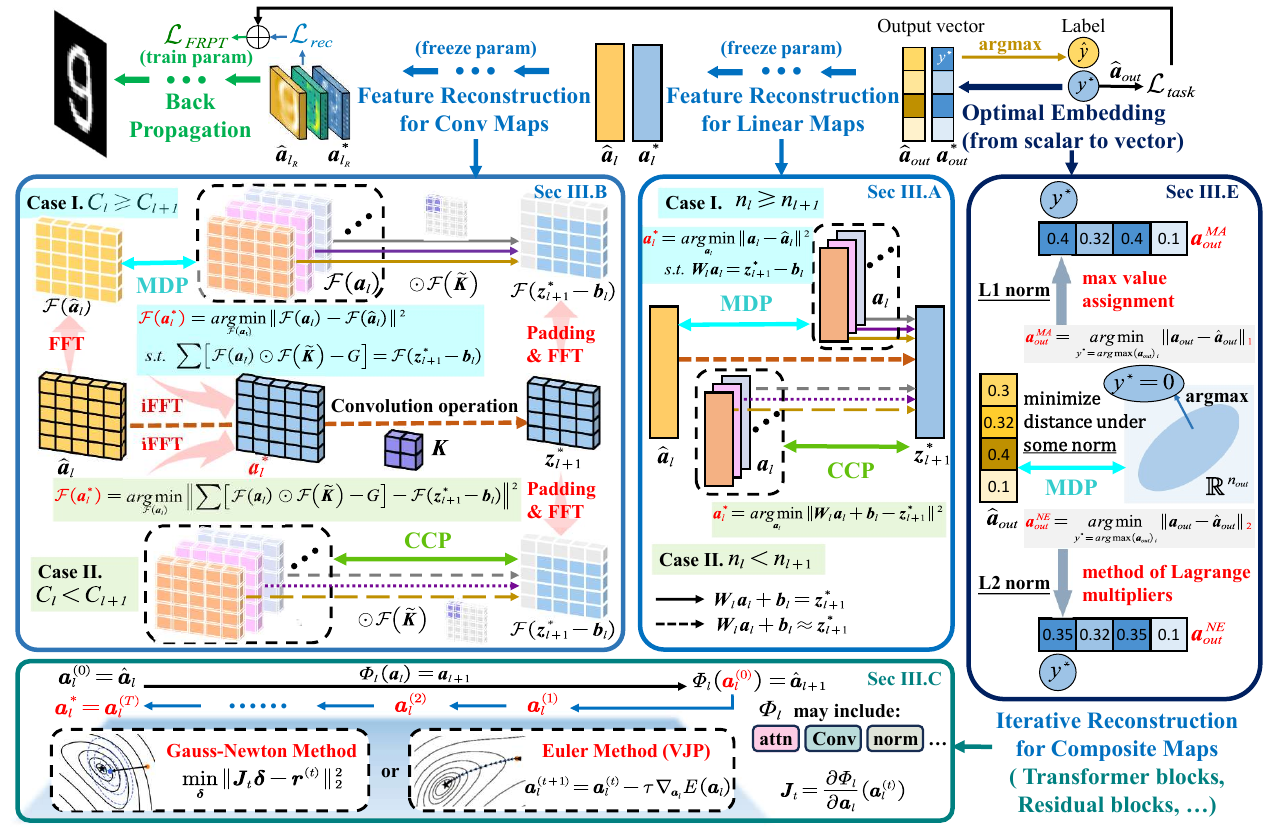}
    \caption{\rev{\textbf{Overview of FR-PT and the feature reconstruction operators.} Starting from the ground-truth labels, FR-PT traverses the frozen downstream path in reverse and reconstructs intermediate target features module by module until reaching the selected reconstruction layer $l_R$ (blue path). The reconstructed $\mathbf{a}_{l_R}^{\ast}$ is then used as feature-level supervision to optimize the remaining trainable modules. Feature reverse propagation is formulated under the CCP and MDP, with operator-specific reconstruction methods.}}
    \label{fig:method_overview}
\end{figure*}\par
\revEnvEnd
\rev{Consider a pretrained network composed of a sequence of modules. Let $\hat{\mathbf{a}}_l$ denote the forward feature at layer $l$ and $\mathbf{a}_{l}^{\ast}$ the reconstructed target feature at layer $l$. As illustrated in Fig.~\ref{fig:method_overview}, FR-PT keeps the downstream part frozen and reconstructs intermediate target features in reverse, rather than propagating task gradients through it. Taking image classification as an example, FR starts from the ground-truth label $\mathbf{y}^{\ast}$ (or the desired output $\mathbf{a}_{out}^{\ast}$ for regression tasks), and reconstructs $\mathbf{a}_{l}^{\ast}$ from $\mathbf{a}_{l+1}^{\ast}$ through each frozen module $\Phi_{l}$ until the selected reconstruction layer $l_R$ is reached.\par
}
\rev{The reconstructed feature $\mathbf{a}_{l_R}^{\ast}$ then provides direct feature-level supervision for post-training. Inspired by loss balancing strategies in multi-objective learning \cite{8578879}, we adaptively scale the reconstruction term according to the moving-average magnitudes of the task and reconstruction losses:
}
\revEnvBegin
\begin{align}
    &\mathcal{L}_{task} = \operatorname{CrossEntropy}(\mathbf{y}^{\ast},\hat{\mathbf{a}}_{out})\\
    &\mathcal{L}_{rec} = \operatorname{MSE}(\mathbf{a}_{l_R}^{\ast},\hat{\mathbf{a}}_{l_R})\\
    &\lambda_{rec}^{(t)}=\operatorname{clip}\!\left(
    \frac{c_{\mathrm{rec}}\bar{\mathcal{L}}_{task}^{(t)}}{\bar{\mathcal{L}}_{rec}^{(t)}+\epsilon},
    \lambda_{\min},\lambda_{\max}\right)\text{ if }c_{\mathrm{rec}}>0\text{ else }0\\
    &\mathcal{L}_{FRPT}=\mathcal{L}_{task}+\lambda_{rec}^{(t)}\mathcal{L}_{rec},
    \label{ltotal}
\end{align}
\revEnvEnd
\rev{where $c_{\mathrm{rec}}\in[0,1]$ controls the strength of reconstruction supervision and is set to zero for the matched standard post-training reference. $\bar{\mathcal{L}}_{task}^{(t)}$ and $\bar{\mathcal{L}}_{rec}^{(t)}$ denote the exponential moving averages of the task and reconstruction losses, respectively, and are updated without gradient propagation. We set $\epsilon=10^{-8}$ and clip $\lambda_{rec}^{(t)}$ to $[10^{-5},10]$.\par
}
\rev{Thus, the problem reduces to how to obtain the target feature $\mathbf{a}_{l}^{\ast}$. As shown in Fig.~\ref{fig:method_overview}, feature reverse propagation through different operators is governed by two principles. The \emph{Computation Consistency Principle} (CCP) favors reconstructed targets that remain consistent with the computation of the frozen downstream operator, i.e., $\min \|\Phi_l(\mathbf{a}_{l})-\mathbf{a}_{l+1}\|_2^2$. In contrast, the \emph{Minimum Deviation Principle} (MDP) minimizes the deviation from the original forward feature $\|\mathbf{a}_l-\hat{\mathbf{a}}_l\|_2^2$, encouraging the reconstructed target to stay close to the representation learned by the upstream modules. Together, the two principles aim to better approximate the feature evolution from the network input toward the ground-truth output while preserving the mappings learned during pre-training.\par
}
\rev{Accordingly, FR prioritizes consistency with the downstream target under CCP, while MDP is used to select among non-unique solutions or regularize the reconstruction toward the original feature when the inverse problem is numerically unstable. The following sections instantiate these principles for different neural network operators and modules.\par
}
\rev{In this work, FR-PT adopts an offline reconstruction protocol: target features are reconstructed once from the pretrained checkpoint for the post-training dataset and stored before optimization, and are subsequently used as fixed feature-level supervision during post-training.
}
\subsection{\rev{Feature Reconstruction for Linear Maps}}
\label{sec:linear layer}
\rev{For an affine module $\Phi_l$ defined by $\mathbf{z}_{l+1}=\Phi_l(\mathbf{a}_l)=\mathbf{W}_l\mathbf{a}_l+\mathbf{b}_l$, with $n_l=\dim(\mathbf{a}_l)$ and $n_{l+1}=\dim(\mathbf{z}_{l+1})$, we solve for the reconstructed feature $\mathbf{a}_{l}^{\ast}$ given the module output target $\mathbf{z}_{l+1}^{\ast}$. We distinguish two reconstruction geometries according to the input and output widths.\par 
}
\rev{\textbf{Case I (width-reducing): $n_l\ge n_{l+1}$.} When the affine map reduces feature width, multiple preimages may satisfy CCP. We therefore impose MDP as the selection principle and seek the exact preimage closest to the former feature,
}
\revEnvBegin
\begin{align}
    \mathbf{a}_{l}^{\ast}=\arg\min_{\mathbf{a}_l}
    &\ \|\mathbf{a}_l-\hat{\mathbf{a}}_l\|_2^2
    \notag\\
    \mathrm{s.t.}\quad&\mathbf{W}_l\mathbf{a}_l=\mathbf{z}_{l+1}^{\ast}-\mathbf{b}_{l}.
    \label{eq:linear_model}
\end{align}
\revEnvEnd
\rev{If $\mathbf{W}_l$ has full row rank, the unique minimizer is}
\revEnvBegin
\begin{equation}
    \mathbf{a}_{l}^{\ast}
    =\hat{\mathbf{a}}_l+\mathbf{W}_l^T
    (\mathbf{W}_l\mathbf{W}_l^T)^{-1}
    (\mathbf{z}_{l+1}^{\ast}-\mathbf{b}_{l}-\mathbf{W}_l\hat{\mathbf{a}}_l),
    \label{eq:linear_caseI_closed}
\end{equation}
\revEnvEnd
\rev{which geometrically projects $\hat{\mathbf{a}}_l$ onto the affine solution set $\{\mathbf{a}:\mathbf{W}_l\mathbf{a}+\mathbf{b}_{l}=\mathbf{z}_{l+1}^{\ast}\}$. The KKT derivation is given in Supplementary Part~A.1.\par }
\rev{\textbf{Case II (width-expanding): $n_l<n_{l+1}$.} When the output width exceeds the input width, an exact preimage may not exist. FR therefore minimizes CCP directly through least squares,
}
\revEnvBegin
\begin{equation}
    \mathbf{a}_{l}^{\ast}
    =\arg\min_{\mathbf{a}_l}
    \|\mathbf{W}_l\mathbf{a}_{l}-(\mathbf{z}^{\ast}_{l+1}-\mathbf{b}_{l})\|_2^2.
    \label{eq:lsp_linear}
\end{equation}
\revEnvEnd
\rev{Geometrically, $\mathbf{W}_l\mathbf{a}_{l}^{\ast}$ is the orthogonal projection of $\mathbf{z}^{\ast}_{l+1}-\mathbf{b}_{l}$ onto the column space of $\mathbf{W}_l$.\par }
\rev{\textbf{Reliability-gated MDP-centered regularization:}
In deep neural networks, ill-conditioned weight matrix is common. However, the conditioning of $\mathbf{W}_l$ directly affects the numerical reliability of feature reconstruction. Explicitly estimating its condition number, however, requires an SVD decomposition and is unnecessarily expensive at every reverse step. We therefore first apply the fast nominal linear-system or least-squares solver above and assess the resulting $\mathbf{a}_l^\ast$ before invoking regularization.  
The reliability gate checks finiteness, the entrywise magnitude $\|\mathbf{a}_l^\ast\|_\infty\le M_{\infty}$, the relative deviation from the original forward feature $\rho_{\mathrm{dev}}\le M_{\mathrm{dev}}$, and computation consistency. The last one is measured by the normalized forward residual $\rho_1 \le M_1$ for Case~I, and the normalized first-order least-squares optimality residual $\rho_2 \le M_2$ for Case~II. 
Only samples that fail the reliability gate invoke the SVD-based MDP-centered Tikhonov fallback 
}
\revEnvBegin
\begin{equation}
    \mathbf{a}_{l,\alpha}^{\ast}
    =\arg\min_{\mathbf{a}_l}
    \|\mathbf{W}_l\mathbf{a}_l+\mathbf{b}_l-\mathbf{z}_{l+1}^{\ast}\|_2^2
    +\alpha\|\mathbf{a}_l-\hat{\mathbf{a}}_l\|_2^2, \alpha>0,
    \label{eq:mdp_tikhonov_main}
\end{equation}
\revEnvEnd
\rev{or equivalently, with $\boldsymbol{\delta}_{\alpha}=\mathbf{a}_{l,\alpha}^{\ast}-\hat{\mathbf{a}}_l$ and $\mathbf{r}=\mathbf{z}_{l+1}^{\ast}-\mathbf{b}_{l}-\mathbf{W}_l\hat{\mathbf{a}}_l$, it suffices to solve}
\revEnvBegin
\begin{equation}
    (\mathbf{W}_l^{T}\mathbf{W}_l+\alpha\mathbb{I})\boldsymbol{\delta}_{\alpha}
    =\mathbf{W}_l^{T}\mathbf{r}.
    \label{eq:linear_tikhonov_main}
\end{equation}
\revEnvEnd
\rev{For each failed sample, the regularization coefficient is
}
\revEnvBegin
\begin{align}
     &\alpha
   =\max\!\left\{
    \left(\frac{\sigma_{\max}}{\kappa_{\mathrm{eff}}}\right)^2,
    \epsilon_{\mathrm{mach}}\max(\sigma_{\max}^2,1),
    \alpha_{\mathrm{mag}}
    \right\},\\
    &\alpha_{\mathrm{mag}}
    =
    \left(\dfrac{\|\mathbf{r}\|_2}{2D_{max}}\right)^2 \text{ if } D_{max}>0 \text{ else } 0.
    \label{eq:linear_alpha_main}
\end{align}
\revEnvEnd
\rev{where $D_{max}=M_{\infty}-\|\hat{\mathbf{a}}_l\|_{\infty}$, $\sigma_{\max}$ is the largest singular value of $\mathbf{W}_l$, and $\kappa_{\mathrm{eff}}$ controls the strength of Tikhonov damping. The three terms provide scale-aware damping, a machine-precision floor, and a sufficient lower bound associated with the feature-magnitude safeguard, respectively. Their derivation, spectral interpretation, and rank-deficient behavior are detailed in Supplementary Part~A.2, with hyperparameter settings provided in Supplementary Part~B.1.2.\par
}
\subsection{\rev{Feature Reconstruction of Convolutional Maps}}
\label{conv_layer}
\rev{For a convolutional operator in a CNN, given the target output $\mathbf{z}_{l+1}^{\ast}\in\mathbb{R}^{BS\times C_{l+1}\times H_{l+1}\times W_{l+1}}$, a frozen convolution kernel $\mathbf{K}_{l}$ with bias $\mathbf{b}_{l}$, and the forward feature $\hat{\mathbf{a}}_l\in\mathbb{R}^{BS\times C_l\times H_l\times W_l}$, we reconstruct the corresponding target feature $\mathbf{a}_{l}^{\ast}$. The implemented reverse operators assume a single group, unit stride, and unit dilation. We use zero padding as an example in the following discussion.\par
}
\rev{\textbf{(1)} Fundamentally, convolution is a linear transformation; thus, its feature reverse propagation can be achieved in a single step. The explicit \textbf{Matrix-Solve} implementation constructs the exact linear map $\mathbf{A}_l\in\mathbb{R}^{N_o\times N_i}$, where $N_i=C_lH_lW_l$ and $N_o=C_{l+1}H_{l+1}W_{l+1}$, by vectorizing the bias-free convolution operator. For each sample $x$, the convolution can then be equivalently expressed in matrix form as
}
\revEnvBegin
\begin{equation}
    \operatorname{vec}(\mathbf{z}_{l+1}^{\ast}[x]-\mathbf{b}_l)
    =\mathbf{A}_l\operatorname{vec}(\mathbf{a}_{l}^{\ast}[x]).
    \label{eq:exact_matrix}
\end{equation}
\revEnvEnd
\rev{Define the residual
$\mathbf{r}_{l}^{x}=\operatorname{vec}(\mathbf{z}_{l+1}^{\ast}[x]-\mathbf{b}_l)
-\mathbf{A}_{l}\operatorname{vec}(\hat{\mathbf{a}}_{l}[x])$. The two reconstruction geometries are then handled in the same manner as in Sec.~\ref{sec:linear layer}.
In \textbf{Case I $N_i\ge N_o$}, it uses
}
\revEnvBegin
\begin{equation}\label{eq:conv_exact_caseI}
    \mathbf{a}_{l}^{\ast}[x]=\hat{\mathbf{a}}_l[x]+\operatorname{unvec}\!\left\{\mathbf{A}_{l}^{T}
    (\mathbf{A}_{l}\mathbf{A}_{l}^{T})^{-1}\mathbf{r}_{l}^{x}\right\}.
\end{equation}
\revEnvEnd
\rev{In \textbf{Case II $N_i<N_o$}, it uses the direct least-squares solution}
\revEnvBegin
\begin{equation}\label{eq:conv_exact_caseII}
    \mathbf{a}_{l}^{\ast}[x]=\operatorname{unvec}\!\left\{\arg\min_{\mathbf{a}\in\mathbb{R}^{N_i}}
    \|\mathbf{A}_{l}\mathbf{a}-\operatorname{vec}(\mathbf{z}_{l+1}^{\ast}[x]-\mathbf{b}_l)\|_2^2\right\}.
\end{equation}
\revEnvEnd
\rev{If the Matrix-Solve candidate fails the reliability check as described in the linear case (Sec.~\ref{sec:linear layer}), we instead solve the MDP-centered damped normal/Gram system based on $\mathbf{A}_l$. Matrix-Solve exactly represents the \texttt{conv2d} operator in neural networks, but explicitly storing $\mathbf{A}_l$ requires $\mathcal{O}(N_oN_i)$ memory and dense matrix factorizations become computationally prohibitive for large feature maps.\par
}
\rev{\textbf{(2)} To avoid materializing $\mathbf{A}_l$, the matrix-free \textbf{Iterative-CG} implementation evaluates $\mathbf{A}_l\mathbf{v}$ and $\mathbf{A}_l^T\mathbf{u}$ using convolution and transposed convolution, respectively. In \textbf{Case I} it solves
}
\revEnvBegin
\begin{equation}\label{eq:conv_cg_caseI}
    (\mathbf{A}_l\mathbf{A}_l^{T})\boldsymbol{\lambda}=\mathbf{r}_{l}^{x},\qquad
    \operatorname{vec}(\mathbf{a}_{l}^{\ast}[x])
    =\operatorname{vec}(\hat{\mathbf{a}}_l[x])+\mathbf{A}_l^{T}\boldsymbol{\lambda},
\end{equation}
\revEnvEnd
\rev{whereas in \textbf{Case~II} it solves the normal equation}
\revEnvBegin
\begin{equation}\label{eq:conv_cg_caseII}
    (\mathbf{A}_l^{T}\mathbf{A}_l)\operatorname{vec}(\mathbf{a}_{l}^{\ast}[x])
    =\mathbf{A}_l^{T}\operatorname{vec}(\mathbf{z}_{l+1}^{\ast}[x]-\mathbf{b}_l).
\end{equation}
\revEnvEnd
\rev{Both systems are solved by batched conjugate gradient without explicitly forming the corresponding Gram operators. If the result is unreliable, Case I adds a shared damping to the $\mathbf{A}_l\mathbf{A}_l^T$ system and recomputes the correction; Case II solves for the anchored correction through the damped $\mathbf{A}_l^T\mathbf{A}_l$ system. These damped formulations are algebraically equivalent to Eq.~\eqref{eq:mdp_tikhonov_main}.\par
}
\rev{\textbf{(3)} For higher scalability, we further develop the FFT-based reconstruction that exploits spectral structure of convolution. The circular convolution theorem converts spatial circular convolution into pointwise multiplication in the Fourier domain, which greatly simplifies the reverse computation. Thus, the spatially coupled convolution is decomposed into $H_l\times W_l$ independent channel-wise linear systems: for each frequency $(u,v)$, $\mathbf{K}^{F}_{u,v}\mathbf{a}^{F,x}_{u,v}
=\mathbf{d}^{F,x}_{u,v}$ holds,
where $\mathbf{K}^{F}_{u,v}\in\mathbb{C}^{C_{l+1}\times C_l}$ is the channel response obtained from the discrete Fourier transform (DFT) of the zero-padded flipped kernel, $\mathbf{a}^{F,x}_{u,v}\in\mathbb{C}^{C_l}$ is the input-feature Fourier coefficient vector, and $\mathbf{d}^{F,x}_{u,v}\in \mathbb{C}^{C_{l+1}}$ is the frequency-domain target after finite-boundary treatment.\par
}
\rev{The finite boundary distinguishes two implementations. \textbf{FFT-Padded} explicitly zero-pads the input feature by $p$ pixels on each side, evaluates circular convolution on the enlarged $(H_l+2p)\times(W_l+2p)$ grid, and selects the output block corresponding to the original CNN convolution. \textbf{FFT-Boundary} instead retains the original $H_l\times W_l$ grid and subtracts from the circular convolution the wrap-around terms. Both constructions reproduce the original forward \texttt{conv2d} operator exactly; their detailed identities and numerical verification are provided in Supplementary Part~A.3.\par
}
\rev{For reverse reconstruction, the two methods differ only in the construction of $\mathbf{d}^{F,x}_{u,v}$ and the former-feature coefficient $\hat{\mathbf{a}}^{F,x}_{u,v}$. For FFT-Padded, they are obtained on the enlarged padded grid. For FFT-Boundary, the target is embedded and spatially aligned on the original grid, while the boundary correction, which would depend on the unknown $\mathbf{a}_l^\ast[x]$, is evaluated at the former forward feature $\hat{\mathbf{a}}_l[x]$ to preserve frequency-wise linearity. Once $\mathbf{d}^{F,x}_{u,v}$ is obtained, both implementations use the same reconstruction models.\par
}
\rev{By Parseval's theorem, minimizing the spatial CCP/MDP discrepancy is equivalent to minimizing its corresponding frequency-domain discrepancy. 
In \textbf{Case I, $C_l\ge C_{l+1}$}, we select the feasible coefficient nearest to the former feature,
}
\revEnvBegin
\begin{equation}\label{eq:conv_fft_caseI}
\mathbf{a}^{F,x,\ast}_{u,v}
=\arg\min_{\mathbf{a}^{F}\in\mathbb{C}^{C_l}}
\|\mathbf{a}^{F}-\hat{\mathbf{a}}^{F,x}_{u,v}\|_2^2 
\quad\mathrm{s.t.}
\mathbf{K}^{F}_{u,v}\mathbf{a}^{F}
=\mathbf{d}^{F,x}_{u,v}.
\end{equation}
\revEnvEnd
\rev{In \textbf{Case II, $C_l<C_{l+1}$}, we solve the frequency-wise CCP least-squares problem}
\revEnvBegin
\begin{equation}\label{eq:conv_fft_caseII}
\mathbf{a}^{F,x,\ast}_{u,v}
=
\arg\min_{\mathbf{a}^{F}\in\mathbb{C}^{C_l}}
\|\mathbf{K}^{F}_{u,v}\mathbf{a}^{F}
-\mathbf{d}^{F,x}_{u,v}\|_2^2.
\end{equation}
\revEnvEnd
\rev{Therefore, instead of solving a system over all spatial variables jointly, the FFT formulation decomposes the reconstruction into multiple independent systems involving only the channel dimensions.
The reconstructed spatial feature is finally obtained by inverse DFT, followed by removal of the artificial padding for FFT-Padded.\par
}
\rev{Finally, numerical stabilization is applied identically to both FFT implementations but independently per frequency--sample pair. The nominal solution of Eq.~\eqref{eq:conv_fft_caseI} or Eq.~\eqref{eq:conv_fft_caseII} is first tested for numerical reliability. A failed pair is replaced by the MDP-centered Tikhonov reconstruction
}
\revEnvBegin
\begin{equation}\label{eq:conv_fft_tikhonov}
(\mathbf{a}^{F,x,\ast}_{u,v})_{\alpha}
=
\arg\min_{\mathbf{a}^{F}}
\|\mathbf{K}^{F}_{u,v}\mathbf{a}^{F}
-\mathbf{d}^{F,x}_{u,v}\|_2^2
+
\alpha_{u,v}^{x}
\|\mathbf{a}^{F}
-\hat{\mathbf{a}}^{F,x}_{u,v}\|_2^2,
\end{equation}
\revEnvEnd
\rev{where $\alpha_{u,v}^{x}$ is selected from the local frequency response using the same spectral-scale, machine-precision, and feature-magnitude criteria as Eq.~\eqref{eq:linear_alpha_main}. Hence, the FFT implementations preserve the CCP/MDP reconstruction geometry of the spatial formulation while replacing the large convolutional system by multiple small, frequency-decoupled channel systems.\par
}
\rev{The two boundary treatments introduce different approximations only in the reverse problem. FFT-Padded treats the enlarged padded grid as reconstruction variables; the recovered padding entries are therefore not constrained to remain zero, and cropping them after the inverse DFT may introduce CCP inconsistency. For FFT-Boundary, it preserves the physical number of feature variables, but freezes the unknown boundary correction at the former feature $\hat{\mathbf{a}}_l[x]$. The resulting spatial CCP residual is bounded by the frequency-system solve residual plus a kernel-dependent constant times the reconstructed boundary deviation $\|P_{\partial_p}(\mathbf{a}_l^{\ast}-\hat{\mathbf{a}}_l)\|_F$. Hence MDP-centered control also limits the boundary-anchoring approximation. Specially, for $p=0$, no circular-boundary term is introduced.
In addition, when $H_{l+1}<H_l$ or $W_{l+1}<W_l$, embedding the observed output into the full $H_l\times W_l$ grid introduces auxiliary constraints outside the retained convolutional output. 
These drawbacks are analyzed in detail in Supplementary Part~A.2.\par
}
\subsection{\rev{Feature Reconstruction of Composite Blocks}}
\rev{Modern architectures contain coupled residual and attention blocks, for which reversing each internal operation independently may violate block-level dependencies and increase the CCP error. This section reconstructs the input feature of the entire differentiable composite map $\mathbf{a}_{l+1}=\Phi_l(\mathbf{a}_l)$ iteratively, initialized from $\mathbf{a}_l^{(0)}=\hat{\mathbf{a}}_l$.\par
}
\rev{Exact CCP solvability for a nonlinear composite block depends on whether the target output lies in the range of the block mapping. For an affine map, full row rank is necessary and sufficient for surjectivity onto the output space. For $\Phi_l(\mathbf{a}_l)=\mathbf{P}\mathbf{a}_l+F(\mathbf{a}_l)$ with a full-row-rank shortcut $\mathbf{P}$ and a continuous bounded residual branch $F$ are sufficient for surjectivity. The stronger condition $\operatorname{Lip}(F)<1$ for $\Phi_l(\mathbf{a}_l)=\mathbf{a}_l+F(\mathbf{a}_l)$ guarantees a unique and stable inverse. These sufficient conditions cover bounded residual branches and contraction-controlled blocks, but not arbitrary ReLU/GELU residual networks; in the latter case the iterative solver pursues the best CCP fit. Proofs are given in Supplementary Part~A.4.\par
}
\rev{The implementation automatically selects either an explicit-Jacobian route or a matrix-free VJP (Vector--Jacobian Product) route according to the estimated number of Jacobian entries.\par
}
\textbf{\rev{(1) Explicit-Jacobian Gauss--Newton.}} \rev{At iteration $t$, define}
\revEnvBegin
\begin{equation}
    \mathbf{r}^{(t)}=\mathbf{a}_{l+1}^{\ast}-\Phi_l(\mathbf{a}_l^{(t)}),\qquad
    \mathbf{J}_t=\left.\frac{\partial \Phi_l(\mathbf{a}_l)}{\partial \mathbf{a}_l}\right|_{\mathbf{a}_l=\mathbf{a}_l^{(t)}}.
\end{equation}
\revEnvEnd
\rev{The nominal Gauss--Newton step solves the linearized CCP problem
}
\revEnvBegin
\begin{equation}
    \boldsymbol{\delta}^{(t)}
    =\arg\min_{\boldsymbol{\delta}}
    \|\mathbf{J}_t\boldsymbol{\delta}-\mathbf{r}^{(t)}\|_2^2.
    \label{eq:gn_nominal_main}
\end{equation}
\revEnvEnd
\rev{Each linearized step is checked for numerical reliability. A reliable step is norm-clipped if necessary and passed to backtracking to determine damping $d$; and then it is accepted and let $\mathbf{a}_l^{(t+1)}=\mathbf{a}_l^{(t)}+d\cdot\boldsymbol{\delta}^{(t)}$ only if the resulting feature satisfies the feature/deviation bounds and strictly decreases the nonlinear CCP residual.\par
}
\rev{If the nominal Gauss--Newton step fails its linearized reliability test, the implementation solves the MDP-centered fallback described below
}
\revEnvBegin
\begin{equation}
\boldsymbol{\delta}^{(t)}
=\arg\min_{\boldsymbol{\delta}}
\|\mathbf{J}_t\boldsymbol{\delta}-\mathbf{r}^{(t)}\|_2^2
+\alpha_t\|\mathbf{a}_l^{(t)}+\boldsymbol{\delta}-\hat{\mathbf{a}}_l\|_2^2,
\label{eq:gn_delta}
\end{equation}
\revEnvEnd
\rev{where $\alpha_t$ is determined by Eq.~\ref{eq:linear_alpha_main}. The regularized step must satisfy its stationarity check and the same clipped backtracking nonlinear-descent test. Reliability rules and coefficient construction are detailed in Supplementary Part~A.2, and iterative hyperparameters in Sec.~B.1.2.\par }
\rev{\textbf{(2) Matrix-free VJP method.} 
If the explicit Jacobian would exceed the prescribed size limit, the implementation directly minimizes the squared relative CCP residual (Eq.~\ref{eq:supp_vjp_ccp}) with Adam and obtains the required derivatives through VJPs. 
}
\revEnvBegin
\begin{equation}
    \min E_{\mathrm{VJP}}(\mathbf{a}_{l}^{(t)})
    =\frac{1}{BS}\sum_{x=1}^{BS}
    \frac{\|\Phi(\mathbf{a}_{l}^{(t)}[x])-\mathbf{a}^{\ast}[x]\|_2^2}
    {\|\mathbf{a}^{\ast}[x]\|_2^2+\epsilon}
    \label{eq:supp_vjp_ccp}
\end{equation}
\revEnvEnd
\rev{Each feature-space update is clipped to a maximum step norm, after which the iterate is projected onto an MDP-centered deviation ball around $\hat{\mathbf{a}}_l$. 
After each optimizer step, the update norm is clipped and the feature is projected onto the ball
}
\revEnvBegin
\begin{equation}
    \|\mathbf{a}[x]-\hat{\mathbf{a}}[x]\|_2
    \le \rho\,\max\{\|\hat{\mathbf{a}}[x]\|_2,\epsilon_{\mathrm{ref}}\},
    \label{eq:supp_vjp_projection}
\end{equation}
\revEnvEnd
\rev{with $\rho=10^2$. This is a hard MDP-centered feasibility projection, not a second penalized optimization. The best finite iterate according to relative CCP is retained. If the final feature still violates the global magnitude/deviation reliability check, the code returns the anchor $\hat{\mathbf{a}}$.\par
}
\subsection{\rev{Feature Reconstruction of Other Maps}}
\rev{Other common auxiliary operators use lightweight reverse rules that match the implementation.\par
}
\rev{\textbf{(a) Nonlinear activations.}
Given the former pre-activation $\hat{\mathbf{z}}_l$ and the target activation $\mathbf{a}_l^{\ast}$, FR reconstructs $\mathbf{z}_l^{\ast}$ such that minimizes CCP error $\|\sigma_l(\mathbf{z}_l^{\ast})-\mathbf{a}_l^{\ast}\|$, and then the MDP error $\|\mathbf{z}_l^{\ast}-\hat{\mathbf{z}}_l\|$. Generally, target values outside the attainable range are projected to a numerically feasible range. Monotone activations, including tanh, sigmoid, and ELU, are reversed analytically with small range clamps to avoid infinite boundary values. Piecewise activations such as ReLU, ReLU6, HardTanh, and HardSigmoid are reversed branch-wise, with saturated regions resolved by selecting the preimage nearest to $\hat{\mathbf{z}}_l$. For non-monotone GELU and SiLU, candidate roots are obtained by bisection on their monotone branches and the nearest feasible root is retained. For Softmax, the target probabilities determine the logits up to an additive constant, which is chosen to minimize their deviation from the former logits. Detailed formulas and examples are provided in Supplementary Part~A.5.\par
}
\textbf{(b) Batch normalization.} BatchNorm is reversed only in evaluation mode using its frozen running mean and variance. For a channel with nonzero affine scale $\gamma_l$,
\revEnvBegin
\begin{equation}
    z_l^{\ast}
    =\frac{a_l^{\ast}-\beta_l}{\gamma_l}
      \sqrt{\operatorname{Var}_l+\epsilon_{BN}}
      +\operatorname{Mean}_l.\notag
\end{equation}
\revEnvEnd
\rev{If $\gamma_l=0$, the channel is not identifiable from the output and the former input value is retained, which is the MDP choice among valid preimages.\par
}
\rev{\textbf{(c) Layer normalization.} Unlike BatchNorm, LayerNorm computes its statistics from each input and is therefore not directly inverted from stored statistics. After safely removing the affine transform, let $\bar{\mathbf{z}}$ denote the target normalized feature over the normalized dimensions, centered to satisfy $\operatorname{Mean}(\bar{\mathbf{z}})=0$, and let $q=\operatorname{Mean}(\bar{\mathbf{z}}^2)$. Since a feasible LayerNorm output satisfies $q<1$, numerically infeasible targets are projected to $q<1$, after which the centered pre-normalization feature is recovered as $ \mathbf{u}^{\ast}=\sqrt{\epsilon_{LN}/(1-q)}$ LayerNorm is invariant to an additive shift; hence the remaining degree of freedom is selected by MDP as $\mu^{\ast}=\operatorname{Mean}(\hat{\mathbf{z}}_l-\mathbf{u}^{\ast})$, giving $\mathbf{z}_l^{\ast}=\mathbf{u}^{\ast}+\mu^{\ast}$. For zero or numerically negligible affine scales, division is avoided and the missing normalized entries are completed under the zero-mean constraint, with the free shift chosen to set the corresponding reconstructed entries to zero.\par
}
\rev{\textbf{(d) Pooling.}
We only support pooling with stride equal to the kernel size, zero padding, and no dilation. Given the target pooled feature and the former input, FR independently computes the minimum-$L_2$ modification within each pooling window. For average pooling, a uniform offset is added to match the target average. For max pooling, values above the target are clipped; if the current maximum is below the target, only one original maximal entry is raised. Uncovered spatial entries remain unchanged.\par
}
\subsection{Optimal Embedding for Category Labels}
\label{sec:optimal_embedding}
\rev{For classification, reverse reconstruction cannot start directly from a scalar label. We therefore construct an output-space target that predicts the ground-truth class while deviating as little as possible from the original output vector. Given the original output \(\hat{\mathbf{a}}_L\) and label \(y^{\ast}\), the optimal embedding solves
}
\revEnvBegin
\begin{align}
    \mathbf{a}_{out}^{\ast}
    =\arg\min_{\mathbf{a}_{out}}\quad
    &\|\mathbf{a}_{out}-\hat{\mathbf{a}}_{out}\|_{N}
    \label{model:oe}\\
    \mathrm{s.t.}\quad
    &(\mathbf{a}_{out})_{y^{\ast}}\ge(\mathbf{a}_{out})_i,
    \quad i\ne y^{\ast}.
    \notag
\end{align}
\revEnvEnd
\rev{Under the \(L_1\) norm, a minimum-deviation solution is the maximum-assignment embedding
}
\revEnvBegin
\begin{equation}\label{ma}
(\mathbf{a}^{MA}_{out})_{j}=
    \begin{cases}
    \max_i(\hat{\mathbf{a}}_{out})_i, & j=y^{\ast},\\
    (\hat{\mathbf{a}}_{out})_j, & j\ne y^{\ast}.
    \end{cases}\notag
\end{equation}\par
\revEnvEnd
\rev{Under the \(L_2\) norm, Eq.~\ref{model:oe} is the Euclidean projection onto the convex set in which the ground-truth component is maximal. Let the competing components larger than
\((\hat{\mathbf{a}}_{out})_{y^{\ast}}\) be sorted as
\(v_1\ge v_2\ge\cdots\ge v_m\), and define}
\revEnvBegin
\begin{equation}
    \theta_k=
    \frac{(\hat{\mathbf{a}}_{out})_{y^{\ast}}+\sum_{s=1}^{k}v_s}{k+1}.
    \label{eq:oe_threshold}
\end{equation}
\revEnvEnd
\rev{The unique active-set size \(k^{\ast}\) satisfies
\(v_{k^{\ast}}\ge\theta_{k^{\ast}}\) and
\(v_{k^{\ast}+1}\le\theta_{k^{\ast}}\), with the second condition omitted when \(k^{\ast}=m\). The nearest embedding is obtained through Alg.~\ref{alg:NE}.
The full Lagrangian and KKT derivation is provided in Supplementary Part~A.5.
}
\revEnvBegin
\begin{algorithm}
\begingroup
\revEnvBegin
\caption{\rev{Parallelizable Nearest-Embedding Algorithm}}
\label{alg:NE}
\begin{algorithmic}[1]
\REQUIRE original output \(\hat{\mathbf{a}}_{out}\) and ground-truth label \(y^{\ast}\)
\ENSURE reconstructed output vector \(\mathbf{a}_{out}^{NE}\)
\STATE Collect and sort the components greater than \((\hat{\mathbf{a}}_{out})_{y^{\ast}}\) to obtain \(v_1\ge\cdots\ge v_m\).
\STATE Compute the prefix sums of \(\{v_s\}_{s=1}^{m}\) and all thresholds \(\{\theta_k\}_{k=1}^{m}\) using Eq.~\ref{eq:oe_threshold}.
\STATE Find the unique \(k^{\ast}\) satisfying the active-set inequalities.
\STATE Set the ground-truth component and the \(k^{\ast}\) active competing components to \(\theta_{k^{\ast}}\).
\STATE Set remaining components to their original values in \(\hat{\mathbf{a}}_{out}\).
\RETURN \(\mathbf{a}_{out}^{NE}\)
\end{algorithmic}
\revEnvEnd
\endgroup
\end{algorithm}
\revEnvEnd
\subsection{\rev{Stability and Error Analysis}}
\label{sec:stability_analysis}
\rev{This section summarizes two theoretical properties that are crucial to FR-PT: the numerical stability of MDP-centered Tikhonov reconstruction, and the propagation of reconstruction errors across multiple reverse layers.\par
}
\revEnvBegin
\begin{theorem}[MDP-centered Tikhonov stability]
\label{thm:stable_regular}
Let $\mathbf{A}$ be a linear transformation, $\mathbf{y}=\mathbf{A}\mathbf{x}$ the bias-free target, and $\hat{\mathbf{x}}$ the original forward feature. For $\alpha>0$, the MDP-centered Tikhonov fallback
\begin{equation}
\mathbf{x}_{\alpha}=\arg\min_{\mathbf{x}}
\|\mathbf{A}\mathbf{x}-\mathbf{y}\|_2^2
+\alpha\|\mathbf{x}-\hat{\mathbf{x}}\|_2^2
\end{equation}
has, for the compact SVD $\mathbf{A}=\mathbf{U}\boldsymbol{\Sigma}\mathbf{V}^{H}$ with singular values $\{\sigma_i\}$, the correction
\begin{equation}
    \mathbf{x}_{\alpha}-\hat{\mathbf{x}}
    =\mathbf{V}\operatorname{diag}\!\left(\frac{\sigma_i}{\sigma_i^2+\alpha}\right)
      \mathbf{U}^H(\mathbf{y}-\mathbf{A}\hat{\mathbf{x}}),
    \label{eq:supp_tikhonov_svd}
\end{equation}
and satisfies
\begin{align}
\|\mathbf{x}_{\alpha}-\hat{\mathbf{x}}\|_2
&\le \frac{\|\mathbf{y}-\mathbf{A}\hat{\mathbf{x}}\|_2}{2\sqrt{\alpha}},
\label{eq:tikhonov_correction_main}\\
\|\mathbf{x}_{\alpha}(\mathbf{y}+\boldsymbol{\xi})-
\mathbf{x}_{\alpha}(\mathbf{y})\|_2
&\le \frac{\|\boldsymbol{\xi}\|_2}{2\sqrt{\alpha}}.
\label{eq:tikhonov_stability}
\end{align}
Moreover, the Hermitian matrix $\mathbf{A}^H\mathbf{A}+\alpha\mathbb{I}$ is positive definite, and its $2$-norm condition number satisfies
\begin{equation}
    \kappa_2(\mathbf{A}^H\mathbf{A}+\alpha\mathbb{I})
    =\frac{\sigma_{\max}^2+\alpha}
    {\lambda_{\min}(\mathbf{A}^H\mathbf{A})+\alpha}
    \le 1+\frac{\sigma_{\max}^2}{\alpha},
    \label{eq:supp_tikhonov_condition}
\end{equation}
where  $\kappa_2(\cdot)$ denotes the matrix condition number induced by the $2$-norm, $\sigma_{\max}$ is the largest singular value of $\mathbf{A}$ and $\lambda_{\min}(\mathbf{A}^H\mathbf{A})$ is the smallest eigenvalue of $\mathbf{A}^H\mathbf{A}$. These bounds remain valid when $\mathbf{A}$ is rank deficient.
\end{theorem}
\revEnvEnd
\rev{Thm.~\ref{thm:stable_regular} applies to all FR procedures using MDP-centered Tikhonov regularization, including linear maps, Matrix-Solve and FFT-based convolutional maps, and the local linearized subproblems arising in the Gauss--Newton reconstruction of composite blocks. The filter gain $\sigma/(\sigma^2+\alpha)$ is uniformly bounded, which suppresses the amplification associated with the nominal inverse gain $1/\sigma$ near small singular values. Supplementary Part~A.2 gives the proof.\par
}
\revEnvBegin
\begin{theorem}[Layer-wise reconstruction error]
Let $\mathcal{R}_l$ be the ideal reverse map, $\widetilde{\mathcal{R}}_l$ its real implementation; $\mathbf{a}_l^{\ast}=
    \mathcal{R}_l(\mathbf{a}_{l+1}^{\ast}), \widetilde{\mathbf{a}}_l
    =\widetilde{\mathcal{R}}_l(\widetilde{\mathbf{a}}_{l+1})$ are ideal and real obtained reconstructed features, respectively. If $\mathcal{R}_l$ is locally $\rho_l$-Lipschitz with respect to its target and
$\|\widetilde{\mathcal{R}}_l(\mathbf{u})-\mathcal{R}_l(\mathbf{u})\|\le\varepsilon_l$, then
\begin{equation}
 e_l=\|\widetilde{\mathbf{a}}_l-\mathbf{a}_l^{\ast}\|\le\rho_l e_{l+1}+\varepsilon_l,
 \label{eq:error_recursion}
\end{equation}
and, for any layer $k \in \{1,\ldots,L_{out}-1\}$,
\begin{equation}
 e_k\le
 \left(\prod_{j=k}^{L_{out}-1}\rho_j\right)e_{L_{out}}+
 \sum_{i=k}^{L_{out}-1}\left(\prod_{j=k}^{i-1}\rho_j\right)\varepsilon_i.
 \label{eq:error_accumulation}
\end{equation}
\end{theorem}\par
\revEnvEnd
\rev{Here $\varepsilon_l$ may include regularization bias $\varepsilon_l^{\mathrm{reg}}$, finite solver tolerance $\varepsilon_l^{\mathrm{solver}}$, FFT-boundary approximation $\varepsilon_l^{\mathrm{boundary}}$, iterative truncation $\varepsilon_l^{\mathrm{iter}}$, and pooling approximation $\varepsilon_l^{\mathrm{model}}$. The product of local sensitivities explains how reconstruction errors may accumulate across reverse operators. Supplementary Part~A.6 gives the proof and discusses the regimes $\rho<1$, $\rho=1$, and $\rho>1$.\par
}
\section{Experiments}
\rev{We empirically investigate FR-PT from four complementary aspects: post-training performance on standard benchmarks and real-world applications; algorithmic design choices for feature reconstruction; feature-level diagnostic analysis via reconstruction; and favorable conditions for reconstruction supervision. Detailed hardware configurations and hyperparameter settings are provided in Supplementary Part~B.1.\par
}
\subsection{\rev{Post-Training Performance}}
\subsubsection{\rev{Image Classification Tasks}}
\label{sec:comparison} 
\rev{For image classification, we evaluate FR-PT using pretrained SimpleCNN, ResNet-18/34, and SimpleViT models across six standard benchmarks, with network architectures detailed in Supplementary Part~B.2. Each model is post-trained for 10 epochs under different reconstruction layers $l_R$ and reconstruction weights $c_{\mathrm{rec}}$. For each FR-PT setting, we introduce a matched BP reference with $c_{\mathrm{rec}}=0$, which differs from FR-PT with $c_{\mathrm{rec}}>0$ only by the removal of the reconstruction term in Eq.~\ref{ltotal}, thereby isolating the contribution of feature-level supervision. Nearest Embedding is used to construct the classification output target, while FFT-Padded reconstruction is adopted for convolutional layers.
Tables~\ref{tab:mnist}--\ref{tab:tin} report the mean $\pm$ standard deviation of three representative metrics over ten random seeds ($0,\ldots,9$). Epoch-5, Mean, and Best denote the test accuracy (\%) at epoch 5, the average accuracy over all 10 post-training epochs, and the maximum accuracy attained during post-training, respectively. Within each $l_R$ group, the highest mean value for each metric is highlighted in bold. For each metric and each reconstruction layer $l_R$, the best-performing setting with $c_{\mathrm{rec}}>0$ is compared with its matched $c_{\mathrm{rec}}=0$ reference using a paired one-sided $t$-test. Significance at $p<0.05$ is indicated by '$\ast$', with further statistical details provided in Supplementary Part~B.3.\par
}
\begin{table}[tp]
    \centering
    \caption{\rev{MNIST post-training results.}}
    \label{tab:mnist}
    \setlength{\tabcolsep}{5.5pt}
    \begin{tabular}{@{}lccccc@{}}
    \toprule
    \revCell{Model} & \revCell{$l_{R}$} & \revCell{$c_{\mathrm{rec}}$} & \revCell{Epoch 5 Acc} & \revCell{Mean Acc} & \revCell{Best Acc} \\
    \midrule
    \revCell{\multirow{9}{*}{\makecell[c]{Simple\\CNN}}} & \revCell{\multirow{3}{*}{out}} & \revCell{0} & \revCell{97.64 $\pm$ 0.02} & \revCell{97.65 $\pm$ 0.01} & \revCell{97.68 $\pm$ 0.01} \\
          &    & \revCell{0.1} & \revCell{97.71 $\pm$ 0.01} & \revCell{97.71 $\pm$ 0.00} & \revCell{97.72 $\pm$ 0.01} \\
          &    & \revCell{0.3} & \revCell{\textbf{97.71 $\pm$ 0.01}$^{*}$} & \revCell{\textbf{97.71 $\pm$ 0.00}$^{*}$} & \revCell{\textbf{97.72 $\pm$ 0.00}$^{*}$} \\
    \cline{2-6}
    \addlinespace[1pt]
          & \revCell{\multirow{3}{*}{2}} & \revCell{0} & \revCell{97.63 $\pm$ 0.02} & \revCell{97.63 $\pm$ 0.00} & \revCell{97.67 $\pm$ 0.01} \\
          &    & \revCell{0.1} & \revCell{97.61 $\pm$ 0.01} & \revCell{97.61 $\pm$ 0.00} & \revCell{97.67 $\pm$ 0.01} \\
          &    & \revCell{0.3} & \revCell{\textbf{97.66 $\pm$ 0.02}$^{*}$} & \revCell{\textbf{97.65 $\pm$ 0.01}$^{*}$} & \revCell{\textbf{97.70 $\pm$ 0.01}$^{*}$} \\
    \cline{2-6}
    \addlinespace[1pt]
          & \revCell{\multirow{3}{*}{1}} & \revCell{0} & \revCell{97.60 $\pm$ 0.02} & \revCell{97.62 $\pm$ 0.01} & \revCell{97.65 $\pm$ 0.00} \\
          &    & \revCell{0.1} & \revCell{\textbf{97.61 $\pm$ 0.01}} & \revCell{97.61 $\pm$ 0.00} & \revCell{97.65 $\pm$ 0.01} \\
          &    & \revCell{0.3} & \revCell{97.60 $\pm$ 0.01} & \revCell{\textbf{97.62 $\pm$ 0.00}} & \revCell{\textbf{97.66 $\pm$ 0.01}$^{*}$} \\
    \bottomrule
    \end{tabular}
    \revTableFootnote{$^{*}$ marks a nominal paired one-sided $t$-test at $p<0.05$ for the bolded block-wise best against $c_{\mathrm{rec}}=0$ when the best uses $c_{\mathrm{rec}}>0$; otherwise it is compared with the best nonzero setting.}
\end{table}\par
\begin{table}[tp]
    \centering
    \caption{\rev{CIFAR-10 post-training results.}}
    \label{tab:cifar10}
    \setlength{\tabcolsep}{5.5pt}
    \begin{tabular}{@{}lccccc@{}}
    \toprule
    \revCell{Model} & \revCell{$l_{R}$} & \revCell{$c_{\mathrm{rec}}$} & \revCell{Epoch 5 Acc} & \revCell{Mean Acc} & \revCell{Best Acc} \\
    \midrule
    \revCell{\multirow{15}{*}{\makecell[c]{Simple\\CNN}}} & \revCell{\multirow{3}{*}{out}} & \revCell{0} & \revCell{65.62 $\pm$ 0.17} & \revCell{65.68 $\pm$ 0.04} & \revCell{66.03 $\pm$ 0.08} \\
          &    & \revCell{0.1} & \revCell{66.28 $\pm$ 0.10} & \revCell{66.30 $\pm$ 0.02} & \revCell{66.45 $\pm$ 0.05} \\
          &    & \revCell{0.3} & \revCell{\textbf{66.43 $\pm$ 0.08}$^{*}$} & \revCell{\textbf{66.42 $\pm$ 0.02}$^{*}$} & \revCell{\textbf{66.52 $\pm$ 0.04}$^{*}$} \\
    \cline{2-6}
    \addlinespace[1pt]
          & \revCell{\multirow{3}{*}{4}} & \revCell{0} & \revCell{65.66 $\pm$ 0.16} & \revCell{65.74 $\pm$ 0.04} & \revCell{66.03 $\pm$ 0.09} \\
          &    & \revCell{0.1} & \revCell{65.88 $\pm$ 0.14} & \revCell{65.91 $\pm$ 0.03} & \revCell{66.09 $\pm$ 0.06} \\
          &    & \revCell{0.3} & \revCell{\textbf{66.00 $\pm$ 0.09}$^{*}$} & \revCell{\textbf{65.98 $\pm$ 0.03}$^{*}$} & \revCell{\textbf{66.21 $\pm$ 0.07}$^{*}$} \\
    \cline{2-6}
    \addlinespace[1pt]
          & \revCell{\multirow{3}{*}{3}} & \revCell{0} & \revCell{65.63 $\pm$ 0.15} & \revCell{65.67 $\pm$ 0.04} & \revCell{65.87 $\pm$ 0.08} \\
          &    & \revCell{0.1} & \revCell{65.75 $\pm$ 0.13} & \revCell{65.72 $\pm$ 0.03} & \revCell{65.92 $\pm$ 0.06} \\
          &    & \revCell{0.3} & \revCell{\textbf{65.77 $\pm$ 0.17}$^{*}$} & \revCell{\textbf{65.75 $\pm$ 0.03}$^{*}$} & \revCell{\textbf{65.98 $\pm$ 0.09}$^{*}$} \\
    \cline{2-6}
    \addlinespace[1pt]
          & \revCell{\multirow{3}{*}{2}} & \revCell{0} & \revCell{65.55 $\pm$ 0.12} & \revCell{65.58 $\pm$ 0.02} & \revCell{65.82 $\pm$ 0.08} \\
          &    & \revCell{0.1} & \revCell{65.58 $\pm$ 0.12} & \revCell{65.62 $\pm$ 0.03} & \revCell{65.85 $\pm$ 0.07} \\
          &    & \revCell{0.3} & \revCell{\textbf{65.64 $\pm$ 0.11}$^{*}$} & \revCell{\textbf{65.70 $\pm$ 0.03}$^{*}$} & \revCell{\textbf{65.88 $\pm$ 0.07}$^{*}$} \\
    \cline{2-6}
    \addlinespace[1pt]
          & \revCell{\multirow{3}{*}{1}} & \revCell{0} & \revCell{65.75 $\pm$ 0.07} & \revCell{65.81 $\pm$ 0.03} & \revCell{66.01 $\pm$ 0.08} \\
          &    & \revCell{0.1} & \revCell{65.76 $\pm$ 0.07} & \revCell{65.82 $\pm$ 0.03} & \revCell{\textbf{66.03 $\pm$ 0.07}$^{*}$} \\
          &    & \revCell{0.3} & \revCell{\textbf{65.78 $\pm$ 0.06}} & \revCell{\textbf{65.83 $\pm$ 0.02}$^{*}$} & \revCell{66.02 $\pm$ 0.07} \\
    \midrule
    \revCell{\multirow{18}{*}{\makecell[c]{ResNet-\\18}}} & \revCell{\multirow{3}{*}{out}} & \revCell{0} & \revCell{94.83 $\pm$ 0.09} & \revCell{94.87 $\pm$ 0.03} & \revCell{94.97 $\pm$ 0.03} \\
          &    & \revCell{0.02} & \revCell{94.83 $\pm$ 0.10} & \revCell{94.87 $\pm$ 0.03} & \revCell{94.99 $\pm$ 0.04} \\
          &    & \revCell{0.3} & \revCell{\textbf{95.08 $\pm$ 0.07}$^{*}$} & \revCell{\textbf{95.07 $\pm$ 0.02}$^{*}$} & \revCell{\textbf{95.17 $\pm$ 0.03}$^{*}$} \\
    \cline{2-6}
    \addlinespace[1pt]
          & \revCell{\multirow{3}{*}{5}} & \revCell{0} & \revCell{94.85 $\pm$ 0.09} & \revCell{94.88 $\pm$ 0.02} & \revCell{94.99 $\pm$ 0.03} \\
          &    & \revCell{0.02} & \revCell{94.85 $\pm$ 0.08} & \revCell{94.89 $\pm$ 0.02} & \revCell{94.99 $\pm$ 0.03} \\
          &    & \revCell{0.3} & \revCell{\textbf{94.97 $\pm$ 0.08}$^{*}$} & \revCell{\textbf{94.96 $\pm$ 0.02}$^{*}$} & \revCell{\textbf{95.05 $\pm$ 0.04}$^{*}$} \\
    \cline{2-6}
    \addlinespace[1pt]
          & \revCell{\multirow{3}{*}{4}} & \revCell{0} & \revCell{94.86 $\pm$ 0.10} & \revCell{94.88 $\pm$ 0.02} & \revCell{94.99 $\pm$ 0.04} \\
          &    & \revCell{0.02} & \revCell{94.85 $\pm$ 0.10} & \revCell{94.88 $\pm$ 0.02} & \revCell{94.99 $\pm$ 0.05} \\
          &    & \revCell{0.3} & \revCell{\textbf{94.91 $\pm$ 0.07}$^{*}$} & \revCell{\textbf{94.93 $\pm$ 0.02}$^{*}$} & \revCell{\textbf{95.02 $\pm$ 0.02}$^{*}$} \\
    \cline{2-6}
    \addlinespace[1pt]
          & \revCell{\multirow{3}{*}{3}} & \revCell{0} & \revCell{94.97 $\pm$ 0.06} & \revCell{94.96 $\pm$ 0.02} & \revCell{95.09 $\pm$ 0.04} \\
          &    & \revCell{0.02} & \revCell{94.96 $\pm$ 0.06} & \revCell{94.96 $\pm$ 0.02} & \revCell{95.09 $\pm$ 0.04} \\
          &    & \revCell{0.3} & \revCell{\textbf{95.03 $\pm$ 0.04}$^{*}$} & \revCell{\textbf{95.03 $\pm$ 0.02}$^{*}$} & \revCell{\textbf{95.13 $\pm$ 0.03}$^{*}$} \\
    \cline{2-6}
    \addlinespace[1pt]
          & \revCell{\multirow{3}{*}{2}} & \revCell{0} & \revCell{95.06 $\pm$ 0.05} & \revCell{95.00 $\pm$ 0.02} & \revCell{95.11 $\pm$ 0.02} \\
          &    & \revCell{0.02} & \revCell{95.06 $\pm$ 0.05} & \revCell{95.03 $\pm$ 0.02} & \revCell{95.12 $\pm$ 0.03} \\
          &    & \revCell{0.3} & \revCell{\textbf{95.08 $\pm$ 0.05}} & \revCell{\textbf{95.08 $\pm$ 0.01}$^{*}$} & \revCell{\textbf{95.18 $\pm$ 0.04}$^{*}$} \\
    \cline{2-6}
    \addlinespace[1pt]
          & \revCell{\multirow{3}{*}{1}} & \revCell{0} & \revCell{\textbf{95.12 $\pm$ 0.03}} & \revCell{\textbf{95.08 $\pm$ 0.02}} & \revCell{\textbf{95.16 $\pm$ 0.03}} \\
          &    & \revCell{0.02} & \revCell{95.12 $\pm$ 0.04} & \revCell{95.08 $\pm$ 0.01} & \revCell{95.15 $\pm$ 0.03} \\
          &    & \revCell{0.3} & \revCell{95.07 $\pm$ 0.04} & \revCell{95.06 $\pm$ 0.02} & \revCell{95.14 $\pm$ 0.02} \\
    \bottomrule
    \end{tabular}
    \revTableFootnote{$^{*}$ marks a nominal paired one-sided $t$-test at $p<0.05$ for the bolded block-wise best against $c_{\mathrm{rec}}=0$ when the best uses $c_{\mathrm{rec}}>0$; otherwise it is compared with the best nonzero setting.}
\end{table}\par
\begin{table}[tp]
    \centering
    \caption{\rev{CIFAR-100 post-training results.}}
    \label{tab:cifar100}
    \setlength{\tabcolsep}{5.5pt}
    \scriptsize
    \begin{tabular}{@{}lccccc@{}}
    \toprule
    \revCell{Model} & \revCell{$l_{R}$} & \revCell{$c_{\mathrm{rec}}$} & \revCell{Epoch 5 Acc} & \revCell{Mean Acc} & \revCell{Best Acc} \\
    \midrule
     \revCell{\multirow{15}{*}{\makecell[c]{Simple\\CNN}}} & \revCell{\multirow{3}{*}{out}} & \revCell{0} & \revCell{\textbf{44.26 $\pm$ 0.06}$^{*}$} & \revCell{\textbf{44.30 $\pm$ 0.04}$^{*}$} & \revCell{\textbf{44.49 $\pm$ 0.08}$^{*}$} \\
          &    & \revCell{0.02} & \revCell{44.18 $\pm$ 0.07} & \revCell{44.19 $\pm$ 0.01} & \revCell{44.32 $\pm$ 0.05} \\
          &    & \revCell{0.1} & \revCell{44.19 $\pm$ 0.07} & \revCell{44.18 $\pm$ 0.02} & \revCell{44.28 $\pm$ 0.03} \\
    \cline{2-6}
    \addlinespace[1pt]
          & \revCell{\multirow{3}{*}{4}} & \revCell{0} & \revCell{\textbf{44.31 $\pm$ 0.10}} & \revCell{44.31 $\pm$ 0.05} & \revCell{44.47 $\pm$ 0.07} \\
          &    & \revCell{0.02} & \revCell{44.31 $\pm$ 0.11} & \revCell{\textbf{44.33 $\pm$ 0.02}$^{*}$} & \revCell{\textbf{44.53 $\pm$ 0.07}$^{*}$} \\
          &    & \revCell{0.1} & \revCell{44.07 $\pm$ 0.07} & \revCell{44.12 $\pm$ 0.03} & \revCell{44.31 $\pm$ 0.05} \\
    \cline{2-6}
    \addlinespace[1pt]
          & \revCell{\multirow{3}{*}{3}} & \revCell{0} & \revCell{44.06 $\pm$ 0.08} & \revCell{44.02 $\pm$ 0.03} & \revCell{44.23 $\pm$ 0.05} \\
          &    & \revCell{0.02} & \revCell{44.05 $\pm$ 0.09} & \revCell{44.02 $\pm$ 0.03} & \revCell{44.25 $\pm$ 0.06} \\
          &    & \revCell{0.1} & \revCell{\textbf{44.16 $\pm$ 0.05}$^{*}$} & \revCell{\textbf{44.08 $\pm$ 0.03}$^{*}$} & \revCell{\textbf{44.28 $\pm$ 0.07}$^{*}$} \\
    \cline{2-6}
    \addlinespace[1pt]
          & \revCell{\multirow{3}{*}{2}} & \revCell{0} & \revCell{43.70 $\pm$ 0.11} & \revCell{43.68 $\pm$ 0.02} & \revCell{43.86 $\pm$ 0.07} \\
          &    & \revCell{0.1} & \revCell{43.69 $\pm$ 0.15} & \revCell{43.68 $\pm$ 0.03} & \revCell{43.88 $\pm$ 0.08} \\
          &    & \revCell{0.3} & \revCell{\textbf{43.75 $\pm$ 0.15}} & \revCell{\textbf{43.72 $\pm$ 0.04}$^{*}$} & \revCell{\textbf{43.92 $\pm$ 0.09}$^{*}$} \\
    \cline{2-6}
    \addlinespace[1pt]
          & \revCell{\multirow{3}{*}{1}} & \revCell{0} & \revCell{43.85 $\pm$ 0.07} & \revCell{43.81 $\pm$ 0.01} & \revCell{43.95 $\pm$ 0.04} \\
          &    & \revCell{0.1} & \revCell{43.85 $\pm$ 0.07} & \revCell{43.81 $\pm$ 0.02} & \revCell{43.95 $\pm$ 0.05} \\
          &    & \revCell{0.3} & \revCell{\textbf{43.87 $\pm$ 0.06}$^{*}$} & \revCell{\textbf{43.82 $\pm$ 0.02}} & \revCell{\textbf{43.96 $\pm$ 0.05}} \\
    \midrule
     \revCell{\multirow{18}{*}{\makecell[c]{ResNet\\-18}}} & \revCell{\multirow{3}{*}{out}} & \revCell{0} & \revCell{79.59 $\pm$ 0.09} & \revCell{79.50 $\pm$ 0.07} & \revCell{79.70 $\pm$ 0.06} \\
          &    & \revCell{0.1} & \revCell{\textbf{79.65 $\pm$ 0.09}$^{*}$} & \revCell{\textbf{79.64 $\pm$ 0.04}$^{*}$} & \revCell{79.80 $\pm$ 0.04} \\
          &    & \revCell{0.3} & \revCell{79.58 $\pm$ 0.09} & \revCell{79.59 $\pm$ 0.04} & \revCell{\textbf{79.80 $\pm$ 0.05}$^{*}$} \\
    \cline{2-6}
    \addlinespace[1pt]
          & \revCell{\multirow{3}{*}{5}} & \revCell{0} & \revCell{79.62 $\pm$ 0.13} & \revCell{79.51 $\pm$ 0.05} & \revCell{79.77 $\pm$ 0.06} \\
          &    & \revCell{0.1} & \revCell{\textbf{79.66 $\pm$ 0.07}} & \revCell{79.65 $\pm$ 0.04} & \revCell{79.82 $\pm$ 0.07} \\
          &    & \revCell{0.3} & \revCell{79.66 $\pm$ 0.08} & \revCell{\textbf{79.66 $\pm$ 0.04}$^{*}$} & \revCell{\textbf{79.84 $\pm$ 0.04}$^{*}$} \\
    \cline{2-6}
    \addlinespace[1pt]
          & \revCell{\multirow{3}{*}{4}} & \revCell{0} & \revCell{79.59 $\pm$ 0.11} & \revCell{79.50 $\pm$ 0.07} & \revCell{79.75 $\pm$ 0.08} \\
          &    & \revCell{0.1} & \revCell{79.67 $\pm$ 0.09} & \revCell{79.67 $\pm$ 0.04} & \revCell{79.83 $\pm$ 0.05} \\
          &    & \revCell{0.3} & \revCell{\textbf{79.70 $\pm$ 0.10}$^{*}$} & \revCell{\textbf{79.69 $\pm$ 0.04}$^{*}$} & \revCell{\textbf{79.86 $\pm$ 0.07}$^{*}$} \\
    \cline{2-6}
    \addlinespace[1pt]
          & \revCell{\multirow{3}{*}{3}} & \revCell{0} & \revCell{\textbf{79.48 $\pm$ 0.09}} & \revCell{79.44 $\pm$ 0.04} & \revCell{79.63 $\pm$ 0.06} \\
          &    & \revCell{0.1} & \revCell{79.48 $\pm$ 0.07} & \revCell{\textbf{79.49 $\pm$ 0.03}$^{*}$} & \revCell{79.64 $\pm$ 0.06} \\
          &    & \revCell{0.3} & \revCell{79.47 $\pm$ 0.09} & \revCell{79.49 $\pm$ 0.04} & \revCell{\textbf{79.64 $\pm$ 0.07}} \\
    \cline{2-6}
    \addlinespace[1pt]
          & \revCell{\multirow{3}{*}{2}} & \revCell{0} & \revCell{79.47 $\pm$ 0.10} & \revCell{79.44 $\pm$ 0.04} & \revCell{79.60 $\pm$ 0.07} \\
          &    & \revCell{0.1} & \revCell{\textbf{79.49 $\pm$ 0.10}} & \revCell{79.49 $\pm$ 0.03} & \revCell{79.64 $\pm$ 0.05} \\
          &    & \revCell{0.3} & \revCell{79.48 $\pm$ 0.08} & \revCell{\textbf{79.50 $\pm$ 0.02}$^{*}$} & \revCell{\textbf{79.64 $\pm$ 0.05}} \\
    \cline{2-6}
    \addlinespace[1pt]
          & \revCell{\multirow{3}{*}{1}} & \revCell{0} & \revCell{79.21 $\pm$ 0.08} & \revCell{79.26 $\pm$ 0.02} & \revCell{\textbf{79.41 $\pm$ 0.05}} \\
          &    & \revCell{0.1} & \revCell{\textbf{79.25 $\pm$ 0.04}$^{*}$} & \revCell{\textbf{79.27 $\pm$ 0.01}} & \revCell{79.38 $\pm$ 0.03} \\
          &    & \revCell{0.3} & \revCell{79.24 $\pm$ 0.06} & \revCell{79.27 $\pm$ 0.02} & \revCell{79.37 $\pm$ 0.03} \\
    \midrule
    \revCell{\multirow{33}{*}{\makecell[c]{Simple\\ViT}}} & \revCell{\multirow{3}{*}{out}} & \revCell{0} & \revCell{61.00 $\pm$ 0.13} & \revCell{61.10 $\pm$ 0.05} & \revCell{63.28 $\pm$ 0.10} \\
          &    & \revCell{0.1} & \revCell{\textbf{63.56 $\pm$ 0.08}$^{*}$} & \revCell{62.82 $\pm$ 0.04} & \revCell{\textbf{63.56 $\pm$ 0.07}$^{*}$} \\
          &    & \revCell{0.3} & \revCell{63.36 $\pm$ 0.08} & \revCell{\textbf{63.35 $\pm$ 0.02}$^{*}$} & \revCell{63.46 $\pm$ 0.06} \\
    \cline{2-6}
    \addlinespace[1pt]
          & \revCell{\multirow{3}{*}{10}} & \revCell{0} & \revCell{60.98 $\pm$ 0.11} & \revCell{61.07 $\pm$ 0.05} & \revCell{63.28 $\pm$ 0.11} \\
          &    & \revCell{0.1} & \revCell{\textbf{63.49 $\pm$ 0.10}$^{*}$} & \revCell{62.69 $\pm$ 0.03} & \revCell{\textbf{63.74 $\pm$ 0.10}$^{*}$} \\
          &    & \revCell{0.3} & \revCell{63.47 $\pm$ 0.07} & \revCell{\textbf{63.50 $\pm$ 0.03}$^{*}$} & \revCell{63.60 $\pm$ 0.03} \\
    \cline{2-6}
    \addlinespace[1pt]
          & \revCell{\multirow{3}{*}{9}} & \revCell{0} & \revCell{60.98 $\pm$ 0.11} & \revCell{61.07 $\pm$ 0.05} & \revCell{63.28 $\pm$ 0.11} \\
          &    & \revCell{0.1} & \revCell{63.48 $\pm$ 0.12} & \revCell{63.00 $\pm$ 0.05} & \revCell{64.02 $\pm$ 0.09} \\
          &    & \revCell{0.3} & \revCell{\textbf{63.93 $\pm$ 0.10}$^{*}$} & \revCell{\textbf{63.91 $\pm$ 0.03}$^{*}$} & \revCell{\textbf{64.07 $\pm$ 0.05}$^{*}$} \\
    \cline{2-6}
    \addlinespace[1pt]
          & \revCell{\multirow{3}{*}{8}} & \revCell{0} & \revCell{61.00 $\pm$ 0.10} & \revCell{61.00 $\pm$ 0.04} & \revCell{63.19 $\pm$ 0.13} \\
          &    & \revCell{0.1} & \revCell{63.42 $\pm$ 0.13} & \revCell{62.99 $\pm$ 0.05} & \revCell{64.02 $\pm$ 0.07} \\
          &    & \revCell{0.3} & \revCell{\textbf{63.83 $\pm$ 0.10}$^{*}$} & \revCell{\textbf{63.85 $\pm$ 0.03}$^{*}$} & \revCell{\textbf{64.02 $\pm$ 0.06}$^{*}$} \\
    \cline{2-6}
    \addlinespace[1pt]
          & \revCell{\multirow{3}{*}{7}} & \revCell{0} & \revCell{61.56 $\pm$ 0.18} & \revCell{61.36 $\pm$ 0.04} & \revCell{63.08 $\pm$ 0.14} \\
          &    & \revCell{0.1} & \revCell{63.37 $\pm$ 0.16} & \revCell{63.08 $\pm$ 0.05} & \revCell{63.71 $\pm$ 0.06} \\
          &    & \revCell{0.3} & \revCell{\textbf{63.48 $\pm$ 0.13}$^{*}$} & \revCell{\textbf{63.49 $\pm$ 0.04}$^{*}$} & \revCell{\textbf{63.72 $\pm$ 0.06}$^{*}$} \\
    \cline{2-6}
    \addlinespace[1pt]
          & \revCell{\multirow{3}{*}{6}} & \revCell{0} & \revCell{61.96 $\pm$ 0.10} & \revCell{61.85 $\pm$ 0.04} & \revCell{63.08 $\pm$ 0.13} \\
          &    & \revCell{0.1} & \revCell{\textbf{63.24 $\pm$ 0.06}$^{*}$} & \revCell{63.13 $\pm$ 0.03} & \revCell{63.49 $\pm$ 0.08} \\
          &    & \revCell{0.3} & \revCell{\textbf{63.24 $\pm$ 0.06}} & \revCell{\textbf{63.23 $\pm$ 0.03}$^{*}$} & \revCell{\textbf{63.49 $\pm$ 0.08}$^{*}$} \\
    \cline{2-6}
    \addlinespace[1pt]
          & \revCell{\multirow{3}{*}{5}} & \revCell{0} & \revCell{62.55 $\pm$ 0.10} & \revCell{62.45 $\pm$ 0.04} & \revCell{63.23 $\pm$ 0.10} \\
          &    & \revCell{0.1} & \revCell{\textbf{63.31 $\pm$ 0.14}$^{*}$} & \revCell{63.29 $\pm$ 0.03} & \revCell{\textbf{63.54 $\pm$ 0.09}$^{*}$} \\
          &    & \revCell{0.3} & \revCell{\textbf{63.31 $\pm$ 0.14}} & \revCell{\textbf{63.29 $\pm$ 0.03}$^{*}$} & \revCell{\textbf{63.54 $\pm$ 0.09}} \\
    \cline{2-6}
    \addlinespace[1pt]
          & \revCell{\multirow{3}{*}{4}} & \revCell{0} & \revCell{63.02 $\pm$ 0.15} & \revCell{62.95 $\pm$ 0.05} & \revCell{63.46 $\pm$ 0.09} \\
          &    & \revCell{0.1} & \revCell{\textbf{63.56 $\pm$ 0.09}$^{*}$} & \revCell{\textbf{63.54 $\pm$ 0.02}$^{*}$} & \revCell{\textbf{63.74 $\pm$ 0.05}$^{*}$} \\
          &    & \revCell{0.3} & \revCell{\textbf{63.56 $\pm$ 0.09}} & \revCell{\textbf{63.54 $\pm$ 0.02}} & \revCell{\textbf{63.74 $\pm$ 0.05}} \\
    \cline{2-6}
    \addlinespace[1pt]
          & \revCell{\multirow{3}{*}{3}} & \revCell{0} & \revCell{63.38 $\pm$ 0.07} & \revCell{63.34 $\pm$ 0.03} & \revCell{63.64 $\pm$ 0.06} \\
          &    & \revCell{0.1} & \revCell{\textbf{63.69 $\pm$ 0.07}$^{*}$} & \revCell{\textbf{63.67 $\pm$ 0.03}$^{*}$} & \revCell{\textbf{63.83 $\pm$ 0.04}$^{*}$} \\
          &    & \revCell{0.3} & \revCell{\textbf{63.69 $\pm$ 0.07}} & \revCell{\textbf{63.67 $\pm$ 0.03}} & \revCell{\textbf{63.83 $\pm$ 0.04}} \\
    \cline{2-6}
    \addlinespace[1pt]
          & \revCell{\multirow{3}{*}{2}} & \revCell{0} & \revCell{63.47 $\pm$ 0.09} & \revCell{63.45 $\pm$ 0.02} & \revCell{63.68 $\pm$ 0.04} \\
          &    & \revCell{0.1} & \revCell{\textbf{63.81 $\pm$ 0.05}$^{*}$} & \revCell{\textbf{63.78 $\pm$ 0.03}$^{*}$} & \revCell{\textbf{63.91 $\pm$ 0.05}$^{*}$} \\
          &    & \revCell{0.3} & \revCell{\textbf{63.81 $\pm$ 0.05}} & \revCell{\textbf{63.78 $\pm$ 0.03}} & \revCell{\textbf{63.91 $\pm$ 0.05}} \\
    \cline{2-6}
    \addlinespace[1pt]
          & \revCell{\multirow{3}{*}{1}} & \revCell{0} & \revCell{63.55 $\pm$ 0.07} & \revCell{63.55 $\pm$ 0.02} & \revCell{63.75 $\pm$ 0.06} \\
          &    & \revCell{0.1} & \revCell{\textbf{63.84 $\pm$ 0.12}$^{*}$} & \revCell{\textbf{63.79 $\pm$ 0.02}$^{*}$} & \revCell{\textbf{63.94 $\pm$ 0.07}$^{*}$} \\
          &    & \revCell{0.3} & \revCell{\textbf{63.84 $\pm$ 0.12}} & \revCell{\textbf{63.79 $\pm$ 0.02}} & \revCell{\textbf{63.94 $\pm$ 0.07}} \\
    \bottomrule
    \end{tabular}
    \revTableFootnote{$^{*}$ marks a nominal paired one-sided $t$-test at $p<0.05$ for the bolded block-wise best against $c_{\mathrm{rec}}=0$ when the best uses $c_{\mathrm{rec}}>0$; otherwise it is compared with the best nonzero setting.}
\end{table}\par
\begin{table}[tp]
    \centering
    \caption{\rev{Imagenette post-training results.}}
    \label{tab:nette}
    \setlength{\tabcolsep}{5.5pt}
    \begin{tabular}{@{}lccccc@{}}
    \toprule
    \revCell{Model} & \revCell{$l_{R}$} & \revCell{$c_{\mathrm{rec}}$} & \revCell{Epoch 5 Acc} & \revCell{Mean Acc} & \revCell{Best Acc} \\
    \midrule
    \revCell{\multirow{21}{*}{\makecell[c]{Simple\\CNN}}} & \revCell{\multirow{3}{*}{out}} & \revCell{0} & \revCell{59.62 $\pm$ 0.20} & \revCell{59.63 $\pm$ 0.13} & \revCell{59.93 $\pm$ 0.14} \\
          &    & \revCell{0.1} & \revCell{\textbf{60.17 $\pm$ 0.15}$^{*}$} & \revCell{\textbf{60.16 $\pm$ 0.06}$^{*}$} & \revCell{\textbf{60.40 $\pm$ 0.11}$^{*}$} \\
          &    & \revCell{0.3} & \revCell{60.08 $\pm$ 0.13} & \revCell{60.11 $\pm$ 0.04} & \revCell{60.28 $\pm$ 0.08} \\
    \cline{2-6}
    \addlinespace[1pt]
          & \revCell{\multirow{3}{*}{6}} & \revCell{0} & \revCell{59.61 $\pm$ 0.20} & \revCell{59.63 $\pm$ 0.11} & \revCell{60.00 $\pm$ 0.15} \\
          &    & \revCell{0.1} & \revCell{59.72 $\pm$ 0.17} & \revCell{59.76 $\pm$ 0.06} & \revCell{60.04 $\pm$ 0.13} \\
          &    & \revCell{0.3} & \revCell{\textbf{59.99 $\pm$ 0.19}$^{*}$} & \revCell{\textbf{59.89 $\pm$ 0.07}$^{*}$} & \revCell{\textbf{60.18 $\pm$ 0.08}$^{*}$} \\
    \cline{2-6}
    \addlinespace[1pt]
          & \revCell{\multirow{3}{*}{5}} & \revCell{0} & \revCell{\textbf{59.62 $\pm$ 0.17}$^{*}$} & \revCell{\textbf{59.66 $\pm$ 0.08}$^{*}$} & \revCell{\textbf{59.93 $\pm$ 0.10}$^{*}$} \\
          &    & \revCell{0.1} & \revCell{59.27 $\pm$ 0.25} & \revCell{59.37 $\pm$ 0.05} & \revCell{59.71 $\pm$ 0.11} \\
          &    & \revCell{0.3} & \revCell{58.97 $\pm$ 0.17} & \revCell{59.05 $\pm$ 0.10} & \revCell{59.46 $\pm$ 0.10} \\
    \cline{2-6}
    \addlinespace[1pt]
          & \revCell{\multirow{3}{*}{4}} & \revCell{0} & \revCell{59.57 $\pm$ 0.18} & \revCell{59.67 $\pm$ 0.07} & \revCell{59.96 $\pm$ 0.10} \\
          &    & \revCell{0.1} & \revCell{59.90 $\pm$ 0.15} & \revCell{59.92 $\pm$ 0.07} & \revCell{60.20 $\pm$ 0.07} \\
          &    & \revCell{0.3} & \revCell{\textbf{60.08 $\pm$ 0.26}$^{*}$} & \revCell{\textbf{60.09 $\pm$ 0.07}$^{*}$} & \revCell{\textbf{60.39 $\pm$ 0.15}$^{*}$} \\
    \cline{2-6}
    \addlinespace[1pt]
          & \revCell{\multirow{3}{*}{3}} & \revCell{0} & \revCell{\textbf{59.88 $\pm$ 0.13}$^{*}$} & \revCell{59.77 $\pm$ 0.08} & \revCell{60.03 $\pm$ 0.10} \\
          &    & \revCell{0.1} & \revCell{59.77 $\pm$ 0.17} & \revCell{59.80 $\pm$ 0.04} & \revCell{60.15 $\pm$ 0.11} \\
          &    & \revCell{0.3} & \revCell{59.68 $\pm$ 0.22} & \revCell{\textbf{59.82 $\pm$ 0.03}$^{*}$} & \revCell{\textbf{60.15 $\pm$ 0.09}$^{*}$} \\
    \cline{2-6}
    \addlinespace[1pt]
          & \revCell{\multirow{3}{*}{2}} & \revCell{0} & \revCell{59.76 $\pm$ 0.18} & \revCell{59.77 $\pm$ 0.03} & \revCell{60.06 $\pm$ 0.05} \\
          &    & \revCell{0.1} & \revCell{59.80 $\pm$ 0.18} & \revCell{59.80 $\pm$ 0.06} & \revCell{60.02 $\pm$ 0.07} \\
          &    & \revCell{0.3} & \revCell{\textbf{59.83 $\pm$ 0.23}} & \revCell{\textbf{59.82 $\pm$ 0.06}$^{*}$} & \revCell{\textbf{60.08 $\pm$ 0.11}} \\
    \cline{2-6}
    \addlinespace[1pt]
          & \revCell{\multirow{3}{*}{1}} & \revCell{0} & \revCell{59.56 $\pm$ 0.17} & \revCell{59.61 $\pm$ 0.06} & \revCell{59.96 $\pm$ 0.11} \\
          &    & \revCell{0.1} & \revCell{59.53 $\pm$ 0.13} & \revCell{59.60 $\pm$ 0.06} & \revCell{59.94 $\pm$ 0.13} \\
          &    & \revCell{0.3} & \revCell{\textbf{59.61 $\pm$ 0.28}} & \revCell{\textbf{59.62 $\pm$ 0.07}} & \revCell{\textbf{59.96 $\pm$ 0.13}} \\
    \midrule
    \revCell{\multirow{18}{*}{\makecell[c]{ResNet\\-18}}} & \revCell{\multirow{3}{*}{out}} & \revCell{0} & \revCell{98.17 $\pm$ 0.05} & \revCell{98.11 $\pm$ 0.04} & \revCell{98.21 $\pm$ 0.03} \\
          &    & \revCell{0.02} & \revCell{98.17 $\pm$ 0.06} & \revCell{98.14 $\pm$ 0.04} & \revCell{98.24 $\pm$ 0.04} \\
          &    & \revCell{0.3} & \revCell{\textbf{98.41 $\pm$ 0.05}$^{*}$} & \revCell{\textbf{98.36 $\pm$ 0.02}$^{*}$} & \revCell{\textbf{98.47 $\pm$ 0.04}$^{*}$} \\
    \cline{2-6}
    \addlinespace[1pt]
          & \revCell{\multirow{3}{*}{5}} & \revCell{0} & \revCell{98.09 $\pm$ 0.07} & \revCell{98.04 $\pm$ 0.04} & \revCell{98.15 $\pm$ 0.05} \\
          &    & \revCell{0.02} & \revCell{98.10 $\pm$ 0.06} & \revCell{98.05 $\pm$ 0.04} & \revCell{98.17 $\pm$ 0.07} \\
          &    & \revCell{0.3} & \revCell{\textbf{98.22 $\pm$ 0.05}$^{*}$} & \revCell{\textbf{98.19 $\pm$ 0.03}$^{*}$} & \revCell{\textbf{98.32 $\pm$ 0.04}$^{*}$} \\
    \cline{2-6}
    \addlinespace[1pt]
          & \revCell{\multirow{3}{*}{4}} & \revCell{0} & \revCell{\textbf{98.08 $\pm$ 0.07}} & \revCell{98.04 $\pm$ 0.04} & \revCell{98.16 $\pm$ 0.05} \\
          &    & \revCell{0.02} & \revCell{\textbf{98.08 $\pm$ 0.07}} & \revCell{98.03 $\pm$ 0.04} & \revCell{\textbf{98.16 $\pm$ 0.05}} \\
          &    & \revCell{0.3} & \revCell{98.08 $\pm$ 0.04} & \revCell{\textbf{98.04 $\pm$ 0.04}} & \revCell{\textbf{98.16 $\pm$ 0.07}} \\
    \cline{2-6}
    \addlinespace[1pt]
          & \revCell{\multirow{3}{*}{3}} & \revCell{0} & \revCell{97.29 $\pm$ 0.08} & \revCell{97.28 $\pm$ 0.03} & \revCell{97.40 $\pm$ 0.04} \\
          &    & \revCell{0.02} & \revCell{97.36 $\pm$ 0.07} & \revCell{97.31 $\pm$ 0.02} & \revCell{97.44 $\pm$ 0.06} \\
          &    & \revCell{0.3} & \revCell{\textbf{97.61 $\pm$ 0.09}$^{*}$} & \revCell{\textbf{97.55 $\pm$ 0.02}$^{*}$} & \revCell{\textbf{97.72 $\pm$ 0.04}$^{*}$} \\
    \cline{2-6}
    \addlinespace[1pt]
          & \revCell{\multirow{3}{*}{2}} & \revCell{0} & \revCell{97.08 $\pm$ 0.06} & \revCell{97.07 $\pm$ 0.03} & \revCell{97.21 $\pm$ 0.03} \\
          &    & \revCell{0.02} & \revCell{97.11 $\pm$ 0.06} & \revCell{97.07 $\pm$ 0.03} & \revCell{97.22 $\pm$ 0.06} \\
          &    & \revCell{0.3} & \revCell{\textbf{97.25 $\pm$ 0.06}$^{*}$} & \revCell{\textbf{97.22 $\pm$ 0.02}$^{*}$} & \revCell{\textbf{97.38 $\pm$ 0.05}$^{*}$} \\
    \cline{2-6}
    \addlinespace[1pt]
          & \revCell{\multirow{3}{*}{1}} & \revCell{0} & \revCell{96.94 $\pm$ 0.06} & \revCell{96.93 $\pm$ 0.01} & \revCell{97.03 $\pm$ 0.03} \\
          &    & \revCell{0.02} & \revCell{\textbf{96.95 $\pm$ 0.05}} & \revCell{\textbf{96.95 $\pm$ 0.02}$^{*}$} & \revCell{\textbf{97.04 $\pm$ 0.04}} \\
          &    & \revCell{0.3} & \revCell{96.94 $\pm$ 0.04} & \revCell{\textbf{96.95 $\pm$ 0.02}} & \revCell{97.04 $\pm$ 0.02} \\
    \bottomrule
    \end{tabular}
    \revTableFootnote{$^{*}$ marks a nominal paired one-sided $t$-test at $p<0.05$ for the bolded block-wise best against $c_{\mathrm{rec}}=0$ when the best uses $c_{\mathrm{rec}}>0$; otherwise it is compared with the best nonzero setting.}
\end{table}\par
\begin{table}[tp]
    \centering
    \caption{\rev{Imagewoof post-training results.}}
    \label{tab:woof}
    \setlength{\tabcolsep}{5.5pt}
    \begin{tabular}{@{}lccccc@{}}
    \toprule
    \revCell{Model} & \revCell{$l_{R}$} & \revCell{$c_{\mathrm{rec}}$} & \revCell{Epoch 5 Acc} & \revCell{Mean Acc} & \revCell{Best Acc} \\
    \midrule
    \revCell{\multirow{21}{*}{\makecell[c]{Simple\\CNN}}}  & \revCell{\multirow{3}{*}{out}} & \revCell{0} & \revCell{33.72 $\pm$ 0.29} & \revCell{33.82 $\pm$ 0.11} & \revCell{34.25 $\pm$ 0.13} \\
          &    & \revCell{0.1} & \revCell{\textbf{34.25 $\pm$ 0.12}$^{*}$} & \revCell{34.22 $\pm$ 0.06} & \revCell{34.44 $\pm$ 0.11} \\
          &    & \revCell{0.3} & \revCell{34.21 $\pm$ 0.10} & \revCell{\textbf{34.22 $\pm$ 0.04}$^{*}$} & \revCell{\textbf{34.45 $\pm$ 0.07}$^{*}$} \\
    \cline{2-6}
    \addlinespace[1pt]
          & \revCell{\multirow{3}{*}{6}} & \revCell{0} & \revCell{33.75 $\pm$ 0.22} & \revCell{33.82 $\pm$ 0.09} & \revCell{34.24 $\pm$ 0.12} \\
          &    & \revCell{0.1} & \revCell{33.71 $\pm$ 0.32} & \revCell{33.71 $\pm$ 0.06} & \revCell{34.06 $\pm$ 0.13} \\
          &    & \revCell{0.3} & \revCell{\textbf{34.04 $\pm$ 0.28}$^{*}$} & \revCell{\textbf{33.98 $\pm$ 0.09}$^{*}$} & \revCell{\textbf{34.34 $\pm$ 0.11}$^{*}$} \\
    \cline{2-6}
    \addlinespace[1pt]
          & \revCell{\multirow{3}{*}{5}} & \revCell{0} & \revCell{\textbf{33.53 $\pm$ 0.17}} & \revCell{\textbf{33.70 $\pm$ 0.09}$^{*}$} & \revCell{\textbf{34.08 $\pm$ 0.14}$^{*}$} \\
          &    & \revCell{0.1} & \revCell{33.37 $\pm$ 0.26} & \revCell{33.42 $\pm$ 0.12} & \revCell{33.78 $\pm$ 0.20} \\
          &    & \revCell{0.3} & \revCell{33.42 $\pm$ 0.31} & \revCell{33.38 $\pm$ 0.08} & \revCell{33.75 $\pm$ 0.09} \\
    \cline{2-6}
    \addlinespace[1pt]
          & \revCell{\multirow{3}{*}{4}} & \revCell{0} & \revCell{33.53 $\pm$ 0.31} & \revCell{33.52 $\pm$ 0.08} & \revCell{33.89 $\pm$ 0.22} \\
          &    & \revCell{0.1} & \revCell{33.35 $\pm$ 0.27} & \revCell{33.53 $\pm$ 0.09} & \revCell{33.91 $\pm$ 0.19} \\
          &    & \revCell{0.3} & \revCell{\textbf{33.62 $\pm$ 0.32}} & \revCell{\textbf{33.64 $\pm$ 0.07}$^{*}$} & \revCell{\textbf{34.17 $\pm$ 0.17}$^{*}$} \\
    \cline{2-6}
    \addlinespace[1pt]
          & \revCell{\multirow{3}{*}{3}} & \revCell{0} & \revCell{33.51 $\pm$ 0.33} & \revCell{33.55 $\pm$ 0.09} & \revCell{34.03 $\pm$ 0.09} \\
          &    & \revCell{0.1} & \revCell{33.49 $\pm$ 0.26} & \revCell{\textbf{33.70 $\pm$ 0.09}$^{*}$} & \revCell{\textbf{34.06 $\pm$ 0.17}} \\
          &    & \revCell{0.3} & \revCell{\textbf{33.59 $\pm$ 0.17}} & \revCell{33.61 $\pm$ 0.06} & \revCell{33.89 $\pm$ 0.09} \\
    \cline{2-6}
    \addlinespace[1pt]
          & \revCell{\multirow{3}{*}{2}} & \revCell{0} & \revCell{33.67 $\pm$ 0.33} & \revCell{\textbf{33.66 $\pm$ 0.10}} & \revCell{\textbf{34.12 $\pm$ 0.13}} \\
          &    & \revCell{0.1} & \revCell{33.63 $\pm$ 0.21} & \revCell{33.65 $\pm$ 0.09} & \revCell{34.00 $\pm$ 0.15} \\
          &    & \revCell{0.3} & \revCell{\textbf{33.68 $\pm$ 0.21}} & \revCell{33.65 $\pm$ 0.10} & \revCell{34.05 $\pm$ 0.13} \\
    \cline{2-6}
    \addlinespace[1pt]
          & \revCell{\multirow{3}{*}{1}} & \revCell{0} & \revCell{33.82 $\pm$ 0.32} & \revCell{33.72 $\pm$ 0.07} & \revCell{34.15 $\pm$ 0.17} \\
          &    & \revCell{0.1} & \revCell{33.96 $\pm$ 0.34} & \revCell{33.81 $\pm$ 0.09} & \revCell{34.26 $\pm$ 0.24} \\
          &    & \revCell{0.3} & \revCell{\textbf{34.02 $\pm$ 0.42}$^{*}$} & \revCell{\textbf{33.88 $\pm$ 0.12}$^{*}$} & \revCell{\textbf{34.43 $\pm$ 0.17}$^{*}$} \\
    \midrule
    \revCell{\multirow{18}{*}{\makecell[c]{ResNet\\-18}}} & \revCell{\multirow{3}{*}{out}} & \revCell{0} & \revCell{94.24 $\pm$ 0.10} & \revCell{94.21 $\pm$ 0.04} & \revCell{94.38 $\pm$ 0.08} \\
          &    & \revCell{0.02} & \revCell{94.25 $\pm$ 0.10} & \revCell{94.24 $\pm$ 0.04} & \revCell{94.42 $\pm$ 0.08} \\
          &    & \revCell{0.3} & \revCell{\textbf{94.48 $\pm$ 0.09}$^{*}$} & \revCell{\textbf{94.48 $\pm$ 0.04}$^{*}$} & \revCell{\textbf{94.62 $\pm$ 0.08}$^{*}$} \\
    \cline{2-6}
    \addlinespace[1pt]
          & \revCell{\multirow{3}{*}{5}} & \revCell{0} & \revCell{94.24 $\pm$ 0.08} & \revCell{94.19 $\pm$ 0.05} & \revCell{94.35 $\pm$ 0.08} \\
          &    & \revCell{0.02} & \revCell{94.24 $\pm$ 0.08} & \revCell{94.21 $\pm$ 0.05} & \revCell{94.38 $\pm$ 0.08} \\
          &    & \revCell{0.3} & \revCell{\textbf{94.43 $\pm$ 0.09}$^{*}$} & \revCell{\textbf{94.40 $\pm$ 0.03}$^{*}$} & \revCell{\textbf{94.55 $\pm$ 0.05}$^{*}$} \\
    \cline{2-6}
    \addlinespace[1pt]
          & \revCell{\multirow{3}{*}{4}} & \revCell{0} & \revCell{\textbf{94.24 $\pm$ 0.09}} & \revCell{94.19 $\pm$ 0.05} & \revCell{\textbf{94.35 $\pm$ 0.07}} \\
          &    & \revCell{0.02} & \revCell{94.23 $\pm$ 0.10} & \revCell{94.19 $\pm$ 0.05} & \revCell{94.35 $\pm$ 0.07} \\
          &    & \revCell{0.3} & \revCell{94.23 $\pm$ 0.07} & \revCell{\textbf{94.19 $\pm$ 0.05}} & \revCell{94.35 $\pm$ 0.09} \\
    \cline{2-6}
    \addlinespace[1pt]
          & \revCell{\multirow{3}{*}{3}} & \revCell{0} & \revCell{94.27 $\pm$ 0.11} & \revCell{94.26 $\pm$ 0.04} & \revCell{94.61 $\pm$ 0.06} \\
          &    & \revCell{0.02} & \revCell{94.30 $\pm$ 0.11} & \revCell{94.29 $\pm$ 0.04} & \revCell{94.63 $\pm$ 0.10} \\
          &    & \revCell{0.3} & \revCell{\textbf{94.46 $\pm$ 0.10}$^{*}$} & \revCell{\textbf{94.48 $\pm$ 0.02}$^{*}$} & \revCell{\textbf{94.66 $\pm$ 0.07}$^{*}$} \\
    \cline{2-6}
    \addlinespace[1pt]
          & \revCell{\multirow{3}{*}{2}} & \revCell{0} & \revCell{94.47 $\pm$ 0.08} & \revCell{94.47 $\pm$ 0.03} & \revCell{94.70 $\pm$ 0.09} \\
          &    & \revCell{0.02} & \revCell{94.53 $\pm$ 0.09} & \revCell{94.51 $\pm$ 0.04} & \revCell{94.70 $\pm$ 0.08} \\
          &    & \revCell{0.3} & \revCell{\textbf{94.63 $\pm$ 0.09}$^{*}$} & \revCell{\textbf{94.61 $\pm$ 0.02}$^{*}$} & \revCell{\textbf{94.73 $\pm$ 0.04}} \\
    \cline{2-6}
    \addlinespace[1pt]
          & \revCell{\multirow{3}{*}{1}} & \revCell{0} & \revCell{\textbf{94.70 $\pm$ 0.06}} & \revCell{94.66 $\pm$ 0.03} & \revCell{\textbf{94.80 $\pm$ 0.06}} \\
          &    & \revCell{0.02} & \revCell{94.68 $\pm$ 0.06} & \revCell{\textbf{94.69 $\pm$ 0.02}$^{*}$} & \revCell{\textbf{94.80 $\pm$ 0.05}} \\
          &    & \revCell{0.3} & \revCell{94.65 $\pm$ 0.06} & \revCell{94.68 $\pm$ 0.02} & \revCell{94.77 $\pm$ 0.04} \\
    \bottomrule
    \end{tabular}
    \revTableFootnote{$^{*}$ marks a nominal paired one-sided $t$-test at $p<0.05$ for the bolded block-wise best against $c_{\mathrm{rec}}=0$ when the best uses $c_{\mathrm{rec}}>0$; otherwise it is compared with the best nonzero setting.}
\end{table}\par
\begin{table}[tp]
    \centering
    \caption{\rev{Tiny-ImageNet post-training results.}}
    \label{tab:tin}
    \renewcommand{\arraystretch}{0.944}
    \setlength{\tabcolsep}{5.5pt}
    \scriptsize
    \begin{tabular}{@{}lccccc@{}}
    \toprule
    \revCell{Model} & \revCell{$l_{R}$} & \revCell{$c_{\mathrm{rec}}$} & \revCell{Epoch 5 Acc} & \revCell{Mean Acc} & \revCell{Best Acc} \\
    \midrule
    \revCell{\multirow{18}{*}{\makecell[c]{Simple\\CNN}}} & \revCell{\multirow{3}{*}{out}} & \revCell{0} & \revCell{33.49 $\pm$ 0.16} & \revCell{\textbf{33.52 $\pm$ 0.06}$^{*}$} & \revCell{\textbf{33.92 $\pm$ 0.16}$^{*}$} \\
          &    & \revCell{0.02} & \revCell{\textbf{33.50 $\pm$ 0.08}} & \revCell{33.46 $\pm$ 0.02} & \revCell{33.62 $\pm$ 0.05} \\
          &    & \revCell{0.1} & \revCell{33.48 $\pm$ 0.04} & \revCell{33.44 $\pm$ 0.03} & \revCell{33.56 $\pm$ 0.04} \\
    \cline{2-6}
    \addlinespace[1pt]
          & \revCell{\multirow{3}{*}{5}} & \revCell{0} & \revCell{33.46 $\pm$ 0.18} & \revCell{33.48 $\pm$ 0.06} & \revCell{33.91 $\pm$ 0.15} \\
          &    & \revCell{0.02} & \revCell{33.48 $\pm$ 0.15} & \revCell{33.50 $\pm$ 0.07} & \revCell{33.91 $\pm$ 0.15} \\
          &    & \revCell{0.1} & \revCell{\textbf{33.61 $\pm$ 0.12}$^{*}$} & \revCell{\textbf{33.67 $\pm$ 0.08}$^{*}$} & \revCell{\textbf{34.04 $\pm$ 0.14}$^{*}$} \\
    \cline{2-6}
    \addlinespace[1pt]
          & \revCell{\multirow{3}{*}{4}} & \revCell{0} & \revCell{\textbf{33.39 $\pm$ 0.19}} & \revCell{\textbf{33.43 $\pm$ 0.07}} & \revCell{33.81 $\pm$ 0.15} \\
          &    & \revCell{0.02} & \revCell{33.34 $\pm$ 0.21} & \revCell{33.42 $\pm$ 0.07} & \revCell{\textbf{33.83 $\pm$ 0.16}} \\
          &    & \revCell{0.1} & \revCell{33.29 $\pm$ 0.19} & \revCell{33.34 $\pm$ 0.07} & \revCell{33.78 $\pm$ 0.16} \\
    \cline{2-6}
    \addlinespace[1pt]
          & \revCell{\multirow{3}{*}{3}} & \revCell{0} & \revCell{33.33 $\pm$ 0.18} & \revCell{33.39 $\pm$ 0.08} & \revCell{33.74 $\pm$ 0.14} \\
          &    & \revCell{0.02} & \revCell{33.32 $\pm$ 0.19} & \revCell{33.39 $\pm$ 0.08} & \revCell{33.76 $\pm$ 0.11} \\
          &    & \revCell{0.1} & \revCell{\textbf{33.37 $\pm$ 0.20}} & \revCell{\textbf{33.40 $\pm$ 0.09}} & \revCell{\textbf{33.76 $\pm$ 0.08}} \\
    \cline{2-6}
    \addlinespace[1pt]
          & \revCell{\multirow{3}{*}{2}} & \revCell{0} & \revCell{33.35 $\pm$ 0.12} & \revCell{33.36 $\pm$ 0.08} & \revCell{33.65 $\pm$ 0.12} \\
          &    & \revCell{0.02} & \revCell{33.40 $\pm$ 0.16} & \revCell{33.42 $\pm$ 0.08} & \revCell{33.71 $\pm$ 0.12} \\
          &    & \revCell{0.1} & \revCell{\textbf{33.46 $\pm$ 0.13}$^{*}$} & \revCell{\textbf{33.48 $\pm$ 0.09}$^{*}$} & \revCell{\textbf{33.80 $\pm$ 0.09}$^{*}$} \\
    \cline{2-6}
    \addlinespace[1pt]
          & \revCell{\multirow{3}{*}{1}} & \revCell{0} & \revCell{\textbf{33.28 $\pm$ 0.09}} & \revCell{\textbf{33.30 $\pm$ 0.04}} & \revCell{33.52 $\pm$ 0.09} \\
          &    & \revCell{0.02} & \revCell{33.26 $\pm$ 0.09} & \revCell{33.29 $\pm$ 0.04} & \revCell{33.52 $\pm$ 0.10} \\
          &    & \revCell{0.1} & \revCell{33.25 $\pm$ 0.11} & \revCell{33.30 $\pm$ 0.04} & \revCell{\textbf{33.55 $\pm$ 0.10}$^{*}$} \\
    \midrule
    \revCell{\multirow{18}{*}{\makecell[c]{ResNet\\-34}}}& \revCell{\multirow{3}{*}{out}} & \revCell{0} & \revCell{58.77 $\pm$ 0.18} & \revCell{58.91 $\pm$ 0.12} & \revCell{59.59 $\pm$ 0.27} \\
          &    & \revCell{0.1} & \revCell{59.00 $\pm$ 0.20} & \revCell{59.30 $\pm$ 0.10} & \revCell{60.08 $\pm$ 0.27} \\
          &    & \revCell{0.3} & \revCell{\textbf{59.53 $\pm$ 0.17}$^{*}$} & \revCell{\textbf{59.71 $\pm$ 0.08}$^{*}$} & \revCell{\textbf{60.21 $\pm$ 0.13}$^{*}$} \\
    \cline{2-6}
    \addlinespace[1pt]
          & \revCell{\multirow{3}{*}{5}} & \revCell{0} & \revCell{58.83 $\pm$ 0.13} & \revCell{58.98 $\pm$ 0.12} & \revCell{59.66 $\pm$ 0.27} \\
          &    & \revCell{0.1} & \revCell{59.06 $\pm$ 0.13} & \revCell{59.34 $\pm$ 0.06} & \revCell{60.12 $\pm$ 0.20} \\
          &    & \revCell{0.3} & \revCell{\textbf{59.58 $\pm$ 0.20}$^{*}$} & \revCell{\textbf{59.71 $\pm$ 0.09}$^{*}$} & \revCell{\textbf{60.17 $\pm$ 0.13}$^{*}$} \\
    \cline{2-6}
    \addlinespace[1pt]
          & \revCell{\multirow{3}{*}{4}} & \revCell{0} & \revCell{58.80 $\pm$ 0.20} & \revCell{58.97 $\pm$ 0.08} & \revCell{59.74 $\pm$ 0.26} \\
          &    & \revCell{0.1} & \revCell{59.02 $\pm$ 0.10} & \revCell{59.17 $\pm$ 0.11} & \revCell{59.96 $\pm$ 0.23} \\
          &    & \revCell{0.3} & \revCell{\textbf{59.29 $\pm$ 0.24}$^{*}$} & \revCell{\textbf{59.47 $\pm$ 0.06}$^{*}$} & \revCell{\textbf{60.08 $\pm$ 0.09}$^{*}$} \\
    \cline{2-6}
    \addlinespace[1pt]
          & \revCell{\multirow{3}{*}{3}} & \revCell{0} & \revCell{58.99 $\pm$ 0.21} & \revCell{58.83 $\pm$ 0.09} & \revCell{59.86 $\pm$ 0.18} \\
          &    & \revCell{0.1} & \revCell{59.44 $\pm$ 0.30} & \revCell{59.22 $\pm$ 0.08} & \revCell{59.95 $\pm$ 0.19} \\
          &    & \revCell{0.3} & \revCell{\textbf{59.64 $\pm$ 0.26}$^{*}$} & \revCell{\textbf{59.49 $\pm$ 0.07}$^{*}$} & \revCell{\textbf{59.96 $\pm$ 0.21}$^{*}$} \\
    \cline{2-6}
    \addlinespace[1pt]
          & \revCell{\multirow{3}{*}{2}} & \revCell{0} & \revCell{59.09 $\pm$ 0.16} & \revCell{58.96 $\pm$ 0.07} & \revCell{59.33 $\pm$ 0.11} \\
          &    & \revCell{0.1} & \revCell{59.19 $\pm$ 0.21} & \revCell{59.16 $\pm$ 0.06} & \revCell{59.45 $\pm$ 0.09} \\
          &    & \revCell{0.3} & \revCell{\textbf{59.20 $\pm$ 0.26}$^{*}$} & \revCell{\textbf{59.17 $\pm$ 0.05}$^{*}$} & \revCell{\textbf{59.47 $\pm$ 0.09}$^{*}$} \\
    \cline{2-6}
    \addlinespace[1pt]
          & \revCell{\multirow{3}{*}{1}} & \revCell{0} & \revCell{58.78 $\pm$ 0.16} & \revCell{58.81 $\pm$ 0.06} & \revCell{59.03 $\pm$ 0.10} \\
          &    & \revCell{0.1} & \revCell{\textbf{58.80 $\pm$ 0.17}} & \revCell{\textbf{58.85 $\pm$ 0.05}$^{*}$} & \revCell{\textbf{59.07 $\pm$ 0.05}} \\
          &    & \revCell{0.3} & \revCell{\textbf{58.80 $\pm$ 0.17}} & \revCell{\textbf{58.85 $\pm$ 0.05}} & \revCell{\textbf{59.07 $\pm$ 0.05}} \\
    \midrule
    \revCell{\multirow{33}{*}{\makecell[c]{Simple\\ViT}}} & \revCell{\multirow{3}{*}{out}} & \revCell{0} & \revCell{43.19 $\pm$ 0.16} & \revCell{43.04 $\pm$ 0.04} & \revCell{44.85 $\pm$ 0.18} \\
          &    & \revCell{0.1} & \revCell{\textbf{45.12 $\pm$ 0.06}$^{*}$} & \revCell{\textbf{45.15 $\pm$ 0.02}$^{*}$} & \revCell{\textbf{45.42 $\pm$ 0.06}$^{*}$} \\
          &    & \revCell{0.3} & \revCell{45.10 $\pm$ 0.09} & \revCell{45.10 $\pm$ 0.03} & \revCell{45.22 $\pm$ 0.05} \\
    \cline{2-6}
    \addlinespace[1pt]
          & \revCell{\multirow{3}{*}{10}} & \revCell{0} & \revCell{43.23 $\pm$ 0.18} & \revCell{43.07 $\pm$ 0.04} & \revCell{44.82 $\pm$ 0.18} \\
          &    & \revCell{0.1} & \revCell{\textbf{45.50 $\pm$ 0.14}$^{*}$} & \revCell{44.76 $\pm$ 0.05} & \revCell{\textbf{45.51 $\pm$ 0.12}$^{*}$} \\
          &    & \revCell{0.3} & \revCell{45.20 $\pm$ 0.08} & \revCell{\textbf{45.20 $\pm$ 0.02}$^{*}$} & \revCell{45.35 $\pm$ 0.03} \\
    \cline{2-6}
    \addlinespace[1pt]
          & \revCell{\multirow{3}{*}{9}} & \revCell{0} & \revCell{43.23 $\pm$ 0.18} & \revCell{43.07 $\pm$ 0.04} & \revCell{44.82 $\pm$ 0.18} \\
          &    & \revCell{0.1} & \revCell{45.38 $\pm$ 0.13} & \revCell{44.99 $\pm$ 0.04} & \revCell{45.64 $\pm$ 0.07} \\
          &    & \revCell{0.3} & \revCell{\textbf{45.56 $\pm$ 0.10}$^{*}$} & \revCell{\textbf{45.52 $\pm$ 0.03}$^{*}$} & \revCell{\textbf{45.68 $\pm$ 0.04}$^{*}$} \\
    \cline{2-6}
    \addlinespace[1pt]
          & \revCell{\multirow{3}{*}{8}} & \revCell{0} & \revCell{43.02 $\pm$ 0.19} & \revCell{42.87 $\pm$ 0.05} & \revCell{44.74 $\pm$ 0.14} \\
          &    & \revCell{0.1} & \revCell{45.23 $\pm$ 0.11} & \revCell{44.78 $\pm$ 0.03} & \revCell{45.47 $\pm$ 0.06} \\
          &    & \revCell{0.3} & \revCell{\textbf{45.38 $\pm$ 0.12}$^{*}$} & \revCell{\textbf{45.32 $\pm$ 0.03}$^{*}$} & \revCell{\textbf{45.50 $\pm$ 0.04}$^{*}$} \\
    \cline{2-6}
    \addlinespace[1pt]
          & \revCell{\multirow{3}{*}{7}} & \revCell{0} & \revCell{43.58 $\pm$ 0.13} & \revCell{43.40 $\pm$ 0.03} & \revCell{44.61 $\pm$ 0.08} \\
          &    & \revCell{0.1} & \revCell{44.98 $\pm$ 0.11} & \revCell{44.79 $\pm$ 0.03} & \revCell{45.22 $\pm$ 0.06} \\
          &    & \revCell{0.3} & \revCell{\textbf{45.00 $\pm$ 0.10}$^{*}$} & \revCell{\textbf{44.97 $\pm$ 0.02}$^{*}$} & \revCell{\textbf{45.22 $\pm$ 0.05}$^{*}$} \\
    \cline{2-6}
    \addlinespace[1pt]
          & \revCell{\multirow{3}{*}{6}} & \revCell{0} & \revCell{43.94 $\pm$ 0.20} & \revCell{43.87 $\pm$ 0.04} & \revCell{44.57 $\pm$ 0.09} \\
          &    & \revCell{0.1} & \revCell{\textbf{44.84 $\pm$ 0.11}$^{*}$} & \revCell{44.70 $\pm$ 0.02} & \revCell{\textbf{44.93 $\pm$ 0.05}$^{*}$} \\
          &    & \revCell{0.3} & \revCell{\textbf{44.84 $\pm$ 0.11}} & \revCell{\textbf{44.74 $\pm$ 0.02}$^{*}$} & \revCell{\textbf{44.93 $\pm$ 0.05}} \\
    \cline{2-6}
    \addlinespace[1pt]
          & \revCell{\multirow{3}{*}{5}} & \revCell{0} & \revCell{44.19 $\pm$ 0.15} & \revCell{44.11 $\pm$ 0.05} & \revCell{\textbf{44.58 $\pm$ 0.09}} \\
          &    & \revCell{0.1} & \revCell{\textbf{44.37 $\pm$ 0.06}$^{*}$} & \revCell{\textbf{44.34 $\pm$ 0.02}$^{*}$} & \revCell{44.54 $\pm$ 0.05} \\
          &    & \revCell{0.3} & \revCell{\textbf{44.37 $\pm$ 0.06}} & \revCell{44.34 $\pm$ 0.02} & \revCell{44.54 $\pm$ 0.05} \\
    \cline{2-6}
    \addlinespace[1pt]
          & \revCell{\multirow{3}{*}{4}} & \revCell{0} & \revCell{\textbf{44.45 $\pm$ 0.10}} & \revCell{44.30 $\pm$ 0.04} & \revCell{44.55 $\pm$ 0.05} \\
          &    & \revCell{0.1} & \revCell{44.44 $\pm$ 0.05} & \revCell{\textbf{44.43 $\pm$ 0.03}$^{*}$} & \revCell{\textbf{44.59 $\pm$ 0.05}$^{*}$} \\
          &    & \revCell{0.3} & \revCell{44.44 $\pm$ 0.05} & \revCell{\textbf{44.43 $\pm$ 0.03}} & \revCell{\textbf{44.59 $\pm$ 0.05}} \\
    \cline{2-6}
    \addlinespace[1pt]
          & \revCell{\multirow{3}{*}{3}} & \revCell{0} & \revCell{44.44 $\pm$ 0.09} & \revCell{44.38 $\pm$ 0.03} & \revCell{44.55 $\pm$ 0.05} \\
          &    & \revCell{0.1} & \revCell{\textbf{44.58 $\pm$ 0.07}$^{*}$} & \revCell{\textbf{44.56 $\pm$ 0.02}$^{*}$} & \revCell{\textbf{44.71 $\pm$ 0.04}$^{*}$} \\
          &    & \revCell{0.3} & \revCell{\textbf{44.58 $\pm$ 0.07}} & \revCell{\textbf{44.56 $\pm$ 0.02}} & \revCell{\textbf{44.71 $\pm$ 0.04}} \\
    \cline{2-6}
    \addlinespace[1pt]
          & \revCell{\multirow{3}{*}{2}} & \revCell{0} & \revCell{44.33 $\pm$ 0.09} & \revCell{44.32 $\pm$ 0.04} & \revCell{44.58 $\pm$ 0.05} \\
          &    & \revCell{0.1} & \revCell{\textbf{44.67 $\pm$ 0.04}$^{*}$} & \revCell{\textbf{44.68 $\pm$ 0.02}$^{*}$} & \revCell{\textbf{44.79 $\pm$ 0.04}$^{*}$} \\
          &    & \revCell{0.3} & \revCell{\textbf{44.67 $\pm$ 0.04}} & \revCell{\textbf{44.68 $\pm$ 0.02}} & \revCell{\textbf{44.79 $\pm$ 0.04}} \\
    \cline{2-6}
    \addlinespace[1pt]
          & \revCell{\multirow{3}{*}{1}} & \revCell{0} & \revCell{44.44 $\pm$ 0.07} & \revCell{44.42 $\pm$ 0.03} & \revCell{44.60 $\pm$ 0.05} \\
          &    & \revCell{0.1} & \revCell{\textbf{44.69 $\pm$ 0.06}$^{*}$} & \revCell{\textbf{44.63 $\pm$ 0.02}$^{*}$} & \revCell{\textbf{44.76 $\pm$ 0.03}$^{*}$} \\
          &    & \revCell{0.3} & \revCell{\textbf{44.69 $\pm$ 0.06}} & \revCell{\textbf{44.63 $\pm$ 0.02}} & \revCell{\textbf{44.76 $\pm$ 0.03}} \\
    \bottomrule
    \end{tabular}
    \revTableFootnote{$^{*}$ marks a nominal paired one-sided $t$-test at $p<0.05$ for the bolded block-wise best against $c_{\mathrm{rec}}=0$ when the best uses $c_{\mathrm{rec}}>0$; otherwise it is compared with the best nonzero setting.}
\end{table}
\rev{Overall, reconstruction supervision yields broad improvements across datasets and architectures. Using Best Acc as a compact summary, the best FR-PT setting with $c_{\mathrm{rec}}>0$ significantly outperforms its matched BP reference in $3/3$, $10/11$, $17/22$, $8/13$, $7/13$, and $18/23$ $l_R$ groups on MNIST (Table~\ref{tab:mnist}), CIFAR-10 (Table~\ref{tab:cifar10}), CIFAR-100 (Table~\ref{tab:cifar100}), Imagenette (Table~\ref{tab:nette}), Imagewoof (Table~\ref{tab:woof}), and Tiny-ImageNet (Table~\ref{tab:tin}), respectively, corresponding to $63/85$ ($74.1\%$) of all $l_R$ groups. In contrast, the matched BP reference significantly outperforms the best $c_{\mathrm{rec}}>0$ setting in only $4/85$ ($4.7\%$) groups. Consistent gains are also evident in individual results. For example, Table~\ref{tab:cifar100} shows that SimpleViT benefits from reconstruction supervision across all evaluated reconstruction layers: at $l_R=9$, the Mean Acc increases from $61.07\%$ with $c_{\mathrm{rec}}=0$ to $63.91\%$ with $c_{\mathrm{rec}}=0.3$, while Best Acc increases from $63.28\%$ to $64.07\%$. Similarly, for ResNet-34 in Table~\ref{tab:tin}, FR-PT improves the Mean/Best Acc from $58.91\%/59.59\%$ to $59.71\%/60.21\%$ at $l_R=\mathrm{out}$. On the real-image benchmarks, ResNet-18 on Imagenette improves from $98.11\%$ to $98.36\%$ in Mean Acc at $l_R=\mathrm{out}$ (Table~\ref{tab:nette}), while the corresponding value on Imagewoof increases from $94.21\%$ to $94.48\%$ (Table~\ref{tab:woof}). These repeated gains across substantially different tasks and architectures support the effectiveness of reconstructed feature targets as additional post-training supervision. Since $\mathbf{a}_{l_R}^{\ast}$ is reconstructed from the ground-truth label through the frozen downstream path under the CCP/MDP formulation, the reconstruction loss guides trainable upstream modules toward GT-conditioned features consistent with the learned downstream mapping. FR-PT thus combines task-level and intermediate feature-level supervision, providing a more direct optimization target for upstream parts.\par }
\rev{The reconstruction layer also has a clear influence on the post-training performance. In many CNN and ResNet configurations, moving $l_R$ toward earlier layers leads to an overall decrease in accuracy regardless of whether reconstruction supervision is used. As shown in Table~\ref{tab:nette}, the Mean Acc of ResNet-18 with $c_{\mathrm{rec}}=0$ decreases from $98.11\%$ at $l_R=\mathrm{out}$ to $96.93\%$ at $l_R=1$, while the best reconstruction-weighted result decreases from $98.36\%$ to $96.95\%$. A similar tendency is observed for ResNet-34 on Tiny-ImageNet (Table~\ref{tab:tin}), where the best Mean Acc decreases from $59.71\%$ at the output layer to $58.85\%$ at $l_R=1$. Table~\ref{tab:cifar100} exhibits the same tendency for ResNet-18: the best Mean Acc reaches $79.69\%$ at $l_R=4$ but drops to $79.27\%$ at $l_R=1$. One possible explanation is that, as $l_R$ moves toward earlier layers, a larger downstream portion is frozen and fewer upstream parameters remain trainable, which progressively restricts the optimization space available to accommodate the post-training objective. This tendency is not universal across architectures; notably, the SimpleViT results in Tables~\ref{tab:cifar100} and~\ref{tab:tin} exhibit a different depth-dependent pattern, suggesting that the effect of $l_R$ may also depend on the underlying network structure and reconstruction geometry.\par
}
\rev{FR-PT does not outperform the matched BP reference in every configuration, particularly at earlier reconstruction layers. For example, for SimpleCNN on Imagewoof at $l_R=2$ (Table~\ref{tab:woof}), the BP reference achieves $33.66\%$ Mean Acc and $34.12\%$ Best Acc, compared with $33.65\%$ and $34.05\%$ for the best $c_{\mathrm{rec}}>0$ setting. Similarly, on Tiny-ImageNet at $l_R=4$ (Table~\ref{tab:tin}), the reference attains $33.43\%$ Mean Acc versus $33.42\%$ with reconstruction supervision. Such small isolated differences may arise from stochastic optimization and finite-seed variability and should not be interpreted as conclusive evidence against reconstruction supervision. More importantly, earlier $l_R$ requires a longer reverse path, along which approximation and numerical errors can accumulate as characterized by Eq.~\ref{eq:error_accumulation}. The reconstructed target may therefore deviate progressively from the ideal CCP/MDP target, making the auxiliary supervision less accurate and potentially reducing or even reversing its post-training benefit.\par
}
\subsubsection{\rev{Trajectory Planning in AD}}
\label{sec:traj_planning_ad}
\rev{To assess FR-PT on a large-scale real-world task, we adopt the NAVSIM trajectory-planning protocol with ResNet-34 and ViT image backbones, whose architectures are detailed in Supplementary Part~B.2. Tables~\ref{tab:navsim_resnet34} and~\ref{tab:navsim_vit} report single-run results after five post-training epochs. The baseline denotes the original pretrained checkpoint; for each reconstruction location $l_R$, $c_{\mathrm{rec}}=0$ is the matched BP reference, while $c_{\mathrm{rec}}>0$ adds feature-level reconstruction supervision under otherwise identical settings. Due to storage constraints, reconstruction is evaluated only at selected late-stage layers rather than exhaustively treating every layer as $l_R$. The reported metrics are NC (no at-fault collisions), DAC (drivable area compliance), EP (ego progress), TTC (time to collision within bound), Comf. (comfort), DDC (driving direction compliance), and PDMS (Predictive Driver Model Score). DDC has zero weight in the NAVSIM v1 PDMS. Boldface marks the best displayed value within each $l_R$ group for each metric.\par
}
\begin{table}[tp]
    \centering
    \caption{\rev{NAVSIM trajectory-planning results using ResNet-34 as the image backbone.}}
    \label{tab:navsim_resnet34}
    \renewcommand{\arraystretch}{1.08}
    \setlength{\tabcolsep}{2pt}
    \scriptsize
    \begin{tabular*}{\columnwidth}{@{\extracolsep{\fill}}clccccccc@{}}
    \toprule
    \revCell{$l_{R}$} & \revCell{$c_{\mathrm{rec}}$} & \revCell{NC$\uparrow$} & \revCell{DAC$\uparrow$} & \revCell{EP$\uparrow$} & \revCell{TTC$\uparrow$} & \revCell{Comf.$\uparrow$} & \revCell{DDC$\uparrow$} & \revCell{PDMS$\uparrow$} \\
    \midrule
    \multicolumn{2}{c}{\revCell{baseline}} & \revCell{96.95} & \revCell{91.14} & \revCell{76.66} & \revCell{91.24} & \revCell{99.97} & \revCell{97.01} & \revCell{81.43} \\
    \midrule
    \revCell{\multirow{4}{*}{\makecell[c]{planning\\$z_2$}}}
            & \revCell{0}    & \revCell{96.48} & \revCell{\textbf{91.43}} & \revCell{76.76} & \revCell{90.62} & \revCell{99.96} & \revCell{97.05} & \revCell{81.34} \\
            & \revCell{0.02} & \revCell{\textbf{97.18}} & \revCell{91.40} & \revCell{76.59} & \revCell{\textbf{91.92}} & \revCell{99.96} & \revCell{\textbf{97.21}} & \revCell{\textbf{81.88}} \\
            & \revCell{0.1}  & \revCell{96.86} & \revCell{91.36} & \revCell{\textbf{77.01}} & \revCell{91.17} & \revCell{\textbf{99.97}} & \revCell{97.19} & \revCell{81.73} \\
            & \revCell{0.3}  & \revCell{96.84} & \revCell{90.78} & \revCell{76.40} & \revCell{91.34} & \revCell{\textbf{99.97}} & \revCell{97.17} & \revCell{81.19} \\
    \midrule
    \revCell{\multirow{4}{*}{\makecell[c]{planning\\$z_1$}}}
            & \revCell{0}    & \revCell{\textbf{97.20}} & \revCell{90.80} & \revCell{76.28} & \revCell{\textbf{92.12}} & \revCell{99.97} & \revCell{97.12} & \revCell{81.48} \\
            & \revCell{0.02} & \revCell{97.00} & \revCell{91.00} & \revCell{76.50} & \revCell{91.78} & \revCell{99.95} & \revCell{97.00} & \revCell{\textbf{82.00}} \\
            & \revCell{0.1}  & \revCell{96.65} & \revCell{\textbf{91.17}} & \revCell{76.52} & \revCell{91.37} & \revCell{\textbf{99.98}} & \revCell{96.92} & \revCell{81.42} \\
            & \revCell{0.3}  & \revCell{97.01} & \revCell{91.13} & \revCell{\textbf{76.67}} & \revCell{91.68} & \revCell{99.97} & \revCell{\textbf{97.19}} & \revCell{81.67} \\
    \midrule
    \revCell{\multirow{4}{*}{\makecell[c]{fusion\\$z$}}}
            & \revCell{0}    & \revCell{96.55} & \revCell{90.97} & \revCell{76.60} & \revCell{90.70} & \revCell{\textbf{99.99}} & \revCell{96.85} & \revCell{81.07} \\
            & \revCell{0.02} & \revCell{96.77} & \revCell{\textbf{91.70}} & \revCell{\textbf{77.12}} & \revCell{91.29} & \revCell{99.98} & \revCell{97.13} & \revCell{\textbf{81.95}} \\
            & \revCell{0.1}  & \revCell{96.85} & \revCell{91.59} & \revCell{76.68} & \revCell{91.60} & \revCell{99.96} & \revCell{\textbf{97.28}} & \revCell{81.81} \\
            & \revCell{0.3}  & \revCell{\textbf{97.13}} & \revCell{91.26} & \revCell{76.69} & \revCell{\textbf{91.76}} & \revCell{99.98} & \revCell{97.05} & \revCell{81.83} \\
    \midrule
    \revCell{\multirow{4}{*}{\makecell[c]{ResNet\\$z_4$}}}
            & \revCell{0}    & \revCell{\textbf{96.88}} & \revCell{\textbf{90.72}} & \revCell{\textbf{76.41}} & \revCell{91.24} & \revCell{99.97} & \revCell{96.96} & \revCell{\textbf{81.16}} \\
            & \revCell{0.02} & \revCell{96.00} & \revCell{90.00} & \revCell{76.27} & \revCell{90.44} & \revCell{\textbf{99.98}} & \revCell{97.00} & \revCell{80.50} \\
            & \revCell{0.1}  & \revCell{96.80} & \revCell{90.44} & \revCell{76.19} & \revCell{\textbf{91.44}} & \revCell{\textbf{99.98}} & \revCell{97.02} & \revCell{80.98} \\
            & \revCell{0.3}  & \revCell{96.36} & \revCell{90.70} & \revCell{76.17} & \revCell{90.57} & \revCell{\textbf{99.98}} & \revCell{\textbf{97.09}} & \revCell{80.66} \\
    \midrule
    \revCell{\multirow{4}{*}{\makecell[c]{ResNet\\$z_3$}}}
            & \revCell{0}    & \revCell{96.84} & \revCell{90.93} & \revCell{76.50} & \revCell{91.22} & \revCell{99.95} & \revCell{\textbf{97.05}} & \revCell{81.22} \\
            & \revCell{0.02} & \revCell{96.97} & \revCell{91.02} & \revCell{76.00} & \revCell{91.40} & \revCell{99.97} & \revCell{96.99} & \revCell{\textbf{81.42}} \\
            & \revCell{0.1}  & \revCell{\textbf{97.01}} & \revCell{90.66} & \revCell{76.24} & \revCell{\textbf{91.43}} & \revCell{\textbf{99.98}} & \revCell{96.95} & \revCell{81.15} \\
            & \revCell{0.3}  & \revCell{96.65} & \revCell{\textbf{91.27}} & \revCell{\textbf{76.82}} & \revCell{90.87} & \revCell{\textbf{99.98}} & \revCell{97.01} & \revCell{\textbf{81.42}} \\
    \bottomrule
    \end{tabular*}
    \vspace{-4pt}
\end{table}\par
\begin{table}[tp]
    \centering
    \caption{\rev{NAVSIM trajectory-planning results using ViT as the image backbone.}}
    \label{tab:navsim_vit}
    \renewcommand{\arraystretch}{1.08}
    \setlength{\tabcolsep}{2pt}
    \scriptsize
    \begin{tabular*}{\columnwidth}{@{\extracolsep{\fill}}clccccccc@{}}
    \toprule
    \revCell{$l_{R}$} & \revCell{$c_{\mathrm{rec}}$} & \revCell{NC$\uparrow$} & \revCell{DAC$\uparrow$} & \revCell{EP$\uparrow$} & \revCell{TTC$\uparrow$} & \revCell{Comf.$\uparrow$} & \revCell{DDC$\uparrow$} & \revCell{PDMS$\uparrow$} \\
    \midrule
    \multicolumn{2}{c}{\revCell{baseline}} & \revCell{95.83} & \revCell{87.45} & \revCell{73.00} & \revCell{89.61} & \revCell{99.97} & \revCell{96.42} & \revCell{77.31} \\
    \midrule
    \revCell{\multirow{4}{*}{\makecell[c]{planning\\$z_2$}}}
            & \revCell{0}    & \revCell{96.04} & \revCell{88.25} & \revCell{73.64} & \revCell{89.80} & \revCell{\textbf{99.98}} & \revCell{96.11} & \revCell{77.97} \\
            & \revCell{0.02} & \revCell{96.15} & \revCell{\textbf{88.70}} & \revCell{\textbf{74.23}} & \revCell{90.04} & \revCell{\textbf{99.98}} & \revCell{96.35} & \revCell{78.55} \\
            & \revCell{0.1}  & \revCell{\textbf{96.38}} & \revCell{88.69} & \revCell{74.18} & \revCell{\textbf{90.66}} & \revCell{\textbf{99.98}} & \revCell{\textbf{96.47}} & \revCell{\textbf{78.84}} \\
            & \revCell{0.3}  & \revCell{96.18} & \revCell{88.12} & \revCell{73.84} & \revCell{90.11} & \revCell{\textbf{99.98}} & \revCell{96.42} & \revCell{78.14} \\
    \midrule
    \revCell{\multirow{4}{*}{\makecell[c]{planning\\$z_1$}}}
            & \revCell{0}    & \revCell{96.43} & \revCell{88.58} & \revCell{\textbf{74.19}} & \revCell{90.42} & \revCell{\textbf{99.98}} & \revCell{96.34} & \revCell{78.68} \\
            & \revCell{0.02} & \revCell{\textbf{96.65}} & \revCell{\textbf{88.79}} & \revCell{73.87} & \revCell{\textbf{91.20}} & \revCell{\textbf{99.98}} & \revCell{96.44} & \revCell{\textbf{79.04}} \\
            & \revCell{0.1}  & \revCell{96.44} & \revCell{88.14} & \revCell{73.33} & \revCell{90.90} & \revCell{\textbf{99.98}} & \revCell{96.24} & \revCell{78.33} \\
            & \revCell{0.3}  & \revCell{96.54} & \revCell{87.75} & \revCell{73.16} & \revCell{91.03} & \revCell{\textbf{99.98}} & \revCell{\textbf{96.49}} & \revCell{78.15} \\
    \midrule
    \revCell{\multirow{4}{*}{\makecell[c]{fusion\\$z$}}}
            & \revCell{0}    & \revCell{96.05} & \revCell{88.19} & \revCell{73.59} & \revCell{90.05} & \revCell{\textbf{99.98}} & \revCell{96.25} & \revCell{78.03} \\
            & \revCell{0.02} & \revCell{96.37} & \revCell{88.35} & \revCell{73.69} & \revCell{90.81} & \revCell{99.97} & \revCell{96.21} & \revCell{78.59} \\
            & \revCell{0.1}  & \revCell{\textbf{96.54}} & \revCell{88.56} & \revCell{73.77} & \revCell{\textbf{91.12}} & \revCell{\textbf{99.98}} & \revCell{\textbf{96.35}} & \revCell{\textbf{78.84}} \\
            & \revCell{0.3}  & \revCell{96.25} & \revCell{\textbf{88.60}} & \revCell{\textbf{74.04}} & \revCell{90.67} & \revCell{\textbf{99.98}} & \revCell{96.32} & \revCell{78.81} \\
    \midrule
    \revCell{\multirow{4}{*}{\makecell[c]{image\\embedding}}}
            & \revCell{0}    & \revCell{96.59} & \revCell{88.13} & \revCell{73.58} & \revCell{91.04} & \revCell{\textbf{99.98}} & \revCell{\textbf{96.72}} & \revCell{78.49} \\
            & \revCell{0.02} & \revCell{96.51} & \revCell{88.01} & \revCell{73.60} & \revCell{90.68} & \revCell{\textbf{99.98}} & \revCell{96.37} & \revCell{78.31} \\
            & \revCell{0.1}  & \revCell{\textbf{96.71}} & \revCell{\textbf{89.07}} & \revCell{\textbf{74.21}} & \revCell{\textbf{91.19}} & \revCell{\textbf{99.98}} & \revCell{96.56} & \revCell{\textbf{79.38}} \\
            & \revCell{0.3}  & \revCell{96.16} & \revCell{88.37} & \revCell{73.64} & \revCell{90.57} & \revCell{\textbf{99.98}} & \revCell{96.55} & \revCell{78.46} \\
    \midrule
    \revCell{\multirow{4}{*}{\makecell[c]{ViT final\\CLS token}}}
            & \revCell{0}    & \revCell{96.19} & \revCell{87.54} & \revCell{73.07} & \revCell{90.18} & \revCell{\textbf{99.98}} & \revCell{96.41} & \revCell{77.55} \\
            & \revCell{0.02} & \revCell{96.23} & \revCell{\textbf{88.11}} & \revCell{\textbf{73.86}} & \revCell{90.15} & \revCell{\textbf{99.98}} & \revCell{96.53} & \revCell{\textbf{78.21}} \\
            & \revCell{0.1}  & \revCell{\textbf{96.52}} & \revCell{87.99} & \revCell{73.37} & \revCell{\textbf{90.84}} & \revCell{\textbf{99.98}} & \revCell{96.48} & \revCell{78.20} \\
            & \revCell{0.3}  & \revCell{96.39} & \revCell{88.05} & \revCell{73.69} & \revCell{90.32} & \revCell{\textbf{99.98}} & \revCell{\textbf{96.56}} & \revCell{78.18} \\
    \midrule
    \revCell{\multirow{4}{*}{\makecell[c]{ViT encoder\\layer 10 output}}}
            & \revCell{0}    & \revCell{96.30} & \revCell{87.32} & \revCell{73.04} & \revCell{90.24} & \revCell{\textbf{99.98}} & \revCell{\textbf{96.48}} & \revCell{77.50} \\
            & \revCell{0.02} & \revCell{96.01} & \revCell{87.49} & \revCell{73.06} & \revCell{90.02} & \revCell{\textbf{99.98}} & \revCell{96.39} & \revCell{77.44} \\
            & \revCell{0.1}  & \revCell{96.29} & \revCell{87.80} & \revCell{73.45} & \revCell{90.32} & \revCell{\textbf{99.98}} & \revCell{96.34} & \revCell{77.96} \\
            & \revCell{0.3}  & \revCell{\textbf{96.48}} & \revCell{\textbf{88.40}} & \revCell{\textbf{73.93}} & \revCell{\textbf{90.86}} & \revCell{\textbf{99.98}} & \revCell{96.46} & \revCell{\textbf{78.67}} \\
    \bottomrule
    \end{tabular*}
    \vspace{-4pt}
\end{table}\par
\rev{
Tables~\ref{tab:navsim_resnet34} and~\ref{tab:navsim_vit} demonstrate that FR-PT improves the matched BP reference at most evaluated reconstruction locations. For example, PDMS increases from $81.07$ ($c_{\mathrm{rec}}=0$) to $81.95$ ($c_{\mathrm{rec}}=0.02$) at \emph{fusion $z$} with ResNet-34, and from $78.49$ ($c_{\mathrm{rec}}=0$) to $79.38$ ($c_{\mathrm{rec}}=0.1$) at \emph{image embedding} with ViT. An exception is ResNet $z_4$, where the best reconstruction-weighted result ($80.98$) remains below the matched BP reference ($81.16$). Overall, the results support the applicability of FR-PT to modern CNN- and Transformer-based planners in large-scale real-world trajectory planning.\par
}
\subsection{\rev{Evaluation of Feature-Reconstruction Components}}
\subsubsection{\rev{Convolutional Reconstruction}}
\label{sec:conv_benchmark}
\revEnvBegin
\begin{figure}[tp]
    \centering
    \includegraphics[width=\linewidth]{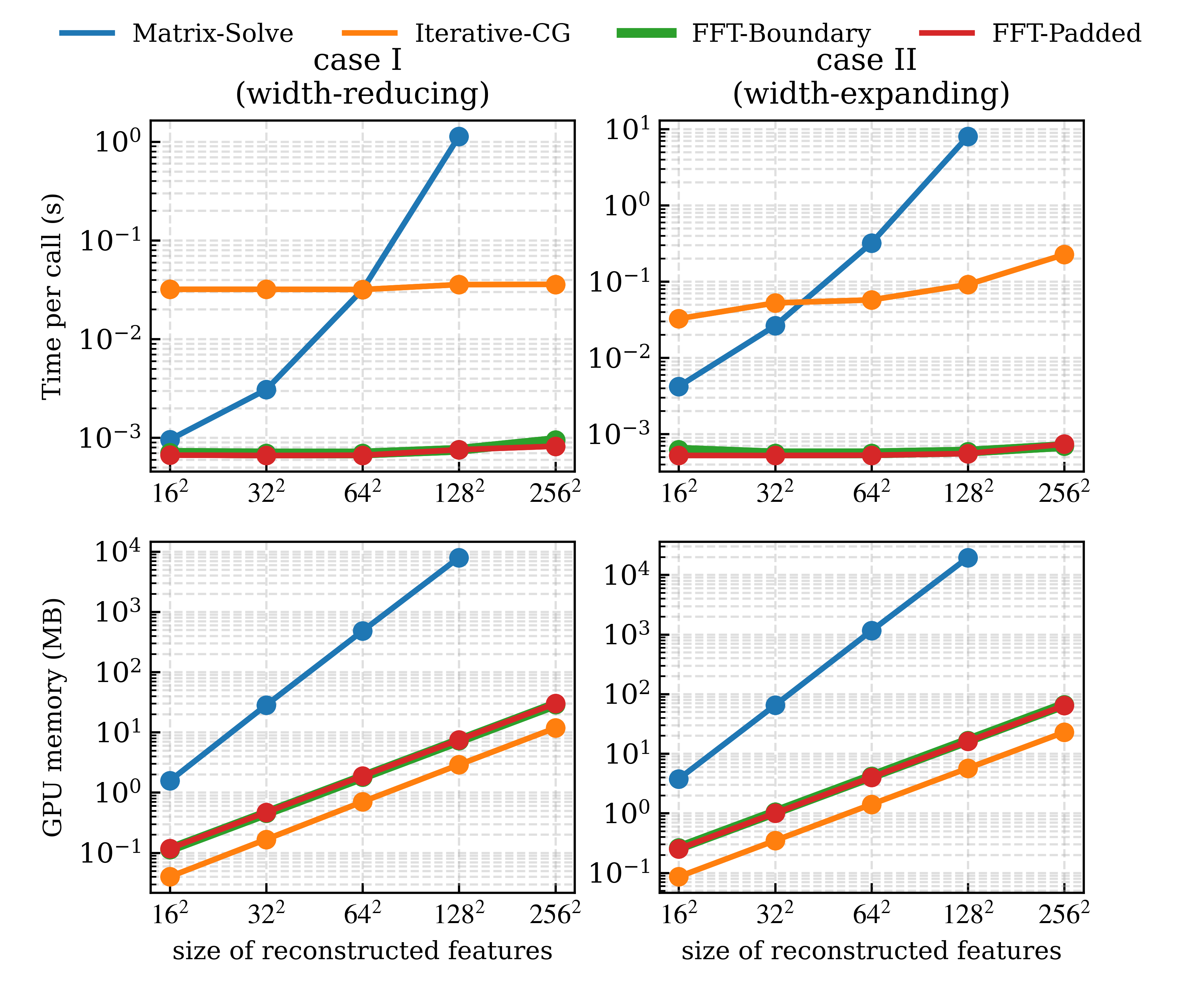}
    \caption{\rev{Runtime and GPU-memory scaling of four convolutional reconstruction solvers on an NVIDIA A100 GPU. Each point is averaged over 500 runs with batch size $2$, input channels $C_{\mathrm{in}}=2$, kernel size $5^2$, zero padding, and unit stride. No regularization check or fallback is applied. The left and right columns use $C_{\mathrm{out}}=1$ and $4$, representing the nominal Case~I and Case~II reconstruction geometries, respectively. Matrix-Solve runs out of memory at a spatial resolution of $256^2$.}}
    \label{fig:conv_reverse_efficiency}
\end{figure}
\revEnvEnd
\rev{We evaluate the four convolutional reconstruction solvers introduced in Sec.~\ref{conv_layer}: Matrix-Solve, Iterative-CG, FFT-Boundary, and FFT-Padded, in terms of computational scalability and reliability.\par
}
\rev{First, Fig.~\ref{fig:conv_reverse_efficiency} compares the runtime and GPU-memory scaling of the four convolutional reconstruction solvers with increasing spatial resolution. Matrix-Solve becomes impractical as resolution grows because it explicitly materializes the convolution matrix and performs dense solves. Iterative-CG avoids this matrix and has the smallest memory footprint among the scalable implementations, but repeated convolution/transposed-convolution products make it slower. The FFT formulations provide the best runtime scaling. In terms of GPU memory, FFT-based methods and Iterative-CG exhibit comparable growth rates with spatial resolution, indicating similar spatial-memory scalability.\par
}
\revEnvBegin
\begin{figure*}[tp]
    \centering
    \includegraphics[width=\linewidth]{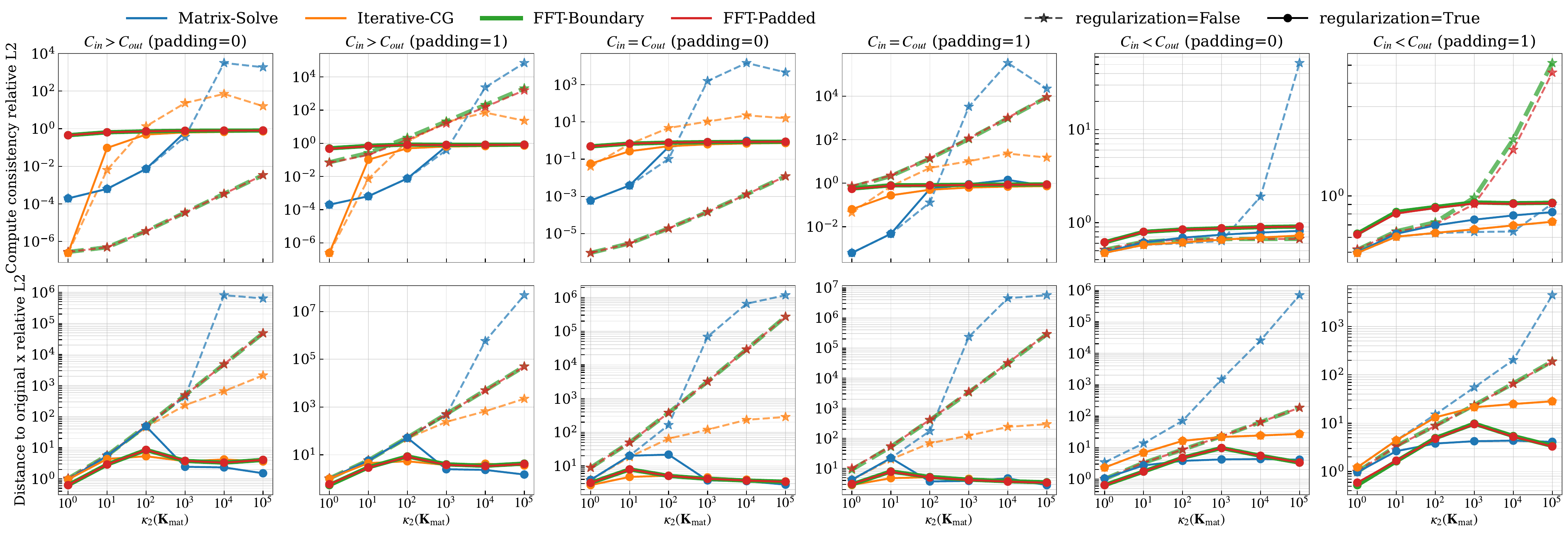}
    \caption{\rev{Numerical behavior of convolutional reconstruction under controlled kernel ill-conditioning. The kernel is reshaped as $\mathbf{K}_{\mathrm{mat}}\in\mathbb{R}^{C_{\mathrm{out}}\times(C_{\mathrm{in}}H_\mathbf{K}W_\mathbf{K})}$, with $\kappa_2(\mathbf{K}_{\mathrm{mat}})$ on the horizontal axis. Each point averages ten random seeds with batch size $2$, feature size $64^2$, kernel size $5^2$, $C_{\mathrm{in}}=8$, $C_{\mathrm{out}}\in\{4,8,16\}$, and padding $0$ or $1$. Two rows show relative computation-consistency error and relative deviation from the original feature, respectively. Dashed curves denote the nominal solver, while solid curves apply the reliability gate with MDP-centered Tikhonov regularization when needed.}}
    \label{fig:conv_reverse_condition}
\end{figure*}\par
\revEnvEnd
\rev{Second, Fig.~\ref{fig:conv_reverse_condition} examines numerical reliability under increasingly ill-conditioned convolution kernels. Without regularization, all four solvers can exhibit rapidly amplified feature deviation and deteriorated computation consistency as $\kappa_2(\mathbf{K}_{\mathrm{mat}})$ increases. Reliability-gated MDP-centered Tikhonov regularization effectively suppresses this instability, keeping both the CCP error and deviation from the original feature bounded across a wide range of condition numbers. This behavior agrees with the stability analysis in Thm.~\ref{thm:stable_regular}, showing that regularization improves numerical reliability while balancing computation consistency and deviation from the pretrained feature.\par
}
\rev{Particularly, the FFT-based solvers are stable after regularization. Across both reconstruction geometries and different padding settings, their CCP and MDP errors remain well controlled as the kernel becomes increasingly ill-conditioned and are generally close to the lowest errors among the four solvers. This suggests that the FFT formulations retain reliable reconstruction quality even under severe kernel ill-conditioning, while preserving their computational scalability.\par
}
\rev{To sum up, these experiments clarify the trade-offs among the four solvers. Matrix-Solve provides a useful exact reference at small scale but scales poorly in runtime and memory; Iterative-CG avoids explicit convolution matrices and is memory efficient, but its runtime and reconstruction accuracy depend on iterative convergence. The FFT-based solvers offer the most favorable balance between computational efficiency and numerical reliability. Accordingly, FFT-Padded is used for convolutional reconstruction in the main experiments.\par
}
\subsubsection{\rev{Output Vector Embedding}}
\label{sec:abl}
\begin{figure*}[tp]
    \centering
    \includegraphics[width=\linewidth]{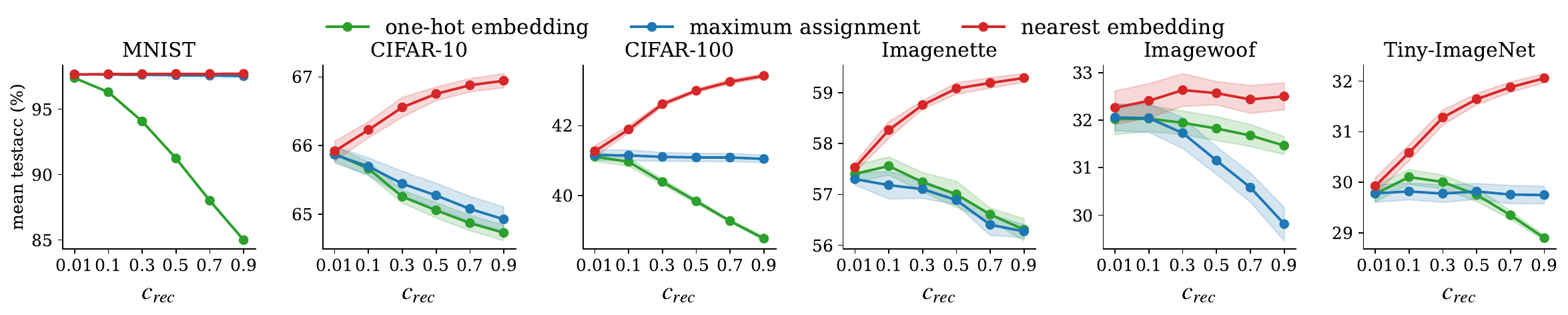}
    \caption{\rev{Mean test accuracy averaged over 10 post-training epochs, with $95\%$ confidence intervals over ten random seeds, for one-hot coding (OH), maximum assignment (MA), and nearest embedding (NE) across six classification datasets as $c_{\mathrm{rec}}$ varies.}}
    \label{fig:optimal}
\end{figure*}

\rev{We compare three label-to-vector embedding strategies: one-hot coding (OH), maximum assignment (MA), and nearest embedding (NE), by testing FR-PT with reconstruction location $l_R=\mathrm{out}$. Fig.~\ref{fig:optimal} reports Mean Acc as $c_{\mathrm{rec}}$ varies from $0.01$ to $0.9$. NE is generally the most robust choice, with accuracy remaining stable or improving as reconstruction supervision becomes stronger. In contrast, OH degrades markedly on the more challenging datasets, while MA is typically less stable. On MNIST, where the pretrained classifier is already near saturation, MA and NE often produce similar reconstructed output targets and therefore exhibit comparable stability. These results support the MDP motivation of modifying the forward output representation only as much as necessary to encode the ground-truth class.
}
\revEnvBegin
\par
\revEnvEnd
\subsection{\rev{Feature-Level Diagnostic Analysis}}
\subsubsection{\rev{Reconstruction Discrepancy and Model Ability}}
\begin{figure*}
    \centering
    \includegraphics[width=\linewidth]{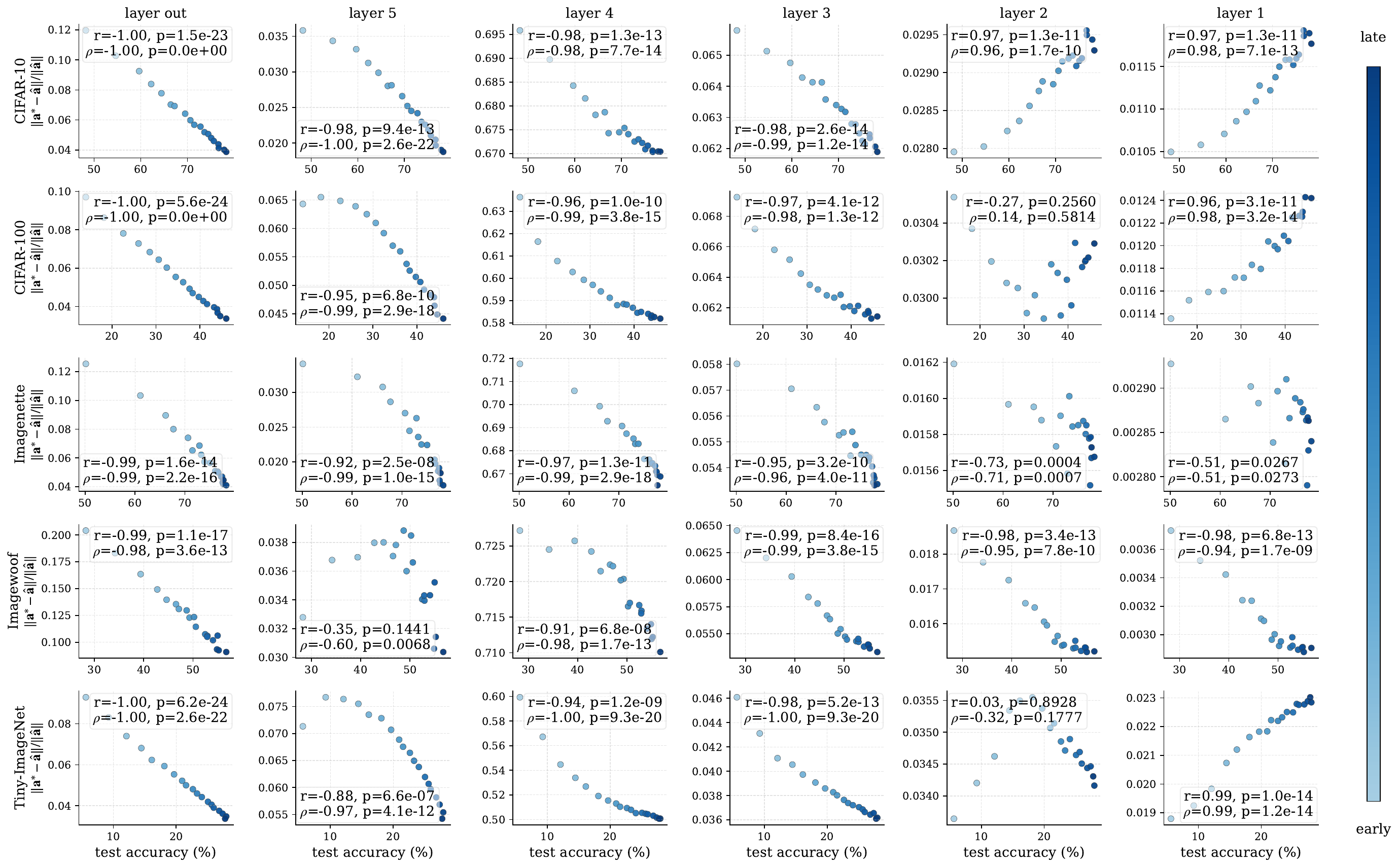}
    \caption{\rev{Relationship between test accuracy and relative reconstruction deviation during standard training. Each point corresponds to a checkpoint from the training trajectory of ResNet-18/34. Columns proceed from the output toward earlier reconstruction layers, and each panel reports Pearson $r$ and Spearman $\rho$ with their corresponding $p$-values.}}
    \label{fig:testacc_relerror}
\end{figure*}

\rev{This section examines how feature reconstruction discrepancy relates to model ability. For ResNet-18/34 models trained on five datasets for 100 epochs, we record the test accuracy and the relative reconstruction deviation $\|\mathbf{a}_{l}^{\ast}-\hat{\mathbf{a}}_{l}\|_2/\|\hat{\mathbf{a}}_{l}\|_2$ every five epochs. Figure~\ref{fig:testacc_relerror} plots their relationship across checkpoints along the training trajectory, with Pearson and Spearman correlations reported for each reconstruction layer.\par
}
\revEnvBegin
\revEnvEnd
\rev{A clear depth-dependent transition emerges in Fig.~\ref{fig:testacc_relerror}. At the output and nearby layers, reconstruction deviation is predominantly negatively correlated with test accuracy across datasets, indicating that better-trained checkpoints generally require smaller label-conditioned feature corrections. Toward earlier layers, however, this relation weakens and can even become positive, as observed at $l_R=1$ on CIFAR-10, CIFAR-100, and Tiny-ImageNet. This pattern aligns with the post-training results in Sec.~\ref{sec:comparison}, where the gain from reconstruction supervision at $l_R=1$ is often smaller than that at later reconstruction layers for ResNet-18/34 (Tables~\ref{tab:cifar10}, \ref{tab:cifar100}, and~\ref{tab:nette}). Importantly, a positive correlation does not imply that reconstruction supervision becomes harmful: early-layer targets may contain larger accumulated approximation errors while still retaining useful supervisory information about the desired forward representation. In summary, this feature-level diagnostic analysis suggests that reconstructed targets are generally more reliable near the output and become less stable as reconstruction proceeds toward earlier layers. This provides practical guidance for selecting $l_R$, favoring later reconstruction layers.\par
}
\subsubsection{\rev{Visualization of Reconstructed Feature Responses}}
\revEnvBegin
\begin{figure*}
     \centering
    \includegraphics[width=0.7\linewidth]{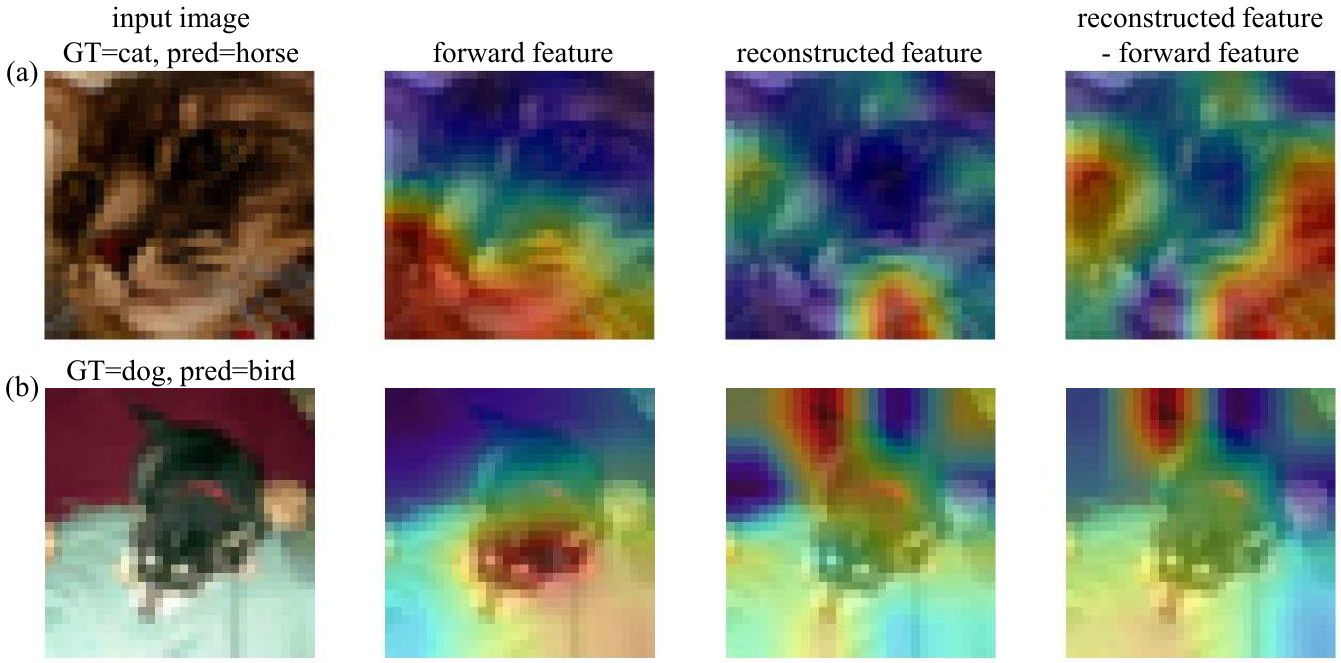}
    \caption{\rev{Grad-CAM visualization of two CIFAR-10 misclassifications using SimpleCNN at layer~3. The panels demonstrate spatial responses from the original forward feature, the label-conditioned reconstructed feature, and their difference.}}
    \label{fig:explain}
\end{figure*}
\revEnvEnd
\rev{Grad-CAM on the forward feature only shows the regions actually emphasized by the current model prediction, whereas Grad-CAM on the reconstructed feature reveals the regions that the frozen downstream mapping would favor for the ground-truth class. In the two CIFAR-10 misclassified examples, the reconstructed targets shift attention toward more class-relevant regions, such as the cat's right eye and whiskers in Fig.~\ref{fig:explain}(a) and the dog's tail in Fig.~\ref{fig:explain}(b). This provides an intuitive interpretability view of feature reconstruction: the reconstructed feature not only serves as supervision for post-training, but also exposes how the intermediate representation should be adjusted to support the correct prediction. Supplementary Part~C provides additional examples across datasets.\par
}
\subsection{\rev{Analysis of Favorable Conditions for FR-PT}}
\subsubsection{\rev{Pretraining Maturity}}
\label{sec:diffstage}
\begin{figure*}[tp]
    \centering
    \includegraphics[width=\linewidth]{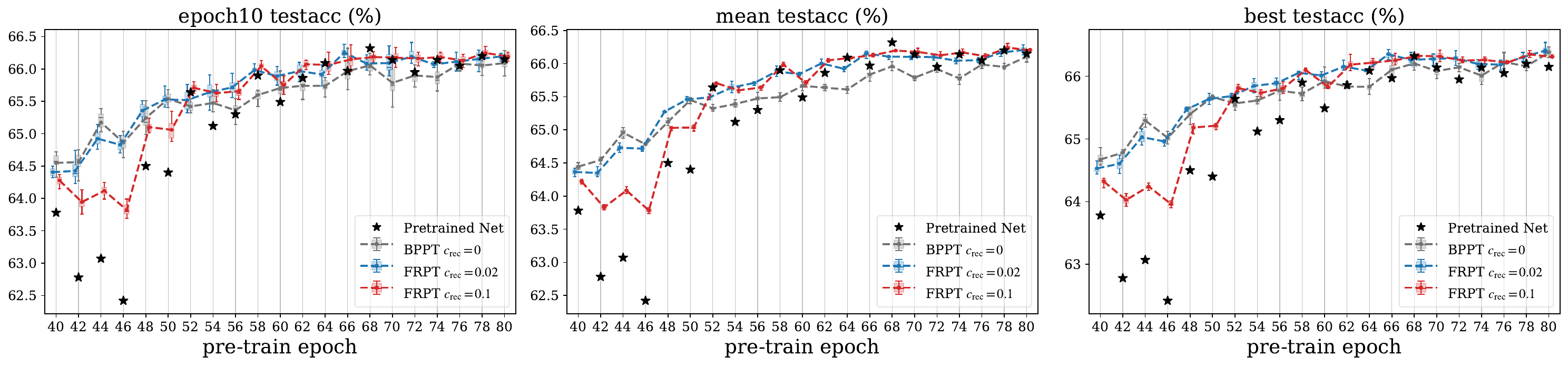}
    \caption{\rev{Ten-epoch post-training results from 21 CIFAR-10 SimpleCNN checkpoints saved between pretraining epochs 40 and 80. Stars indicate pretrained accuracy. Curves compare the matched BP reference ($c_{\mathrm{rec}}=0$) with FR-PT using $c_{\mathrm{rec}}=0.02$ and $0.1$ at $l_R=\mathrm{out}$. Each point is shown as a box plot over ten random seeds ($0,\ldots,9$).}}
    \label{fig:pretraining_stage}
\end{figure*}

\rev{To examine the effect of model maturity, we apply FR-PT to checkpoints sampled along the same CIFAR-10 SimpleCNN pretraining trajectory. Figure~\ref{fig:pretraining_stage} compares the matched BP reference with FR-PT at $c_{\mathrm{rec}}=0.02$ and $0.1$. At earlier training stages, reconstruction supervision is often neutral or detrimental, especially with the larger $c_{\mathrm{rec}}$. Around pretraining epoch 52, the checkpoint accuracy begins to plateau, indicating that the pretrained model is approaching convergence, from which FR-PT more frequently outperforms the matched reference in Epoch-10, Mean, and Best accuracy. 
}
\revEnvBegin
\revEnvEnd
\rev{This behavior agrees with the intended role of FR-PT. The reconstructed targets are defined with respect to the entire network: MDP favors features close to those produced by upstream modules, while CCP constrains them through the frozen downstream computation. Consequently, effective reconstruction supervision generally benefits from sufficiently mature representations and mappings throughout the pretrained network. The experiment does not imply a universal convergence threshold, but suggests that FR-PT is better viewed as a localized correction of a reasonably trained model than as a substitute for initial end-to-end training.
}
\subsubsection{\rev{Reconstruction Geometry}}
\label{sec:diffarch}
\rev{This section examines whether the Case~I/II reconstruction geometry is reflected in post-training behavior using two CIFAR-10 SimpleCNNs with the same overall topology, kernel sizes, pooling schedule, activation, and initialization scheme, but opposite convolutional channel profiles. The first architecture ($5\!\rightarrow\!10\!\rightarrow\!15$) is dominated by Case~II reconstruction along its reverse path, whereas the second ($15\!\rightarrow\!10\!\rightarrow\!5$) is dominated by Case~I. Nearest Embedding is used to construct the output target, and convolutional feature reconstruction is performed with FFT-Padded.
}
\revEnvBegin
\begin{figure*}
    \centering
    \includegraphics[width=\linewidth]{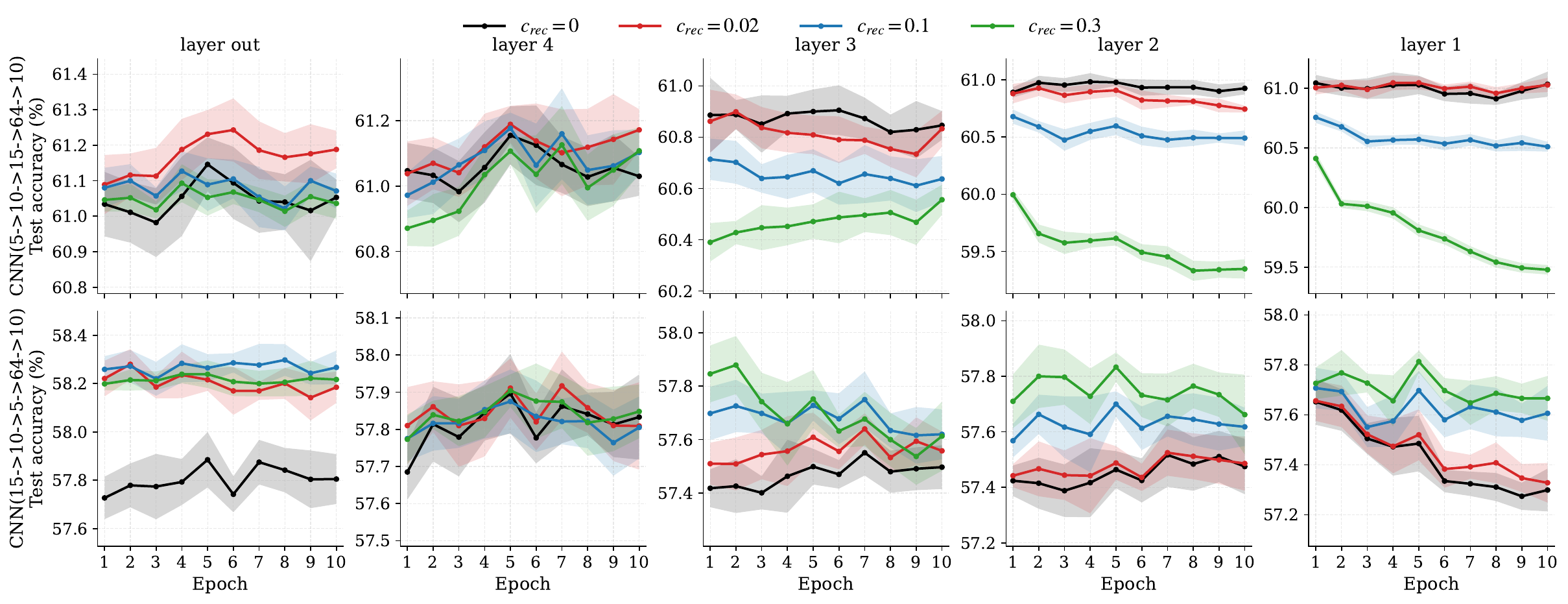}
    \caption{\rev{FR-PT on two CIFAR-10 SimpleCNN architectures with comparable scales but opposite convolutional channel profiles: $5\!\rightarrow\!10\!\rightarrow\!15$ (top) and $15\!\rightarrow\!10\!\rightarrow\!5$ (bottom). Both networks are followed by fully connected layers of dimensions $64$ and $10$.}}
    \label{fig:architecture_geometry}
\end{figure*}\par
\revEnvEnd
\rev{As shown in Fig.~\ref{fig:architecture_geometry}, FR-PT more readily outperforms the matched $c_{\mathrm{rec}}=0$ reference in the second architecture, with positive gains sustained across a broader range of reconstruction depths and coefficients. In contrast, the first architecture becomes more sensitive as reconstruction proceeds through Case~II-dominated steps. This agrees with the modeling expectation: Case~I can exactly satisfy CCP while MDP selects the closest preimage, whereas Case~II generally minimizes a nonzero least-squares CCP residual.
}
\section{Conclusion}
\rev{This work introduced FR-PT, a systematic localized post-training framework that reconstructs label-conditioned intermediate features through the frozen downstream computation to supervise upstream modules. FR is formulated under the CCP and MDP principles and instantiated with operator-specific algorithms, including FFT-based convolutional solvers, iterative reconstruction for composite residual and Transformer-style blocks, and nearest embedding for efficient classification target construction. The framework is further supported by theoretical analyses of MDP-centered Tikhonov regularization for ill-conditioned weight matrices and layer-wise error propagation, providing a principled understanding of numerical stability and accumulated reverse errors.
Extensive experiments on image-classification benchmarks show that FR-PT generally improve localized post-training over matched BP references across diverse architectures and reconstruction locations, while NAVSIM experiments demonstrate its applicability to CNN- and Transformer-based trajectory-planning models. Feature-diagnostic and favorable-condition analyses improve process transparency by revealing how reconstructed targets differ from the original representations and identifying the settings in which feature-level supervision is most effective.\par}
\rev{
FR-PT nevertheless has several practical limitations. Its gains are configuration dependent and may diminish for immature pretrained models or unreliable reverse paths. Feature reconstruction also introduces additional computation and storage relative to matched BP-based localized post-training.\par}
\rev{
Future work will investigate online FR-PT, where reconstruction targets are dynamically updated as the network evolves during post-training rather than being precomputed from the initial pretrained model. More principled selection of reconstruction locations and loss weights, together with repeated evaluation on larger-scale autonomous-driving systems, will also be important for improving the robustness and practical applicability of FR-PT.
}

\bibliographystyle{plain}
\bibliography{references}

\clearpage
\onecolumn
\setcounter{section}{0}
\setcounter{subsection}{0}
\setcounter{subsubsection}{0}
\setcounter{figure}{0}
\setcounter{table}{0}
\setcounter{equation}{0}
\renewcommand{\thesection}{\Alph{section}}
\renewcommand{\thesubsection}{\Alph{section}.\arabic{subsection}}
\renewcommand{\thesubsubsection}{\Alph{section}.\arabic{subsection}.\arabic{subsubsection}}
\renewcommand{\thefigure}{S\arabic{figure}}
\renewcommand{\thetable}{S\arabic{table}}
\renewcommand{\theequation}{S\arabic{equation}}
\section*{Supplementary Material}
\begin{center}
  {\large Hierarchical Feature-level Reverse Propagation for Post-Training Neural Networks}
\end{center}
\noindent This supplementary material provides the theoretical derivations, implementation details, statistical analyses, and extended feature-level diagnostics supporting the main manuscript. We derive the feature-reconstruction operators under the computation-consistency principle (CCP) and minimum-deviation principle (MDP), including reliability-gated MDP-centered Tikhonov regularization, FFT-based convolutional reconstruction, exact-solvability conditions for composite blocks, nearest embedding, and layer-wise reconstruction-error propagation. Detailed architectures, reconstruction hyperparameters, and paired multi-seed statistical analyses are provided for reproducibility. Extended Grad-CAM visualizations further compare forward and reconstructed feature responses across multiple classification datasets, illustrating how label-conditioned reconstruction can expose representation discrepancies and provide diagnostic feature-level supervision.
\par\noindent\textbf{Keywords:} Feature reconstruction, post-training, numerical stability, feature-level diagnosis.
\section{Mathematical Derivations and Theoretical Proofs}
\label{supp:sec:theory_derivations}
This part supplies the derivations and proofs supporting the reverse operators summarized in the main paper. The methodological roles of the computation-consistency principle (CCP) and minimum-deviation principle (MDP) are defined in the main paper and are recalled here only where needed to specify a numerical solver. We derive the nominal linear reconstruction, the reliability-gated MDP-centered Tikhonov fallback, the spatial and Fourier-domain convolution operators, the boundary-anchoring error, sufficient architectural conditions for exact CCP solutions, the output-space projection, and layer-wise error propagation. Throughout this part, vectors are column vectors, $\|\cdot\|_2$ denotes the Euclidean norm, $\|\cdot\|_F$ denotes the Frobenius norm, $\|\cdot\|_\infty$ denotes the infinity norm, and $(\cdot)^H$ denotes the Hermitian transpose.

\subsection{Linear Reconstruction: Closed-Form and Least-Squares Solutions}
\label{supp:sec:linear_mdp_kkt}
The following derivation applies to all feature reconstruction (FR) procedures reducible to linear subproblems, including affine-layer reconstruction, convolutional Matrix-Solve, frequency-wise FFT systems, and Gauss--Newton linearized subproblems for composite blocks. We use the generic model $\mathbf{A}\mathbf{x}=\mathbf{y}$, where $\mathbf{A}\in\mathbb{F}^{m\times n}$ is the linear transformation, $\mathbf{y}\in\mathbb{F}^{m}$ the bias-free target, $\mathbf{x}\in\mathbb{F}^{n}$ the reconstruction variable, and $\hat{\mathbf{x}}$ the original forward feature, with field $\mathbb{F}=\mathbb{R}$ or $\mathbb{C}$. Two nominal reconstruction geometries are distinguished according to dimensions $n$ and $m$.

\textbf{Case I (width-reducing): $n\ge m$.} When exact consistency with the target is achievable, MDP selects the closest preimage,
\begin{align}
    \min_{\vx}\quad
    &\frac{1}{2}\|\vx-\hat{\vx}\|_2^2
    \nonumber\\
    \mathrm{s.t.}\quad
    &\mA\vx=\vy.
    \label{supp:eq:supp_mdp_primal}
\end{align}
Assume first that $\mA$ has full row rank. The Lagrangian
\begin{equation}
    \mathcal{L}(\vx,\boldsymbol{\lambda})
    =\frac{1}{2}\|\vx-\hat{\vx}\|_2^2
    +\boldsymbol{\lambda}^{T}(\mA\vx-\vy)
\end{equation}
yields the stationarity and feasibility conditions
\begin{align}
    \frac{\partial \mathcal{L}}{\partial \vx}(\vx^{\ast}, \boldsymbol{\lambda}^{\ast})=\vx^{\ast}-\hat{\vx}+\mA^T\boldsymbol{\lambda}^{\ast}&=\mathbf{0},
    \label{supp:eq:supp_mdp_stationarity}\\
    \frac{\partial \mathcal{L}}{\partial \boldsymbol{\lambda}}(\vx^{\ast})=\mA\vx^{\ast}-\vy&=\mathbf{0}.
    \label{supp:eq:supp_mdp_feasibility}
\end{align}
Substituting Eq.~\ref{supp:eq:supp_mdp_stationarity} into Eq.~\ref{supp:eq:supp_mdp_feasibility} gives
\begin{equation} \label{supp:eq:gram}
    \mA\mA^T\boldsymbol{\lambda}^{\ast}
    =
    \mA\hat{\vx}-\vy,
\end{equation}
and therefore
\begin{equation}
    \vx^{\ast}=\hat{\vx}-\mA^{T}\boldsymbol{\lambda}^{\ast}
    =\hat{\vx}+\mA^T(\mA\mA^T)^{-1}
    (\vy-\mA\hat{\vx}).
    \label{supp:eq:supp_mdp_closed}
\end{equation}
The correction lies in $\operatorname{Range}(\mA^T)$ and is orthogonal to $\operatorname{Null}(\mA)$, so Eq.~\ref{supp:eq:supp_mdp_closed} is the unique feasible point closest to $\hat{\vx}$. In the complex Fourier case, transposes are replaced by Hermitian transposes. In practice, we first solve the system of linear equations in Eq.~\ref{supp:eq:gram} and then recover $\vx^{\ast}$ using Eq.~\ref{supp:eq:supp_mdp_closed}, thereby avoiding explicit matrix inversion.

\textbf{Case II (width-expanding): $n<m$.} The nominal CCP solution is
\begin{equation}
    \arg\min_{\vx}\|\mA\vx-\vy\|_2^2.
    \label{supp:eq:supp_caseII_ls}
\end{equation}
The code uses direct least squares. When $\mA$ is full column rank the minimizer is unique.

\subsection{Numerical Reliability and MDP-Centered Tikhonov Regularization}
\label{supp:sec:proof_tikhonov_stability}

In deep neural networks, severe ill-conditioning is often more relevant than exact rank deficiency: a weight matrix may be formally full rank yet contain singular values sufficiently close to zero to make inverse or least-squares reconstruction numerically unreliable. The reliability-gated regularization introduced below is designed to address such cases.

For a unified analysis, we use common notation in this section for the linear subproblems arising in FR through affine maps, FFT-based convolutional maps, and composite blocks. Specifically, $\mathbf{A}$ denotes the corresponding linear matrix, $\mathbf{y}$ the bias-free target, and $\hat{\mathbf{x}}$ and $\mathbf{x}^{\ast}$ the original and reconstructed features, respectively, with the nominal relation $\mathbf{y}=\mathbf{A}\mathbf{x}$.
\subsubsection{Reliability Check}
The reliability check considers (i) finiteness, (ii) the entrywise magnitude bound in Eq.~\ref{supp:eq:gate_bound}, (iii) the relative deviation from the original feature in Eq.~\ref{supp:eq:gate_dev}, and (iv) computation consistency, measured by Eq.~\ref{supp:eq:gate_res} for Case~I and Eq.~\ref{supp:eq:gate_opt} for Case~II:
\begin{equation}
    \|\mathbf{x}^{\ast}\|_{\infty} \le M_{\infty},
    \label{supp:eq:gate_bound}
\end{equation}
\begin{equation}
    \rho_{\mathrm{dev}}=
    \frac{\|\mathbf{x}^\ast-\hat{\mathbf{x}}\|_2}
    {\|\hat{\mathbf{x}}\|_2}
    \le M_{\mathrm{dev}}.
    \label{supp:eq:gate_dev}
\end{equation}
\begin{subequations}\label{supp:eq:gate_consistency}
\begin{align}
    \rho_{1}&=
    \frac{\|\mA\vx-\vy\|_2}
    {\|\mA\|_F\|\vx\|_2+\|\vy\|_2}
    \le M_{1},
    &&\text{(Case I)},
    \label{supp:eq:gate_res}\\
    \rho_{2}&=
    \frac{\|\mA^H(\mA\vx-\vy)\|_2}
    {\|\mA\|_F(\|\mA\|_F\|\vx\|_2+\|\vy\|_2)}
    \le M_{2},
    &&\text{(Case II)}.
    \label{supp:eq:gate_opt}
\end{align}
\end{subequations}
Here, $M_{\infty}$, $M_{\mathrm{dev}}$, $M_1$, and $M_2$ are the corresponding reliability thresholds.

\subsubsection{Regularized Model}
For an unreliable linear subproblem, the code solves
\begin{equation}
    \min_{\vx}
    \|\mA\vx-\vy\|_2^2+\alpha\|\vx-\hat{\vx}\|_2^2,
    \qquad \alpha>0,
    \label{supp:eq:supp_tikhonov_model}
\end{equation}
which is strongly convex because $\alpha>0$, so there exists a unique minimizer $\vx_{\alpha}$ even for rank-deficient $\mA$.
With $\boldsymbol{\delta}=\vx-\hat{\vx}$ and $\vr=\vy-\mA\hat{\vx}$,
\begin{equation}
    \boldsymbol{\delta}_{\alpha}
    =
    \arg\min_{\boldsymbol{\delta}}
    \|\mA\boldsymbol{\delta}-\vr\|_2^2
    +
    \alpha\|\boldsymbol{\delta}\|_2^2,
    \qquad
    \vx_{\alpha}=\hat{\vx}+\boldsymbol{\delta}_{\alpha},
    \label{supp:eq:supp_tikhonov_correction}
\end{equation}
Centering the penalty at $\hat{\vx}$ does not weaken the numerical regularization. It only changes the prior from ``the reconstructed feature should be close to zero'' to the method-consistent prior ``the reconstructed feature should remain close to its former forward value.''

Setting the gradient of Eq.~\eqref{supp:eq:supp_tikhonov_correction} to zero gives
\begin{equation}
    (\mA^H\mA+\alpha\mathbb{I})\boldsymbol{\delta}_{\alpha}
    =
    \mA^H\vr,
    \label{supp:eq:supp_tikhonov_normal}
\end{equation}
Thus, the equivalent dual correction is
\begin{equation}
    \boldsymbol{\delta}_{\alpha}
    =(\mA^H\mA+\alpha\mathbb{I})^{-1}\mA^H\vr.
    \label{supp:eq:supp_tikhonov_dual}
\end{equation}
To obtain $\boldsymbol{\delta}_{\alpha}$, the dense linear-layer fallback is computed through an SVD filter, while the large explicit-convolution and matrix-free CG implementations use algebraically equivalent Gram/normal systems to avoid an SVD of the full convolution operator.

\begin{suppTheorem}[MDP-centered Tikhonov stability]
\label{supp:thm:supp_tikhonov_stability}
Let $\mA\in\mathbb{C}^{m\times n}$ and let $\vx_{\alpha}$ be the unique minimizer of Eq.~\eqref{supp:eq:supp_tikhonov_model} for $\alpha>0$. If $\mA=\mU\mSigma\mV^H$ is a compact SVD over its nonzero singular directions, then the correction is
\begin{equation}
    \vx_{\alpha}-\hat{\vx}
    =\mV\operatorname{diag}\!\left(\frac{\sigma_i}{\sigma_i^2+\alpha}\right)
      \mU^H(\vy-\mA\hat{\vx}).
    \label{supp:eq:supp_tikhonov_svd}
\end{equation}
Moreover,
\begin{align}
    \|\vx_{\alpha}-\hat{\vx}\|_2
    &\le \frac{1}{2\sqrt{\alpha}}\|\vy-\mA\hat{\vx}\|_2,
    \label{supp:eq:supp_tikhonov_correction_bound}\\
    \|\vx_{\alpha}(\vy+\boldsymbol{\xi})-\vx_{\alpha}(\vy)\|_2
    &\le \frac{1}{2\sqrt{\alpha}}\|\boldsymbol{\xi}\|_2.
    \label{supp:eq:supp_tikhonov_stability}
\end{align}
The Hermitian matrix $\mA^H\mA+\alpha\mathbb{I}$ is positive definite and
\begin{equation}
    \kappa_2(\mA^H\mA+\alpha\mathbb{I})
    =\frac{\sigma_{\max}^2+\alpha}{\lambda_{\min}(\mA^H\mA)+\alpha}
    \le 1+\frac{\sigma_{\max}^2}{\alpha},
    \label{supp:eq:supp_tikhonov_condition}
\end{equation}
where  $\kappa_2(\cdot)$ denotes the matrix condition number induced by the $2$-norm, $\sigma_{\max}$ is the largest singular value of $\mathbf{A}$ and $\lambda_{\min}(\mathbf{A}^H\mathbf{A})$ is the smallest eigenvalue of $\mathbf{A}^H\mathbf{A}$. All statements remain valid when $\mA$ is rank deficient.
\end{suppTheorem}

\begin{proof} Utilizing the compact SVD form of $\mA$ yields
\begin{equation}
    (\mA^H\mA+\alpha\mI)^{-1}\mA^H = \mV \operatorname{diag}\!\left( \frac{\sigma_i}{\sigma_i^2+\alpha} \right) \mU^H.
    \label{supp:eq:svdsubtitute}
\end{equation}
Thus Eq.~\ref{supp:eq:supp_tikhonov_dual} immediately gives Eq.~\eqref{supp:eq:supp_tikhonov_svd}; null-space components of the correction are zero.

The scalar filter $f_{\alpha}(\sigma)=\sigma/(\sigma^2+\alpha)$ reaches its maximum at $\sigma=\sqrt{\alpha}$, where $f_{\alpha}=1/(2\sqrt{\alpha})$. Applying this operator-norm bound proves the correction bound Eq.~\ref{supp:eq:supp_tikhonov_correction_bound}.
\begin{equation}
    \|\vx_\alpha-\hat{\vx}\|_2 \le \left\| \operatorname{diag}\!\left( \frac{\sigma_i}{\sigma_i^2+\alpha} \right) \right\|_2 \|\vy-\mA\hat{\vx}\|_2
    = \max_i \frac{\sigma_i}{\sigma_i^2+\alpha} \|\vy-\mA\hat{\vx}\|_2
    \le \frac{1}{2\sqrt{\alpha}}\|\vy-\mA\hat{\vx}\|_2
    \notag
\end{equation}

Replacing $\vy$ by $\vy+\boldsymbol{\xi}$ in Eq.~\ref{supp:eq:supp_tikhonov_svd} and subtracting the two solutions yields the same filtered operator Eq.~\ref{supp:eq:svdsubtitute} applied to $\boldsymbol{\xi}$, proving target stability Eq.~\ref{supp:eq:supp_tikhonov_stability}.

Finally, the eigenvalues of $\mA^H\mA+\alpha\mathbb{I}$ are $\lambda_i(\mA^H\mA)+\alpha>0$, which gives Eq.~\eqref{supp:eq:supp_tikhonov_condition} and the stated upper bound.
\end{proof}
The corresponding CCP residual is also explicit. Let $\{\mathbf{u}_i\}_{i=1}^{r}$ be the left singular vectors spanning
$\operatorname{Range}(\mA)$. Decompose
$\vr=\vy-\mA\hat{\vx}$ as
\begin{equation}
    \vr
    =
    \sum_{i=1}^{r}c_i\mathbf{u}_i+\vr_{\perp},
    \qquad
    \vr_{\perp}\perp\operatorname{Range}(\mA),
    \qquad
    c_i=\mathbf{u}_i^H\vr,
    \notag
\end{equation}
then
\begin{equation}
    \vy-\mA\vx_{\alpha}
    =
    \vr_{\perp}
    +
    \sum_{i=1}^{r}
    \frac{\alpha}{\sigma_i^2+\alpha}
    c_i\mathbf{u}_i.
    \notag
\end{equation}
Thus, residual components along weakly observable singular directions
are smoothly retained rather than being abruptly truncated.

\subsubsection{Choice of the Tikhonov Coefficient}
\label{supp:sec:supp_alpha_selection}
The final code uses one coefficient evaluation per failed subproblem; it does not perform geometric regularization trials. When $\alpha$ is not supplied explicitly, the spectral term is
\begin{equation}
    \alpha_{\mathrm{spec}}
    =\left(\frac{\sigma_{\max}}{\kappa_{\mathrm{eff}}}\right)^2,
    \qquad \kappa_{\mathrm{eff}}=10^3,
    \label{supp:eq:supp_alpha_selection}
\end{equation}
with a machine-precision floor
\begin{equation}
    \alpha_{\mathrm{floor}}
    =\epsilon_{\mathrm{mach}}\max\{\sigma_{\max}^2,1\}.
    \label{supp:eq:supp_alpha_floor}
\end{equation}
The corresponding inverse gain is bounded by
\begin{equation}
    \max_{\sigma\ge0}\frac{\sigma}{\sigma^2+\alpha_{\mathrm{spec}}}
    =\frac{\kappa_{\mathrm{eff}}}{2\sigma_{\max}}.
    \notag
\end{equation}
Moreover, Eq.~\ref{supp:eq:supp_tikhonov_correction} can be written as the augmented least-squares system
\begin{equation}
    \min_{\boldsymbol{\delta}}
    \left\|
    \begin{bmatrix}
        \mA\\
        \sqrt{\alpha}\mI
    \end{bmatrix}
    \boldsymbol{\delta}
    -
    \begin{bmatrix}
        \vr\\
        \mathbf{0}
    \end{bmatrix}
    \right\|_2^2.
    \notag
\end{equation}
Let
$\widetilde{\mA}_{\alpha}
=
[\mA^T,\sqrt{\alpha}\mI]^T$
denote the augmented Tikhonov operator. Since
$\widetilde{\mA}_{\alpha}^{H}\widetilde{\mA}_{\alpha}
=\mA^H\mA+\alpha\mI$, its condition number is
\begin{equation}
    \kappa_2(\widetilde{\mA}_{\alpha})
    =
    \sqrt{
    \frac{\sigma_{\max}^2+\alpha}
    {\lambda_{\min}(\mA^H\mA)+\alpha}
    }
    \le \sqrt{1+\frac{\sigma_{\max}^2}{\alpha}},
    \notag
\end{equation}
Therefore, choosing
$\alpha\ge(\sigma_{\max}/\kappa_{\mathrm{eff}})^2$
gives the condition number of the linear operator in MDP-centered Tikhonov problem $\kappa_2(\widetilde{\mA}_{\alpha}) \le \sqrt{1+\kappa_{\mathrm{eff}}^2}$ even in the singular limit case $\lambda_{\min}(\mA^H\mA)\rightarrow0$.

The implementation additionally uses Theorem~\ref{supp:thm:supp_tikhonov_stability} to obtain a sufficient feature-magnitude lower bound $\alpha_{\mathrm{mag}}$.
\begin{equation}
    D_{\max}=M_{\infty}-\|\hat{\vx}\|_{\infty},\qquad
    \alpha_{\mathrm{mag}}
    =
    \begin{cases}
    \left(\frac{\|\vr\|_2}{2D_{\max}}\right)^2, &D_{\max}>0 \\
    0, &\text{otherwise}
    \end{cases}
    \label{supp:eq:supp_alpha_deviation_bound}
\end{equation}
For $D_{\max}>0$, since $\|\boldsymbol{\delta}_{\alpha}\|_{\infty}\le\|\boldsymbol{\delta}_{\alpha}\|_2\le\frac{1}{2\sqrt{\alpha}}\|\vr\|_2$, this is sufficient to keep every entry of the correction within the remaining magnitude margin as follows:
\begin{equation}
    \|\vx_{\alpha}\|_{\infty}\le\|\hat{\vx}\|_{\infty} + \|\boldsymbol{\delta}_{\alpha}\|_{\infty}\le\|\hat{\vx}\|_{\infty}+\frac{1}{2\sqrt{\alpha}}\|\vr\|_2\le\|\hat{\vx}\|_{\infty}+\frac{1}{2\sqrt{\alpha_{\mathrm{mag}}}}\|\vr\|_2=M_{\infty}, \quad \forall \alpha\ge\alpha_{\mathrm{mag}}.
    \notag
\end{equation}
For $D_{\max}\le 0$, this bound is invalid. Consequently, the dense SVD fallback uses
\begin{equation}
    \alpha=\max\{\alpha_{\mathrm{spec}},\alpha_{\mathrm{floor}},\alpha_{\mathrm{mag}}\}.
    \label{supp:eq:supp_alpha_combined}
\end{equation}
For explicit matrix and CG convolution reconstruction, $\sigma_{\max}$ is replaced by a safe Frobenius upper bound or a power-iteration estimate, respectively, while the same magnitude lower bound is retained.

\subsubsection{Implementation for matrix, convolutional, and composite operators}
Table~\ref{supp:tab:supp_regularized_solver_policy} summarizes the implementation. The same objective in Eq.~\eqref{supp:eq:supp_tikhonov_model} is used in every linear fallback, although the algebraic system is chosen to reduce computation and memory.
\begin{table}[H]
    \centering
    \small
    \caption{Reliability-gated reverse solvers and MDP-centered fallbacks.}
    \label{supp:tab:supp_regularized_solver_policy}
    \begin{tabular}{
        p{0.14\linewidth}
        p{0.18\linewidth}
        p{0.27\linewidth}
        p{0.31\linewidth}}
        \toprule
        Reverse module & Method & Nominal solver & Fallback after an unreliable result \\
        \midrule
        Linear (Case I)
        & Linear modeling
        & Gram/KKT nearest exact solve
        & Per-sample SVD Tikhonov correction \\

        Linear (Case II)
        & Linear modeling
        & Direct least squares (\texttt{gels})
        & Per-sample SVD Tikhonov correction \\

        Convolution
        & Matrix-Solve
        & Gram/KKT nearest exact solve or direct least squares
        & Damped normal-equation correction about $\hat{\vx}$\\

        Convolution
        & Iterative-CG
        & CG on $\mA\mA^T$ or $\mA^T\mA$
        & Damped CG with MDP center \\

        Convolution
        & FFT-Boundary or FFT-padded
        & Frequency-wise linear systems or least squares
        & Tikhonov only for failed frequency--batch pairs \\

        Composite block
        & Explicit-Jacobian
        & Gauss--Newton least-squares step
        & MDP-centered local Tikhonov + backtracking \\

        Composite block
        & Matrix-free VJP
        & Relative-CCP Adam optimization
        & MDP-centered projection; return anchor if final check fails \\
        \bottomrule
    \end{tabular}
\end{table}
\subsection{FFT-Based Feature Reconstruction for Convolution}
\label{supp:sec:fft_conv_derivation}
We follow the notation in Sec.~III.B of the main paper. For a single sample $x$, let the activated forward feature at layer $l$ be
$\va_l \in \R^{C_l\times H_l\times W_l}$,
and the reconstructed target pre-activation at layer $l+1$ be
$\vz_{l+1}^{\ast}\in \R^{C_{l+1}\times H_{l+1}\times W_{l+1}}$.
The convolution kernel is
$\mathbf{K}_l\in\mathbb{R}^{C_{l+1}\times C_l\times H_K\times W_K}$,
where $\mathbf{K}_l[n,m]$ denotes the spatial kernel mapping input channel $m$ to output channel $n$. We absorb the bias into the target and define the bias-free target as
$\vy_l^{\ast}=\vz_{l+1}^{\ast}-\vb_l$. The reconstructed feature $\va_{l}^{\ast}$ at layer $l$ is pursued.

In this section, $m$ and $n$ denote the indices of the input and output channels, respectively. The indices $i$ and $j$ denote spatial row and column positions, while $u$ and $v$ represent frequency indices. The following derivation considers the FFT implementations used in this work, for which stride and dilation are both one and grouped convolution is not used.
\subsubsection{Detailed Derivation of Two FFT-Based Methods}

For spatial padding $p$, the linear part of the CNN cross-correlation is
\begin{equation}
    [\mathcal{A}_p(\va_l)]_{n}(i,j)
    =
    \sum_{m=1}^{C_l}
    \sum_{s=0}^{H_K-1}\sum_{t=0}^{W_K-1}
    \mathbf{K}_l[n,m](s,t)
    \va_l[m](i+s-p,j+t-p),
    \label{supp:eq:supp_exact_cnn_conv}
\end{equation}
where samples outside $[0,H_l-1]\times[0,W_l-1]$ are defined as zero. Hence, reconstructing $\va_l^{\ast}$ under CCP corresponds to matching
$\mathcal{A}_p(\va_l^{\ast})$ to $\vy_l^{\ast}$.

Direct vectorization of Eq.~\eqref{supp:eq:supp_exact_cnn_conv} yields a large structured linear system that couples channel and spatial variables. Instead, the spatial convolution can be diagonalized in the Fourier basis. Since the forward operation implemented by deep-learning libraries is cross-correlation, we first spatially flip the kernel,
\begin{equation}
    \widetilde{\mathbf{K}}_l[n,m](s,t)
    =
    \mathbf{K}_l[n,m](H_K-1-s,W_K-1-t).
    \label{supp:eq:supp_flipped_kernel}
\end{equation}
On an $H\times W$ Fourier grid, zero-padding
$\widetilde{\mathbf{K}}_l[n,m]$ at its right and bottom sides gives the circular-convolution representation
\begin{equation}
    [\widetilde{\mathcal{A}}(\va_l)]_{n}(i,j)
    =
    \sum_{m=1}^{C_l}
    \widetilde{\mathbf{K}}_l[n,m]\circledast \va_l[m]
    =\sum_{m=1}^{C_l}
    \sum_{s=0}^{H_{K}-1}\sum_{t=0}^{W_{K}-1}
    \widetilde{\mathbf{K}}_{l}[n,m](s,t)\va_l[m]((i-s)\bmod H_{l},(j-t)\bmod W_{l}), (i,j)\in\Omega_{l}.
    \label{supp:eq:supp_circular_operator}
\end{equation}
By the two-dimensional circular convolution theorem,
\begin{equation}
    \mathcal{F}(f\circledast g)
    =
    \mathcal{F}(f)\odot\mathcal{F}(g),
\end{equation}

the spatial variables are decoupled in the Fourier domain. For every frequency $(u,v)$, define the channel-wise feature spectrum by
\begin{equation}
    \mathcal{F}(\widetilde{\mathcal{A}}(\va_l)[n])(u,v)
    =\sum_{m=1}^{C_l}
    \mathcal{F}(\mathcal{Z}^{(rb)}\widetilde{\mathbf{K}}_l[m,n])(u,v)\mathcal{F}(\va_l[m])(u,v),
    \quad (u,v)\in\Omega_{l}.
    \label{supp:eq:supp_frequency_decoupling}
\end{equation}
where $\mathcal{Z}^{(rb)}$ zero-pads the flipped kernel on the right and bottom sides to the same size as $\va_l$. The DFT decouples the spatial positions, so that each frequency $(u,v)$ can be treated independently. For each spatial frequency $(u,v)\in\Omega_l$, stack the Fourier coefficients over channels as
\begin{align}
    \mathcal{F}_{u,v}(\va_l)
    &=
    [\mathcal{F}(\va_l[1])(u,v),\ldots,
    \mathcal{F}(\va_l[C_l])(u,v)]^T
    \in \mathbb{C}^{C_l}, \\
    \mathcal{F}_{u,v}(\widetilde{\mathcal{A}}(\va_l))
    &=
    [\mathcal{F}(\widetilde{\mathcal{A}}(\va_l)[1])(u,v),\ldots,
    \mathcal{F}(\widetilde{\mathcal{A}}(\va_l)[C_{l+1}])(u,v)]^T
    \in \mathbb{C}^{C_{l+1}}
\end{align}
where an output-sized tensor is embedded at the locations corresponding to the spatial cross-correlation output and zero-padded to the $H_l\times W_l$ grid before the DFT. The channel-to-channel frequency response is
\begin{equation}
    [\mathbf{K}^{F}_{u,v}]_{n,m}
    =
    \mathcal{F}(\mathcal{Z}^{(aux)}\widetilde{\mathbf{K}}_l[n,m])(u,v),
    \quad
    \mathbf{K}^{F}_{u,v}\in\mathbb{C}^{C_{l+1}\times C_l}, \quad(u,v)\in\Omega_{l}.
    \label{supp:eq:supp_transfer_matrix}
\end{equation}
Circular convolution is therefore reduced to:
\begin{equation}
    \mathcal{F}_{u,v}(\widetilde{\mathcal{A}}(\va_l))
    =
    \mathbf{K}^{F}_{u,v}\cdot\mathcal{F}_{u,v}(\va_l),
    \quad (u,v)\in\Omega_{l}.
    \label{supp:eq:supp_circular_matrix}
\end{equation}
Thus, rather than solving one large spatial system, the FFT representation decomposes the convolution into $H_lW_l$ independent systems, each involving only $C_l$ unknown Fourier coefficients. The remaining difference from Eq.~\eqref{supp:eq:supp_exact_cnn_conv} is the boundary condition: circular convolution wraps samples across the feature boundary, whereas CNN convolution uses zero padding. We employ two different treatments of this boundary discrepancy.

\textbf{(1) FFT-padded.}
FFT-padded eliminates circular wrap-around in the retained output by explicitly padding the input feature before applying the FFT. Let $\mathcal{Z}_p$ zero-pad $\va_l$ by $p$ entries on all four spatial sides, resulting in a grid
\begin{equation}
    \Omega_l'
    =
    [0,H_l+2p-1]\times[0,W_l+2p-1].
    \notag
\end{equation}
The flipped kernel is zero-padded to the same grid when constructing
$\mathbf{K}^{F}_{u,v}$ by Eq.~\ref{supp:eq:supp_transfer_matrix}.
Let $\mathcal{S}_p$ select the bottom-right
$H_{l+1}\times W_{l+1}$ block of the inverse-DFT result. Then

\begin{equation}
    \mathcal{A}_p(\va_l)
    = \mathcal{S}_p \widetilde{\mathcal{A}}(\va_l)=
    \mathcal{S}_p\mathcal{F}^{-1}
    \left([\mathbf{K}^{F}_{u,v} \cdot
    \mathcal{F}_{u,v}(\mathcal{Z}_p\va_l)]_{(u,v)\in\Omega_{l}^{\prime}}
    \right),
    \label{supp:eq:supp_fft_padded_operator}
\end{equation}
where the multiplication by $\mathbf{K}^{F}$ is performed independently at each frequency position $(u,v)$ on the padded grid. Because the retained block contains only samples whose corresponding linear convolution is supported by the explicitly zero-padded feature, Eq.~\eqref{supp:eq:supp_fft_padded_operator} exactly reproduces the CNN cross-correlation in Eq.~\eqref{supp:eq:supp_exact_cnn_conv}. Listing~\ref{supp:lst:fft_padded_python} verifies this identity numerically.

\begin{lstlisting}[style=pythonstyle,caption={PyTorch verification code for the FFT-padded convolution operator.},label={supp:lst:fft_padded_python}]
import torch
torch.manual_seed(42)

def FFT_padded_conv(x, kernel, PADDING):
    """Verify the correctness of the FFT-padded method."""
    y_size = (x.shape[-2] + 2 * PADDING - kernel.shape[-2] + 1, \
                x.shape[-1] + 2 * PADDING - kernel.shape[-1] + 1)
    pad_x = torch.nn.ZeroPad2d((PADDING, PADDING, PADDING, PADDING))
    x_padded = pad_x(x.to(torch.float64))
    fft_x = torch.fft.fft2(x_padded, dim=[-2, -1])
    kernel_flip = torch.flip(kernel, dims=[-2, -1])
    # zero-pad kernel_flip at bottom and right by default
    fft_kernel_flip = torch.fft.fft2(kernel_flip, s=fft_x.shape[-2:], dim=[-2, -1])
    other_fft_y = torch.einsum('bcuv,dcuv->bduv', fft_x, fft_kernel_flip)
    other_y = torch.fft.ifft2(other_fft_y)[:,:,-y_size[-2]:, -y_size[-1]:]
    return other_y.real

if __name__ == '__main__':
    X_SHAPE = (2, 2, 16, 16)  # (BS, C1, H1, W1)
    KERNEL_SHAPE = (4, X_SHAPE[1], 5, 5) # (C2, C1, kernel_H, kernel_W)
    PADDING:int = 2
    assert 2*PADDING <= min(KERNEL_SHAPE[-2], KERNEL_SHAPE[-1])-1 # We only support y_size <= x_size

    x = torch.rand(X_SHAPE, dtype=torch.float64)    # (BS, C1, H1, W1)
    kernel = torch.rand(KERNEL_SHAPE, dtype=torch.float64)  # (C2, C1, kernel_H, kernel_W)
    y = torch.nn.functional.conv2d(x, kernel, padding=PADDING, stride=1, dilation=1) # (BS, C2, H2, W2)
    other_y = FFT_padded_conv(x, kernel, PADDING)  # get the result using FFT methods
    print(f"FFT_padded conv result: max_absolute_error = {torch.max(torch.abs(y - other_y))}\n")
\end{lstlisting}

The Python script output is reproduced below. The maximum absolute error is at the level of double-precision roundoff, verifying the correctness of Eq.~\ref{supp:eq:supp_fft_padded_operator}.
\begin{verbatim}
FFT_padded conv result: max_absolute_error = 1.0658141036401503e-14
\end{verbatim}

For reverse reconstruction, $\vy_l^{\ast}$ is padded at its left and top sides to the $(H_l+2p)\times(W_l+2p)$ grid; this operation is denoted by $\mathcal{Z}^{(lt)}$.
The former forward feature provides the MDP anchor
$\mathcal{F}_{u,v}(\mathcal{Z}_p\hat{\va}_l)$.

For Case I (width-reducing), $C_l\ge C_{l+1}$, the nominal frequency-wise reconstruction gives priority to the exact CCP constraint and selects the spectrum nearest to the former feature:
\begin{align}
    \mathcal{F}_{u,v}(\va_l^{\ast})
    =
    \arg\min_{\mathbf{a}^{F}\in\mathbb{C}^{C_l}}
    \quad&
    \left\|
    \mathbf{a}^{F}
    -
    \mathcal{F}_{u,v}(\mathcal{Z}_p\hat{\va}_l)
    \right\|_2^2
    \nonumber\\
    \mathrm{s.t.}\quad&
    \mathbf{K}^{F}_{u,v}\mathbf{a}^{F}
    =
    \mathcal{F}_{u,v}
    (\mathcal{Z}^{(lt)}\vy_l^{\ast}),
    \qquad
    (u,v)\in\Omega_l'.
    \label{supp:eq:supp_fft_padded_caseI}
\end{align}

For Case II, $C_l< C_{l+1}$,
\begin{equation}\label{supp:eq:conv_fft_caseII}
    \mathbf{a}^{F,x,\ast}_{u,v}=
    \arg\min_{\mathbf{a}^{F}\in\mathbb{C}^{C_l}}
    \|\mathbf{K}^{F}_{u,v}\mathbf{a}^{F,x}_{u,v}-\mathcal{F}_{u,v}(\mathcal{Z}^{(lt)}\vy_l^{\ast})\|_2^2.
\end{equation}
After solving all frequency systems, the inverse DFT is applied and the $p$ padded entries are removed from each spatial side to obtain the final reconstructed feature $\mathbf{a}^{x, \ast}_{u,v} = \mathcal{Z}_{p}^{-1}\mathcal{F}^{-1}(\mathbf{a}^{F,x, \ast})$ of the FFT-padded method.

Although Eq.~\eqref{supp:eq:supp_fft_padded_operator} is an exact implementation of the \emph{forward} CNN convolution, the corresponding reverse model is approximate when $p>0$. In particular, the inversion treats the entire padded feature $\mathcal{Z}_p\va_l$ as an unknown and therefore introduces $(H_l+2p)(W_l+2p)$ spatial variables per channel instead of the physical $H_lW_l$ variables. The reconstructed values in the artificial padding region are not constrained to remain zero, and removing them after the inverse DFT generally changes the forward response of the retained feature. Consequently, the final cropped reconstruction need not satisfy the original CCP exactly. FFT-padded is therefore simple and highly efficient, but obtains its frequency-wise decoupling by relaxing the zero-padding structure during reverse reconstruction.

\begin{figure}[H]
    \centering
    \includegraphics[width=0.5\textwidth]{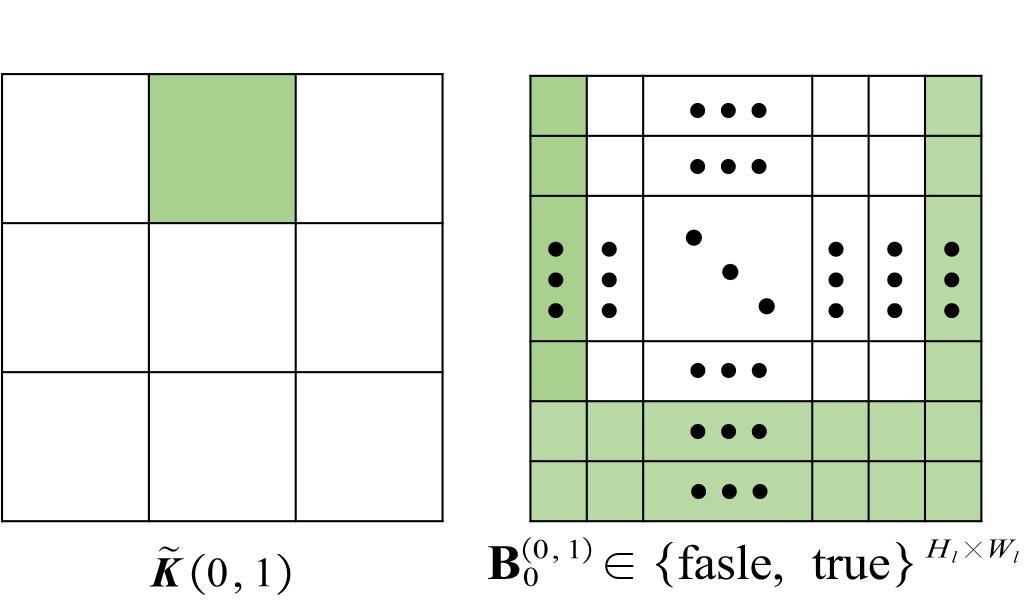}
    \caption{An example for $\mathbf{B}^{(i,j)}_p$, where $(i,j)=(0,1)$ and $padding=0$.}
    \label{supp:fig:bound}
\end{figure}

\textbf{(2) FFT-boundary.}
FFT-boundary instead keeps the original $H_l\times W_l$ feature as the reconstruction variable and explicitly removes the wrap-around terms introduced by circular convolution.
We next derive the correction exactly as it is evaluated by \texttt{FFT\_boundary\_conv}. For $(i,j)\in [H_K]\times[W_K]$, let
$\mathbf{B}^{(i,j)}_p\in\{0,1\}^{H_l\times W_l}$
mark those source positions whose circular shift associated with
$\widetilde{\mathbf{K}}_l(i,j)$ would contribute through a wrap-around path that is absent from the $p$-padded CNN convolution. Thus,
$\va_l[m]\odot\mathbf{B}^{(i,j)}_p$ contains precisely the source values that must be removed for this kernel position. Fig.~\ref{supp:fig:bound} illustrates such a boundary mask.

The complete boundary correction in the Fourier domain is
\begin{equation}
    [\mathcal{G}_p(\va_l)]_n(u,v)
    =
    \sum_{m=1}^{C_l}
    \sum_{i=0}^{H_K-1}
    \sum_{j=0}^{W_K-1}
    \widetilde{\mathbf{K}}_l[n,m](i,j)
    \exp\!\left[
        -2\pi\mathrm{i}
        \left(
        \frac{ui}{H_l}
        +
        \frac{vj}{W_l}
        \right)
    \right]
    \times
    \mathcal{F}
    \left(
        \va_l[m]\odot
        \mathbf{B}^{(i,j)}_p
    \right)(u,v).
    \label{supp:eq:supp_fft_boundary_correction}
\end{equation}
The exponential factor is the Fourier-domain phase shift associated with the spatial offset $(i,j)$. Therefore,
\begin{equation}
    \mathbf{K}^{F}_{u,v}
    \mathcal{F}_{u,v}(\va_l)
    -
    \mathcal{G}_{p,u,v}(\va_l)
    \notag
\end{equation}
is exactly the spectrum of the circular convolution after removing all invalid wrap-around contributions.

The resulting spatial tensor is circularly shifted by $(-p,-p)$ and its trailing
$H_{l+1}\times W_{l+1}$ block is retained. Hence the exact CNN operator can be evaluated as
\begin{equation}
    \mathcal{A}_p(\va_l)
    =
    \mathcal{S}_p
    \operatorname{roll}_{(-p,-p)}
    \mathcal{F}^{-1}
    \left(
        \left[
        \mathbf{K}^{F}_{u,v}
        \mathcal{F}_{u,v}(\va_l)
        -
        \mathcal{G}_{p,u,v}(\va_l)
        \right]_{(u,v)\in\Omega_l}
    \right),
    \label{supp:eq:supp_fft_boundary_operator}
\end{equation}
where
$\Omega_l=[0,H_l-1]\times[0,W_l-1]$.
Unlike FFT-padded, Eq.~\eqref{supp:eq:supp_fft_boundary_operator} operates directly on the physical feature grid without enlarging the input variable. Listing~\ref{supp:lst:fft_boundary_conv_verification} confirms that this boundary-corrected FFT expression reproduces the spatial CNN convolution to numerical precision.

The implementation assumes $H_{l+1}\le H_l$ and $W_{l+1}\le W_l$, as enforced by the assertion in Listing~\ref{supp:lst:fft_boundary_conv_verification}.

\begin{lstlisting}[style=pythonstyle,caption={PyTorch verification code for the FFT-boundary convolution operator.},label={supp:lst:fft_boundary_conv_verification}]
import torch
torch.manual_seed(42)

def FFT_boundary_conv(x, kernel, padding):
    """Verify the correctness of the FFT-boundary method."""
    H, W = x.shape[-2:]
    kernel_H, kernel_W = kernel.shape[-2:]
    y_size = (H + 2 * padding - kernel_H + 1, W + 2 * padding - kernel_W + 1)
    fft_x = torch.fft.fft2(x)
    kernel_flip = torch.flip(kernel, dims=(-2, -1))
    fft_kernel = torch.fft.fft2(kernel_flip, s=(H, W))
    fft_y = torch.einsum('dcuv, bcuv->bduv', fft_kernel, fft_x)

    if padding > 0:
        def get_boundary_mask(xx, yy):
            """the boundary active set B(xx,yy) (Fig.S1 gives the example of B(0,1) with padding=0)"""
            top_row = max(xx-padding, 0)
            left_col = max(yy-padding, 0)
            bottom_row = max(kernel_H-1-xx-padding, 0)
            right_col = max(kernel_W-1-yy-padding, 0)
            bound = torch.zeros((H, W), dtype=torch.bool, device=x.device)
            bound[:top_row,:] = True
            bound[:,:left_col] = True
            if bottom_row>0:  bound[-bottom_row:,:] = True
            if right_col>0:  bound[:,-right_col:] = True
            return bound

        DFT_H = torch.fft.fft(torch.eye(H, dtype=x.dtype, device=x.device), dim=0)
        DFT_W = torch.fft.fft(torch.eye(W, dtype=x.dtype, device=x.device), dim=0)
        G = torch.zeros_like(fft_y)
        for i in range(kernel_H):
            for j in range(kernel_W):
                mask = get_boundary_mask(kernel_H - i - 1, kernel_W - j - 1)
                phase = DFT_H[:, i, None] * DFT_W[:, j]  # u, v
                fft_boundary = torch.fft.fft2(x * mask)  # bs, c1, u, v
                weights = kernel_flip[:, :, i, j].to(fft_boundary.dtype)  # c2, c1
                G += torch.einsum('dc,bcuv->bduv', weights, fft_boundary) * phase
        fft_y -= G

    y = torch.fft.ifft2(fft_y).real
    y = torch.roll(y, shifts=(-padding, -padding), dims=(-2, -1))
    return y[..., -y_size[-2]:, -y_size[-1]:]

if __name__ == '__main__':
    X_SHAPE = (2, 2, 16, 16)  # (BS, C1, H1, W1)
    KERNEL_SHAPE = (4, X_SHAPE[1], 5, 5) # (C2, C1, kernel_H, kernel_W)
    PADDING: int = 1
    assert 2*PADDING <= min(KERNEL_SHAPE[-2], KERNEL_SHAPE[-1])-1 # We only support y_size <= x_size
    x = torch.rand(X_SHAPE, dtype=torch.float64)    # (BS, C1, H1, W1)
    kernel = torch.rand(KERNEL_SHAPE, dtype=torch.float64)  # (C2, C1, kernel_H, kernel_W)
    y = torch.nn.functional.conv2d(x, kernel, padding=PADDING, stride=1, dilation=1) # (BS, C2, H2, W2)

    other_y = FFT_boundary_conv(x, kernel, PADDING)  # get the result using FFT methods
    print(f"FFT_boundary conv result: max_absolute_error = {torch.max(torch.abs(y - other_y))}\n")
\end{lstlisting}

The notebook output is reproduced below. The small residual confirms that the boundary-corrected FFT operator reproduces the spatial convolution to the numerical precision of this implementation.
\begin{verbatim}
FFT_boundary conv result: max_absolute_error = 1.2434497875801753e-14
\end{verbatim}

We next derive its reverse model. The retained target $\vy_l^{\ast}$ is first placed in the trailing $H_{l+1}\times W_{l+1}$ block of an
$H_l\times W_l$ tensor, with the remaining entries set to zero, and is then circularly shifted by $(p,p)$ to undo the output alignment in Eq.~\eqref{supp:eq:supp_fft_boundary_operator}. Using the same
$\mathcal{Z}^{(lt)}$ notation, the corresponding spectrum is
$    \mathcal{F}_{u,v}
    \left(
        \operatorname{roll}_{(p,p)}
        \mathcal{Z}^{(lt)}\vy_l^{\ast}
    \right)$.

If the complete full-grid output and the reconstructed boundary feature were known, Eq.~\eqref{supp:eq:supp_fft_boundary_operator} would imply
\begin{equation}
    \mathbf{K}^{F}_{u,v}
    \mathcal{F}_{u,v}(\va_l^{\ast})
    =
    \mathcal{F}_{u,v}
    \left(
        \operatorname{roll}_{(p,p)}
        \mathcal{Z}^{(lt)}\vy_l^{\ast}
    \right)
    +
    \mathcal{G}_{p,u,v}(\va_l^{\ast}).
    \notag
\end{equation}
The difficulty is that the boundary correction on the right-hand side depends on the unknown reconstruction itself. To preserve independent linear systems at every frequency, FFT-boundary evaluates this correction from the former forward feature,
\begin{equation}
    \mathcal{G}_{p,u,v}(\va_l^{\ast})
    \approx
    \mathcal{G}_{p,u,v}(\hat{\va}_l).
    \label{supp:eq:supp_fft_boundary_approx}
\end{equation}
Thus, technically, the reverse model inverts the approximate operator $\widetilde{\mathcal{A}}_p$ as follows:
\begin{equation}
    \widetilde{\mathcal{A}}_p(\va_l;\hat{\va}_l)
    =
    \mathcal{S}_p
    \operatorname{roll}_{(-p,-p)}
    \mathcal{F}^{-1}
    \left(
        \left[
        \mathbf{K}^{F}_{u,v}
        \mathcal{F}_{u,v}(\va_l)
        -
        \mathcal{G}_{p,u,v}(\hat{\va}_l)
        \right]_{(u,v)\in\Omega_l}
    \right),
    \label{supp:eq:supp_fft_boundary_operator_tuta}
\end{equation}
For Case I (width-reducing), $C_l\ge C_{l+1}$, FFT-boundary solves
\begin{align}
    \mathcal{F}_{u,v}(\va_l^{\ast})
    =
    \arg\min_{\mathbf{a}^{F}\in\mathbb{C}^{C_l}}
    \quad&
    \left\|
        \mathbf{a}^{F}
        -
        \mathcal{F}_{u,v}(\hat{\va}_l)
    \right\|_2^2
    \nonumber\\
    \mathrm{s.t.}\quad&
    \mathbf{K}^{F}_{u,v}\mathbf{a}^{F}
    =
    \mathcal{F}_{u,v}
    \left(
        \operatorname{roll}_{(p,p)}
        \mathcal{Z}^{(lt)}\vy_l^{\ast}
    \right)
    +
    \mathcal{G}_{p,u,v}(\hat{\va}_l).
    \label{supp:eq:supp_fft_boundary_caseI}
\end{align}
For Case II (width-expanding), $C_l<C_{l+1}$, it solves
\begin{equation}
    \mathcal{F}_{u,v}(\va_l^{\ast})
    =
    \arg\min_{\mathbf{a}^{F}\in\mathbb{C}^{C_l}}
    \Big\|
    \mathbf{K}^{F}_{u,v}\mathbf{a}^{F}
    -
    \mathcal{F}_{u,v}
    \left(
        \operatorname{roll}_{(p,p)}
        \mathcal{Z}^{(lt)}\vy_l^{\ast}
    \right)- \mathcal{G}_{p,u,v}(\hat{\va}_l)
    \Big\|_2^2.
    \label{supp:eq:supp_fft_boundary_caseII}
\end{equation}
The reconstructed spatial feature $\mathbf{a}_{l}^{\ast}$ is obtained directly by inverse DFT on the $H_l\times W_l$ grid.

FFT-boundary therefore preserves the physical number of reconstruction variables and avoids the artificial padded variables introduced by FFT-padded. Its approximation instead enters through the boundary treatment during \emph{reverse} reconstruction. Specifically, Eq.~\eqref{supp:eq:supp_fft_boundary_approx} freezes a correction that would exactly depend on $\va_l^{\ast}$ at the former feature $\hat{\va}_l$, so deviations between their boundary values translate into CCP error. In addition, when
$H_{l+1}<H_l$ or $W_{l+1}<W_l$, only the retained output block is specified by the CNN target. Zero-filling the remaining entries in
$\mathcal{Z}^{(lt)}\vy_l^{\ast}$
introduces auxiliary full-grid constraints that are not present in the original cropped convolutional equation. This second approximation disappears for spatially size-preserving convolutions,
$H_{l+1}=H_l$ and $W_{l+1}=W_l$.

The two FFT variants consequently trade different approximations for frequency-wise decoupling. FFT-padded evaluates the forward convolution exactly by enlarging the spatial grid, but its reverse reconstruction introduces artificial padded variables whose subsequent removal may violate CCP. FFT-boundary also evaluates the forward convolution exactly, while retaining the physical feature grid; its reverse approximation arises from freezing the unknown boundary correction and, for spatially reducing convolutions, from completing an only partially observed full-grid output.

\textbf{Unified reliability-gated regularization.}

Numerical reliability is then checked independently for each frequency--sample pair. A failed pair is replaced by the MDP-centered Tikhonov reconstruction
\begin{equation}
    \mathbf{a}^{F,\alpha}_{u,v}
    =
    \arg\min_{\mathbf{a}^{F}}
    \left\|
        \mathbf{K}^{F}_{u,v}\mathbf{a}^{F}
        -
        \mathbf{d}^{F}_{u,v}
    \right\|_2^2
    +
    \alpha_{u,v}
    \left\|
        \mathbf{a}^{F}
        -
        \hat{\mathbf{a}}^{F}_{u,v}
    \right\|_2^2.
    \label{supp:eq:supp_fft_mdp_reg_problem}
\end{equation}
where $\mathbf{d}^{F}_{u,v}$ denotes the corresponding frequency-domain right-hand side defined in Eq.~\ref{supp:eq:conv_fft_caseII} or Eq.~\ref{supp:eq:supp_fft_boundary_caseII}.
Thus, the FFT implementations share the same reliability-gated MDP-centered stabilization; only their treatment of the convolutional boundary differs. After all frequency systems have been processed, the inverse DFT yields the reconstructed real-valued feature.

\subsubsection{Boundary-anchoring error of FFT-boundary}
\label{supp:sec:proof_fft_boundary}

The forward boundary correction $\mathcal{G}_p(\va)$ is linear in the boundary values of $\va$. The exact frequency constraint would use $\mathcal{G}_p(\va_l^{\ast})$, whereas the practical reverse solver freezes the correction at the former feature $\mathcal{G}_p(\hat{\va}_l)$. The resulting approximation error is therefore controlled by the boundary component of the feature deviation, which is also the quantity discouraged by the MDP anchor.

\begin{suppProposition}[Boundary error of the anchored FFT reconstruction]
\label{supp:prop:supp_fft_boundary}
Let $\mathcal{A}_p$ denote the exact $p$-padded convolution as defined in Eq.~\ref{supp:eq:supp_fft_boundary_operator}, and let $\widetilde{\mathcal{A}}_p(\va;\hat{\va})$ denote the FFT-boundary operator whose boundary correction is evaluated at $\hat{\va}$ as defined in Eq.~\ref{supp:eq:supp_fft_boundary_operator_tuta}. Let $\mathcal{P}_{\partial_p}$ select the union of input entries participating in boundary-crossing terms that can affect the retained convolutional output. Then, for any reconstructed feature $\va$,
\begin{equation}
    \|\mathcal{A}_p(\va)
      -\widetilde{\mathcal{A}}_p(\va;\hat{\va})\|_F
    \le
    \left(
    \sum_{n=1}^{C_{l+1}}
    \sum_{m=1}^{C_l}
    \|\mathbf{K}_l[n,m]\|_1^2
    \right)^{1/2}
    \|\mathcal{P}_{\partial_p}(\va-\hat{\va})\|_F,
    \label{supp:eq:supp_fft_padding_bound}
\end{equation}
For $p=0$ on the retained valid-convolution region, no boundary-crossing sample is used and the right-hand side is zero.
\end{suppProposition}

\begin{proof}
Let $\Delta\va=\va-\hat{\va}$. We first express explicitly the discrepancy introduced by freezing the
boundary correction at $\hat{\va}$. From
Eqs.~\eqref{supp:eq:supp_fft_boundary_operator} and~\eqref{supp:eq:supp_fft_boundary_operator_tuta},
\begin{equation}
    \mathcal{A}_p(\va)
    -\widetilde{\mathcal{A}}_p(\va;\hat{\va})
    =
    \mathcal{S}_p
    \operatorname{roll}_{(-p,-p)}
    \mathcal{F}^{-1}
    \left(
        \left[
        -\mathcal{G}_{p,u,v}(\va)
        +\mathcal{G}_{p,u,v}(\hat{\va})
        \right]_{(u,v)\in\Omega_l}
    \right).
    \notag
\end{equation}
The boundary correction $\mathcal{G}_p(\va)$ is linear in $\va$ according to Eq.~\ref{supp:eq:supp_fft_boundary_correction}, so $\mathcal{G}_p(\va)-\mathcal{G}_p(\hat{\va})=\mathcal{G}_p(\Delta\va)$ holds, and the above formula becomes
\begin{equation}
    \mathcal{A}_p(\va)
    -\widetilde{\mathcal{A}}_p(\va;\hat{\va})
    =
    -\mathcal{S}_p
    \operatorname{roll}_{(-p,-p)}
    \mathcal{F}^{-1}
    \left(
        \mathcal{G}_p(\Delta\va)
    \right).
    \label{supp:eq:supp_boundary_error_step2}
\end{equation}

To bound this quantity, we consider first the full-grid boundary error $\overline{\mathbf{E}}_p =\mathcal{F}^{-1}\left(\mathcal{G}_p(\Delta\va)\right)$ before the final roll and spatial selection.

From Eq.~\eqref{supp:eq:supp_fft_boundary_correction}, its $n$-th output
channel is generated by
\begin{equation}
    [\mathcal{G}_p(\Delta\va)]_n(u,v)
    =
    \sum_{m=1}^{C_l}
    \sum_{i=0}^{H_K-1}
    \sum_{j=0}^{W_K-1}
    \widetilde{\mathbf{K}}_l[n,m](i,j)
    \times
    \exp\!\left[
        -2\pi\mathrm{i}
        \left(
            \frac{ui}{H_l}
            +
            \frac{vj}{W_l}
        \right)
    \right]
    \mathcal{F}
    \left(
        \Delta\va[m]\odot
        \mathbf{B}^{(i,j)}_p
    \right)(u,v).
    \label{supp:eq:supp_boundary_error_frequency_expansion}
\end{equation}
Let $\mathcal{T}_{i,j}$ denote the circular spatial shift associated
with the phase factor
\[
    \exp\!\left[
        -2\pi\mathrm{i}
        \left(
            \frac{ui}{H_l}
            +
            \frac{vj}{W_l}
        \right)
    \right].
\]
By the DFT shift property,
\begin{equation}
    \mathcal{T}_{i,j}(\mathbf{X})=\mathcal{F}^{-1}
    \left(
        e^{-2\pi\mathrm{i}
        (ui/H_l+vj/W_l)}
        \mathcal{F}(\mathbf{X})(u,v)
    \right),
\end{equation}
where the exact shift direction depends only on the adopted DFT
convention and is irrelevant to the norm estimate below. Circular
shifts are permutations of spatial entries and therefore preserve the
Frobenius norm:
\begin{equation}
    \|\mathcal{T}_{i,j}(\mathbf{X})\|_F
    =
    \|\mathbf{X}\|_F.
    \label{supp:eq:supp_shift_norm_preserving}
\end{equation}
Taking the inverse DFT of
Eq.~\eqref{supp:eq:supp_boundary_error_frequency_expansion} thus yields
\begin{align}
    \overline{\mathbf{E}}_p[n]
    =
    \sum_{m=1}^{C_l}
    \sum_{i=0}^{H_K-1}
    \sum_{j=0}^{W_K-1}
    \widetilde{\mathbf{K}}_l[n,m](i,j)
    \mathcal{T}_{i,j}
    \left(
        \Delta\va[m]\odot
        \mathbf{B}^{(i,j)}_p
    \right).
    \label{supp:eq:supp_boundary_error_spatial_expansion}
\end{align}

By definition, $\mathcal{P}_{\partial_p}$ contains all input entries
that can participate in these boundary-crossing terms. Hence, for
every $(i,j)$ and input channel $m$,
\begin{equation}
    \left\|
        \Delta\va[m]\odot\mathbf{B}^{(i,j)}_p
    \right\|_F
    \le
    \left\|
        \mathcal{P}_{\partial_p}\Delta\va[m]
    \right\|_F.
    \label{supp:eq:supp_boundary_mask_bound}
\end{equation}
Applying the triangle inequality to
Eq.~\eqref{supp:eq:supp_boundary_error_spatial_expansion}, followed by
Eqs.~\eqref{supp:eq:supp_shift_norm_preserving} and~\eqref{supp:eq:supp_boundary_mask_bound}, gives
\begin{equation}
    \|\overline{\mathbf{E}}_p[n]\|_F
    \le
    \sum_{m=1}^{C_l}
    \sum_{i=0}^{H_K-1}
    \sum_{j=0}^{W_K-1}
    \left|
        \widetilde{\mathbf{K}}_l[n,m](i,j)
    \right|
    \left\|
        \mathcal{P}_{\partial_p}\Delta\va[m]
    \right\|_F
    =
    \sum_{m=1}^{C_l}
    \|\widetilde{\mathbf{K}}_l[n,m]\|_1
    \left\|
        \mathcal{P}_{\partial_p}\Delta\va[m]
    \right\|_F.
    \label{supp:eq:supp_boundary_channel_triangle}
\end{equation}
Spatial flipping does not change the $\ell_1$ norm of a kernel, so
    $\|\widetilde{\mathbf{K}}_l[n,m]\|_1
    =
    \|\mathbf{K}_l[n,m]\|_1$.
Applying the Cauchy--Schwarz inequality over the input-channel index
$m$ therefore gives
\begin{equation}
    \|\overline{\mathbf{E}}_p[n]\|_F^2
    \le
    \left(
        \sum_{m=1}^{C_l}
        \|\mathbf{K}_l[n,m]\|_1^2
    \right)
    \left(
        \sum_{m=1}^{C_l}
        \left\|
            \mathcal{P}_{\partial_p}
            \Delta\va[m]
        \right\|_F^2
    \right)
    =
    \left(
        \sum_{m=1}^{C_l}
        \|\mathbf{K}_l[n,m]\|_1^2
    \right)
    \left\|
        \mathcal{P}_{\partial_p}\Delta\va
    \right\|_F^2.
    \label{supp:eq:supp_boundary_input_channel_cs}
\end{equation}
Summing Eq.~\eqref{supp:eq:supp_boundary_input_channel_cs} over all output
channels yields
\begin{equation}
    \|\overline{\mathbf{E}}_p\|_F^2
    =
    \sum_{n=1}^{C_{l+1}}
    \|\overline{\mathbf{E}}_p[n]\|_F^2
    \le
    \left(
        \sum_{n=1}^{C_{l+1}}
        \sum_{m=1}^{C_l}
        \|\mathbf{K}_l[n,m]\|_1^2
    \right)
    \left\|
        \mathcal{P}_{\partial_p}
        (\va-\hat{\va})
    \right\|_F^2.
    \label{supp:eq:supp_boundary_output_channel_sum}
\end{equation}

Finally, the circular roll in
Eq.~\eqref{supp:eq:supp_boundary_error_step2} preserves the Frobenius norm,
while $\mathcal{S}_p$ only selects a subset of spatial entries and
therefore cannot increase it. Consequently,
\begin{equation}
    \left\|
        \mathcal{A}_p(\va)
        -
        \widetilde{\mathcal{A}}_p(\va;\hat{\va})
    \right\|_F
    \le
    \|\overline{\mathbf{E}}_p\|_F.
\end{equation}
Taking the square root of
Eq.~\eqref{supp:eq:supp_boundary_output_channel_sum} gives
\begin{equation}
    \left\|
        \mathcal{A}_p(\va)
        -
        \widetilde{\mathcal{A}}_p(\va;\hat{\va})
    \right\|_F
    \le
    \left(
        \sum_{n=1}^{C_{l+1}}
        \sum_{m=1}^{C_l}
        \|\mathbf{K}_l[n,m]\|_1^2
    \right)^{1/2}
    \left\|
        \mathcal{P}_{\partial_p}
        (\va-\hat{\va})
    \right\|_F,
\end{equation}
which proves Eq.~\eqref{supp:eq:supp_fft_padding_bound}.

For $p=0$, the retained valid-convolution outputs involve only spatial
windows fully contained in the finite input domain. Hence no
wrap-around term contributes to the retained output, and the boundary
correction vanishes after spatial selection. Equivalently,
$\mathcal{P}_{\partial_0}$ is empty when restricted to the retained
valid-convolution region, and both sides of
Eq.~\eqref{supp:eq:supp_fft_padding_bound} are zero.
\end{proof}


\subsection{Architectural Conditions for Exact Computation Consistency}
\label{supp:sec:proof_composite_blocks}

For a frozen composite block $\Phi_l:\mathbb{R}^{n}\rightarrow\mathbb{R}^{m}$ and a target $\va_{l+1}^{\ast}$, an exact CCP reconstruction exists precisely when $\va_{l+1}^{\ast}\in\operatorname{Range}(\Phi_l)$. This section gives useful architectural conditions under which the range covers the entire admissible target space. These results concern existence of a perfect CCP solution; they do not assert that a finite-step numerical solver always finds that solution.

\begin{suppProposition}[Affine reverse unit]
\label{supp:prop:supp_affine_surjectivity}
Let $\Phi_l(\va_l)=\mA\va_l+\vb$, where $\mA\in\mathbb{R}^{m\times n}$. An exact CCP solution exists for every $\va^{\ast}\in\mathbb{R}^{m}$ if and only if $\operatorname{rank}(\mA)=m$. In particular, $n\ge m$ is necessary but is not by itself sufficient.
\end{suppProposition}

\begin{proof}
The range of $\Phi_l$ is the affine set $\vb+\operatorname{Range}(\mA)$. This set equals $\mathbb{R}^{m}$ exactly when the column space of $\mA$ has dimension $m$.
\end{proof}

The next result covers residual blocks whose shortcut is surjective and whose residual branch is bounded.

\begin{suppTheorem}[Surjectivity of a bounded residual branch]
\label{supp:prop:supp_bounded_residual_surjectivity}
Let $\Phi_l(\va_l)=\mP\va_l+F(\va_l)$, where the projection $\mP\in\mathbb{R}^{m\times n}$ has full row rank and the residual branch $F:\mathbb{R}^{n}\rightarrow\mathbb{R}^{m}$ is continuous and bounded: $\|F(\va_l)\|_2\le M$ for all $\va_l$. Then $\Phi_l$ is surjective. Consequently, for every target $\va^{\ast}_{l+1}\in\mathbb{R}^{m}$ there exists at least one $\va_l^{\ast}\in \mathbb{R}^n$ satisfying $\Phi_l(\va_l^{\ast})=\va^{\ast}_{l+1}$.
\end{suppTheorem}

\begin{proof}
Because $\mP$ has full row rank, it admits the right inverse
    $\mR=\mP^T(\mP\mP^T)^{-1}$, so that
    $\mP\mR=\mI_m$.
Fix an arbitrary target $\va_{l+1}^{\ast}$ and define
\begin{equation}
    G_{\va_{l+1}^{\ast}}(\va_l)
    =
    \mR(\va_{l+1}^{\ast}-F(\va_l)).
    \notag
\end{equation}
The map $G_{\va_{l+1}^{\ast}}$ is continuous and satisfies
\begin{equation}
    \|G_{\va_{l+1}^{\ast}}(\va_l)\|_2
    \le
    \|\mR\|_2(\|\va_{l+1}^{\ast}\|_2+M)
    =:R_{\va_{l+1}^{\ast}}.
    \notag
\end{equation}
Hence $G_{\va_{l+1}^{\ast}}$ maps the closed ball $\overline{B}(0,R_{\va_{l+1}^{\ast}})$ into itself. Brouwer's fixed-point theorem gives a point $\va_l^{\ast}$ such that $\va_l^{\ast}=G_{\va_{l+1}^{\ast}}(\va_l^{\ast})$. Applying $\mP$ to both sides yields
\begin{equation}
    \mP\va_l^{\ast}
    =
    \va_{l+1}^{\ast}-F(\va_l^{\ast}),
    \notag
\end{equation}
which is equivalent to $\Phi_l(\va_l^{\ast})=\va_{l+1}^{\ast}$.
\end{proof}

For an identity shortcut, $\mP=\mathbb{I}$, Theorem~\ref{supp:prop:supp_bounded_residual_surjectivity} shows that
\begin{equation}
    \Phi_l(\va_l)=\va_l+F(\va_l)
    \notag
\end{equation}
is surjective whenever the residual branch is continuous and bounded. For example, the residual MLP
\begin{equation}
    \Phi_l(\va_l)
    =
    \va_l+
    \mW_2\sigma(\mW_1\va_l+\vb_1)+\vb_2
    \notag
\end{equation}
is surjective if $\sigma$ is bounded, such as $\tanh$ and sigmoid. The result does not require $\mW_1$ or $\mW_2$ to be invertible.
A pre-normalized Transformer residual sublayer can satisfy the same condition. For a fixed token count and feature dimension, the output of LayerNorm belongs to a bounded set (with the usual positive numerical constant in its variance normalization). A frozen attention or MLP module is continuous and therefore bounded on this compact set. Thus maps of the form
\begin{equation}
    \Phi_l(\va_l)
    =\va_l+\operatorname{MSA}(\operatorname{LN}(\va_l))
    \quad\text{or}\quad
    \Phi_l(\va_l)
    =\va_l+\operatorname{MLP}(\operatorname{LN}(\va_l))
    \notag
\end{equation}
are surjective under the stated finite-dimensional continuity assumptions. A composition of surjective sublayers remains surjective. This observation applies to pre-normalized residual sublayers; it does not automatically apply to an unnormalized Transformer block. For such a block, exact solvability requires additional conditions such as the contraction condition below.

\begin{suppProposition}[Bijectivity of a contractive residual branch]
\label{supp:prop:supp_residual_bijection}
Let $\Phi_l(\va_l)=\va_l+F(\va_l)$, where $F$ is Lipschitz continuous with $\operatorname{Lip}(F)=q<1$. Then $\Phi_l$ is bijective and
\begin{equation}
    \|\Phi_l^{-1}(\va_{l+1}^{(1)})-\Phi_l^{-1}(\va_{l+1}^{(2)})\|_2
    \le
    \frac{1}{1-q}\|\va_{l+1}^{(1)}-\va_{l+1}^{(2)}\|_2.
    \label{supp:eq:supp_inverse_lipschitz}
\end{equation}
\end{suppProposition}

\begin{proof}
For a fixed $\va_{l+1}^{\ast}$, define $G_{\va_{l+1}^{\ast}}(\va_l)=\va_{l+1}^{\ast}-F(\va_l)$. Since $\operatorname{Lip}(G_{\va_{l+1}^{\ast}})=q<1$, Banach's fixed-point theorem gives a unique $\va_{l}^{\ast}$ satisfying $\va_{l}^{\ast}=G_{\va_{l+1}^{\ast}}(\va_l^{\ast})$, equivalently $\Phi_l(\va_l^{\ast})=\va_{l+1}^{\ast}$. Hence $\Phi_l$ is bijective. For $\Phi_l(\va_l^{(i)})=\va_{l+1}^{(i)}$,
\begin{equation}
    \|\va_l^{(1)}-\va_l^{(2)}\|_2
    \le
    \|\va_{l+1}^{(1)}-\va_{l+1}^{(2)}\|_2
    +
    \|F(\va_l^{(1)})-F(\va_l^{(2)})\|_2
    \le
    \|\va_{l+1}^{(1)}-\va_{l+1}^{(2)}\|_2
    +q\|\va_l^{(1)}-\va_l^{(2)}\|_2.
\end{equation}
Rearranging the above formula proves Eq.~\eqref{supp:eq:supp_inverse_lipschitz}.
\end{proof}
Compared with Theorem~\ref{supp:prop:supp_bounded_residual_surjectivity}, this condition is stronger: boundedness guarantees existence but may allow multiple preimages, whereas contraction guarantees existence, uniqueness, and inverse stability.

For this residual MLP, a sufficient condition is
\begin{equation}
    \|\mW_2\|_2\,
    \operatorname{Lip}(\sigma)\,
    \|\mW_1\|_2
    <1.
    \notag
\end{equation}
Neither an identity shortcut nor dimensional equality alone guarantees surjectivity for an arbitrary unbounded residual branch. For example, in one dimension,
\begin{equation}
    T(x)=x-2\operatorname{ReLU}(x)
    =
    \begin{cases}
        x, & x<0,\\
        -x, & x\ge0,
    \end{cases}
    \notag
\end{equation}
has range $(-\infty,0]$ and is therefore not surjective. Consequently, blocks using unbounded activations such as ReLU, GELU, or SiLU do not admit a universal exact-CCP guarantee without additional spectral, monotonicity, or normalization assumptions.

When the architectural conditions above are not available, the iterative solver seeks an exact solution when one is reachable and otherwise returns a numerically stable approximate CCP reconstruction.

\subsection{Reverse Computation of Activations}
\label{supp:sec:supp_activation_reverse}
Most activation functions can be reversed independently at each feature entry. Let
$\hat{\vz}$ denote the pre-activation recorded during the original forward pass and
$\va^{\ast}$ denote the reconstructed target activation. Since reconstructed $\va^{\ast}$ may exceed the range of the activation or be numerically abnormal, the implementation first defines a safe target
\begin{equation}
    \va_{\mathrm{safe}}^{\ast}
    =
    \mathcal{Q}_{\sigma}(\va^{\ast}),
    \notag
\end{equation}
where $\mathcal{Q}_{\sigma}$ denotes a minimal feasibility or numerical-stabilization correction for activation $\sigma$, such as clipping, lower-bound enforcement, or renormalization.
The activation reverse computation is modeled by the following problem:
\begin{equation}
    \vz^{\ast}
    =
    \arg\min
    \|\vz-\hat{\vz}\|_2^2, \quad \text{s.t. }\sigma(\vz)=\va_{\mathrm{safe}}^{\ast},
    \label{supp:eq:activation-mdp}
\end{equation}
where $\hat{\vz}$ is the recorded pre-activation feature map and $\va^{\ast}_{\mathrm{safe}}$ is the reconstructed activation target after pre-processing. Thus, the usual result is $\sigma(\vz^{\ast})=\va_{\mathrm{safe}}^{\ast}$, rather than necessarily $\sigma(\vz^{\ast})=\va^{\ast}$. Exact CCP with respect to the original target can be claimed only when the target is attainable, is unchanged by numerical stabilization, and the reverse rule is exact up to negligible numerical error. The resulting reverse rules can be grouped according to the invertibility of the corresponding activation.

\textbf{(1) Monotone activations with analytical inverse.}
For strictly monotone activations, the reverse computation is entry-wise and is obtained directly from the analytical inverse.

For the hyperbolic tangent,
\begin{equation}
    a=\tanh(z),\qquad a\in(-1,1),
    \notag
\end{equation}
the implementation first clips the target to
$[-1+\epsilon,1-\epsilon]$ with $\epsilon=10^{-6}$ and computes
\begin{equation}
    \va_{\mathrm{safe}}^{\ast}
    =\operatorname{clip}(\va^{\ast},-1+\epsilon,1-\epsilon),
    \qquad
    \vz^{\ast}
    =
    \operatorname{arctanh}(\va_{\mathrm{safe}}^{\ast}).
    \notag
\end{equation}
The clipping avoids the infinite inverse values associated with
$\va^{\ast}=\pm1$.

Similarly, for the sigmoid activation
\begin{equation}
    a=\frac{1}{1+\exp(-z)},\qquad a\in(0,1),
    \notag
\end{equation}
the target is clipped to $[\epsilon,1-\epsilon]$ and
\begin{equation}
    \va_{\mathrm{safe}}^{\ast}
    =\operatorname{clip}(\va^{\ast},\epsilon,1-\epsilon),
    \qquad
    \vz^{\ast}
    =
    \log\frac{\va_{\mathrm{safe}}^{\ast}}{1-\va_{\mathrm{safe}}^{\ast}}.
    \notag
\end{equation}

For ELU with parameter $\alpha>0$,
\begin{equation}
    \operatorname{ELU}(z)
    =
    \begin{cases}
        z, & z>0,\\
        \alpha(\exp(z)-1), & z\le0 ,
    \end{cases}
    \notag
\end{equation}
whose attainable range is $(-\alpha,\infty)$, the implementation clips
$\va^{\ast}$ from below at $-\alpha+\epsilon$ and uses
\begin{equation}
    \va_{\mathrm{safe}}^{\ast}
    =\max(\va^{\ast},-\alpha+\epsilon),
    \qquad
    \vz^{\ast}
    =
    \begin{cases}
        \va^{\ast}_{\mathrm{safe}},
        & \va^{\ast}_{\mathrm{safe}}>0,\\[1mm]
        \log\left(1+\dfrac{\va^{\ast}_{\mathrm{safe}}}{\alpha}\right),
        & \va^{\ast}_{\mathrm{safe}}\le0,
    \end{cases}
    \notag
\end{equation}
The implementation raises a \texttt{ValueError} when $\alpha\le0$ and evaluates the logarithm with a $\operatorname{log1p}$ form.

LeakyReLU is theoretically one-to-one when the negative-slope parameter satisfies $s>0$,
\begin{equation}
    \operatorname{LeakyReLU}(z)
    =
    \begin{cases}
        z, & z\ge0,\\
        sz, & z<0.
    \end{cases}
    \notag
\end{equation}
For $s=0$, the implementation reduces to the ReLU rule described below.
For $s>0$, the current implementation uses
\begin{equation}
    \vz^{\ast}
    =
    \begin{cases}
        \va^{\ast}, & \va^{\ast}\ge0,\\
        \va^{\ast}/0.5, & \va^{\ast}<0,
    \end{cases}
    \notag
\end{equation}
where the fixed denominator $0.5$ is used instead of directly dividing by a potentially small negative slope, thereby limiting amplification of negative target values. The theoretical exact inverse on the negative branch is $\va^{\ast}/s$; hence the displayed rule is exact only when $s=0.5$. When $s\ne0.5$, it is a stabilized approximate rule and its negative-branch CCP residual is
\begin{equation}
    \operatorname{LeakyReLU}_{s}(\vz^{\ast})-\va^{\ast}
    =
    \left(\frac{s}{0.5}-1\right)\va^{\ast}.
    \notag
\end{equation}

\textbf{(2) Piecewise activations with non-unique saturated preimages.}
For piecewise saturating activations, boundary output values correspond to an interval of valid pre-activations. In these cases, Eq.~\eqref{supp:eq:activation-mdp} admits a simple projection of the former pre-activation onto the corresponding preimage set.

For ReLU,
\begin{equation}
    a=\max(z,0),
    \notag
\end{equation}
a positive target has the unique inverse $\vz^{\ast}=\va^{\ast}$, whereas
$\va^{\ast}=0$ is generated by every $\vz\le0$. The implementation therefore uses
\begin{equation}
    \va_{\mathrm{safe}}^{\ast}=\max(\va^{\ast},0),
    \vz^{\ast}
    =
    \begin{cases}
        \va_{\mathrm{safe}}^{\ast}, & \va_{\mathrm{safe}}^{\ast}>0,\\
        \min(\hat{\vz},0), & \va_{\mathrm{safe}}^{\ast}=0.
    \end{cases}
    \notag
\end{equation}
Thus, negative targets are mapped to the feasible zero output, and the closest feasible pre-activation to $\hat{\vz}$ is retained.

For ReLU6,
\begin{equation}
    a=\min(\max(z,0),6),
    \notag
\end{equation}
the target is first projected to
\begin{equation}
    \va^{\ast}_{\mathrm{safe}}
    =
    \operatorname{clip}(\va^{\ast},0,6).
    \notag
\end{equation}
The reverse rule is
\begin{equation}
    \vz^{\ast}
    =
    \begin{cases}
        \min(\hat{\vz},0),
        & \va^{\ast}_{\mathrm{safe}}\le0,\\
        \va^{\ast}_{\mathrm{safe}},
        & 0<\va^{\ast}_{\mathrm{safe}}<6,\\
        \max(\hat{\vz},6),
        & \va^{\ast}_{\mathrm{safe}}\ge6.
    \end{cases}
    \notag
\end{equation}
The two saturated branches therefore retain the preimage nearest to the former pre-activation.

The same construction applies to HardTanh,
\begin{equation}
    a=\operatorname{clip}(z,a_{\min},a_{\max}).
    \notag
\end{equation}
After
\begin{equation}
    \va^{\ast}_{\mathrm{safe}}
    =
    \operatorname{clip}
    (\va^{\ast},a_{\min},a_{\max}),
    \notag
\end{equation}
the implemented inverse is
\begin{equation}
    \vz^{\ast}
    =
    \begin{cases}
        \min(\hat{\vz},a_{\min}),
        & \va^{\ast}_{\mathrm{safe}}\le a_{\min},\\
        \va^{\ast}_{\mathrm{safe}},
        & a_{\min}<\va^{\ast}_{\mathrm{safe}}<a_{\max},\\
        \max(\hat{\vz},a_{\max}),
        & \va^{\ast}_{\mathrm{safe}}\ge a_{\max}.
    \end{cases}
    \notag
\end{equation}

For PyTorch HardSigmoid,
\begin{equation}
    a
    =
    \begin{cases}
        0, & z\le-3,\\
        z/6+1/2, & -3<z<3,\\
        1, & z\ge3,
    \end{cases}
    \notag
\end{equation}
the target is first clipped to $[0,1]$:
\begin{equation}
    \va^{\ast}_{\mathrm{safe}}
    =\operatorname{clip}(\va^{\ast},0,1).
    \notag
\end{equation}
Its reverse computation is
\begin{equation}
    \vz^{\ast}
    =
    \begin{cases}
        \min(\hat{\vz},-3),
        & \va^{\ast}_{\mathrm{safe}}\le0,\\
        6\va^{\ast}_{\mathrm{safe}}-3,
        & 0<\va^{\ast}_{\mathrm{safe}}<1,\\
        \max(\hat{\vz},3),
        & \va^{\ast}_{\mathrm{safe}}\ge1.
    \end{cases}
    \notag
\end{equation}

\textbf{(3) Non-monotone activations with multiple isolated preimages.}
GELU and SiLU are not globally one-to-one. For a negative target strictly above their minima, two distinct finite pre-activations may therefore satisfy the same CCP constraint. At the minimum, the two branch roots coincide. The implementation explicitly searches both monotone branches and applies MDP to select the root closest to $\hat z$.

This subsection uses only the exact GELU, corresponding to PyTorch's \texttt{approximate="none"}; the implementation does not handle the tanh approximation here.
For the exact GELU,
\begin{equation}
    \operatorname{GELU}(z)
    =
    z\Phi(z),
    \notag
\end{equation}
where $\Phi(\cdot)$ is the standard-normal cumulative distribution function, the unique minimum occurs numerically at
\begin{equation}
    z_{\min}^{G}
    =
    -0.7517915246935645,
    \qquad
    a_{\min}^{G}
    =
    \operatorname{GELU}(z_{\min}^{G})
    \approx -0.16997.
    \notag
\end{equation}
Targets below $a_{\min}^{G}$ are first projected to this minimum:
\begin{equation}
    \va^{\ast}_{\mathrm{safe}}
    =
    \max(\va^{\ast},a_{\min}^{G}).
    \notag
\end{equation}
For $\va^{\ast}_{\mathrm{safe}}\ge0$, GELU is monotone on the relevant non-negative branch and the implementation obtains its unique root by bisection on
\begin{equation}
    [0,h],
    \qquad
    h=\max\left\{\max_i a_{\mathrm{safe},i}^{\ast}+8,\,8\right\}.
    \notag
\end{equation}
For
$a_{\min}^{G}\le \va^{\ast}_{\mathrm{safe}}<0$,
two roots are computed: a left root
$z_{\mathrm{left}}^{G}$
from the monotone interval
\begin{equation}
    [-30,z_{\min}^{G}]
    \notag
\end{equation}
and a right root
$z_{\mathrm{right}}^{G}$
from
\begin{equation}
    [z_{\min}^{G},h].
    \notag
\end{equation}
Equivalently, the right-root interval is described by
\begin{equation}
    \ell_{\mathrm{right}}^{G}
    =
    \begin{cases}
        z_{\min}^{G}, & \va_{\mathrm{safe}}^{\ast}<0,\\
        0, & \va_{\mathrm{safe}}^{\ast}\ge0,
    \end{cases}
    \qquad
    u_{\mathrm{right}}^{G}=h.
    \notag
\end{equation}
Both roots are obtained by 80 bisection iterations by default (the iteration count is configurable), and MDP selects
\begin{equation}
    z^{\ast}
    =
    \arg\min_{
        z\in
        \{z_{\mathrm{left}}^{G},
        z_{\mathrm{right}}^{G}\}}
    |z-\hat z|,
    \qquad
    a_{\min}^{G}\le \va^{\ast}_{\mathrm{safe}}<0.
    \notag
\end{equation}
For a non-negative target, only the right root is used. The left endpoint $-30$ is a numerical truncation of the theoretically unbounded left branch, not its complete boundary. At equal candidate distances, the implementation selects the right root.

SiLU (Swish) is treated analogously:
\begin{equation}
    \operatorname{SiLU}(z)
    =
    z\,\operatorname{sigmoid}(z).
    \notag
\end{equation}
Its numerical minimum used by the implementation is
\begin{equation}
    z_{\min}^{S}
    =
    -1.2784645427610738,
    \qquad
    a_{\min}^{S}
    =
    \operatorname{SiLU}(z_{\min}^{S})
    \approx -0.27846.
    \notag
\end{equation}
The target is first projected as
\begin{equation}
    \va^{\ast}_{\mathrm{safe}}
    =
    \max(\va^{\ast},a_{\min}^{S}).
    \notag
\end{equation}
For $\va^{\ast}_{\mathrm{safe}}\ge0$, the unique non-negative root is found by bisection on $[0,h]$, with the same adaptive upper endpoint
$h=\max\{\max_i a_{\mathrm{safe},i}^{\ast}+8,8\}$.
For
$a_{\min}^{S}\le \va^{\ast}_{\mathrm{safe}}<0$,
the implementation separately searches
\begin{equation}
    [-60,z_{\min}^{S}]
    \qquad\text{and the actual right-root interval}\qquad
    [z_{\min}^{S},h]
    \notag
\end{equation}
for the two roots
$z_{\mathrm{left}}^{S}$ and
$z_{\mathrm{right}}^{S}$, respectively. The reconstructed pre-activation is then
\begin{equation}
    z^{\ast}
    =
    \arg\min_{
        z\in
        \{z_{\mathrm{left}}^{S},
        z_{\mathrm{right}}^{S}\}}
    |z-\hat z|.
    \notag
\end{equation}
Equivalently, the SiLU right-root interval has
\begin{equation}
    \ell_{\mathrm{right}}^{S}
    =
    \begin{cases}
        z_{\min}^{S}, & \va_{\mathrm{safe}}^{\ast}<0,\\
        0, & \va_{\mathrm{safe}}^{\ast}\ge0,
    \end{cases}
    \qquad
    u_{\mathrm{right}}^{S}=h.
    \notag
\end{equation}
The left endpoint $-60$ truncates the theoretically unbounded left branch; it cannot be claimed to cover every target arbitrarily close to $0^-$, and the default 80 bisection iterations are configurable. At equal candidate distances, the implementation selects the right root. Within the adopted numerical search intervals, GELU and SiLU satisfy CCP with respect to the safe target up to bisection and floating-point errors.

HardSwish is defined by
\begin{equation}
    \operatorname{HardSwish}(z)
    =z\frac{\operatorname{ReLU6}(z+3)}{6}
    =
    \begin{cases}
        0, & z\le-3,\\
        \dfrac{z(z+3)}{6}, & -3<z<3,\\
        z, & z\ge3.
    \end{cases}
    \notag
\end{equation}
Its minimum is attained at $z_{\min}^{H}=-3/2$ and equals $a_{\min}^{H}=-3/8$. The implementation first sets
$\va_{\mathrm{safe}}^{\ast}=\max(\va^{\ast},-3/8)$. For the middle quadratic branch, the candidate roots are
\begin{equation}
    z_{\pm}^{H}
    =
    \frac{-3\pm\sqrt{9+24\va_{\mathrm{safe}}^{\ast}}}{2}.
    \notag
\end{equation}
The complete reverse rule is
\begin{equation}
\vz^{\ast}=\begin{cases}
\displaystyle
\arg\min_{z\in\{z_-^H,z_+^H\}}|z-\hat z|,
& -\dfrac{3}{8}\le \va_{\mathrm{safe}}^{\ast}<0,\\[3mm]
\displaystyle
\arg\min_{z\in\{\min(\hat z,-3),0\}}|z-\hat z|,
& \va_{\mathrm{safe}}^{\ast}=0,\\[3mm]
z_+^H,
& 0<\va_{\mathrm{safe}}^{\ast}<3,\\
\va_{\mathrm{safe}}^{\ast},
& \va_{\mathrm{safe}}^{\ast}\ge3.
\end{cases}
    \notag
\end{equation}
The zero output has the non-unique preimage $z\le-3$ together with $z=0$; the first candidate is the closest point in the left half-line to $\hat z$. If the two candidates are equidistant, the strict comparison in the implementation selects $z=0$.

\textbf{(4) Coupled activation: Softmax.}
Unlike the preceding element-wise activations, Softmax couples all entries along its normalization dimension $\mathcal D$:
\begin{equation}
    a_i
    =
    \frac{\exp(z_i)}
    {\sum_{j\in\mathcal D}\exp(z_j)}.
    \notag
\end{equation}
For a strictly positive normalized target $\va^{\ast}$, every exact preimage has the form
\begin{equation}
    z_i^{\ast}
    =
    \log a_i^{\ast}+c,
    \notag
\end{equation}
because adding a common constant does not change the Softmax output. MDP determines the free shift by
\begin{equation}
    c^{\ast}
    =
    \arg\min_c
    \sum_{i\in\mathcal D}
    \left(
        \log a_i^{\ast}+c-\hat z_i
    \right)^2
    =
    \operatorname{Mean}_{i\in\mathcal D}
    \left(
        \hat z_i-\log a_i^{\ast}
    \right).
    \notag
\end{equation}
Consequently,
\begin{equation}
    z_i^{\ast}
    =
    \log a_i^{\ast}
    +
    \operatorname{Mean}_{j\in\mathcal D}
    \left(
        \hat z_j-\log a_j^{\ast}
    \right).
    \label{supp:eq:supp_reverse_softmax}
\end{equation}

In the numerical implementation, any negative target element is invalid and raises a \texttt{ValueError}; a Softmax normalization slice whose total is not positive also raises a \texttt{ValueError}. Otherwise, each probability is first lower-bounded by
$\epsilon=10^{-12}$ and the resulting vector is renormalized,
\begin{equation}
    \widetilde a_i^{\ast}
    =
    \frac{\max(a_i^{\ast},\epsilon)}
    {\sum_{j\in\mathcal D}\max(a_j^{\ast},\epsilon)}.
    \notag
\end{equation}
Equation~\eqref{supp:eq:supp_reverse_softmax} is then applied with
$\widetilde a_i^{\ast}$ in place of $a_i^{\ast}$. This produces a finite approximation when the desired Softmax target contains exact zeros.

The constants used by these operations inherit the working dtype of the target tensor; the activation reverse functions do not themselves force FP32. In particular, in FP16, $1-10^{-6}$ may round to $1$ and $10^{-12}$ may underflow to zero. If the caller does not enforce FP32, these effects are part of the numerical implementation limits. Overall, the activation reverse operators exploit analytical inverses whenever the activation is one-to-one, use the former forward pre-activation to resolve non-unique preimages, and map unattainable targets to feasible and numerically stable surrogates. These reverse operators satisfy CCP exactly with respect to the original target when that target is attainable, remains unchanged by numerical stabilization, and the adopted reverse rule is exact. When clipping, renormalization, finite-interval root search, or the stabilized LeakyReLU rule is applied, CCP is satisfied with respect to the resulting numerically safe target, up to the corresponding numerical approximation error. MDP is then used to resolve non-unique preimages whenever multiple valid candidates are available.

\subsection{Detailed Derivation for Nearest Embedding Solution}
\label{supp:sec:kkt_derivation}

For a classification output $\hat{\va}_{out}\in\R^{n_{out}}$ and ground-truth class $y^{\ast}$, the $L_2$ nearest-embedding problem in the main paper is
\begin{align}
    \min_{\va_{out}}\quad
    &\frac{1}{2}\|\va_{out}-\hat{\va}_{out}\|_2^2
    \label{supp:eq:kkt_primal}\\
    \mathrm{s.t.}\quad
    &(\va_{out})_i-(\va_{out})_{y^{\ast}}\le 0,
    \quad i\ne y^{\ast}.
    \nonumber
\end{align}
The objective is strictly convex and the feasible set is a nonempty closed convex polyhedron; hence, Eq.~\eqref{supp:eq:kkt_primal} admits a unique minimizer. Introduce multipliers $\mu_i\ge0$ for the ordering constraints. The Lagrangian is
\begin{equation}
    L(\va_{out},\boldsymbol{\mu})
    =
    \frac{1}{2}\|\va_{out}-\hat{\va}_{out}\|_2^2
    +
    \sum_{i\ne y^{\ast}}
    \mu_i\big((\va_{out})_i-(\va_{out})_{y^{\ast}}\big).
\end{equation}
Since the gradients of any active constraints, $\{\mathbf{e}_i-\mathbf{e}_{y^{\ast}}\}$, are linearly independent, there exists a unique KKT pair $(\va_{out}^{NE},\boldsymbol{\mu}^{\ast})$ satisfying
\begin{align}
    (\va_{out}^{NE})_{y^{\ast}}-(\hat{\va}_{out})_{y^{\ast}}
    -\sum_{i\ne y^{\ast}}\mu_i^{\ast}&=0,
    \label{supp:eq:kkt_stationarity_y}\\
    (\va_{out}^{NE})_i-(\hat{\va}_{out})_i+\mu_i^{\ast}&=0,
    \quad i\ne y^{\ast},
    \label{supp:eq:kkt_stationarity_i}\\
    (\va_{out}^{NE})_i-(\va_{out}^{NE})_{y^{\ast}}&\le0,
    \quad i\ne y^{\ast},
    \label{supp:eq:kkt_feasibility}\\
    \mu_i^{\ast}&\ge0,
    \quad i\ne y^{\ast},
    \label{supp:eq:kkt_dual_feasibility}\\
    \mu_i^{\ast}\big((\va_{out}^{NE})_i-(\va_{out}^{NE})_{y^{\ast}}\big)&=0,
    \quad i\ne y^{\ast}.
    \label{supp:eq:kkt_complementarity}
\end{align}

Let $\mathcal{A}$ denote the active set of competing classes satisfying
$(\va_{out}^{NE})_i=(\va_{out}^{NE})_{y^{\ast}}$. For $i\notin\mathcal{A}$, complementary slackness gives $\mu_i^{\ast}=0$, and Eq.~\eqref{supp:eq:kkt_stationarity_i} yields $(\va_{out}^{NE})_i=(\hat{\va}_{out})_i$. For $i\in\mathcal{A}$, denote the shared projected value $(\va_{out}^{NE})_{y^{\ast}}$ by $\theta$. Equations~\eqref{supp:eq:kkt_stationarity_y} and~\eqref{supp:eq:kkt_stationarity_i} then give
\begin{equation}
    \theta
    =
    \frac{
    (\hat{\va}_{out})_{y^{\ast}}
    +
    \sum_{i\in\mathcal{A}}(\hat{\va}_{out})_i
    }{|\mathcal{A}|+1}.
    \label{supp:eq:supp_oe_threshold}
\end{equation}
Dual feasibility requires $(\hat{\va}_{out})_i\ge\theta$ for $i\in\mathcal{A}$, whereas primal feasibility of the unchanged components requires $(\hat{\va}_{out})_i\le\theta$ for $i\notin\mathcal{A}$.

To determine the active set, sort the components exceeding the original ground-truth component as $v_1\ge\cdots\ge v_m$ and define
\begin{equation}
    \theta_k
    =
    \frac{(\hat{\va}_{out})_{y^{\ast}}+\sum_{s=1}^{k}v_s}{k+1}.
\end{equation}
If $m=0$, $\hat{\va}_{out}$ is already feasible. Otherwise, define $k^{\ast}$ as the maximal index satisfying
\begin{equation}
    v_{k^{\ast}}\ge\theta_{k^{\ast}}.
    \label{supp:eq:supp_oe_active_set}
\end{equation}
By maximality,
\begin{equation}
    v_{k^{\ast}+1}<\theta_{k^{\ast}},
\end{equation}
whenever $k^{\ast}<m$. Thus, $k^{\ast}$ uniquely identifies the active set, consistently with the unique KKT solution. The ground-truth component and the $k^{\ast}$ active competitors are set to $\theta_{k^{\ast}}$, while all remaining components retain their original values. This yields exactly the nearest-embedding solution implemented by Algorithm~1 in the main paper.

\subsection{Layer-Wise Error Propagation}
\label{supp:sec:proof_error_accumulation}
We now distinguish the ideal reverse map from its actual implementation. Let $\mathcal{R}_l$ denote the ideal reconstruction map for reverse unit $l$, and let $\widetilde{\mathcal{R}}_l$ denote the implemented map. Define the ideal and implemented reverse chains by
$\va_l^{\ast}=\mathcal{R}_l(\va_{l+1}^{\ast})$ and $\widetilde{\va}_l =\widetilde{\mathcal{R}}_l(\widetilde{\va}_{l+1})$.
Let $e_l =\|\widetilde{\va}_l-\va_l^{\ast}\|$.

\begin{suppTheorem}[Layer-wise reverse-reconstruction error]
\label{supp:cor:supp_error_accumulation}
Assume that, on the relevant target neighborhood, the ideal reverse map satisfies
\begin{equation}
    \|\mathcal{R}_l(\mathbf{u})-\mathcal{R}_l(\mathbf{v})\| \le \rho_l\|\mathbf{u}-\mathbf{v}\|
    \notag
\end{equation}
and that the implementation error is uniformly bounded by
\begin{equation}
    \|\widetilde{\mathcal{R}}_l(\mathbf{u})-\mathcal{R}_l(\mathbf{u})\|
    \le
    \varepsilon_l.
    \notag
\end{equation}
Then
\begin{equation}
    e_l
    \le
    \rho_l e_{l+1}+\varepsilon_l.
    \label{supp:eq:supp_error_recursion}
\end{equation}
Consequently, for any $k\in\{1, \ldots,L_{out}\}$,
\begin{equation}
    e_k
    \le
    \left(\prod_{j=k}^{L_{out}-1}\rho_j\right)e_{L_{out}}
    +
    \sum_{i=k}^{L_{out}-1}
    \left(\prod_{j=k}^{i-1}\rho_j\right)\varepsilon_i,
    \label{supp:eq:supp_error_accumulation}
\end{equation}
where an empty product equals one.
\end{suppTheorem}

\begin{proof}
Adding and subtracting $\mathcal{R}_l(\widetilde{\va}_{l+1})$ gives
\begin{equation}
    e_l
    =
    \|\widetilde{\mathcal{R}}_l(\widetilde{\va}_{l+1})
      -\mathcal{R}_l(\va_{l+1}^{\ast})\|
    \le
    \|\widetilde{\mathcal{R}}_l(\widetilde{\va}_{l+1})
      -\mathcal{R}_l(\widetilde{\va}_{l+1})\|
     +
    \|\mathcal{R}_l(\widetilde{\va}_{l+1})
      -\mathcal{R}_l(\va_{l+1}^{\ast})\|
    \le
    \varepsilon_l+\rho_l e_{l+1},
    \notag
\end{equation}
which proves Eq.~\eqref{supp:eq:supp_error_recursion}. Repeated substitution of this recursion yields Eq.~\eqref{supp:eq:supp_error_accumulation}.
\end{proof}

For an MDP-centered Tikhonov linear reverse map, the target-sensitivity factor $\rho_l$ is bounded by Theorem~\ref{supp:thm:supp_tikhonov_stability}:
\begin{equation}
    \rho_l
    \le
    \frac{1}{2\sqrt{\alpha_l}}.
    \notag
\end{equation}
For an unregularized full-rank linear reverse, the corresponding factor is $1/\sigma_{\min,l}$, which can be arbitrarily large for a nearly singular operator.

The implementation error $\varepsilon_l$ can be decomposed according to its source:
\begin{equation}
    \varepsilon_l
    \le
    \varepsilon_l^{\mathrm{reg}}
    +\varepsilon_l^{\mathrm{solve}}
    +\varepsilon_l^{\mathrm{boundary}}
    +\varepsilon_l^{\mathrm{iter}}
    +\varepsilon_l^{\mathrm{model}}.
    \notag
\end{equation}
Here $\varepsilon_l^{\mathrm{reg}}$ is the bias introduced by the MDP penalty, $\varepsilon_l^{\mathrm{solve}}$ covers floating-point and finite-CG errors, $\varepsilon_l^{\mathrm{boundary}}$ is the FFT boundary approximation, $\varepsilon_l^{\mathrm{iter}}$ is the finite Gauss--Newton or VJP optimization error, and $\varepsilon_l^{\mathrm{model}}$ covers approximate reverse rules such as pooling interpolation. The decomposition is additive and need not be unique; its purpose is to identify the distinct mechanisms that contribute to $\varepsilon_l$.

If $\rho_l\le\bar\rho$ and $\varepsilon_l\le\bar\varepsilon$ along a reverse path of depth $D$, then
\begin{equation}
    e_{L_{out}-D}
    \le
    \bar\rho^D e_{L_{out}}
    +
    \bar\varepsilon
    \sum_{q=0}^{D-1}\bar\rho^q.
    \notag
\end{equation}
Thus $\bar\rho<1$ yields a uniformly bounded geometric series, $\bar\rho=1$ yields at most linear accumulation, and $\bar\rho>1$ permits geometric amplification. This result motivates limiting the reverse depth and activating regularization at layers whose nominal inverse is numerically unstable.

Finally, suppose the forward unit $\Phi_l$ is locally $\rho_l$-Lipschitz and the ideal reconstruction satisfies exact CCP, $\Phi_l(\va_l^{\ast})=\va_{l+1}^{\ast}$. Then the implemented CCP error is bounded by
\begin{equation}
    \|\Phi_l(\widetilde{\va}_l)-\widetilde{\va}_{l+1}\|
    \le\|\Phi_l(\widetilde{\va}_l)-\Phi_l(\va_l^{\ast})\| + \|\va_{l+1}^{\ast}-\widetilde{\va}_{l+1}\|\le
    \rho_l e_l+e_{l+1}.
    \notag
\end{equation}
The implemented MDP deviation from the former feature obeys
\begin{equation}
    \|\widetilde{\va}_l-\hat{\va}_l\|
    \le
    \|\va_l^{\ast}-\hat{\va}_l\|+e_l.
    \notag
\end{equation}

\section{Detailed Statistical Analysis on Image Classification Post-Training Experiments}
\label{supp:sec:experimental_settings_statistics}

This part complements the compact classification tables in the main paper. It reports the complete architectures, paired-seed protocol, paired $t$-tests, Wilcoxon signed-rank tests, bootstrap confidence intervals, and paired effect sizes. The reconstruction-loss coefficient is denoted by $c_{\mathrm{rec}}$ throughout this part to distinguish it from the Tikhonov coefficient $\alpha$ used in Part~A.
\subsection{Implementation Details}
\label{supp:sec:supp_implementation_details}

\subsubsection{Hardware configuration}
All experiments were executed on a Supermicro SYS-420GP-TNAR+ server running Ubuntu 22.04.5 LTS. The machine contains two Intel Xeon Platinum 8358P processors (32 physical cores per socket, 128 logical CPUs in total), approximately 2~TiB of system memory, and eight NVIDIA A100-SXM4 GPUs with 80~GB memory per GPU. The recorded NVIDIA driver is 570.211.01 with CUDA 12.8 compatibility. These are the hardware characteristics relevant to the reported training and reconstruction workloads.

\subsubsection{Feature-reconstruction hyperparameters}
The following settings are taken from the final reverse-computation code; when a function default differs from the value supplied by its dispatcher, the dispatched value is reported.

\textbf{Adaptive reconstruction-loss weighting.} The exponential moving averages (EMAs) of the task and reconstruction losses are updated as
\begin{equation}
    \bar{\mathcal{L}}_{q}^{(t)}
    =
    \beta \bar{\mathcal{L}}_{q}^{(t-1)}
    +(1-\beta)\mathcal{L}_{q}^{(t)},
    \qquad q\in\{\mathrm{task},\mathrm{rec}\},
\end{equation}
where $\beta=0.9$ is the EMA decay factor. The resulting $\bar{\mathcal{L}}_{task}^{(t)}$ and $\bar{\mathcal{L}}_{rec}^{(t)}$ are used only to compute $\lambda_{rec}^{(t)}$ and are detached from gradient propagation. This smoothing reduces batch-wise fluctuations in the relative loss scale.

\textbf{Common reliability and regularization settings.} For the linear models used to reconstruct features through linear maps, FFT-based convolutional maps, and composite blocks, we use $\mathbf{A}$ as the linear transformation, $\mathbf{y}=\mathbf{A}\mathbf{x}$ as the bias-free target, and $\hat{\mathbf{x}}$ and $\mathbf{x}^{\ast}$ as the original and reconstructed features, respectively. We use a common absolute feature bound of $M_{\infty}=10^{3}$.
In addition, the maximum deviation/amplification ratio in Eq.~\ref{supp:eq:gate_dev} is $M_{\mathrm{dev}}=10^{2}$ wherever the corresponding reliability checks are applicable.

The normalized consistency thresholds in Eqs.~\ref{supp:eq:gate_res} and~\ref{supp:eq:gate_opt} are set to $M_{1}=10^{-4}$ and $M_{2}=10^{-3}$ in double precision, and to $M_{1}=10^{-3}$ and $M_{2}=10^{-2}$ in single precision.
The common Tikhonov scale is $\kappa_{\mathrm{eff}}=10^{3}$. No geometric retries for increasing $\alpha$ are used. For the dense SVD fallback,
\begin{equation}
    \alpha=\max\left\{
    (\sigma_{\max}/10^3)^2,
    \epsilon_{\mathrm{mach}}\max(\sigma_{\max}^2,1),
    \left(\frac{\|\vy-\mA\hat{\vx}\|_2}
    {2(10^3-\|\hat{\vx}\|_{\infty})}\right)^2
    \right\},
    \label{supp:eq:supp_feature_alpha}
\end{equation}
where the last term is used only when the magnitude margin is positive. The same construction is applied independently to every failed FFT frequency--batch pair. The linear/FFT matrix reliability tests use $\eta_{\mathrm{eq}}\le10^{-4}$ and $\eta_{\mathrm{ls}}\le10^{-3}$ in float64/complex128, and $10^{-3}$ and $10^{-2}$, respectively, in float32/complex64. Their reference-norm floor for the relative-deviation test is $10^{-2}\sqrt{n}$.

\textbf{Linear layers.} Case I uses the direct Gram correction $\mA^T(\mA\mA^T)^{-1}(\vy-\mA\hat{\vx})$; Case II uses \texttt{torch.linalg.lstsq} with the \texttt{gels} driver. Failed samples are replaced by the per-sample MDP-centered SVD Tikhonov solution with the common settings above.

\textbf{Convolutional Matrix-Solve.}
Its Case-I nominal solve uses $\mA\mA^T$ directly, while Case II first attempts direct least squares and uses the centered normal equation if the least-squares backend raises an error. For operator-form reliability checks, the absolute threshold and deviation ratio are again $10^3$ and $10^2$, the reference floor is $10^{-6}\sqrt{N_i}$, the equality-residual tolerance is $10^{-4}$, and the least-squares optimality tolerance is $10^{-3}$.

\textbf{Convolutional Iterative-CG.}
The convolution operator norm is estimated with 10 power iterations. Iterative-CG is dispatched with at most 100 iterations and relative CG tolerance $10^{-9}$. If the nominal CG candidate is rejected, damping is the larger of $(\widehat{\|A\|}_2/10^3)^2$ and the feature-magnitude lower bound in Eq.~\eqref{supp:eq:supp_alpha_deviation_bound}.

\textbf{FFT-Boundary and FFT-Padded.} FFT arithmetic is performed in float64/complex128 before the reconstructed real feature is cast back to its original dtype. Each frequency uses a $C_{l+1}\times C_l$ channel system. Case I uses a direct solve when square and an $AA^H$ nearest-anchor correction otherwise; Case II uses complex least squares with \texttt{gels}. Reliability is tested independently for each frequency--batch pair with $\max|q|\le10^3$, deviation ratio $10^2$, equality tolerance $10^{-4}$, and least-squares optimality tolerance $10^{-3}$. Failed pairs alone use the SVD Tikhonov filter with $\kappa_{\mathrm{eff}}=10^3$; there is no additional frequency-specific cap and no post-IFFT retry.

\textbf{Composite blocks.} The automatic dispatcher materializes the Jacobian only when the estimated number of entries does not exceed $5\times10^7$; otherwise it selects VJP/Adam. The explicit-Jacobian route uses at most 20 iterations, initial damping $0.8$, maximum step norm 10, relative stopping tolerance $10^{-6}$, linearized least-squares optimality tolerance $10^{-3}$, maximum elementwise step magnitude $10^3$, maximum normalized step amplification $10^2$, maximum feature magnitude $10^3$, and maximum feature-deviation ratio $10^2$. Backtracking performs at most 8 trials and multiplies the step scale by $0.8$ after each rejected trial. Its Tikhonov fallback uses $\kappa_{\mathrm{eff}}=10^3$ and a regularized optimality tolerance of $10^{-2}$. The VJP route uses Adam with learning rate $10^{-2}$ (the dispatcher passes $\min(0.8,10^{-2})$), at most 20 iterations, maximum per-update norm 10, relative stopping tolerance $10^{-6}$, and MDP-centered projection radius $10^2$ times the anchor norm with a floor of $10^{-4}\sqrt{n}$. The returned VJP feature is finally checked with $\max|a|\le10^3$ and the $10^2$ deviation ratio; failure returns the former feature rather than launching a second penalized optimization.

\textbf{Activation, normalization, and pooling auxiliaries.} Tanh, sigmoid, ELU, and Softplus inverses use numerical range clamps with $\epsilon=10^{-6}$; GELU and SiLU use 80 bisection iterations; Softmax uses probability floor $10^{-12}$. BatchNorm inversion uses the layer's stored $\epsilon_{BN}$ and running statistics in evaluation mode. Pooling reverse computation requires stride equal to kernel size and uses nearest-neighbor interpolation to the original spatial size.

\subsection{All network architectures}

This section reports the network architectures used in Sec.~IV.A of the main paper. All tensor sizes are written without the batch dimension unless otherwise stated.

\paragraph{SimpleCNNs for image classification.}
The SimpleCNN models are plain convolutional networks with valid convolutions, max-pooling, and fully connected classifiers. Here, Conv$(c_{\mathrm{in}},c_{\mathrm{out}},k)$ denotes a 2D convolution with stride $1$, padding $0$, and kernel size $k$. FC$(d_{\mathrm{in}},d_{\mathrm{out}})$ denotes a fully connected layer. Every convolutional layer and hidden fully connected layer is followed by a ReLU activation. Max-pooling uses kernel size $2$ and stride $2$. The final linear classifier produces logits and has no activation. All convolutional and linear weights are initialized by Xavier uniform initialization in the code. The complete layer-wise architectures of the SimpleCNN models for all
evaluated datasets are reported in
Table~\ref{supp:tab:simplecnn_arch}.

\begin{table}[H]
\centering
\small
\caption{SimpleCNN architectures.}
\label{supp:tab:simplecnn_arch}
\begin{tabular}{l l l}
\toprule
Dataset & Layer & Output feature size \\
\midrule
MNIST & Input & $1 \times 28 \times 28$ \\
 & Conv$(1,3,5)$ + ReLU & $3 \times 24 \times 24$ \\
 & MaxPool$(2,2)$ & $3 \times 12 \times 12$ \\
 & Conv$(3,3,5)$ + ReLU & $3 \times 8 \times 8$ \\
 & MaxPool$(2,2)$ & $3 \times 4 \times 4$ \\
 & Flatten & $48$ \\
 & FC$(48,10)$ & $10$ \\
\midrule
CIFAR-10 & Input & $3 \times 32 \times 32$ \\
 & Conv$(3,10,5)$ + ReLU & $10 \times 28 \times 28$ \\
 & MaxPool$(2,2)$ & $10 \times 14 \times 14$ \\
 & Conv$(10,10,3)$ + ReLU & $10 \times 12 \times 12$ \\
 & MaxPool$(2,2)$ & $10 \times 6 \times 6$ \\
 & Conv$(10,10,3)$ + ReLU & $10 \times 4 \times 4$ \\
 & Flatten & $160$ \\
 & FC$(160,128)$ + ReLU & $128$ \\
 & FC$(128,10)$ & $10$ \\
\midrule
CIFAR-100 & Input & $3 \times 32 \times 32$ \\
 & Conv$(3,18,5)$ + ReLU & $18 \times 28 \times 28$ \\
 & MaxPool$(2,2)$ & $18 \times 14 \times 14$ \\
 & Conv$(18,18,3)$ + ReLU & $18 \times 12 \times 12$ \\
 & MaxPool$(2,2)$ & $18 \times 6 \times 6$ \\
 & Conv$(18,18,3)$ + ReLU & $18 \times 4 \times 4$ \\
 & Flatten & $288$ \\
 & FC$(288,256)$ + ReLU & $256$ \\
 & FC$(256,100)$ & $100$ \\
\midrule
Imagenette/Imagewoof & Input & $3 \times 224 \times 224$ \\
 & Conv$(3,16,25)$ + ReLU & $16 \times 200 \times 200$ \\
 & MaxPool$(2,2)$ & $16 \times 100 \times 100$ \\
 & Conv$(16,16,17)$ + ReLU & $16 \times 84 \times 84$ \\
 & MaxPool$(2,2)$ & $16 \times 42 \times 42$ \\
 & Conv$(16,16,11)$ + ReLU & $16 \times 32 \times 32$ \\
 & MaxPool$(2,2)$ & $16 \times 16 \times 16$ \\
 & Conv$(16,16,5)$ + ReLU & $16 \times 12 \times 12$ \\
 & MaxPool$(2,2)$ & $16 \times 6 \times 6$ \\
 & Conv$(16,16,3)$ + ReLU & $16 \times 4 \times 4$ \\
 & Flatten & $256$ \\
 & FC$(256,128)$ + ReLU & $128$ \\
 & FC$(128,10)$ & $10$ \\
\midrule
Tiny-ImageNet & Input & $3 \times 64 \times 64$ \\
 & Conv$(3,45,5)$ + ReLU & $45 \times 60 \times 60$ \\
 & MaxPool$(2,2)$ & $45 \times 30 \times 30$ \\
 & Conv$(45,45,3)$ + ReLU & $45 \times 28 \times 28$ \\
 & MaxPool$(2,2)$ & $45 \times 14 \times 14$ \\
 & Conv$(45,45,3)$ + ReLU & $45 \times 12 \times 12$ \\
 & MaxPool$(2,2)$ & $45 \times 6 \times 6$ \\
 & Conv$(45,45,3)$ + ReLU & $45 \times 4 \times 4$ \\
 & Flatten & $720$ \\
 & FC$(720,512)$ + ReLU & $512$ \\
 & FC$(512,200)$ & $200$ \\
\bottomrule
\end{tabular}
\end{table}

\paragraph{ResNets for image classification.}
The ResNet models use the torchvision ResNet implementation with \texttt{BasicBlock}. ResNet-18 contains $[2,2,2,2]$ residual blocks in stages 1--4, and ResNet-34 contains $[3,4,6,3]$ residual blocks. Each residual block has two $3 \times 3$ convolutional layers, each followed by batch normalization; the first convolution is followed by ReLU, and another ReLU is applied after adding the residual branch. When the spatial size or channel dimension changes, the skip branch uses a $1 \times 1$ convolution followed by batch normalization. The classifier is a global average pooling layer followed by a linear layer.

For CIFAR-10 and CIFAR-100, the code uses a CIFAR-style stem: Conv$(3,64,3)$ with stride $1$ and padding $1$, followed by batch normalization and ReLU, and the max-pooling layer is replaced by identity. For Imagenette, Imagewoof, and Tiny-ImageNet, the standard ImageNet-style stem is used: Conv$(3,64,7)$ with stride $2$ and padding $3$, followed by batch normalization, ReLU, and MaxPool$(3,2)$ with padding $1$.

\paragraph{SimpleViTs for image classification.}
The SimpleViT models are used for CIFAR-100 and Tiny-ImageNet. Both use patch embedding implemented as a convolution whose kernel size and stride equal the patch size, a learnable class token, learnable positional embeddings, nine Transformer blocks, and a linear classification head. The embedding dimension is $192$, the number of attention heads is $12$, the MLP ratio is $2.0$, and dropout is $0$. Unlike the standard pre-normalized ViT block, the implemented block has no layer normalization. Each block applies multi-head self-attention with a residual connection, followed by an MLP with Linear$(192,384)$, GELU, Dropout$(0)$, Linear$(384,192)$, and a second residual connection. The complete layer-wise architectures of the SimpleViT
models for CIFAR-100 and Tiny-ImageNet are reported in
Table~\ref{supp:tab:simplevit_arch}.

\begin{table}[H]
\centering
\small
\caption{SimpleViT architectures.}
\label{supp:tab:simplevit_arch}
\begin{tabular}{l l l}
\toprule
Dataset & Layer & Output feature size \\
\midrule
CIFAR-100 & Input & $3 \times 32 \times 32$ \\
 & Patch embedding Conv$(3,192,4)$, stride $4$ & $192 \times 8 \times 8$ \\
 & Flatten patches and transpose & $64 \times 192$ \\
 & Add class token and positional embedding & $65 \times 192$ \\
 & Transformer blocks 1--9 & $65 \times 192$ after each block \\
 & Select class token & $192$ \\
 & FC$(192,100)$ & $100$ \\
\midrule
Tiny-ImageNet & Input & $3 \times 64 \times 64$ \\
 & Patch embedding Conv$(3,192,8)$, stride $8$ & $192 \times 8 \times 8$ \\
 & Flatten patches and transpose & $64 \times 192$ \\
 & Add class token and positional embedding & $65 \times 192$ \\
 & Transformer blocks 1--9 & $65 \times 192$ after each block \\
 & Select class token & $192$ \\
 & FC$(192,200)$ & $200$ \\
\bottomrule
\end{tabular}
\end{table}

\paragraph{Trajectory planners for NAVSIM}
The complete layer-wise architectures of the ResNet-34 and ViT-based
NAVSIM trajectory planners are summarized in
Table~\ref{supp:tab:navsim_planner_architectures}.
Here, $\operatorname{Conv}(c_{\mathrm{in}},c_{\mathrm{out}},k,s,p)$ denotes a
2D convolution with input channels $c_{\mathrm{in}}$, output channels
$c_{\mathrm{out}}$, kernel size $k$, stride $s$, and padding $p$.
The two planners receive the same inputs and use the same status-conditioned
fusion module and trajectory-prediction head, but employ different visual
backbones and image-projection layers. Each model receives a stitched camera
image of size $3\times256\times1024$ and an 8-dimensional ego-status vector.
Before visual feature extraction, the camera image is normalized channel-wise
using the configured image mean and standard deviation.

The ResNet-based planner uses an explicitly implemented ResNet-34 with
$[3,4,6,3]$ BasicBlocks in stages 1--4. The stem consists of a $7\times7$
convolution with 64 output channels, stride 2, and padding 3, followed by
batch normalization, ReLU, and $3\times3$ max-pooling with stride 2 and
padding 1. Each BasicBlock contains two $3\times3$ convolutions followed by
batch normalization. ReLU is applied after the first convolutional branch
and after residual addition. When the spatial resolution or channel
dimension changes, the shortcut branch uses a $1\times1$ convolution followed
by batch normalization. The first blocks of stages 2--4 use stride 2 for
spatial downsampling.

For the $3\times256\times1024$ camera input, the final ResNet feature map has
size $512\times8\times32$. It is reduced to a fixed spatial grid by adaptive
average pooling and then flattened. A linear layer followed by layer
normalization and GELU maps the pooled visual feature to a 512-dimensional
image embedding. The pretrained classification head is retained only for
checkpoint compatibility and is bypassed during trajectory prediction. The
visual backbone is initialized from publicly available ResNet-34 weights
pretrained on ImageNet-1K.

The ViT-based planner uses an explicitly implemented ViT-B/16. Patch
embedding is implemented as a convolution with kernel size and stride
$16\times16$. The $3\times256\times1024$ input therefore produces an
embedding map of size $768\times16\times64$, corresponding to 1024 raw patch
tokens. To reduce the memory cost of self-attention while preserving the
wide field of view of the stitched camera input, the embedding map is
adaptively average-pooled to $768\times8\times32$. The resulting 256 patch
tokens are flattened and transposed, after which a learnable class token is
prepended to form a sequence of 257 tokens with embedding dimension 768.

Learnable absolute positional embeddings are added to the token sequence.
When the pretrained checkpoint is loaded, its patch positional embeddings
are bicubically interpolated to the $8\times32$ token grid, while the
class-token positional embedding is retained. The encoder contains 12
pre-normalized Transformer blocks. Each block applies layer normalization,
12-head self-attention, and a residual connection, followed by a second
layer normalization and an MLP with Linear$(768,3072)$, GELU, and
Linear$(3072,768)$. Each attention head has dimension 64. After the final
layer normalization, the 768-dimensional class-token representation is
selected and projected to a 512-dimensional image embedding using a single
linear layer. No additional normalization or activation is applied after
this projection. The pretrained classification head is retained only for
checkpoint compatibility and is bypassed during trajectory prediction. The
visual backbone is initialized from publicly available ViT-B/16 weights
pretrained on ImageNet-21K and subsequently fine-tuned on ImageNet-1K.

In both planners, the 8-dimensional ego-status vector is processed by a
two-layer status encoder. Each layer consists of a linear transformation,
layer normalization, and GELU. The resulting status context is passed through
two independent linear heads to generate the 512-dimensional modulation
vectors $\boldsymbol{\gamma}$ and $\boldsymbol{\beta}$. The image and status
features are fused through the element-wise affine modulation
\begin{equation}
\mathbf{z}_{\mathrm{fusion}}
=
\left[
\mathbf{1}
+
\alpha_{\mathrm{mod}}\tanh(\boldsymbol{\gamma})
\right]
\odot\mathbf{e}_{\mathrm{img}}
+
\boldsymbol{\beta},
    \notag
\end{equation}
where $\mathbf{e}_{\mathrm{img}}\in\mathbb{R}^{512}$ is the image embedding,
$\alpha_{\mathrm{mod}}$ is the configured modulation scale, and $\odot$
denotes element-wise multiplication. The weights of the
$\boldsymbol{\gamma}$ and $\boldsymbol{\beta}$ heads are initialized from a
zero-mean Gaussian distribution with standard deviation $10^{-3}$, and their
biases are initialized to zero. This initialization makes the initial fused
representation close to the image-only representation while retaining
gradients through the status branch.

The fused feature is processed by two planning layers with hidden dimension
256. Each planning layer contains a linear transformation, layer
normalization, and LeakyReLU with negative slope 0.5. The final linear
trajectory head produces 24 scalar values without an additional activation.
These values are reshaped into eight future poses:
\begin{equation}
\widehat{\boldsymbol{\tau}}
=
\left[
(\hat{x}_1,\hat{y}_1,\hat{\theta}_1),\ldots,
(\hat{x}_8,\hat{y}_8,\hat{\theta}_8)
\right]
\in\mathbb{R}^{8\times3}.
    \notag
\end{equation}
The eight predicted poses cover a future horizon of $4\,\mathrm{s}$ with a
sampling interval of $0.5\,\mathrm{s}$.

\begin{table}[H]
\centering
\caption{NAVSIM trajectory-planner architectures.
$\operatorname{FC}(d_{\mathrm{in}},d_{\mathrm{out}})$ denotes a fully
connected layer. $H_p$ and $W_p$ denote the configured output size of the
ResNet adaptive pooling layer, while $d_m$ and $d_c$ denote the intermediate
and output dimensions of the status encoder. All tensor sizes exclude the
batch dimension.}
\label{supp:tab:navsim_planner_architectures}
\small
\setlength{\tabcolsep}{5pt}
\renewcommand{\arraystretch}{0.96}
\begin{tabular}{llll}
\toprule
Model & Module & Layer & Output size \\
\midrule
\multirow{20}{*}{\makecell[l]{Model I\\ResNet-34 as image backbone}} & Image input & Camera input & $3\times256\times1024$ \\
& \multirow{8}{*}{Image backbone} & Conv$(3,64,7,2,3)$ + BN + ReLU & $64\times128\times512$ \\
& & MaxPool$(3,2,1)$ & $64\times64\times256$ \\
& & Stage 1: BasicBlock$(64)\times3$ & $64\times64\times256$ \\
& & Stage 2: BasicBlock$(128)\times4$ & $128\times32\times128$ \\
& & Stage 3: BasicBlock$(256)\times6$ & $256\times16\times64$ \\
& & Stage 4: BasicBlock$(512)\times3$ & $512\times8\times32$ \\
& & AdaptiveAvgPool$(H_p,W_p)$ & $512\times H_p\times W_p$ \\
& & Flatten & $512H_pW_p$ \\
& Image projection & FC$(512H_pW_p,512)$ + LN + GELU & $512$ \\
& \multirow{3}{*}{Status encoder} & Ego-status input & $8$ \\
& & FC$(8,d_m)$ + LN + GELU & $d_m$ \\
& & FC$(d_m,d_c)$ + LN + GELU & $d_c$ \\
& \multirow{2}{*}{Modulation heads} & $\gamma$ head: FC$(d_c,512)$ & $512$ \\
& & $\beta$ head: FC$(d_c,512)$ & $512$ \\
& Feature fusion & Status-conditioned affine modulation & $512$ \\
& \multirow{2}{*}{Planning layers} & FC$(512,256)$ + LN + LeakyReLU$(0.5)$ & $256$ \\
& & FC$(256,256)$ + LN + LeakyReLU$(0.5)$ & $256$ \\
& \multirow{2}{*}{Trajectory head} & FC$(256,24)$ & $24$ \\
& & Reshape & $8\times3$ \\
\midrule
\multirow{20}{*}{\makecell[l]{Model II\\ViT-B/16 as image backbone}} & Image input & Camera input & $3\times256\times1024$ \\
& \multirow{8}{*}{Image backbone} & Patch embedding: Conv$(3,768,16,16,0)$ & $768\times16\times64$ \\
& & AdaptiveAvgPool$(8,32)$ & $768\times8\times32$ \\
& & Flatten and transpose & $256\times768$ \\
& & Prepend learnable class token & $257\times768$ \\
& & Add learnable positional embeddings & $257\times768$ \\
& & Transformer block $\times12$ & $257\times768$ \\
& & Final layer normalization & $257\times768$ \\
& & Class-token extraction & $768$ \\
& Image projection & FC$(768,512)$ & $512$ \\
& \multirow{3}{*}{Status encoder} & Ego-status input & $8$ \\
& & FC$(8,d_m)$ + LN + GELU & $d_m$ \\
& & FC$(d_m,d_c)$ + LN + GELU & $d_c$ \\
& \multirow{2}{*}{Modulation heads} & $\gamma$ head: FC$(d_c,512)$ & $512$ \\
& & $\beta$ head: FC$(d_c,512)$ & $512$ \\
& Feature fusion & Status-conditioned affine modulation & $512$ \\
& \multirow{2}{*}{Planning layers} & FC$(512,256)$ + LN + LeakyReLU$(0.5)$ & $256$ \\
& & FC$(256,256)$ + LN + LeakyReLU$(0.5)$ & $256$ \\
& \multirow{2}{*}{Trajectory head} & FC$(256,24)$ & $24$ \\
& & Reshape & $8\times3$ \\
\bottomrule
\end{tabular}
\end{table}

\subsection{Multi-Seed Statistical Tests on FR-PT Experiments}
\label{supp:sec:multi_seed_statistics}

The main paper reports the mean and standard deviation of each classification setting over ten random seeds. This section complements these aggregate summaries with seed-wise paired inference. For every reconstruction layer $l_R$, evaluation metric, and seed, the reconstruction-weighted run and its reference share the pretrained checkpoint, trainable module, and random seed. Pairing removes between-seed variation that is common to both settings and directly measures the paired change associated with introducing the reconstruction term under the matched protocol.

Tables~\ref{supp:tab:mnist-mn-simplecnnv2-relu-0.9771-10e-paired-tests}--\ref{supp:tab:tin-tin-simplevit-p8-d192-l9-e60-0.4491-10e-paired-tests} provide the complete detailed analysis for 13 model--checkpoint combinations. They cover MNIST, CIFAR-10, CIFAR-100, Imagenette, Imagewoof, and Tiny-ImageNet, using SimpleCNNv2, ResNet-18, ResNet-34, and SimpleViT where applicable. Each table reports five evaluation metrics: the best accuracy observed during post-training, the mean accuracy over the post-training trajectory, and the accuracies at epochs 1, 5, and 10. All reported comparisons use $n=10$ paired seeds.

Let $a_s(c_{\mathrm{rec}})$ denote the accuracy for seed $s$ and reconstruction strength $c_{\mathrm{rec}}$. For each layer--metric pair,  $c_{\mathrm{rec}}^{\ast}$ denotes the setting with the highest mean accuracy. If \(c_{\mathrm{rec}}^{\ast}>0\), the reference is \(c_{\mathrm{rec}}=0\). If \(c_{\mathrm{rec}}^{\ast}=0\), the reference is the best-performing nonzero setting. 

The primary quantity in the table is the paired difference
\begin{equation}
    d_s  = a_s(c_{\mathrm{rec}}^{\ast})-a_s(c_{\mathrm{rec}}^{\mathrm{ref}}),
\end{equation}
computed independently for each seed. The reported $\Delta=\bar d$ is the sample mean of $\{d_s\}_{s=1}^{n}$, expressed in percentage points. A positive $\Delta$ means that the selected setting outperformed the stated reference on average, with the substantive direction determined by whether $c_{\mathrm{rec}}^{\ast}$ is nonzero or zero. The 95\% confidence interval is the bootstrap percentile interval for the mean paired difference. An interval entirely above zero indicates that the positive paired difference is consistently supported under bootstrap resampling. An interval that includes zero does not establish a consistently positive difference at the present sample size.

We additionally report two one-sided paired significance tests. The paired $t$-test evaluates
\begin{equation}
    H_0:\mu_d\le 0
    \quad \mathrm{versus}\quad
    H_1:\mu_d>0,
\end{equation}
where $\mu_d$ is the population mean paired difference. Its statistic is
\begin{equation}
    t
    =
    \frac{\bar{d}}{s_d/\sqrt{n}},
    \quad df=n-1,
\end{equation}
where $\bar{d}$ and $s_d$ are the sample mean and standard deviation of the paired differences. Here $s_d$ is distinct from the per-setting standard deviations reported in the main tables. The paired $t$-test assumes that the seed-wise differences are approximately normally distributed. The Wilcoxon signed-rank test provides a complementary rank-based analysis that does not require normality of the paired differences, although its conventional location-shift interpretation assumes approximate symmetry of the nonzero differences. In each table, $W$ is the signed-rank statistic and $p_W$ is its one-sided $p$-value. Agreement between small $p_t$ and $p_W$ values provides stronger evidence that the result is not driven solely by the parametric mean-based test.

The effect size $d_z$ is the paired-sample standardized mean difference,
\begin{equation}
    d_z
    =
    \frac{\bar{d}}{s_d}
    =
    \frac{t}{\sqrt{n}}.
\end{equation}
Unlike a $p$-value, which depends strongly on $n$, $d_z$ expresses the paired gain relative to the seed-to-seed variability of that gain. The tables should therefore be read using four complementary quantities. The value of $\Delta$ gives practical magnitude in percentage points, the confidence interval describes uncertainty, $p_t$ and $p_W$ assess directional consistency, and $d_z$ gives standardized magnitude. Stronger evidence is provided when $\Delta$ is practically meaningful, the confidence interval is above zero, both tests give small $p$-values, and $d_z$ is appreciable. A small $p$-value or large $d_z$ does not by itself imply a practically important accuracy increase.

These analyses are descriptive rather than confirmatory. The value of $c_{\mathrm{rec}}^{\ast}$ is selected from the same ten seeds used for inference, and the tables contain many comparisons across layers, metrics, and models. The reported $p$-values are not adjusted for this multiplicity. Selection can therefore make nominal confidence intervals and $p$-values optimistic. We use the tests to characterize the consistency of the observed paired differences, not to claim family-wise significance across the full search space. Confirmatory use would require an independent validation set or a nested selection procedure followed by multiplicity control.

\begin{table}[H]
    \centering
    \small
    \caption{mn\_simplecnnv2\_relu\_0.9771}
    \label{supp:tab:mnist-mn-simplecnnv2-relu-0.9771-10e-paired-tests}
    \begin{tabular*}{\textwidth}{@{\extracolsep{\fill}}llclrrcrrrrr}
    \toprule
    $l_{r}$ & Metric & $c_{\mathrm{rec}}^{\ast}$ & Reference & $n$ & $\Delta$ & 95\% CI & $t(df)$ & $p_t$ & $W$ & $p_W$ & $d_z$ \\
    \midrule
    out & best & 0.3 & $c_{\mathrm{rec}}=0$ & 10 & 0.04 & [0.03, 0.04] & 15.49(9) & $<10^{-4}$ & 55.00 & 0.00179 & 4.90 \\
    out & epoch 10 & 0.3 & $c_{\mathrm{rec}}=0$ & 10 & 0.06 & [0.04, 0.07] & 6.65(9) & $<10^{-4}$ & 55.00 & 0.00249 & 2.10 \\
    out & epoch 1 & 0.1 & $c_{\mathrm{rec}}=0$ & 10 & 0.06 & [0.04, 0.07] & 9.00(9) & $<10^{-4}$ & 55.00 & 0.00245 & 2.85 \\
    out & epoch 5 & 0.3 & $c_{\mathrm{rec}}=0$ & 10 & 0.07 & [0.06, 0.08] & 12.12(9) & $<10^{-4}$ & 55.00 & 0.00247 & 3.83 \\
    out & mean & 0.3 & $c_{\mathrm{rec}}=0$ & 10 & 0.06 & [0.06, 0.06] & 23.12(9) & $<10^{-4}$ & 55.00 & 0.00253 & 7.31 \\
    2 & best & 0.3 & $c_{\mathrm{rec}}=0$ & 10 & 0.02 & [0.01, 0.03] & 4.44(9) & 0.000809 & 44.00 & 0.00502 & 1.40 \\
    2 & epoch 10 & 0.3 & $c_{\mathrm{rec}}=0$ & 10 & 0.01 & [0.00, 0.02] & 2.75(9) & 0.0112 & 40.00 & 0.0183 & 0.87 \\
    2 & epoch 1 & 0.3 & $c_{\mathrm{rec}}=0$ & 10 & 0.01 & [0.00, 0.02] & 2.26(9) & 0.0249 & 38.50 & 0.028 & 0.72 \\
    2 & epoch 5 & 0.3 & $c_{\mathrm{rec}}=0$ & 10 & 0.03 & [0.02, 0.04] & 6.28(9) & $<10^{-4}$ & 45.00 & 0.00358 & 1.99 \\
    2 & mean & 0.3 & $c_{\mathrm{rec}}=0$ & 10 & 0.02 & [0.02, 0.03] & 11.73(9) & $<10^{-4}$ & 55.00 & 0.00253 & 3.71 \\
    1 & best & 0.3 & $c_{\mathrm{rec}}=0$ & 10 & 0.01 & [0.00, 0.01] & 3.86(9) & 0.00193 & 28.00 & 0.00694 & 1.22 \\
    1 & epoch 10 & 0.3 & $c_{\mathrm{rec}}=0$ & 10 & 0.00 & [-0.00, 0.01] & 0.76(9) & 0.234 & 14.00 & 0.226 & 0.24 \\
    1 & epoch 1 & 0 & second-best $c_{\mathrm{rec}}=0.1$ & 10 & 0.00 & [-0.00, 0.01] & 1.00(9) & 0.172 & 14.50 & 0.198 & 0.32 \\
    1 & epoch 5 & 0.1 & $c_{\mathrm{rec}}=0$ & 10 & 0.00 & [-0.01, 0.01] & 0.25(9) & 0.406 & 12.00 & 0.374 & 0.08 \\
    1 & mean & 0.3 & $c_{\mathrm{rec}}=0$ & 10 & 0.00 & [-0.00, 0.01] & 1.56(9) & 0.0764 & 44.00 & 0.0462 & 0.49 \\
    \bottomrule
    \end{tabular*}
    \tablefootnote{$\Delta$ is the paired mean difference in percentage points. CI is a bootstrap percentile confidence interval for $\Delta$. $p_t$ is the one-sided paired t-test p-value, $W$ and $p_W$ are from the one-sided Wilcoxon signed-rank test, and $d_z$ is the paired-sample standardized effect size.}
\end{table}

\begin{table}[H]
    \centering
    \small
    \caption{ci10\_simplecnnv2\_relu\_0.6663}
    \label{supp:tab:cifar10-ci10-simplecnnv2-relu-0.6663-10-paired-tests}
    \begin{tabular*}{\textwidth}{@{\extracolsep{\fill}}llclrrcrrrrr}
    \toprule
    $l_{r}$ & Metric & $c_{\mathrm{rec}}^{\ast}$ & Reference & $n$ & $\Delta$ & 95\% CI & $t(df)$ & $p_t$ & $W$ & $p_W$ & $d_z$ \\
    \midrule
    out & best & 0.3 & $c_{\mathrm{rec}}=0$ & 10 & 0.48 & [0.44, 0.53] & 19.30(9) & $<10^{-4}$ & 55.00 & 0.00252 & 6.10 \\
    out & epoch 10 & 0.3 & $c_{\mathrm{rec}}=0$ & 10 & 1.00 & [0.92, 1.08] & 24.20(9) & $<10^{-4}$ & 55.00 & 0.0025 & 7.65 \\
    out & epoch 1 & 0.3 & $c_{\mathrm{rec}}=0$ & 10 & 0.49 & [0.40, 0.58] & 10.20(9) & $<10^{-4}$ & 55.00 & 0.0025 & 3.22 \\
    out & epoch 5 & 0.3 & $c_{\mathrm{rec}}=0$ & 10 & 0.81 & [0.70, 0.91] & 13.92(9) & $<10^{-4}$ & 55.00 & 0.00253 & 4.40 \\
    out & mean & 0.3 & $c_{\mathrm{rec}}=0$ & 10 & 0.74 & [0.71, 0.77] & 47.04(9) & $<10^{-4}$ & 55.00 & 0.00253 & 14.88 \\
    4 & best & 0.3 & $c_{\mathrm{rec}}=0$ & 10 & 0.18 & [0.10, 0.25] & 4.36(9) & 0.000911 & 54.00 & 0.00346 & 1.38 \\
    4 & epoch 10 & 0.3 & $c_{\mathrm{rec}}=0$ & 10 & 0.57 & [0.51, 0.64] & 16.48(9) & $<10^{-4}$ & 55.00 & 0.0025 & 5.21 \\
    4 & epoch 1 & 0.1 & $c_{\mathrm{rec}}=0$ & 10 & 0.02 & [-0.04, 0.07] & 0.71(9) & 0.248 & 42.50 & 0.063 & 0.22 \\
    4 & epoch 5 & 0.3 & $c_{\mathrm{rec}}=0$ & 10 & 0.33 & [0.23, 0.44] & 5.93(9) & 0.000111 & 54.00 & 0.00344 & 1.87 \\
    4 & mean & 0.3 & $c_{\mathrm{rec}}=0$ & 10 & 0.24 & [0.22, 0.26] & 18.81(9) & $<10^{-4}$ & 55.00 & 0.00253 & 5.95 \\
    3 & best & 0.3 & $c_{\mathrm{rec}}=0$ & 10 & 0.12 & [0.06, 0.16] & 4.44(9) & 0.000811 & 52.00 & 0.00612 & 1.40 \\
    3 & epoch 10 & 0.1 & $c_{\mathrm{rec}}=0$ & 10 & 0.03 & [-0.04, 0.08] & 0.80(9) & 0.223 & 36.00 & 0.193 & 0.25 \\
    3 & epoch 1 & 0.1 & $c_{\mathrm{rec}}=0$ & 10 & 0.03 & [-0.02, 0.08] & 1.15(9) & 0.14 & 38.50 & 0.131 & 0.36 \\
    3 & epoch 5 & 0.3 & $c_{\mathrm{rec}}=0$ & 10 & 0.14 & [0.08, 0.19] & 4.54(9) & 0.000705 & 53.00 & 0.00467 & 1.44 \\
    3 & mean & 0.3 & $c_{\mathrm{rec}}=0$ & 10 & 0.08 & [0.05, 0.10] & 5.06(9) & 0.000341 & 53.00 & 0.00465 & 1.60 \\
    2 & best & 0.3 & $c_{\mathrm{rec}}=0$ & 10 & 0.06 & [0.03, 0.09] & 3.41(9) & 0.00388 & 50.00 & 0.0109 & 1.08 \\
    2 & epoch 10 & 0.3 & $c_{\mathrm{rec}}=0$ & 10 & 0.26 & [0.21, 0.31] & 9.67(9) & $<10^{-4}$ & 55.00 & 0.00253 & 3.06 \\
    2 & epoch 1 & 0.3 & $c_{\mathrm{rec}}=0$ & 10 & 0.02 & [-0.01, 0.05] & 1.24(9) & 0.123 & 30.50 & 0.171 & 0.39 \\
    2 & epoch 5 & 0.3 & $c_{\mathrm{rec}}=0$ & 10 & 0.09 & [0.03, 0.15] & 3.04(9) & 0.00697 & 49.00 & 0.0142 & 0.96 \\
    2 & mean & 0.3 & $c_{\mathrm{rec}}=0$ & 10 & 0.12 & [0.10, 0.13] & 15.53(9) & $<10^{-4}$ & 55.00 & 0.00253 & 4.91 \\
    1 & best & 0.1 & $c_{\mathrm{rec}}=0$ & 10 & 0.01 & [0.00, 0.02] & 2.45(9) & 0.0184 & 40.00 & 0.0182 & 0.77 \\
    1 & epoch 10 & 0.3 & $c_{\mathrm{rec}}=0$ & 10 & 0.03 & [0.00, 0.06] & 1.83(9) & 0.0504 & 43.00 & 0.0571 & 0.58 \\
    1 & epoch 1 & 0.1 & $c_{\mathrm{rec}}=0$ & 10 & 0.01 & [0.00, 0.02] & 2.02(9) & 0.0369 & 47.50 & 0.02 & 0.64 \\
    1 & epoch 5 & 0.3 & $c_{\mathrm{rec}}=0$ & 10 & 0.02 & [-0.00, 0.05] & 1.78(9) & 0.0548 & 35.50 & 0.0616 & 0.56 \\
    1 & mean & 0.3 & $c_{\mathrm{rec}}=0$ & 10 & 0.02 & [0.01, 0.03] & 3.77(9) & 0.00222 & 55.00 & 0.00253 & 1.19 \\
    \bottomrule
    \end{tabular*}
    \tablefootnote{$\Delta$ is the paired mean difference in percentage points. CI is a bootstrap percentile confidence interval for $\Delta$. $p_t$ is the one-sided paired t-test p-value, $W$ and $p_W$ are from the one-sided Wilcoxon signed-rank test, and $d_z$ is the paired-sample standardized effect size.}
\end{table}

\begin{table}[H]
    \centering
    \small
    \caption{ci10\_resnet18\_e5\_0.9493}
    \label{supp:tab:cifar10-ci10-resnet18-e5-0.9493-10e-paired-tests}
    \begin{tabular*}{\textwidth}{@{\extracolsep{\fill}}llclrrcrrrrr}
    \toprule
    $l_{r}$ & Metric & $c_{\mathrm{rec}}^{\ast}$ & Reference & $n$ & $\Delta$ & 95\% CI & $t(df)$ & $p_t$ & $W$ & $p_W$ & $d_z$ \\
    \midrule
    out & best & 0.3 & $c_{\mathrm{rec}}=0$ & 10 & 0.20 & [0.17, 0.23] & 12.56(9) & $<10^{-4}$ & 55.00 & 0.00247 & 3.97 \\
    out & epoch 10 & 0.3 & $c_{\mathrm{rec}}=0$ & 10 & 0.15 & [0.07, 0.26] & 2.93(9) & 0.00844 & 55.00 & 0.00253 & 0.93 \\
    out & epoch 1 & 0.3 & $c_{\mathrm{rec}}=0$ & 10 & 0.14 & [0.09, 0.20] & 4.70(9) & 0.000562 & 55.00 & 0.0025 & 1.49 \\
    out & epoch 5 & 0.3 & $c_{\mathrm{rec}}=0$ & 10 & 0.26 & [0.19, 0.31] & 7.65(9) & $<10^{-4}$ & 55.00 & 0.00252 & 2.42 \\
    out & mean & 0.3 & $c_{\mathrm{rec}}=0$ & 10 & 0.20 & [0.19, 0.21] & 28.26(9) & $<10^{-4}$ & 55.00 & 0.00253 & 8.94 \\
    5 & best & 0.3 & $c_{\mathrm{rec}}=0$ & 10 & 0.06 & [0.04, 0.08] & 4.47(9) & 0.000779 & 45.00 & 0.00382 & 1.41 \\
    5 & epoch 10 & 0.3 & $c_{\mathrm{rec}}=0$ & 10 & 0.05 & [0.00, 0.10] & 2.07(9) & 0.0343 & 38.00 & 0.033 & 0.65 \\
    5 & epoch 1 & 0.3 & $c_{\mathrm{rec}}=0$ & 10 & 0.03 & [0.02, 0.05] & 3.43(9) & 0.00375 & 43.50 & 0.00631 & 1.08 \\
    5 & epoch 5 & 0.3 & $c_{\mathrm{rec}}=0$ & 10 & 0.12 & [0.09, 0.15] & 6.57(9) & $<10^{-4}$ & 55.00 & 0.00252 & 2.08 \\
    5 & mean & 0.3 & $c_{\mathrm{rec}}=0$ & 10 & 0.08 & [0.07, 0.08] & 21.99(9) & $<10^{-4}$ & 55.00 & 0.00252 & 6.95 \\
    4 & best & 0.3 & $c_{\mathrm{rec}}=0$ & 10 & 0.03 & [0.00, 0.05] & 2.35(9) & 0.0217 & 46.50 & 0.0262 & 0.74 \\
    4 & epoch 10 & 0.3 & $c_{\mathrm{rec}}=0$ & 10 & 0.05 & [0.03, 0.07] & 5.15(9) & 0.0003 & 55.00 & 0.00246 & 1.63 \\
    4 & epoch 1 & 0.3 & $c_{\mathrm{rec}}=0$ & 10 & 0.04 & [0.02, 0.06] & 2.95(9) & 0.00806 & 50.50 & 0.00949 & 0.93 \\
    4 & epoch 5 & 0.3 & $c_{\mathrm{rec}}=0$ & 10 & 0.06 & [0.01, 0.11] & 2.12(9) & 0.0313 & 38.00 & 0.033 & 0.67 \\
    4 & mean & 0.3 & $c_{\mathrm{rec}}=0$ & 10 & 0.05 & [0.04, 0.06] & 8.77(9) & $<10^{-4}$ & 55.00 & 0.00253 & 2.77 \\
    3 & best & 0.3 & $c_{\mathrm{rec}}=0$ & 10 & 0.04 & [0.01, 0.06] & 2.92(9) & 0.00846 & 48.50 & 0.016 & 0.92 \\
    3 & epoch 10 & 0.3 & $c_{\mathrm{rec}}=0$ & 10 & 0.17 & [0.12, 0.21] & 6.48(9) & $<10^{-4}$ & 55.00 & 0.00253 & 2.05 \\
    3 & epoch 1 & 0.3 & $c_{\mathrm{rec}}=0$ & 10 & 0.00 & [-0.02, 0.02] & 0.09(9) & 0.463 & 16.00 & 0.611 & 0.03 \\
    3 & epoch 5 & 0.3 & $c_{\mathrm{rec}}=0$ & 10 & 0.06 & [0.03, 0.11] & 2.92(9) & 0.00852 & 50.50 & 0.00949 & 0.92 \\
    3 & mean & 0.3 & $c_{\mathrm{rec}}=0$ & 10 & 0.08 & [0.07, 0.09] & 11.80(9) & $<10^{-4}$ & 55.00 & 0.00253 & 3.73 \\
    2 & best & 0.3 & $c_{\mathrm{rec}}=0$ & 10 & 0.07 & [0.04, 0.09] & 4.29(9) & 0.00101 & 45.00 & 0.00384 & 1.36 \\
    2 & epoch 10 & 0.3 & $c_{\mathrm{rec}}=0$ & 10 & 0.19 & [0.16, 0.23] & 9.42(9) & $<10^{-4}$ & 55.00 & 0.00246 & 2.98 \\
    2 & epoch 1 & 0 & second-best $c_{\mathrm{rec}}=0.02$ & 10 & 0.01 & [-0.00, 0.02] & 1.50(9) & 0.0839 & 21.50 & 0.1 & 0.47 \\
    2 & epoch 5 & 0.3 & $c_{\mathrm{rec}}=0$ & 10 & 0.02 & [-0.01, 0.06] & 1.16(9) & 0.137 & 37.00 & 0.166 & 0.37 \\
    2 & mean & 0.3 & $c_{\mathrm{rec}}=0$ & 10 & 0.08 & [0.07, 0.09] & 15.27(9) & $<10^{-4}$ & 55.00 & 0.00252 & 4.83 \\
    1 & best & 0 & second-best $c_{\mathrm{rec}}=0.02$ & 10 & 0.01 & [-0.01, 0.03] & 1.34(9) & 0.107 & 34.00 & 0.0862 & 0.42 \\
    1 & epoch 10 & 0.3 & $c_{\mathrm{rec}}=0$ & 10 & 0.04 & [-0.00, 0.09] & 1.66(9) & 0.0659 & 35.00 & 0.0693 & 0.52 \\
    1 & epoch 1 & 0 & second-best $c_{\mathrm{rec}}=0.02$ & 10 & 0.00 & [-0.01, 0.01] & 0.16(9) & 0.44 & 19.00 & 0.444 & 0.05 \\
    1 & epoch 5 & 0 & second-best $c_{\mathrm{rec}}=0.02$ & 10 & 0.00 & [-0.02, 0.03] & 0.34(9) & 0.37 & 19.00 & 0.444 & 0.11 \\
    1 & mean & 0 & second-best $c_{\mathrm{rec}}=0.02$ & 10 & 0.00 & [-0.00, 0.01] & 0.61(9) & 0.28 & 33.50 & 0.27 & 0.19 \\
    \bottomrule
    \end{tabular*}
    \tablefootnote{$\Delta$ is the paired mean difference in percentage points. CI is a bootstrap percentile confidence interval for $\Delta$. $p_t$ is the one-sided paired t-test p-value, $W$ and $p_W$ are from the one-sided Wilcoxon signed-rank test, and $d_z$ is the paired-sample standardized effect size.}
\end{table}

\begin{table}[H]
    \centering
    \small
    \caption{ci100\_simplecnnv2\_relu\_0.4411}
    \label{supp:tab:cifar100-ci100-simplecnnv2-relu-0.4411-10e-paired-tests}
    \begin{tabular*}{\textwidth}{@{\extracolsep{\fill}}llclrrcrrrrr}
    \toprule
    $l_{r}$ & Metric & $c_{\mathrm{rec}}^{\ast}$ & Reference & $n$ & $\Delta$ & 95\% CI & $t(df)$ & $p_t$ & $W$ & $p_W$ & $d_z$ \\
    \midrule
    out & best & 0 & second-best $c_{\mathrm{rec}}=0.02$ & 10 & 0.17 & [0.12, 0.23] & 5.38(9) & 0.000221 & 55.00 & 0.00252 & 1.70 \\
    out & epoch 10 & 0 & second-best $c_{\mathrm{rec}}=0.02$ & 10 & 0.11 & [0.03, 0.21] & 2.21(9) & 0.0273 & 44.00 & 0.0461 & 0.70 \\
    out & epoch 1 & 0 & second-best $c_{\mathrm{rec}}=0.02$ & 10 & 0.06 & [-0.03, 0.16] & 1.29(9) & 0.115 & 40.00 & 0.101 & 0.41 \\
    out & epoch 5 & 0 & second-best $c_{\mathrm{rec}}=0.1$ & 10 & 0.06 & [0.01, 0.12] & 2.21(9) & 0.0271 & 38.00 & 0.0327 & 0.70 \\
    out & mean & 0 & second-best $c_{\mathrm{rec}}=0.02$ & 10 & 0.11 & [0.09, 0.13] & 9.23(9) & $<10^{-4}$ & 55.00 & 0.00253 & 2.92 \\
    4 & best & 0.02 & $c_{\mathrm{rec}}=0$ & 10 & 0.06 & [0.03, 0.10] & 3.34(9) & 0.00431 & 52.00 & 0.00623 & 1.06 \\
    4 & epoch 10 & 0.02 & $c_{\mathrm{rec}}=0$ & 10 & 0.02 & [-0.07, 0.10] & 0.53(9) & 0.305 & 39.00 & 0.12 & 0.17 \\
    4 & epoch 1 & 0.02 & $c_{\mathrm{rec}}=0$ & 10 & 0.02 & [-0.01, 0.05] & 1.33(9) & 0.109 & 34.50 & 0.0767 & 0.42 \\
    4 & epoch 5 & 0 & second-best $c_{\mathrm{rec}}=0.02$ & 10 & 0.00 & [-0.07, 0.08] & 0.07(9) & 0.472 & 25.00 & 0.383 & 0.02 \\
    4 & mean & 0.02 & $c_{\mathrm{rec}}=0$ & 10 & 0.02 & [0.00, 0.04] & 1.97(9) & 0.0403 & 45.00 & 0.0372 & 0.62 \\
    3 & best & 0.1 & $c_{\mathrm{rec}}=0$ & 10 & 0.05 & [0.01, 0.09] & 2.12(9) & 0.0317 & 46.00 & 0.0293 & 0.67 \\
    3 & epoch 10 & 0.1 & $c_{\mathrm{rec}}=0$ & 10 & 0.06 & [-0.00, 0.12] & 1.91(9) & 0.0441 & 38.00 & 0.0332 & 0.60 \\
    3 & epoch 1 & 0 & second-best $c_{\mathrm{rec}}=0.1$ & 10 & 0.01 & [-0.03, 0.04] & 0.30(9) & 0.386 & 28.00 & 0.48 & 0.09 \\
    3 & epoch 5 & 0.1 & $c_{\mathrm{rec}}=0$ & 10 & 0.10 & [0.06, 0.15] & 4.48(9) & 0.000768 & 54.00 & 0.00346 & 1.42 \\
    3 & mean & 0.1 & $c_{\mathrm{rec}}=0$ & 10 & 0.06 & [0.05, 0.07] & 10.10(9) & $<10^{-4}$ & 55.00 & 0.00253 & 3.19 \\
    2 & best & 0.3 & $c_{\mathrm{rec}}=0$ & 10 & 0.05 & [0.01, 0.10] & 2.25(9) & 0.0255 & 48.50 & 0.016 & 0.71 \\
    2 & epoch 10 & 0.3 & $c_{\mathrm{rec}}=0$ & 10 & 0.02 & [-0.03, 0.06] & 0.62(9) & 0.277 & 34.00 & 0.254 & 0.19 \\
    2 & epoch 1 & 0.02 & $c_{\mathrm{rec}}=0$ & 10 & 0.01 & [0.00, 0.02] & 2.62(9) & 0.0138 & 41.00 & 0.0136 & 0.83 \\
    2 & epoch 5 & 0.3 & $c_{\mathrm{rec}}=0$ & 10 & 0.05 & [-0.00, 0.11] & 1.74(9) & 0.0577 & 42.00 & 0.0697 & 0.55 \\
    2 & mean & 0.3 & $c_{\mathrm{rec}}=0$ & 10 & 0.04 & [0.03, 0.05] & 6.47(9) & $<10^{-4}$ & 55.00 & 0.00253 & 2.05 \\
    1 & best & 0.3 & $c_{\mathrm{rec}}=0$ & 10 & 0.02 & [-0.00, 0.03] & 1.61(9) & 0.0714 & 41.50 & 0.0761 & 0.51 \\
    1 & epoch 10 & 0.3 & $c_{\mathrm{rec}}=0$ & 10 & 0.03 & [0.01, 0.06] & 2.91(9) & 0.0087 & 41.50 & 0.0121 & 0.92 \\
    1 & epoch 1 & 0.1 & $c_{\mathrm{rec}}=0$ & 10 & 0.00 & [-0.01, 0.01] & 0.41(9) & 0.346 & 20.50 & 0.36 & 0.13 \\
    1 & epoch 5 & 0.3 & $c_{\mathrm{rec}}=0$ & 10 & 0.02 & [0.01, 0.04] & 2.60(9) & 0.0145 & 35.00 & 0.00858 & 0.82 \\
    1 & mean & 0.3 & $c_{\mathrm{rec}}=0$ & 10 & 0.01 & [-0.00, 0.01] & 1.69(9) & 0.0629 & 42.00 & 0.0697 & 0.53 \\
    \bottomrule
    \end{tabular*}
    \tablefootnote{$\Delta$ is the paired mean difference in percentage points. CI is a bootstrap percentile confidence interval for $\Delta$. $p_t$ is the one-sided paired t-test p-value, $W$ and $p_W$ are from the one-sided Wilcoxon signed-rank test, and $d_z$ is the paired-sample standardized effect size.}
\end{table}

\begin{table}[H]
    \centering
    \small
    \caption{ci100\_resnet18\_epoch200\_0.7926}
    \label{supp:tab:cifar100-ci100-resnet18-epoch200-0.7926-10e-paired-tests}
    \begin{tabular*}{\textwidth}{@{\extracolsep{\fill}}llclrrcrrrrr}
    \toprule
    $l_{r}$ & Metric & $c_{\mathrm{rec}}^{\ast}$ & Reference & $n$ & $\Delta$ & 95\% CI & $t(df)$ & $p_t$ & $W$ & $p_W$ & $d_z$ \\
    \midrule
    out & best & 0.3 & $c_{\mathrm{rec}}=0$ & 10 & 0.11 & [0.07, 0.14] & 5.36(9) & 0.000227 & 55.00 & 0.0025 & 1.70 \\
    out & epoch 10 & 0.1 & $c_{\mathrm{rec}}=0$ & 10 & 0.18 & [0.12, 0.23] & 6.23(9) & $<10^{-4}$ & 55.00 & 0.00252 & 1.97 \\
    out & epoch 1 & 0.1 & $c_{\mathrm{rec}}=0$ & 10 & 0.34 & [0.25, 0.42] & 7.39(9) & $<10^{-4}$ & 55.00 & 0.00253 & 2.34 \\
    out & epoch 5 & 0.1 & $c_{\mathrm{rec}}=0$ & 10 & 0.06 & [0.00, 0.12] & 1.98(9) & 0.0397 & 45.00 & 0.0369 & 0.63 \\
    out & mean & 0.1 & $c_{\mathrm{rec}}=0$ & 10 & 0.14 & [0.10, 0.17] & 7.26(9) & $<10^{-4}$ & 54.00 & 0.00346 & 2.30 \\
    5 & best & 0.3 & $c_{\mathrm{rec}}=0$ & 10 & 0.07 & [0.02, 0.11] & 3.07(9) & 0.00669 & 51.00 & 0.00823 & 0.97 \\
    5 & epoch 10 & 0.3 & $c_{\mathrm{rec}}=0$ & 10 & 0.23 & [0.16, 0.31] & 5.69(9) & 0.000149 & 55.00 & 0.00253 & 1.80 \\
    5 & epoch 1 & 0.1 & $c_{\mathrm{rec}}=0$ & 10 & 0.36 & [0.27, 0.44] & 7.58(9) & $<10^{-4}$ & 55.00 & 0.00253 & 2.40 \\
    5 & epoch 5 & 0.1 & $c_{\mathrm{rec}}=0$ & 10 & 0.04 & [-0.03, 0.11] & 0.99(9) & 0.173 & 30.00 & 0.187 & 0.31 \\
    5 & mean & 0.3 & $c_{\mathrm{rec}}=0$ & 10 & 0.15 & [0.13, 0.17] & 13.25(9) & $<10^{-4}$ & 55.00 & 0.00253 & 4.19 \\
    4 & best & 0.3 & $c_{\mathrm{rec}}=0$ & 10 & 0.11 & [0.05, 0.16] & 3.38(9) & 0.00404 & 52.00 & 0.00626 & 1.07 \\
    4 & epoch 10 & 0.3 & $c_{\mathrm{rec}}=0$ & 10 & 0.25 & [0.19, 0.31] & 7.42(9) & $<10^{-4}$ & 55.00 & 0.0025 & 2.35 \\
    4 & epoch 1 & 0.1 & $c_{\mathrm{rec}}=0$ & 10 & 0.38 & [0.27, 0.49] & 6.15(9) & $<10^{-4}$ & 55.00 & 0.00253 & 1.95 \\
    4 & epoch 5 & 0.3 & $c_{\mathrm{rec}}=0$ & 10 & 0.11 & [0.05, 0.16] & 3.52(9) & 0.00324 & 51.00 & 0.00827 & 1.11 \\
    4 & mean & 0.3 & $c_{\mathrm{rec}}=0$ & 10 & 0.19 & [0.15, 0.22] & 9.13(9) & $<10^{-4}$ & 55.00 & 0.00253 & 2.89 \\
    3 & best & 0.3 & $c_{\mathrm{rec}}=0$ & 10 & 0.02 & [-0.02, 0.06] & 0.76(9) & 0.234 & 34.00 & 0.254 & 0.24 \\
    3 & epoch 10 & 0.1 & $c_{\mathrm{rec}}=0$ & 10 & 0.15 & [0.07, 0.22] & 3.80(9) & 0.00212 & 53.00 & 0.00465 & 1.20 \\
    3 & epoch 1 & 0.3 & $c_{\mathrm{rec}}=0$ & 10 & 0.04 & [-0.02, 0.10] & 1.14(9) & 0.141 & 37.50 & 0.154 & 0.36 \\
    3 & epoch 5 & 0 & second-best $c_{\mathrm{rec}}=0.1$ & 10 & 0.00 & [-0.07, 0.07] & 0.06(9) & 0.479 & 31.00 & 0.36 & 0.02 \\
    3 & mean & 0.1 & $c_{\mathrm{rec}}=0$ & 10 & 0.05 & [0.03, 0.07] & 4.68(9) & 0.000576 & 54.00 & 0.00346 & 1.48 \\
    2 & best & 0.3 & $c_{\mathrm{rec}}=0$ & 10 & 0.04 & [-0.01, 0.09] & 1.42(9) & 0.0942 & 38.00 & 0.142 & 0.45 \\
    2 & epoch 10 & 0.1 & $c_{\mathrm{rec}}=0$ & 10 & 0.10 & [0.03, 0.18] & 2.54(9) & 0.0158 & 49.00 & 0.0141 & 0.80 \\
    2 & epoch 1 & 0.3 & $c_{\mathrm{rec}}=0$ & 10 & 0.04 & [-0.01, 0.10] & 1.34(9) & 0.106 & 36.00 & 0.193 & 0.43 \\
    2 & epoch 5 & 0.1 & $c_{\mathrm{rec}}=0$ & 10 & 0.02 & [-0.05, 0.09] & 0.42(9) & 0.341 & 32.50 & 0.305 & 0.13 \\
    2 & mean & 0.3 & $c_{\mathrm{rec}}=0$ & 10 & 0.05 & [0.03, 0.08] & 3.82(9) & 0.00206 & 53.00 & 0.00467 & 1.21 \\
    1 & best & 0 & second-best $c_{\mathrm{rec}}=0.1$ & 10 & 0.02 & [-0.01, 0.06] & 1.24(9) & 0.124 & 33.50 & 0.0961 & 0.39 \\
    1 & epoch 10 & 0.1 & $c_{\mathrm{rec}}=0$ & 10 & 0.03 & [0.00, 0.07] & 1.65(9) & 0.0665 & 37.50 & 0.0364 & 0.52 \\
    1 & epoch 1 & 0.3 & $c_{\mathrm{rec}}=0$ & 10 & 0.02 & [-0.02, 0.06] & 0.92(9) & 0.191 & 36.50 & 0.179 & 0.29 \\
    1 & epoch 5 & 0.1 & $c_{\mathrm{rec}}=0$ & 10 & 0.04 & [0.00, 0.07] & 1.87(9) & 0.0469 & 37.00 & 0.0428 & 0.59 \\
    1 & mean & 0.1 & $c_{\mathrm{rec}}=0$ & 10 & 0.00 & [-0.01, 0.02] & 0.51(9) & 0.311 & 33.00 & 0.288 & 0.16 \\
    \bottomrule
    \end{tabular*}
    \tablefootnote{$\Delta$ is the paired mean difference in percentage points. CI is a bootstrap percentile confidence interval for $\Delta$. $p_t$ is the one-sided paired t-test p-value, $W$ and $p_W$ are from the one-sided Wilcoxon signed-rank test, and $d_z$ is the paired-sample standardized effect size.}
\end{table}

\begin{table}[H]
    \centering
    \small
    \caption{ci100\_simplevit\_p4\_d192\_l9\_e30\_0.6320}
    \label{supp:tab:cifar100-ci100-simplevit-p4-d192-l9-e30-0.6320-10e-paired-tests}
    \begin{tabular*}{\textwidth}{@{\extracolsep{\fill}}llclrrcrrrrr}
    \toprule
    $l_{r}$ & Metric & $c_{\mathrm{rec}}^{\ast}$ & Reference & $n$ & $\Delta$ & 95\% CI & $t(df)$ & $p_t$ & $W$ & $p_W$ & $d_z$ \\
    \midrule
    out & best & 0.1 & $c_{\mathrm{rec}}=0$ & 10 & 0.28 & [0.21, 0.35] & 7.43(9) & $<10^{-4}$ & 55.00 & 0.00253 & 2.35 \\
    out & epoch 10 & 0.3 & $c_{\mathrm{rec}}=0$ & 10 & 3.65 & [3.58, 3.73] & 90.28(9) & $<10^{-4}$ & 55.00 & 0.00252 & 28.55 \\
    out & epoch 1 & 0.3 & $c_{\mathrm{rec}}=0$ & 10 & 0.10 & [0.02, 0.18] & 2.33(9) & 0.0222 & 46.50 & 0.0263 & 0.74 \\
    out & epoch 5 & 0.1 & $c_{\mathrm{rec}}=0$ & 10 & 2.56 & [2.47, 2.64] & 54.38(9) & $<10^{-4}$ & 55.00 & 0.00253 & 17.20 \\
    out & mean & 0.3 & $c_{\mathrm{rec}}=0$ & 10 & 2.25 & [2.22, 2.27] & 156.80(9) & $<10^{-4}$ & 55.00 & 0.00253 & 49.58 \\
    10 & best & 0.1 & $c_{\mathrm{rec}}=0$ & 10 & 0.46 & [0.36, 0.56] & 8.43(9) & $<10^{-4}$ & 55.00 & 0.00252 & 2.67 \\
    10 & epoch 10 & 0.3 & $c_{\mathrm{rec}}=0$ & 10 & 3.99 & [3.91, 4.05] & 113.48(9) & $<10^{-4}$ & 55.00 & 0.0025 & 35.89 \\
    10 & epoch 1 & 0.1 & $c_{\mathrm{rec}}=0$ & 10 & 0.21 & [0.14, 0.28] & 5.42(9) & 0.000211 & 55.00 & 0.00252 & 1.71 \\
    10 & epoch 5 & 0.1 & $c_{\mathrm{rec}}=0$ & 10 & 2.50 & [2.44, 2.57] & 69.29(9) & $<10^{-4}$ & 55.00 & 0.00252 & 21.91 \\
    10 & mean & 0.3 & $c_{\mathrm{rec}}=0$ & 10 & 2.43 & [2.40, 2.46] & 143.05(9) & $<10^{-4}$ & 55.00 & 0.00253 & 45.24 \\
    9 & best & 0.3 & $c_{\mathrm{rec}}=0$ & 10 & 0.78 & [0.70, 0.86] & 18.08(9) & $<10^{-4}$ & 55.00 & 0.00253 & 5.72 \\
    9 & epoch 10 & 0.3 & $c_{\mathrm{rec}}=0$ & 10 & 4.35 & [4.27, 4.42] & 112.32(9) & $<10^{-4}$ & 55.00 & 0.00253 & 35.52 \\
    9 & epoch 1 & 0.1 & $c_{\mathrm{rec}}=0$ & 10 & 0.66 & [0.59, 0.73] & 18.05(9) & $<10^{-4}$ & 55.00 & 0.0025 & 5.71 \\
    9 & epoch 5 & 0.3 & $c_{\mathrm{rec}}=0$ & 10 & 2.94 & [2.84, 3.04] & 56.24(9) & $<10^{-4}$ & 55.00 & 0.00253 & 17.79 \\
    9 & mean & 0.3 & $c_{\mathrm{rec}}=0$ & 10 & 2.84 & [2.80, 2.88] & 142.91(9) & $<10^{-4}$ & 55.00 & 0.00253 & 45.19 \\
    8 & best & 0.3 & $c_{\mathrm{rec}}=0$ & 10 & 0.83 & [0.75, 0.92] & 18.29(9) & $<10^{-4}$ & 55.00 & 0.00253 & 5.79 \\
    8 & epoch 10 & 0.3 & $c_{\mathrm{rec}}=0$ & 10 & 4.42 & [4.35, 4.49] & 114.97(9) & $<10^{-4}$ & 55.00 & 0.00252 & 36.36 \\
    8 & epoch 1 & 0.1 & $c_{\mathrm{rec}}=0$ & 10 & 0.74 & [0.67, 0.83] & 16.45(9) & $<10^{-4}$ & 55.00 & 0.00253 & 5.20 \\
    8 & epoch 5 & 0.3 & $c_{\mathrm{rec}}=0$ & 10 & 2.83 & [2.79, 2.86] & 146.61(9) & $<10^{-4}$ & 55.00 & 0.0025 & 46.36 \\
    8 & mean & 0.3 & $c_{\mathrm{rec}}=0$ & 10 & 2.85 & [2.83, 2.88] & 219.06(9) & $<10^{-4}$ & 55.00 & 0.00253 & 69.27 \\
    7 & best & 0.3 & $c_{\mathrm{rec}}=0$ & 10 & 0.64 & [0.57, 0.70] & 17.53(9) & $<10^{-4}$ & 55.00 & 0.00253 & 5.54 \\
    7 & epoch 10 & 0.3 & $c_{\mathrm{rec}}=0$ & 10 & 3.70 & [3.63, 3.77] & 95.74(9) & $<10^{-4}$ & 55.00 & 0.00253 & 30.27 \\
    7 & epoch 1 & 0.1 & $c_{\mathrm{rec}}=0$ & 10 & 0.59 & [0.49, 0.72] & 9.52(9) & $<10^{-4}$ & 55.00 & 0.00252 & 3.01 \\
    7 & epoch 5 & 0.3 & $c_{\mathrm{rec}}=0$ & 10 & 1.93 & [1.80, 2.06] & 27.21(9) & $<10^{-4}$ & 55.00 & 0.00252 & 8.60 \\
    7 & mean & 0.3 & $c_{\mathrm{rec}}=0$ & 10 & 2.13 & [2.11, 2.16] & 195.50(9) & $<10^{-4}$ & 55.00 & 0.00253 & 61.82 \\
    6 & best & 0.3 & $c_{\mathrm{rec}}=0$ & 10 & 0.41 & [0.35, 0.47] & 13.22(9) & $<10^{-4}$ & 55.00 & 0.00252 & 4.18 \\
    6 & epoch 10 & 0.3 & $c_{\mathrm{rec}}=0$ & 10 & 2.35 & [2.24, 2.47] & 37.73(9) & $<10^{-4}$ & 55.00 & 0.00252 & 11.93 \\
    6 & epoch 1 & 0.1 & $c_{\mathrm{rec}}=0$ & 10 & 0.44 & [0.35, 0.54] & 8.54(9) & $<10^{-4}$ & 55.00 & 0.00252 & 2.70 \\
    6 & epoch 5 & 0.1 & $c_{\mathrm{rec}}=0$ & 10 & 1.28 & [1.22, 1.35] & 37.16(9) & $<10^{-4}$ & 55.00 & 0.00247 & 11.75 \\
    6 & mean & 0.3 & $c_{\mathrm{rec}}=0$ & 10 & 1.38 & [1.36, 1.40] & 107.79(9) & $<10^{-4}$ & 55.00 & 0.00253 & 34.09 \\
    5 & best & 0.1 & $c_{\mathrm{rec}}=0$ & 10 & 0.31 & [0.26, 0.36] & 11.30(9) & $<10^{-4}$ & 55.00 & 0.00252 & 3.57 \\
    5 & epoch 10 & 0.3 & $c_{\mathrm{rec}}=0$ & 10 & 1.34 & [1.29, 1.40] & 47.95(9) & $<10^{-4}$ & 55.00 & 0.00252 & 15.16 \\
    5 & epoch 1 & 0.1 & $c_{\mathrm{rec}}=0$ & 10 & 0.30 & [0.24, 0.36] & 9.22(9) & $<10^{-4}$ & 55.00 & 0.00252 & 2.92 \\
    5 & epoch 5 & 0.1 & $c_{\mathrm{rec}}=0$ & 10 & 0.76 & [0.67, 0.86] & 14.93(9) & $<10^{-4}$ & 55.00 & 0.00253 & 4.72 \\
    5 & mean & 0.3 & $c_{\mathrm{rec}}=0$ & 10 & 0.84 & [0.82, 0.86] & 71.33(9) & $<10^{-4}$ & 55.00 & 0.00253 & 22.56 \\
    4 & best & 0.1 & $c_{\mathrm{rec}}=0$ & 10 & 0.27 & [0.22, 0.32] & 10.58(9) & $<10^{-4}$ & 55.00 & 0.00253 & 3.34 \\
    4 & epoch 10 & 0.1 & $c_{\mathrm{rec}}=0$ & 10 & 0.90 & [0.83, 0.96] & 25.38(9) & $<10^{-4}$ & 55.00 & 0.00252 & 8.03 \\
    4 & epoch 1 & 0.1 & $c_{\mathrm{rec}}=0$ & 10 & 0.23 & [0.17, 0.28] & 7.97(9) & $<10^{-4}$ & 55.00 & 0.00253 & 2.52 \\
    4 & epoch 5 & 0.1 & $c_{\mathrm{rec}}=0$ & 10 & 0.55 & [0.47, 0.61] & 14.78(9) & $<10^{-4}$ & 55.00 & 0.00252 & 4.68 \\
    4 & mean & 0.1 & $c_{\mathrm{rec}}=0$ & 10 & 0.59 & [0.57, 0.61] & 48.58(9) & $<10^{-4}$ & 55.00 & 0.00252 & 15.36 \\
    3 & best & 0.1 & $c_{\mathrm{rec}}=0$ & 10 & 0.18 & [0.15, 0.22] & 8.60(9) & $<10^{-4}$ & 55.00 & 0.00253 & 2.72 \\
    3 & epoch 10 & 0.1 & $c_{\mathrm{rec}}=0$ & 10 & 0.51 & [0.43, 0.59] & 12.10(9) & $<10^{-4}$ & 55.00 & 0.00247 & 3.83 \\
    3 & epoch 1 & 0.1 & $c_{\mathrm{rec}}=0$ & 10 & 0.06 & [0.03, 0.10] & 3.46(9) & 0.00357 & 52.00 & 0.00614 & 1.09 \\
    3 & epoch 5 & 0.1 & $c_{\mathrm{rec}}=0$ & 10 & 0.31 & [0.26, 0.35] & 11.93(9) & $<10^{-4}$ & 55.00 & 0.0025 & 3.77 \\
    3 & mean & 0.1 & $c_{\mathrm{rec}}=0$ & 10 & 0.33 & [0.31, 0.34] & 37.23(9) & $<10^{-4}$ & 55.00 & 0.00253 & 11.77 \\
    2 & best & 0.1 & $c_{\mathrm{rec}}=0$ & 10 & 0.23 & [0.19, 0.27] & 11.31(9) & $<10^{-4}$ & 55.00 & 0.0025 & 3.58 \\
    2 & epoch 10 & 0.1 & $c_{\mathrm{rec}}=0$ & 10 & 0.46 & [0.42, 0.50] & 22.18(9) & $<10^{-4}$ & 55.00 & 0.0025 & 7.02 \\
    2 & epoch 1 & 0.1 & $c_{\mathrm{rec}}=0$ & 10 & 0.06 & [0.03, 0.10] & 3.61(9) & 0.00285 & 52.00 & 0.00623 & 1.14 \\
    2 & epoch 5 & 0.1 & $c_{\mathrm{rec}}=0$ & 10 & 0.34 & [0.31, 0.38] & 17.49(9) & $<10^{-4}$ & 55.00 & 0.00252 & 5.53 \\
    2 & mean & 0.1 & $c_{\mathrm{rec}}=0$ & 10 & 0.33 & [0.32, 0.35] & 37.40(9) & $<10^{-4}$ & 55.00 & 0.00253 & 11.83 \\
    1 & best & 0.1 & $c_{\mathrm{rec}}=0$ & 10 & 0.19 & [0.14, 0.25] & 6.97(9) & $<10^{-4}$ & 55.00 & 0.0025 & 2.21 \\
    1 & epoch 10 & 0.1 & $c_{\mathrm{rec}}=0$ & 10 & 0.30 & [0.26, 0.35] & 12.32(9) & $<10^{-4}$ & 55.00 & 0.00253 & 3.90 \\
    1 & epoch 1 & 0.1 & $c_{\mathrm{rec}}=0$ & 10 & 0.03 & [-0.01, 0.07] & 1.35(9) & 0.104 & 41.00 & 0.0838 & 0.43 \\
    1 & epoch 5 & 0.1 & $c_{\mathrm{rec}}=0$ & 10 & 0.29 & [0.22, 0.36] & 7.62(9) & $<10^{-4}$ & 55.00 & 0.00253 & 2.41 \\
    1 & mean & 0.1 & $c_{\mathrm{rec}}=0$ & 10 & 0.25 & [0.23, 0.27] & 24.21(9) & $<10^{-4}$ & 55.00 & 0.00253 & 7.65 \\
    \bottomrule
    \end{tabular*}
    \tablefootnote{$\Delta$ is the paired mean difference in percentage points. CI is a bootstrap percentile confidence interval for $\Delta$. $p_t$ is the one-sided paired t-test p-value, $W$ and $p_W$ are from the one-sided Wilcoxon signed-rank test, and $d_z$ is the paired-sample standardized effect size.}
\end{table}

\begin{table}[H]
    \centering
    \small
    \caption{nette\_simplecnnv2\_relu\_0.6005}
    \label{supp:tab:nette-nette-simplecnnv2-relu-0.6005-10e-paired-tests}
    \begin{tabular*}{\textwidth}{@{\extracolsep{\fill}}llclrrcrrrrr}
    \toprule
    $l_{r}$ & Metric & $c_{\mathrm{rec}}^{\ast}$ & Reference & $n$ & $\Delta$ & 95\% CI & $t(df)$ & $p_t$ & $W$ & $p_W$ & $d_z$ \\
    \midrule
    out & best & 0.1 & $c_{\mathrm{rec}}=0$ & 10 & 0.47 & [0.37, 0.58] & 8.31(9) & $<10^{-4}$ & 55.00 & 0.0025 & 2.63 \\
    out & epoch 10 & 0.1 & $c_{\mathrm{rec}}=0$ & 10 & 0.60 & [0.44, 0.76] & 6.66(9) & $<10^{-4}$ & 55.00 & 0.00252 & 2.11 \\
    out & epoch 1 & 0.3 & $c_{\mathrm{rec}}=0$ & 10 & 0.48 & [0.31, 0.66] & 5.05(9) & 0.000344 & 55.00 & 0.00253 & 1.60 \\
    out & epoch 5 & 0.1 & $c_{\mathrm{rec}}=0$ & 10 & 0.55 & [0.41, 0.68] & 7.16(9) & $<10^{-4}$ & 55.00 & 0.00253 & 2.26 \\
    out & mean & 0.1 & $c_{\mathrm{rec}}=0$ & 10 & 0.53 & [0.45, 0.61] & 11.69(9) & $<10^{-4}$ & 55.00 & 0.00253 & 3.70 \\
    6 & best & 0.3 & $c_{\mathrm{rec}}=0$ & 10 & 0.19 & [0.10, 0.29] & 3.73(9) & 0.00234 & 45.00 & 0.00384 & 1.18 \\
    6 & epoch 10 & 0.3 & $c_{\mathrm{rec}}=0$ & 10 & 0.42 & [0.30, 0.54] & 6.55(9) & $<10^{-4}$ & 55.00 & 0.00252 & 2.07 \\
    6 & epoch 1 & 0 & second-best $c_{\mathrm{rec}}=0.1$ & 10 & 0.05 & [-0.05, 0.14] & 0.89(9) & 0.198 & 30.00 & 0.187 & 0.28 \\
    6 & epoch 5 & 0.3 & $c_{\mathrm{rec}}=0$ & 10 & 0.37 & [0.27, 0.47] & 6.85(9) & $<10^{-4}$ & 45.00 & 0.00384 & 2.17 \\
    6 & mean & 0.3 & $c_{\mathrm{rec}}=0$ & 10 & 0.27 & [0.23, 0.31] & 11.71(9) & $<10^{-4}$ & 55.00 & 0.00252 & 3.70 \\
    5 & best & 0 & second-best $c_{\mathrm{rec}}=0.1$ & 10 & 0.22 & [0.13, 0.31] & 4.44(9) & 0.000818 & 44.00 & 0.00543 & 1.40 \\
    5 & epoch 10 & 0.1 & $c_{\mathrm{rec}}=0$ & 10 & 0.02 & [-0.14, 0.19] & 0.17(9) & 0.433 & 21.50 & 0.547 & 0.05 \\
    5 & epoch 1 & 0 & second-best $c_{\mathrm{rec}}=0.1$ & 10 & 0.34 & [0.17, 0.52] & 3.58(9) & 0.00296 & 52.00 & 0.00626 & 1.13 \\
    5 & epoch 5 & 0 & second-best $c_{\mathrm{rec}}=0.1$ & 10 & 0.34 & [0.20, 0.50] & 4.26(9) & 0.00106 & 55.00 & 0.00253 & 1.35 \\
    5 & mean & 0 & second-best $c_{\mathrm{rec}}=0.1$ & 10 & 0.29 & [0.25, 0.33] & 13.46(9) & $<10^{-4}$ & 55.00 & 0.00253 & 4.26 \\
    4 & best & 0.3 & $c_{\mathrm{rec}}=0$ & 10 & 0.43 & [0.36, 0.52] & 9.67(9) & $<10^{-4}$ & 55.00 & 0.00252 & 3.06 \\
    4 & epoch 10 & 0.3 & $c_{\mathrm{rec}}=0$ & 10 & 0.57 & [0.42, 0.69] & 7.64(9) & $<10^{-4}$ & 55.00 & 0.00253 & 2.42 \\
    4 & epoch 1 & 0.3 & $c_{\mathrm{rec}}=0$ & 10 & 0.21 & [0.12, 0.28] & 4.78(9) & 0.000503 & 53.00 & 0.00465 & 1.51 \\
    4 & epoch 5 & 0.3 & $c_{\mathrm{rec}}=0$ & 10 & 0.50 & [0.35, 0.65] & 6.07(9) & $<10^{-4}$ & 55.00 & 0.00252 & 1.92 \\
    4 & mean & 0.3 & $c_{\mathrm{rec}}=0$ & 10 & 0.42 & [0.37, 0.46] & 17.55(9) & $<10^{-4}$ & 55.00 & 0.00253 & 5.55 \\
    3 & best & 0.3 & $c_{\mathrm{rec}}=0$ & 10 & 0.11 & [0.04, 0.20] & 2.58(9) & 0.0149 & 47.00 & 0.0231 & 0.81 \\
    3 & epoch 10 & 0 & second-best $c_{\mathrm{rec}}=0.3$ & 10 & 0.04 & [-0.11, 0.18] & 0.49(9) & 0.317 & 27.50 & 0.277 & 0.16 \\
    3 & epoch 1 & 0.1 & $c_{\mathrm{rec}}=0$ & 10 & 0.26 & [0.11, 0.43] & 3.17(9) & 0.00571 & 54.00 & 0.00342 & 1.00 \\
    3 & epoch 5 & 0 & second-best $c_{\mathrm{rec}}=0.1$ & 10 & 0.11 & [0.02, 0.20] & 2.32(9) & 0.0229 & 45.50 & 0.0328 & 0.73 \\
    3 & mean & 0.3 & $c_{\mathrm{rec}}=0$ & 10 & 0.05 & [0.00, 0.09] & 2.14(9) & 0.0304 & 46.00 & 0.0297 & 0.68 \\
    2 & best & 0.3 & $c_{\mathrm{rec}}=0$ & 10 & 0.03 & [-0.06, 0.10] & 0.57(9) & 0.291 & 28.00 & 0.257 & 0.18 \\
    2 & epoch 10 & 0 & second-best $c_{\mathrm{rec}}=0.1$ & 10 & 0.06 & [0.00, 0.13] & 1.84(9) & 0.0496 & 35.50 & 0.0616 & 0.58 \\
    2 & epoch 1 & 0.3 & $c_{\mathrm{rec}}=0$ & 10 & 0.14 & [-0.02, 0.30] & 1.62(9) & 0.0702 & 40.50 & 0.0924 & 0.51 \\
    2 & epoch 5 & 0.3 & $c_{\mathrm{rec}}=0$ & 10 & 0.07 & [-0.10, 0.23] & 0.79(9) & 0.224 & 36.50 & 0.179 & 0.25 \\
    2 & mean & 0.3 & $c_{\mathrm{rec}}=0$ & 10 & 0.05 & [0.01, 0.09] & 2.06(9) & 0.0347 & 43.50 & 0.0512 & 0.65 \\
    1 & best & 0.3 & $c_{\mathrm{rec}}=0$ & 10 & 0.00 & [-0.09, 0.10] & 0.05(9) & 0.48 & 20.50 & 0.594 & 0.02 \\
    1 & epoch 10 & 0.1 & $c_{\mathrm{rec}}=0$ & 10 & 0.02 & [-0.04, 0.09] & 0.63(9) & 0.273 & 26.00 & 0.339 & 0.20 \\
    1 & epoch 1 & 0.3 & $c_{\mathrm{rec}}=0$ & 10 & 0.01 & [-0.05, 0.07] & 0.23(9) & 0.411 & 21.50 & 0.547 & 0.07 \\
    1 & epoch 5 & 0.3 & $c_{\mathrm{rec}}=0$ & 10 & 0.05 & [-0.13, 0.23] & 0.55(9) & 0.297 & 34.00 & 0.254 & 0.17 \\
    1 & mean & 0.3 & $c_{\mathrm{rec}}=0$ & 10 & 0.01 & [-0.03, 0.05] & 0.69(9) & 0.254 & 35.00 & 0.222 & 0.22 \\
    \bottomrule
    \end{tabular*}
    \tablefootnote{$\Delta$ is the paired mean difference in percentage points. CI is a bootstrap percentile confidence interval for $\Delta$. $p_t$ is the one-sided paired t-test p-value, $W$ and $p_W$ are from the one-sided Wilcoxon signed-rank test, and $d_z$ is the paired-sample standardized effect size.}
\end{table}

\begin{table}[H]
    \centering
    \small
    \caption{nette\_resnet18\_epoch5\_0.9814}
    \label{supp:tab:nette-nette-resnet18-epoch5-0.9814-10e-paired-tests}
    \begin{tabular*}{\textwidth}{@{\extracolsep{\fill}}llclrrcrrrrr}
    \toprule
    $l_{r}$ & Metric & $c_{\mathrm{rec}}^{\ast}$ & Reference & $n$ & $\Delta$ & 95\% CI & $t(df)$ & $p_t$ & $W$ & $p_W$ & $d_z$ \\
    \midrule
    out & best & 0.3 & $c_{\mathrm{rec}}=0$ & 10 & 0.25 & [0.22, 0.29] & 14.30(9) & $<10^{-4}$ & 55.00 & 0.00245 & 4.52 \\
    out & epoch 10 & 0.3 & $c_{\mathrm{rec}}=0$ & 10 & 0.19 & [0.14, 0.24] & 7.14(9) & $<10^{-4}$ & 55.00 & 0.00252 & 2.26 \\
    out & epoch 1 & 0.3 & $c_{\mathrm{rec}}=0$ & 10 & 0.53 & [0.47, 0.58] & 17.47(9) & $<10^{-4}$ & 55.00 & 0.00252 & 5.52 \\
    out & epoch 5 & 0.3 & $c_{\mathrm{rec}}=0$ & 10 & 0.24 & [0.19, 0.28] & 9.74(9) & $<10^{-4}$ & 55.00 & 0.0025 & 3.08 \\
    out & mean & 0.3 & $c_{\mathrm{rec}}=0$ & 10 & 0.25 & [0.23, 0.28] & 17.55(9) & $<10^{-4}$ & 55.00 & 0.00253 & 5.55 \\
    5 & best & 0.3 & $c_{\mathrm{rec}}=0$ & 10 & 0.17 & [0.13, 0.19] & 9.94(9) & $<10^{-4}$ & 55.00 & 0.00249 & 3.14 \\
    5 & epoch 10 & 0.3 & $c_{\mathrm{rec}}=0$ & 10 & 0.14 & [0.11, 0.16] & 9.16(9) & $<10^{-4}$ & 55.00 & 0.00246 & 2.90 \\
    5 & epoch 1 & 0.3 & $c_{\mathrm{rec}}=0$ & 10 & 0.27 & [0.23, 0.31] & 12.12(9) & $<10^{-4}$ & 55.00 & 0.00253 & 3.83 \\
    5 & epoch 5 & 0.3 & $c_{\mathrm{rec}}=0$ & 10 & 0.13 & [0.09, 0.18] & 5.91(9) & 0.000113 & 55.00 & 0.00245 & 1.87 \\
    5 & mean & 0.3 & $c_{\mathrm{rec}}=0$ & 10 & 0.16 & [0.14, 0.17] & 16.06(9) & $<10^{-4}$ & 55.00 & 0.00253 & 5.08 \\
    4 & best & 0.02 & $c_{\mathrm{rec}}=0$ & 10 & 0.00 & [-0.01, 0.01] & 0.43(9) & 0.339 & 11.00 & 0.167 & 0.14 \\
    4 & epoch 10 & 0 & second-best $c_{\mathrm{rec}}=0.02$ & 10 & 0.00 & [-0.01, 0.01] & 0.43(9) & 0.339 & 8.00 & 0.444 & 0.14 \\
    4 & epoch 1 & 0.3 & $c_{\mathrm{rec}}=0$ & 10 & 0.00 & [-0.02, 0.03] & 0.18(9) & 0.429 & 19.00 & 0.666 & 0.06 \\
    4 & epoch 5 & 0 & second-best $c_{\mathrm{rec}}=0.02$ & 10 & -0.00 & [-0.02, 0.02] & -0.00(9) & 0.5 & 9.00 & 0.624 & -0.00 \\
    4 & mean & 0.3 & $c_{\mathrm{rec}}=0$ & 10 & 0.01 & [-0.00, 0.02] & 1.17(9) & 0.137 & 38.50 & 0.131 & 0.37 \\
    3 & best & 0.3 & $c_{\mathrm{rec}}=0$ & 10 & 0.32 & [0.30, 0.34] & 29.05(9) & $<10^{-4}$ & 55.00 & 0.00246 & 9.19 \\
    3 & epoch 10 & 0.3 & $c_{\mathrm{rec}}=0$ & 10 & 0.40 & [0.38, 0.43] & 30.85(9) & $<10^{-4}$ & 55.00 & 0.00249 & 9.76 \\
    3 & epoch 1 & 0.3 & $c_{\mathrm{rec}}=0$ & 10 & 0.12 & [0.05, 0.19] & 3.04(9) & 0.00704 & 49.00 & 0.0139 & 0.96 \\
    3 & epoch 5 & 0.3 & $c_{\mathrm{rec}}=0$ & 10 & 0.32 & [0.26, 0.38] & 10.00(9) & $<10^{-4}$ & 55.00 & 0.00247 & 3.16 \\
    3 & mean & 0.3 & $c_{\mathrm{rec}}=0$ & 10 & 0.27 & [0.25, 0.28] & 30.42(9) & $<10^{-4}$ & 55.00 & 0.00253 & 9.62 \\
    2 & best & 0.3 & $c_{\mathrm{rec}}=0$ & 10 & 0.17 & [0.15, 0.20] & 12.28(9) & $<10^{-4}$ & 55.00 & 0.00225 & 3.88 \\
    2 & epoch 10 & 0.3 & $c_{\mathrm{rec}}=0$ & 10 & 0.19 & [0.14, 0.24] & 6.52(9) & $<10^{-4}$ & 55.00 & 0.0025 & 2.06 \\
    2 & epoch 1 & 0 & second-best $c_{\mathrm{rec}}=0.02$ & 10 & 0.02 & [-0.01, 0.04] & 1.41(9) & 0.0957 & 28.00 & 0.0779 & 0.45 \\
    2 & epoch 5 & 0.3 & $c_{\mathrm{rec}}=0$ & 10 & 0.17 & [0.12, 0.21] & 7.14(9) & $<10^{-4}$ & 55.00 & 0.00249 & 2.26 \\
    2 & mean & 0.3 & $c_{\mathrm{rec}}=0$ & 10 & 0.15 & [0.13, 0.17] & 16.73(9) & $<10^{-4}$ & 55.00 & 0.00253 & 5.29 \\
    1 & best & 0.02 & $c_{\mathrm{rec}}=0$ & 10 & 0.01 & [-0.02, 0.04] & 0.71(9) & 0.247 & 18.50 & 0.221 & 0.23 \\
    1 & epoch 10 & 0.02 & $c_{\mathrm{rec}}=0$ & 10 & 0.02 & [-0.02, 0.06] & 1.08(9) & 0.155 & 24.50 & 0.181 & 0.34 \\
    1 & epoch 1 & 0.02 & $c_{\mathrm{rec}}=0$ & 10 & 0.04 & [0.02, 0.06] & 3.00(9) & 0.00748 & 34.00 & 0.0122 & 0.95 \\
    1 & epoch 5 & 0.02 & $c_{\mathrm{rec}}=0$ & 10 & 0.01 & [-0.02, 0.04] & 0.74(9) & 0.239 & 17.50 & 0.273 & 0.23 \\
    1 & mean & 0.02 & $c_{\mathrm{rec}}=0$ & 10 & 0.02 & [0.01, 0.03] & 4.17(9) & 0.0012 & 55.00 & 0.00247 & 1.32 \\
    \bottomrule
    \end{tabular*}
    \tablefootnote{$\Delta$ is the paired mean difference in percentage points. CI is a bootstrap percentile confidence interval for $\Delta$. $p_t$ is the one-sided paired t-test p-value, $W$ and $p_W$ are from the one-sided Wilcoxon signed-rank test, and $d_z$ is the paired-sample standardized effect size.}
\end{table}

\begin{table}[H]
    \centering
    \small
    \caption{woof\_simplecnnv2\_relu\_e2\_0.3421}
    \label{supp:tab:woof-woof-simplecnnv2-relu-e2-0.3421-10e-paired-tests}
    \begin{tabular*}{\textwidth}{@{\extracolsep{\fill}}llclrrcrrrrr}
    \toprule
    $l_{r}$ & Metric & $c_{\mathrm{rec}}^{\ast}$ & Reference & $n$ & $\Delta$ & 95\% CI & $t(df)$ & $p_t$ & $W$ & $p_W$ & $d_z$ \\
    \midrule
    out & best & 0.3 & $c_{\mathrm{rec}}=0$ & 10 & 0.20 & [0.10, 0.31] & 3.49(9) & 0.00341 & 52.00 & 0.0062 & 1.10 \\
    out & epoch 10 & 0.1 & $c_{\mathrm{rec}}=0$ & 10 & 0.20 & [0.10, 0.30] & 3.90(9) & 0.0018 & 52.00 & 0.00623 & 1.23 \\
    out & epoch 1 & 0.3 & $c_{\mathrm{rec}}=0$ & 10 & 0.75 & [0.64, 0.85] & 13.13(9) & $<10^{-4}$ & 55.00 & 0.00253 & 4.15 \\
    out & epoch 5 & 0.1 & $c_{\mathrm{rec}}=0$ & 10 & 0.53 & [0.34, 0.70] & 5.44(9) & 0.000204 & 54.00 & 0.00344 & 1.72 \\
    out & mean & 0.3 & $c_{\mathrm{rec}}=0$ & 10 & 0.40 & [0.34, 0.46] & 11.96(9) & $<10^{-4}$ & 55.00 & 0.00252 & 3.78 \\
    6 & best & 0.3 & $c_{\mathrm{rec}}=0$ & 10 & 0.10 & [0.00, 0.20] & 1.97(9) & 0.04 & 43.50 & 0.0512 & 0.62 \\
    6 & epoch 10 & 0 & second-best $c_{\mathrm{rec}}=0.3$ & 10 & 0.16 & [-0.04, 0.34] & 1.48(9) & 0.0862 & 39.50 & 0.11 & 0.47 \\
    6 & epoch 1 & 0.3 & $c_{\mathrm{rec}}=0$ & 10 & 0.39 & [0.23, 0.53] & 4.78(9) & 0.000497 & 53.00 & 0.00465 & 1.51 \\
    6 & epoch 5 & 0.3 & $c_{\mathrm{rec}}=0$ & 10 & 0.29 & [0.04, 0.57] & 1.97(9) & 0.0399 & 45.00 & 0.0368 & 0.62 \\
    6 & mean & 0.3 & $c_{\mathrm{rec}}=0$ & 10 & 0.16 & [0.08, 0.24] & 3.71(9) & 0.00241 & 51.00 & 0.0083 & 1.17 \\
    5 & best & 0 & second-best $c_{\mathrm{rec}}=0.1$ & 10 & 0.30 & [0.19, 0.41] & 4.95(9) & 0.000394 & 55.00 & 0.00246 & 1.57 \\
    5 & epoch 10 & 0 & second-best $c_{\mathrm{rec}}=0.3$ & 10 & 0.16 & [0.04, 0.28] & 2.39(9) & 0.0203 & 32.50 & 0.0209 & 0.76 \\
    5 & epoch 1 & 0 & second-best $c_{\mathrm{rec}}=0.1$ & 10 & 0.47 & [0.29, 0.64] & 4.85(9) & 0.000453 & 55.00 & 0.00252 & 1.53 \\
    5 & epoch 5 & 0 & second-best $c_{\mathrm{rec}}=0.3$ & 10 & 0.10 & [-0.03, 0.27] & 1.29(9) & 0.115 & 22.00 & 0.288 & 0.41 \\
    5 & mean & 0 & second-best $c_{\mathrm{rec}}=0.1$ & 10 & 0.28 & [0.22, 0.33] & 9.29(9) & $<10^{-4}$ & 55.00 & 0.00253 & 2.94 \\
    4 & best & 0.3 & $c_{\mathrm{rec}}=0$ & 10 & 0.28 & [0.13, 0.42] & 3.54(9) & 0.00317 & 51.00 & 0.0083 & 1.12 \\
    4 & epoch 10 & 0.3 & $c_{\mathrm{rec}}=0$ & 10 & 0.07 & [-0.05, 0.19] & 1.01(9) & 0.17 & 36.50 & 0.179 & 0.32 \\
    4 & epoch 1 & 0.3 & $c_{\mathrm{rec}}=0$ & 10 & 0.48 & [0.33, 0.62] & 6.09(9) & $<10^{-4}$ & 55.00 & 0.00252 & 1.93 \\
    4 & epoch 5 & 0.3 & $c_{\mathrm{rec}}=0$ & 10 & 0.09 & [-0.21, 0.40] & 0.55(9) & 0.297 & 32.00 & 0.323 & 0.17 \\
    4 & mean & 0.3 & $c_{\mathrm{rec}}=0$ & 10 & 0.13 & [0.08, 0.18] & 4.80(9) & 0.000487 & 55.00 & 0.00253 & 1.52 \\
    3 & best & 0.1 & $c_{\mathrm{rec}}=0$ & 10 & 0.03 & [-0.07, 0.14] & 0.58(9) & 0.287 & 32.00 & 0.323 & 0.18 \\
    3 & epoch 10 & 0.1 & $c_{\mathrm{rec}}=0$ & 10 & 0.52 & [0.39, 0.69] & 6.49(9) & $<10^{-4}$ & 55.00 & 0.00252 & 2.05 \\
    3 & epoch 1 & 0.1 & $c_{\mathrm{rec}}=0$ & 10 & 0.13 & [-0.10, 0.34] & 1.11(9) & 0.148 & 40.00 & 0.101 & 0.35 \\
    3 & epoch 5 & 0.3 & $c_{\mathrm{rec}}=0$ & 10 & 0.08 & [-0.16, 0.32] & 0.62(9) & 0.276 & 34.00 & 0.254 & 0.20 \\
    3 & mean & 0.1 & $c_{\mathrm{rec}}=0$ & 10 & 0.15 & [0.07, 0.23] & 3.24(9) & 0.00508 & 55.00 & 0.00253 & 1.02 \\
    2 & best & 0 & second-best $c_{\mathrm{rec}}=0.3$ & 10 & 0.07 & [-0.02, 0.15] & 1.62(9) & 0.0701 & 36.00 & 0.0547 & 0.51 \\
    2 & epoch 10 & 0 & second-best $c_{\mathrm{rec}}=0.3$ & 10 & 0.12 & [-0.03, 0.26] & 1.50(9) & 0.0842 & 40.00 & 0.101 & 0.47 \\
    2 & epoch 1 & 0.1 & $c_{\mathrm{rec}}=0$ & 10 & 0.12 & [0.01, 0.24] & 1.86(9) & 0.0481 & 29.00 & 0.0615 & 0.59 \\
    2 & epoch 5 & 0.3 & $c_{\mathrm{rec}}=0$ & 10 & 0.01 & [-0.11, 0.12] & 0.16(9) & 0.439 & 28.50 & 0.459 & 0.05 \\
    2 & mean & 0 & second-best $c_{\mathrm{rec}}=0.3$ & 10 & 0.01 & [-0.04, 0.05] & 0.45(9) & 0.331 & 33.00 & 0.288 & 0.14 \\
    1 & best & 0.3 & $c_{\mathrm{rec}}=0$ & 10 & 0.29 & [0.22, 0.36] & 7.14(9) & $<10^{-4}$ & 55.00 & 0.00246 & 2.26 \\
    1 & epoch 10 & 0.3 & $c_{\mathrm{rec}}=0$ & 10 & 0.28 & [0.17, 0.39] & 4.52(9) & 0.000721 & 55.00 & 0.0025 & 1.43 \\
    1 & epoch 1 & 0 & second-best $c_{\mathrm{rec}}=0.1$ & 10 & 0.03 & [-0.02, 0.08] & 1.05(9) & 0.161 & 25.00 & 0.163 & 0.33 \\
    1 & epoch 5 & 0.3 & $c_{\mathrm{rec}}=0$ & 10 & 0.20 & [0.08, 0.32] & 3.10(9) & 0.00636 & 50.50 & 0.00949 & 0.98 \\
    1 & mean & 0.3 & $c_{\mathrm{rec}}=0$ & 10 & 0.16 & [0.10, 0.21] & 5.47(9) & 0.000198 & 55.00 & 0.00253 & 1.73 \\
    \bottomrule
    \end{tabular*}
    \tablefootnote{$\Delta$ is the paired mean difference in percentage points. CI is a bootstrap percentile confidence interval for $\Delta$. $p_t$ is the one-sided paired t-test p-value, $W$ and $p_W$ are from the one-sided Wilcoxon signed-rank test, and $d_z$ is the paired-sample standardized effect size.}
\end{table}

\begin{table}[H]
    \centering
    \small
    \caption{woof\_resnet18\_epoch20\_0.9458}
    \label{supp:tab:woof-woof-resnet18-epoch20-0.9458-10e-paired-tests}
    \begin{tabular*}{\textwidth}{@{\extracolsep{\fill}}llclrrcrrrrr}
    \toprule
    $l_{r}$ & Metric & $c_{\mathrm{rec}}^{\ast}$ & Reference & $n$ & $\Delta$ & 95\% CI & $t(df)$ & $p_t$ & $W$ & $p_W$ & $d_z$ \\
    \midrule
    out & best & 0.3 & $c_{\mathrm{rec}}=0$ & 10 & 0.24 & [0.19, 0.31] & 7.42(9) & $<10^{-4}$ & 55.00 & 0.00239 & 2.35 \\
    out & epoch 10 & 0.3 & $c_{\mathrm{rec}}=0$ & 10 & 0.26 & [0.19, 0.31] & 7.70(9) & $<10^{-4}$ & 55.00 & 0.00239 & 2.43 \\
    out & epoch 1 & 0.3 & $c_{\mathrm{rec}}=0$ & 10 & 0.34 & [0.21, 0.49] & 4.53(9) & 0.000716 & 54.00 & 0.00344 & 1.43 \\
    out & epoch 5 & 0.3 & $c_{\mathrm{rec}}=0$ & 10 & 0.23 & [0.18, 0.30] & 6.91(9) & $<10^{-4}$ & 55.00 & 0.00252 & 2.18 \\
    out & mean & 0.3 & $c_{\mathrm{rec}}=0$ & 10 & 0.27 & [0.24, 0.30] & 19.77(9) & $<10^{-4}$ & 55.00 & 0.00252 & 6.25 \\
    5 & best & 0.3 & $c_{\mathrm{rec}}=0$ & 10 & 0.19 & [0.14, 0.25] & 6.27(9) & $<10^{-4}$ & 55.00 & 0.00247 & 1.98 \\
    5 & epoch 10 & 0.3 & $c_{\mathrm{rec}}=0$ & 10 & 0.20 & [0.15, 0.26] & 6.99(9) & $<10^{-4}$ & 55.00 & 0.00246 & 2.21 \\
    5 & epoch 1 & 0.3 & $c_{\mathrm{rec}}=0$ & 10 & 0.29 & [0.19, 0.41] & 4.94(9) & 0.000403 & 55.00 & 0.0025 & 1.56 \\
    5 & epoch 5 & 0.3 & $c_{\mathrm{rec}}=0$ & 10 & 0.20 & [0.15, 0.25] & 6.77(9) & $<10^{-4}$ & 55.00 & 0.0025 & 2.14 \\
    5 & mean & 0.3 & $c_{\mathrm{rec}}=0$ & 10 & 0.21 & [0.18, 0.25] & 11.21(9) & $<10^{-4}$ & 55.00 & 0.00253 & 3.55 \\
    4 & best & 0 & second-best $c_{\mathrm{rec}}=0.3$ & 10 & 0.00 & [-0.03, 0.04] & 0.14(9) & 0.445 & 16.50 & 0.335 & 0.04 \\
    4 & epoch 10 & 0.3 & $c_{\mathrm{rec}}=0$ & 10 & 0.02 & [-0.01, 0.04] & 1.03(9) & 0.164 & 31.00 & 0.155 & 0.33 \\
    4 & epoch 1 & 0.3 & $c_{\mathrm{rec}}=0$ & 10 & 0.04 & [-0.00, 0.08] & 1.71(9) & 0.0608 & 34.00 & 0.0854 & 0.54 \\
    4 & epoch 5 & 0 & second-best $c_{\mathrm{rec}}=0.02$ & 10 & 0.01 & [-0.01, 0.03] & 1.00(9) & 0.172 & 20.50 & 0.133 & 0.32 \\
    4 & mean & 0.3 & $c_{\mathrm{rec}}=0$ & 10 & 0.00 & [-0.01, 0.01] & 0.40(9) & 0.35 & 36.00 & 0.193 & 0.13 \\
    3 & best & 0.3 & $c_{\mathrm{rec}}=0$ & 10 & 0.04 & [0.01, 0.07] & 2.45(9) & 0.0184 & 34.00 & 0.0121 & 0.77 \\
    3 & epoch 10 & 0.3 & $c_{\mathrm{rec}}=0$ & 10 & 0.43 & [0.37, 0.49] & 14.22(9) & $<10^{-4}$ & 55.00 & 0.0025 & 4.50 \\
    3 & epoch 1 & 0.3 & $c_{\mathrm{rec}}=0$ & 10 & 0.05 & [0.01, 0.08] & 2.53(9) & 0.0162 & 48.50 & 0.0142 & 0.80 \\
    3 & epoch 5 & 0.3 & $c_{\mathrm{rec}}=0$ & 10 & 0.19 & [0.13, 0.26] & 5.69(9) & 0.000148 & 55.00 & 0.00252 & 1.80 \\
    3 & mean & 0.3 & $c_{\mathrm{rec}}=0$ & 10 & 0.23 & [0.21, 0.25] & 20.07(9) & $<10^{-4}$ & 55.00 & 0.00253 & 6.35 \\
    2 & best & 0.3 & $c_{\mathrm{rec}}=0$ & 10 & 0.03 & [-0.03, 0.07] & 1.08(9) & 0.154 & 40.50 & 0.0917 & 0.34 \\
    2 & epoch 10 & 0.3 & $c_{\mathrm{rec}}=0$ & 10 & 0.33 & [0.24, 0.40] & 7.73(9) & $<10^{-4}$ & 55.00 & 0.0025 & 2.44 \\
    2 & epoch 1 & 0 & second-best $c_{\mathrm{rec}}=0.02$ & 10 & 0.01 & [-0.02, 0.03] & 0.39(9) & 0.353 & 17.00 & 0.303 & 0.12 \\
    2 & epoch 5 & 0.3 & $c_{\mathrm{rec}}=0$ & 10 & 0.16 & [0.12, 0.20] & 7.58(9) & $<10^{-4}$ & 55.00 & 0.0025 & 2.40 \\
    2 & mean & 0.3 & $c_{\mathrm{rec}}=0$ & 10 & 0.14 & [0.12, 0.16] & 14.59(9) & $<10^{-4}$ & 55.00 & 0.00253 & 4.61 \\
    1 & best & 0 & second-best $c_{\mathrm{rec}}=0.02$ & 10 & 0.00 & [-0.04, 0.04] & 0.00(9) & 0.5 & 26.00 & 0.561 & 0.00 \\
    1 & epoch 10 & 0.3 & $c_{\mathrm{rec}}=0$ & 10 & 0.12 & [0.07, 0.18] & 4.13(9) & 0.00128 & 45.00 & 0.00379 & 1.31 \\
    1 & epoch 1 & 0.3 & $c_{\mathrm{rec}}=0$ & 10 & 0.02 & [-0.02, 0.06] & 0.88(9) & 0.2 & 25.00 & 0.162 & 0.28 \\
    1 & epoch 5 & 0 & second-best $c_{\mathrm{rec}}=0.02$ & 10 & 0.02 & [-0.02, 0.05] & 1.00(9) & 0.172 & 20.00 & 0.15 & 0.32 \\
    1 & mean & 0.02 & $c_{\mathrm{rec}}=0$ & 10 & 0.02 & [-0.00, 0.04] & 1.93(9) & 0.0429 & 44.00 & 0.0463 & 0.61 \\
    \bottomrule
    \end{tabular*}
    \tablefootnote{$\Delta$ is the paired mean difference in percentage points. CI is a bootstrap percentile confidence interval for $\Delta$. $p_t$ is the one-sided paired t-test p-value, $W$ and $p_W$ are from the one-sided Wilcoxon signed-rank test, and $d_z$ is the paired-sample standardized effect size.}
\end{table}

\begin{table}[H]
    \centering
    \small
    \caption{tin\_simplecnnv2\_relu\_e20\_0.3324}
    \label{supp:tab:tin-tin-simplecnnv2-relu-e20-0.3324-10e-paired-tests}
    \begin{tabular*}{\textwidth}{@{\extracolsep{\fill}}llclrrcrrrrr}
    \toprule
    $l_{r}$ & Metric & $c_{\mathrm{rec}}^{\ast}$ & Reference & $n$ & $\Delta$ & 95\% CI & $t(df)$ & $p_t$ & $W$ & $p_W$ & $d_z$ \\
    \midrule
    out & best & 0 & second-best $c_{\mathrm{rec}}=0.02$ & 10 & 0.31 & [0.23, 0.38] & 7.26(9) & $<10^{-4}$ & 55.00 & 0.00252 & 2.30 \\
    out & epoch 10 & 0.02 & $c_{\mathrm{rec}}=0$ & 10 & 0.13 & [-0.02, 0.30] & 1.49(9) & 0.0857 & 38.00 & 0.142 & 0.47 \\
    out & epoch 1 & 0 & second-best $c_{\mathrm{rec}}=0.02$ & 10 & 0.31 & [0.15, 0.45] & 3.89(9) & 0.00184 & 52.00 & 0.00614 & 1.23 \\
    out & epoch 5 & 0.02 & $c_{\mathrm{rec}}=0$ & 10 & 0.01 & [-0.09, 0.11] & 0.17(9) & 0.436 & 30.50 & 0.38 & 0.05 \\
    out & mean & 0 & second-best $c_{\mathrm{rec}}=0.02$ & 10 & 0.06 & [0.02, 0.09] & 2.83(9) & 0.0099 & 49.00 & 0.0142 & 0.89 \\
    5 & best & 0.1 & $c_{\mathrm{rec}}=0$ & 10 & 0.13 & [0.06, 0.21] & 3.20(9) & 0.00544 & 50.50 & 0.00949 & 1.01 \\
    5 & epoch 10 & 0.1 & $c_{\mathrm{rec}}=0$ & 10 & 0.37 & [0.29, 0.45] & 8.82(9) & $<10^{-4}$ & 55.00 & 0.00253 & 2.79 \\
    5 & epoch 1 & 0 & second-best $c_{\mathrm{rec}}=0.02$ & 10 & 0.10 & [0.04, 0.17] & 2.80(9) & 0.0104 & 49.00 & 0.0142 & 0.88 \\
    5 & epoch 5 & 0.1 & $c_{\mathrm{rec}}=0$ & 10 & 0.15 & [0.04, 0.25] & 2.63(9) & 0.0138 & 48.00 & 0.0183 & 0.83 \\
    5 & mean & 0.1 & $c_{\mathrm{rec}}=0$ & 10 & 0.18 & [0.15, 0.22] & 10.50(9) & $<10^{-4}$ & 55.00 & 0.00253 & 3.32 \\
    4 & best & 0.02 & $c_{\mathrm{rec}}=0$ & 10 & 0.02 & [-0.02, 0.06] & 0.94(9) & 0.187 & 29.50 & 0.203 & 0.30 \\
    4 & epoch 10 & 0.02 & $c_{\mathrm{rec}}=0$ & 10 & 0.01 & [-0.05, 0.06] & 0.16(9) & 0.44 & 28.50 & 0.459 & 0.05 \\
    4 & epoch 1 & 0 & second-best $c_{\mathrm{rec}}=0.02$ & 10 & 0.03 & [-0.02, 0.08] & 1.31(9) & 0.111 & 37.00 & 0.0426 & 0.42 \\
    4 & epoch 5 & 0 & second-best $c_{\mathrm{rec}}=0.02$ & 10 & 0.05 & [-0.01, 0.11] & 1.59(9) & 0.0734 & 36.00 & 0.0549 & 0.50 \\
    4 & mean & 0 & second-best $c_{\mathrm{rec}}=0.02$ & 10 & 0.01 & [-0.01, 0.03] & 1.33(9) & 0.108 & 39.00 & 0.121 & 0.42 \\
    3 & best & 0.1 & $c_{\mathrm{rec}}=0$ & 10 & 0.02 & [-0.04, 0.08] & 0.52(9) & 0.308 & 25.50 & 0.361 & 0.16 \\
    3 & epoch 10 & 0.02 & $c_{\mathrm{rec}}=0$ & 10 & 0.01 & [-0.03, 0.05] & 0.75(9) & 0.238 & 41.00 & 0.0838 & 0.24 \\
    3 & epoch 1 & 0 & second-best $c_{\mathrm{rec}}=0.02$ & 10 & 0.00 & [-0.03, 0.03] & 0.19(9) & 0.427 & 30.00 & 0.399 & 0.06 \\
    3 & epoch 5 & 0.1 & $c_{\mathrm{rec}}=0$ & 10 & 0.04 & [-0.02, 0.09] & 1.21(9) & 0.129 & 31.00 & 0.157 & 0.38 \\
    3 & mean & 0.1 & $c_{\mathrm{rec}}=0$ & 10 & 0.01 & [-0.01, 0.03] & 0.82(9) & 0.217 & 34.00 & 0.254 & 0.26 \\
    2 & best & 0.1 & $c_{\mathrm{rec}}=0$ & 10 & 0.15 & [0.10, 0.19] & 5.64(9) & 0.000158 & 55.00 & 0.0025 & 1.78 \\
    2 & epoch 10 & 0.1 & $c_{\mathrm{rec}}=0$ & 10 & 0.14 & [0.11, 0.17] & 8.96(9) & $<10^{-4}$ & 55.00 & 0.00246 & 2.83 \\
    2 & epoch 1 & 0.1 & $c_{\mathrm{rec}}=0$ & 10 & 0.04 & [-0.00, 0.08] & 1.81(9) & 0.0519 & 44.00 & 0.0461 & 0.57 \\
    2 & epoch 5 & 0.1 & $c_{\mathrm{rec}}=0$ & 10 & 0.11 & [0.04, 0.18] & 2.94(9) & 0.00821 & 50.00 & 0.0109 & 0.93 \\
    2 & mean & 0.1 & $c_{\mathrm{rec}}=0$ & 10 & 0.12 & [0.10, 0.13] & 17.08(9) & $<10^{-4}$ & 55.00 & 0.00253 & 5.40 \\
    1 & best & 0.1 & $c_{\mathrm{rec}}=0$ & 10 & 0.03 & [0.01, 0.04] & 2.86(9) & 0.00936 & 51.00 & 0.00809 & 0.91 \\
    1 & epoch 10 & 0.1 & $c_{\mathrm{rec}}=0$ & 10 & 0.00 & [-0.03, 0.03] & 0.13(9) & 0.45 & 24.50 & 0.406 & 0.04 \\
    1 & epoch 1 & 0.1 & $c_{\mathrm{rec}}=0$ & 10 & 0.02 & [-0.00, 0.04] & 1.85(9) & 0.0489 & 43.50 & 0.0508 & 0.58 \\
    1 & epoch 5 & 0 & second-best $c_{\mathrm{rec}}=0.02$ & 10 & 0.02 & [-0.01, 0.05] & 1.18(9) & 0.134 & 34.50 & 0.077 & 0.37 \\
    1 & mean & 0 & second-best $c_{\mathrm{rec}}=0.1$ & 10 & 0.00 & [-0.00, 0.01] & 0.88(9) & 0.2 & 35.00 & 0.222 & 0.28 \\
    \bottomrule
    \end{tabular*}
    \tablefootnote{$\Delta$ is the paired mean difference in percentage points. CI is a bootstrap percentile confidence interval for $\Delta$. $p_t$ is the one-sided paired t-test p-value, $W$ and $p_W$ are from the one-sided Wilcoxon signed-rank test, and $d_z$ is the paired-sample standardized effect size.}
\end{table}

\begin{table}[H]
    \centering
    \small
    \caption{tin\_resnet34\_epoch20\_0.5872}
    \label{supp:tab:tin-tin-resnet34-epoch20-0.5872-10e-paired-tests}
    \begin{tabular*}{\textwidth}{@{\extracolsep{\fill}}llclrrcrrrrr}
    \toprule
    $l_{r}$ & Metric & $c_{\mathrm{rec}}^{\ast}$ & Reference & $n$ & $\Delta$ & 95\% CI & $t(df)$ & $p_t$ & $W$ & $p_W$ & $d_z$ \\
    \midrule
    out & best & 0.3 & $c_{\mathrm{rec}}=0$ & 10 & 0.62 & [0.45, 0.78] & 7.01(9) & $<10^{-4}$ & 55.00 & 0.00253 & 2.22 \\
    out & epoch 10 & 0.3 & $c_{\mathrm{rec}}=0$ & 10 & 0.74 & [0.59, 0.87] & 9.65(9) & $<10^{-4}$ & 55.00 & 0.00253 & 3.05 \\
    out & epoch 1 & 0.1 & $c_{\mathrm{rec}}=0$ & 10 & 0.52 & [0.41, 0.63] & 8.64(9) & $<10^{-4}$ & 55.00 & 0.00252 & 2.73 \\
    out & epoch 5 & 0.3 & $c_{\mathrm{rec}}=0$ & 10 & 0.76 & [0.65, 0.87] & 12.74(9) & $<10^{-4}$ & 55.00 & 0.00253 & 4.03 \\
    out & mean & 0.3 & $c_{\mathrm{rec}}=0$ & 10 & 0.80 & [0.71, 0.88] & 17.09(9) & $<10^{-4}$ & 55.00 & 0.00253 & 5.40 \\
    5 & best & 0.3 & $c_{\mathrm{rec}}=0$ & 10 & 0.51 & [0.33, 0.68] & 5.38(9) & 0.000222 & 54.00 & 0.00346 & 1.70 \\
    5 & epoch 10 & 0.3 & $c_{\mathrm{rec}}=0$ & 10 & 0.75 & [0.53, 0.97] & 6.37(9) & $<10^{-4}$ & 55.00 & 0.00253 & 2.01 \\
    5 & epoch 1 & 0.1 & $c_{\mathrm{rec}}=0$ & 10 & 0.47 & [0.34, 0.62] & 6.27(9) & $<10^{-4}$ & 55.00 & 0.00253 & 1.98 \\
    5 & epoch 5 & 0.3 & $c_{\mathrm{rec}}=0$ & 10 & 0.75 & [0.63, 0.88] & 10.83(9) & $<10^{-4}$ & 55.00 & 0.00252 & 3.43 \\
    5 & mean & 0.3 & $c_{\mathrm{rec}}=0$ & 10 & 0.73 & [0.63, 0.83] & 13.44(9) & $<10^{-4}$ & 55.00 & 0.00253 & 4.25 \\
    4 & best & 0.3 & $c_{\mathrm{rec}}=0$ & 10 & 0.35 & [0.18, 0.50] & 4.01(9) & 0.00154 & 52.00 & 0.00623 & 1.27 \\
    4 & epoch 10 & 0.3 & $c_{\mathrm{rec}}=0$ & 10 & 0.52 & [0.35, 0.70] & 5.48(9) & 0.000196 & 55.00 & 0.00253 & 1.73 \\
    4 & epoch 1 & 0.3 & $c_{\mathrm{rec}}=0$ & 10 & 0.39 & [0.22, 0.54] & 4.58(9) & 0.000667 & 53.00 & 0.00467 & 1.45 \\
    4 & epoch 5 & 0.3 & $c_{\mathrm{rec}}=0$ & 10 & 0.49 & [0.35, 0.62] & 6.86(9) & $<10^{-4}$ & 55.00 & 0.00252 & 2.17 \\
    4 & mean & 0.3 & $c_{\mathrm{rec}}=0$ & 10 & 0.50 & [0.45, 0.56] & 17.41(9) & $<10^{-4}$ & 55.00 & 0.00253 & 5.51 \\
    3 & best & 0.3 & $c_{\mathrm{rec}}=0$ & 10 & 0.10 & [0.02, 0.20] & 2.05(9) & 0.0351 & 48.00 & 0.0183 & 0.65 \\
    3 & epoch 10 & 0.3 & $c_{\mathrm{rec}}=0$ & 10 & 1.08 & [0.92, 1.24] & 12.84(9) & $<10^{-4}$ & 55.00 & 0.00252 & 4.06 \\
    3 & epoch 1 & 0.3 & $c_{\mathrm{rec}}=0$ & 10 & 0.06 & [-0.02, 0.15] & 1.26(9) & 0.119 & 38.50 & 0.131 & 0.40 \\
    3 & epoch 5 & 0.3 & $c_{\mathrm{rec}}=0$ & 10 & 0.65 & [0.55, 0.75] & 11.84(9) & $<10^{-4}$ & 55.00 & 0.00253 & 3.74 \\
    3 & mean & 0.3 & $c_{\mathrm{rec}}=0$ & 10 & 0.66 & [0.61, 0.71] & 26.23(9) & $<10^{-4}$ & 55.00 & 0.00253 & 8.30 \\
    2 & best & 0.3 & $c_{\mathrm{rec}}=0$ & 10 & 0.14 & [0.09, 0.21] & 4.21(9) & 0.00114 & 54.00 & 0.00346 & 1.33 \\
    2 & epoch 10 & 0.3 & $c_{\mathrm{rec}}=0$ & 10 & 0.43 & [0.30, 0.54] & 6.80(9) & $<10^{-4}$ & 54.00 & 0.00344 & 2.15 \\
    2 & epoch 1 & 0 & second-best $c_{\mathrm{rec}}=0.1$ & 10 & 0.08 & [-0.00, 0.17] & 1.78(9) & 0.0547 & 42.00 & 0.0697 & 0.56 \\
    2 & epoch 5 & 0.3 & $c_{\mathrm{rec}}=0$ & 10 & 0.11 & [0.03, 0.21] & 2.31(9) & 0.023 & 43.50 & 0.0513 & 0.73 \\
    2 & mean & 0.3 & $c_{\mathrm{rec}}=0$ & 10 & 0.21 & [0.17, 0.25] & 10.11(9) & $<10^{-4}$ & 55.00 & 0.00253 & 3.20 \\
    1 & best & 0.1 & $c_{\mathrm{rec}}=0$ & 10 & 0.03 & [-0.03, 0.10] & 0.97(9) & 0.179 & 39.00 & 0.12 & 0.31 \\
    1 & epoch 10 & 0.1 & $c_{\mathrm{rec}}=0$ & 10 & 0.21 & [0.13, 0.29] & 4.73(9) & 0.000539 & 54.00 & 0.00346 & 1.49 \\
    1 & epoch 1 & 0 & second-best $c_{\mathrm{rec}}=0.1$ & 10 & 0.09 & [0.02, 0.14] & 2.75(9) & 0.0112 & 47.00 & 0.0234 & 0.87 \\
    1 & epoch 5 & 0.1 & $c_{\mathrm{rec}}=0$ & 10 & 0.03 & [-0.02, 0.08] & 0.90(9) & 0.196 & 38.00 & 0.142 & 0.28 \\
    1 & mean & 0.1 & $c_{\mathrm{rec}}=0$ & 10 & 0.04 & [0.01, 0.07] & 2.18(9) & 0.0285 & 49.00 & 0.0142 & 0.69 \\
    \bottomrule
    \end{tabular*}
    \tablefootnote{$\Delta$ is the paired mean difference in percentage points. CI is a bootstrap percentile confidence interval for $\Delta$. $p_t$ is the one-sided paired t-test p-value, $W$ and $p_W$ are from the one-sided Wilcoxon signed-rank test, and $d_z$ is the paired-sample standardized effect size.}
\end{table}

\begin{table}[H]
    \centering
    \small
    \caption{tin\_simplevit\_p8\_d192\_l9\_e60\_0.4491}
    \label{supp:tab:tin-tin-simplevit-p8-d192-l9-e60-0.4491-10e-paired-tests}
    \begin{tabular*}{\textwidth}{@{\extracolsep{\fill}}llclrrcrrrrr}
    \toprule
    $l_{r}$ & Metric & $c_{\mathrm{rec}}^{\ast}$ & Reference & $n$ & $\Delta$ & 95\% CI & $t(df)$ & $p_t$ & $W$ & $p_W$ & $d_z$ \\
    \midrule
    out & best & 0.1 & $c_{\mathrm{rec}}=0$ & 10 & 0.57 & [0.46, 0.68] & 9.47(9) & $<10^{-4}$ & 55.00 & 0.00253 & 3.00 \\
    out & epoch 10 & 0.3 & $c_{\mathrm{rec}}=0$ & 10 & 3.66 & [3.59, 3.73] & 99.40(9) & $<10^{-4}$ & 55.00 & 0.00253 & 31.43 \\
    out & epoch 1 & 0.1 & $c_{\mathrm{rec}}=0$ & 10 & 0.14 & [0.04, 0.23] & 2.62(9) & 0.014 & 48.00 & 0.0183 & 0.83 \\
    out & epoch 5 & 0.1 & $c_{\mathrm{rec}}=0$ & 10 & 1.93 & [1.80, 2.05] & 28.73(9) & $<10^{-4}$ & 55.00 & 0.00253 & 9.09 \\
    out & mean & 0.1 & $c_{\mathrm{rec}}=0$ & 10 & 2.11 & [2.08, 2.14] & 130.66(9) & $<10^{-4}$ & 55.00 & 0.00253 & 41.32 \\
    10 & best & 0.1 & $c_{\mathrm{rec}}=0$ & 10 & 0.69 & [0.55, 0.82] & 9.62(9) & $<10^{-4}$ & 55.00 & 0.00253 & 3.04 \\
    10 & epoch 10 & 0.3 & $c_{\mathrm{rec}}=0$ & 10 & 3.71 & [3.67, 3.76] & 149.18(9) & $<10^{-4}$ & 55.00 & 0.00253 & 47.18 \\
    10 & epoch 1 & 0.1 & $c_{\mathrm{rec}}=0$ & 10 & 0.24 & [0.12, 0.35] & 3.91(9) & 0.00177 & 52.00 & 0.00623 & 1.24 \\
    10 & epoch 5 & 0.1 & $c_{\mathrm{rec}}=0$ & 10 & 2.27 & [2.15, 2.40] & 32.74(9) & $<10^{-4}$ & 55.00 & 0.00253 & 10.35 \\
    10 & mean & 0.3 & $c_{\mathrm{rec}}=0$ & 10 & 2.13 & [2.10, 2.16] & 134.21(9) & $<10^{-4}$ & 55.00 & 0.00253 & 42.44 \\
    9 & best & 0.3 & $c_{\mathrm{rec}}=0$ & 10 & 0.85 & [0.73, 0.97] & 13.56(9) & $<10^{-4}$ & 55.00 & 0.00253 & 4.29 \\
    9 & epoch 10 & 0.3 & $c_{\mathrm{rec}}=0$ & 10 & 3.89 & [3.83, 3.94] & 131.76(9) & $<10^{-4}$ & 55.00 & 0.0025 & 41.67 \\
    9 & epoch 1 & 0.1 & $c_{\mathrm{rec}}=0$ & 10 & 0.70 & [0.57, 0.83] & 10.10(9) & $<10^{-4}$ & 55.00 & 0.00252 & 3.19 \\
    9 & epoch 5 & 0.3 & $c_{\mathrm{rec}}=0$ & 10 & 2.33 & [2.22, 2.45] & 35.21(9) & $<10^{-4}$ & 55.00 & 0.00253 & 11.13 \\
    9 & mean & 0.3 & $c_{\mathrm{rec}}=0$ & 10 & 2.45 & [2.42, 2.47] & 190.09(9) & $<10^{-4}$ & 55.00 & 0.00253 & 60.11 \\
    8 & best & 0.3 & $c_{\mathrm{rec}}=0$ & 10 & 0.76 & [0.67, 0.85] & 15.44(9) & $<10^{-4}$ & 55.00 & 0.00253 & 4.88 \\
    8 & epoch 10 & 0.3 & $c_{\mathrm{rec}}=0$ & 10 & 3.82 & [3.73, 3.89] & 89.91(9) & $<10^{-4}$ & 55.00 & 0.00253 & 28.43 \\
    8 & epoch 1 & 0.1 & $c_{\mathrm{rec}}=0$ & 10 & 0.61 & [0.51, 0.73] & 10.32(9) & $<10^{-4}$ & 55.00 & 0.00252 & 3.26 \\
    8 & epoch 5 & 0.3 & $c_{\mathrm{rec}}=0$ & 10 & 2.35 & [2.27, 2.45] & 46.30(9) & $<10^{-4}$ & 55.00 & 0.00252 & 14.64 \\
    8 & mean & 0.3 & $c_{\mathrm{rec}}=0$ & 10 & 2.44 & [2.40, 2.48] & 116.65(9) & $<10^{-4}$ & 55.00 & 0.00253 & 36.89 \\
    7 & best & 0.3 & $c_{\mathrm{rec}}=0$ & 10 & 0.61 & [0.55, 0.68] & 17.60(9) & $<10^{-4}$ & 55.00 & 0.00253 & 5.57 \\
    7 & epoch 10 & 0.3 & $c_{\mathrm{rec}}=0$ & 10 & 2.68 & [2.60, 2.76] & 59.67(9) & $<10^{-4}$ & 55.00 & 0.00252 & 18.87 \\
    7 & epoch 1 & 0.1 & $c_{\mathrm{rec}}=0$ & 10 & 0.59 & [0.51, 0.65] & 14.82(9) & $<10^{-4}$ & 55.00 & 0.00252 & 4.69 \\
    7 & epoch 5 & 0.3 & $c_{\mathrm{rec}}=0$ & 10 & 1.42 & [1.33, 1.52] & 27.10(9) & $<10^{-4}$ & 55.00 & 0.00253 & 8.57 \\
    7 & mean & 0.3 & $c_{\mathrm{rec}}=0$ & 10 & 1.58 & [1.56, 1.60] & 141.06(9) & $<10^{-4}$ & 55.00 & 0.00253 & 44.61 \\
    6 & best & 0.1 & $c_{\mathrm{rec}}=0$ & 10 & 0.36 & [0.32, 0.40] & 15.44(9) & $<10^{-4}$ & 55.00 & 0.00253 & 4.88 \\
    6 & epoch 10 & 0.3 & $c_{\mathrm{rec}}=0$ & 10 & 1.47 & [1.41, 1.55] & 39.23(9) & $<10^{-4}$ & 55.00 & 0.00252 & 12.40 \\
    6 & epoch 1 & 0.1 & $c_{\mathrm{rec}}=0$ & 10 & 0.29 & [0.20, 0.38] & 5.70(9) & 0.000146 & 55.00 & 0.00252 & 1.80 \\
    6 & epoch 5 & 0.1 & $c_{\mathrm{rec}}=0$ & 10 & 0.90 & [0.80, 1.01] & 15.98(9) & $<10^{-4}$ & 55.00 & 0.00253 & 5.05 \\
    6 & mean & 0.3 & $c_{\mathrm{rec}}=0$ & 10 & 0.87 & [0.85, 0.89] & 76.57(9) & $<10^{-4}$ & 55.00 & 0.00253 & 24.22 \\
    5 & best & 0 & second-best $c_{\mathrm{rec}}=0.1$ & 10 & 0.04 & [-0.04, 0.11] & 0.89(9) & 0.197 & 35.00 & 0.222 & 0.28 \\
    5 & epoch 10 & 0.1 & $c_{\mathrm{rec}}=0$ & 10 & 0.59 & [0.51, 0.68] & 12.21(9) & $<10^{-4}$ & 55.00 & 0.00252 & 3.86 \\
    5 & epoch 1 & 0.1 & $c_{\mathrm{rec}}=0$ & 10 & 0.05 & [-0.03, 0.13] & 1.15(9) & 0.139 & 38.50 & 0.131 & 0.36 \\
    5 & epoch 5 & 0.1 & $c_{\mathrm{rec}}=0$ & 10 & 0.18 & [0.09, 0.25] & 4.02(9) & 0.0015 & 52.00 & 0.00623 & 1.27 \\
    5 & mean & 0.1 & $c_{\mathrm{rec}}=0$ & 10 & 0.23 & [0.19, 0.26] & 12.59(9) & $<10^{-4}$ & 55.00 & 0.00252 & 3.98 \\
    4 & best & 0.1 & $c_{\mathrm{rec}}=0$ & 10 & 0.04 & [0.00, 0.08] & 2.00(9) & 0.0384 & 38.00 & 0.033 & 0.63 \\
    4 & epoch 10 & 0.1 & $c_{\mathrm{rec}}=0$ & 10 & 0.46 & [0.41, 0.51] & 17.23(9) & $<10^{-4}$ & 55.00 & 0.0025 & 5.45 \\
    4 & epoch 1 & 0.1 & $c_{\mathrm{rec}}=0$ & 10 & 0.14 & [0.07, 0.20] & 4.09(9) & 0.00136 & 52.00 & 0.00626 & 1.29 \\
    4 & epoch 5 & 0 & second-best $c_{\mathrm{rec}}=0.1$ & 10 & 0.01 & [-0.05, 0.07] & 0.40(9) & 0.35 & 27.00 & 0.297 & 0.13 \\
    4 & mean & 0.1 & $c_{\mathrm{rec}}=0$ & 10 & 0.14 & [0.11, 0.17] & 9.03(9) & $<10^{-4}$ & 55.00 & 0.00253 & 2.86 \\
    3 & best & 0.1 & $c_{\mathrm{rec}}=0$ & 10 & 0.15 & [0.11, 0.20] & 6.54(9) & $<10^{-4}$ & 55.00 & 0.00253 & 2.07 \\
    3 & epoch 10 & 0.1 & $c_{\mathrm{rec}}=0$ & 10 & 0.22 & [0.16, 0.28] & 7.00(9) & $<10^{-4}$ & 55.00 & 0.00252 & 2.21 \\
    3 & epoch 1 & 0.1 & $c_{\mathrm{rec}}=0$ & 10 & 0.18 & [0.15, 0.22] & 10.03(9) & $<10^{-4}$ & 55.00 & 0.00252 & 3.17 \\
    3 & epoch 5 & 0.1 & $c_{\mathrm{rec}}=0$ & 10 & 0.15 & [0.09, 0.20] & 4.83(9) & 0.000466 & 55.00 & 0.00253 & 1.53 \\
    3 & mean & 0.1 & $c_{\mathrm{rec}}=0$ & 10 & 0.18 & [0.16, 0.20] & 15.33(9) & $<10^{-4}$ & 55.00 & 0.00253 & 4.85 \\
    2 & best & 0.1 & $c_{\mathrm{rec}}=0$ & 10 & 0.21 & [0.18, 0.24] & 12.06(9) & $<10^{-4}$ & 55.00 & 0.00252 & 3.81 \\
    2 & epoch 10 & 0.1 & $c_{\mathrm{rec}}=0$ & 10 & 0.57 & [0.52, 0.62] & 19.27(9) & $<10^{-4}$ & 55.00 & 0.00253 & 6.10 \\
    2 & epoch 1 & 0.1 & $c_{\mathrm{rec}}=0$ & 10 & 0.18 & [0.14, 0.23] & 7.64(9) & $<10^{-4}$ & 55.00 & 0.00253 & 2.42 \\
    2 & epoch 5 & 0.1 & $c_{\mathrm{rec}}=0$ & 10 & 0.34 & [0.27, 0.40] & 9.80(9) & $<10^{-4}$ & 55.00 & 0.00246 & 3.10 \\
    2 & mean & 0.1 & $c_{\mathrm{rec}}=0$ & 10 & 0.36 & [0.34, 0.38] & 32.81(9) & $<10^{-4}$ & 55.00 & 0.00253 & 10.38 \\
    1 & best & 0.1 & $c_{\mathrm{rec}}=0$ & 10 & 0.15 & [0.12, 0.19] & 8.32(9) & $<10^{-4}$ & 55.00 & 0.00252 & 2.63 \\
    1 & epoch 10 & 0.1 & $c_{\mathrm{rec}}=0$ & 10 & 0.19 & [0.15, 0.24] & 7.54(9) & $<10^{-4}$ & 55.00 & 0.00253 & 2.38 \\
    1 & epoch 1 & 0.1 & $c_{\mathrm{rec}}=0$ & 10 & 0.14 & [0.11, 0.18] & 7.02(9) & $<10^{-4}$ & 55.00 & 0.00247 & 2.22 \\
    1 & epoch 5 & 0.1 & $c_{\mathrm{rec}}=0$ & 10 & 0.25 & [0.18, 0.30] & 7.85(9) & $<10^{-4}$ & 55.00 & 0.00252 & 2.48 \\
    1 & mean & 0.1 & $c_{\mathrm{rec}}=0$ & 10 & 0.20 & [0.18, 0.22] & 19.33(9) & $<10^{-4}$ & 55.00 & 0.00253 & 6.11 \\
    \bottomrule
    \end{tabular*}
    \tablefootnote{$\Delta$ is the paired mean difference in percentage points. CI is a bootstrap percentile confidence interval for $\Delta$. $p_t$ is the one-sided paired t-test p-value, $W$ and $p_W$ are from the one-sided Wilcoxon signed-rank test, and $d_z$ is the paired-sample standardized effect size.}
\end{table}

Across the complete set of tables, the evidence is heterogeneous rather than uniformly positive across layers and metrics. The largest difference in best accuracy is observed for the Tiny-ImageNet ResNet-34 output reconstruction, with $\Delta=1.93$ percentage points, a 95\% CI of $[1.80,2.07]$, and $d_z=8.70$. The deeper SimpleViT reconstruction layers also show consistent best-accuracy differences on CIFAR-100 and Tiny-ImageNet, reaching $0.76$ and $0.83$ percentage points, respectively. These results demonstrate that the paired improvements extend beyond shallow CNNs.

The tables also expose important negative evidence. Several rows select $c_{\mathrm{rec}}^{\ast}=0$, for which a positive $\Delta$ favors the matched BP post-training reference over the strongest nonzero reconstruction setting. Other rows on CIFAR-10, CIFAR-100, Imagenette, Imagewoof, and Tiny-ImageNet have confidence intervals that include zero or yield disagreement between the parametric and rank-based tests. The best-accuracy differences for several shallow or early reconstruction layers are close to zero, and statistical consistency can vary across post-training epochs. The detailed analyses therefore support configuration-dependent improvements and validation-based selection of $l_R$, $c_{\mathrm{rec}}$, and post-training duration; they do not support uniform superiority over the matched $c_{\mathrm{rec}}=0$ reference. The asterisks in the main-paper tables retain their stated block-wise paired-$t$-test meaning, while the confidence intervals and effect sizes here provide the corresponding magnitude and uncertainty.

\section{Extended Explainable Examples}
\label{supp:sec:visual_interpretability}
This part provides additional misclassification examples from MNIST, CIFAR-10, Imagenette, Tiny-ImageNet, and Imagewoof to further examine the diagnostic and supervisory roles of reconstructed intermediate features. For each example, we compare the Grad-CAM response of the original forward feature with that of the reconstructed target feature at selected internal locations, together with their difference. The former reflects the regions emphasized by the current model representation, whereas the latter provides a label-conditioned reference indicating how the representation would need to shift under the frozen downstream mapping toward the ground-truth prediction. Their spatial discrepancy therefore enables a more concrete diagnosis of missing or misplaced semantic attention and helps assess whether the reconstructed feature highlights task-relevant regions that can provide meaningful guidance for correcting the prediction.

In this context, warm-colored regions indicate relatively strong spatial responses in the current response map, while cool-colored regions indicate relatively weak responses. It should be noted that each heatmap has been normalized independently. Therefore, the colors primarily reflect the relative distribution within a single response map and should not be used directly to compare absolute response intensities across different samples or feature maps. Furthermore, the fourth column does not simply subtract the normalized heatmaps from the second and third columns on a pixel-by-pixel basis. Instead, it first calculates the difference between the reconstructed features and the original forward features in the feature space, and then generates the corresponding spatial responses based on that difference to highlight the feature regions that were primarily adjusted during the post-training phase.

\subsection{MNIST}
On the MNIST dataset, both selected samples belong to the ground-truth class ``4'' but are misclassified by the pretrained model as ``5''. As illustrated in the input images, the handwritten digits exhibit substantial intra-class variation, particularly in the configurations of the diagonal strokes, horizontal junctions, and vertical segments. These atypical local structures partially resemble those of the digit ``5'', thereby increasing the ambiguity between the two classes and contributing to the observed misclassifications.

The original forward features exhibit a relatively broad response across the digit and its surrounding areas, indicating that while the model can extract the overall stroke morphology, it fails to adequately capture the local connectivity relationships that determine the class. After feature reconstruction via post-training, the response distribution undergoes significant changes near the digit's diagonal strokes, stroke intersections, and the vertical structure on the right side. The differential responses in the fourth column are further concentrated in the key stroke regions that are prone to causing confusion between ``4'' and ``5'', rather than covering the entire digit uniformly.

\begin{figure}[H]
    \centering
    \includegraphics[width=\textwidth]{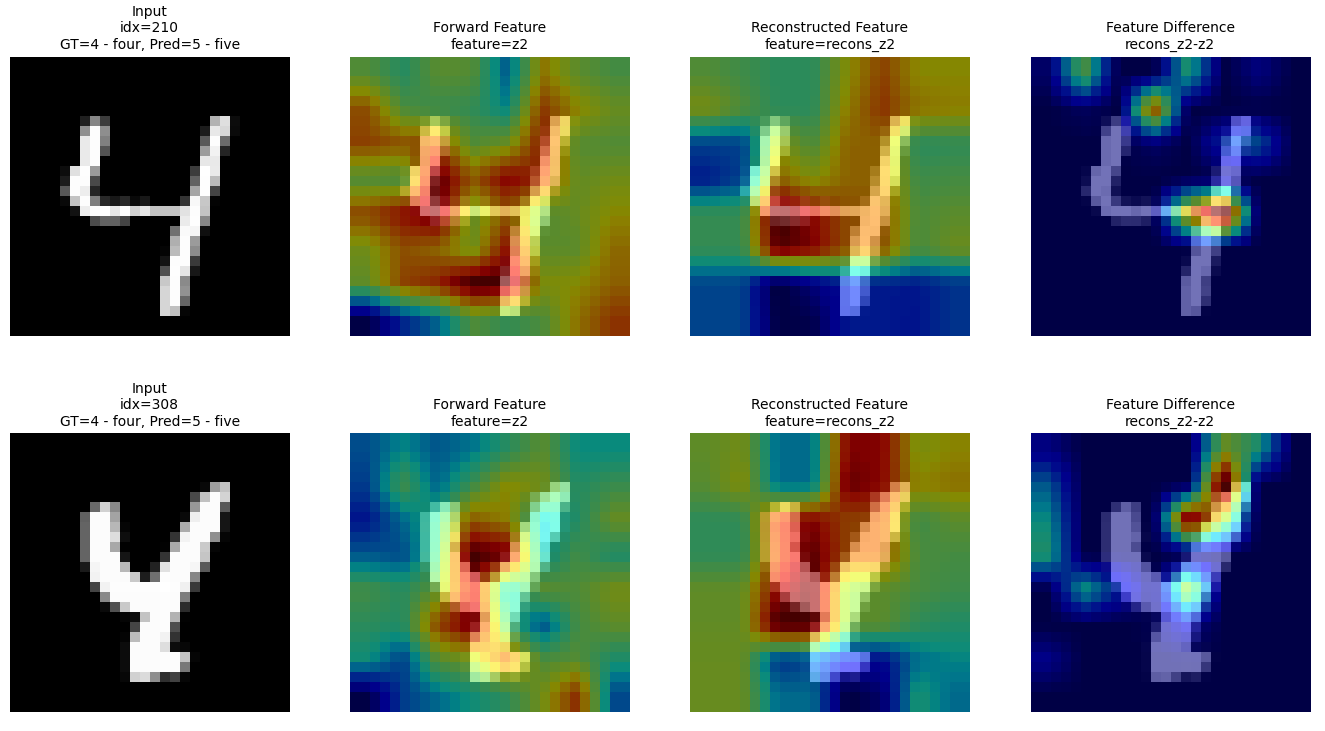}
    \caption{Two examples of Grad-CAM visualization for feature reconstruction comparison at layer $z2$ on MNIST.}
    \label{supp:fig:mnist_interpretability}
\end{figure}

\subsection{CIFAR-10}
On CIFAR-10, the two cat images are misclassified as deer and horse, respectively. In the first example, the forward response mainly emphasizes generic quadruped features around the head and upper body, which are shared across several animal classes and provide limited evidence for distinguishing a cat from a deer. In the second example, the forward response concentrates on the cat's left face, whose profile may resemble the side view of a horse head and contribute to the incorrect prediction.

The reconstructed features shift the responses toward more class-discriminative regions in both cases. For the first image, the stronger emphasis on the shorter ear structure provides a more category-specific cue for separating the cat from deer-like appearance. For the second, the reconstructed response moves toward the central facial region, particularly the cat's nose, reducing reliance on the misleading side-face profile. These label-conditioned changes suggest that reconstruction supervision can guide the model away from generic or confounding animal features and toward visual cues that are more informative for the correct cat prediction.

\begin{figure}[H]
    \centering
    \includegraphics[width=\textwidth]{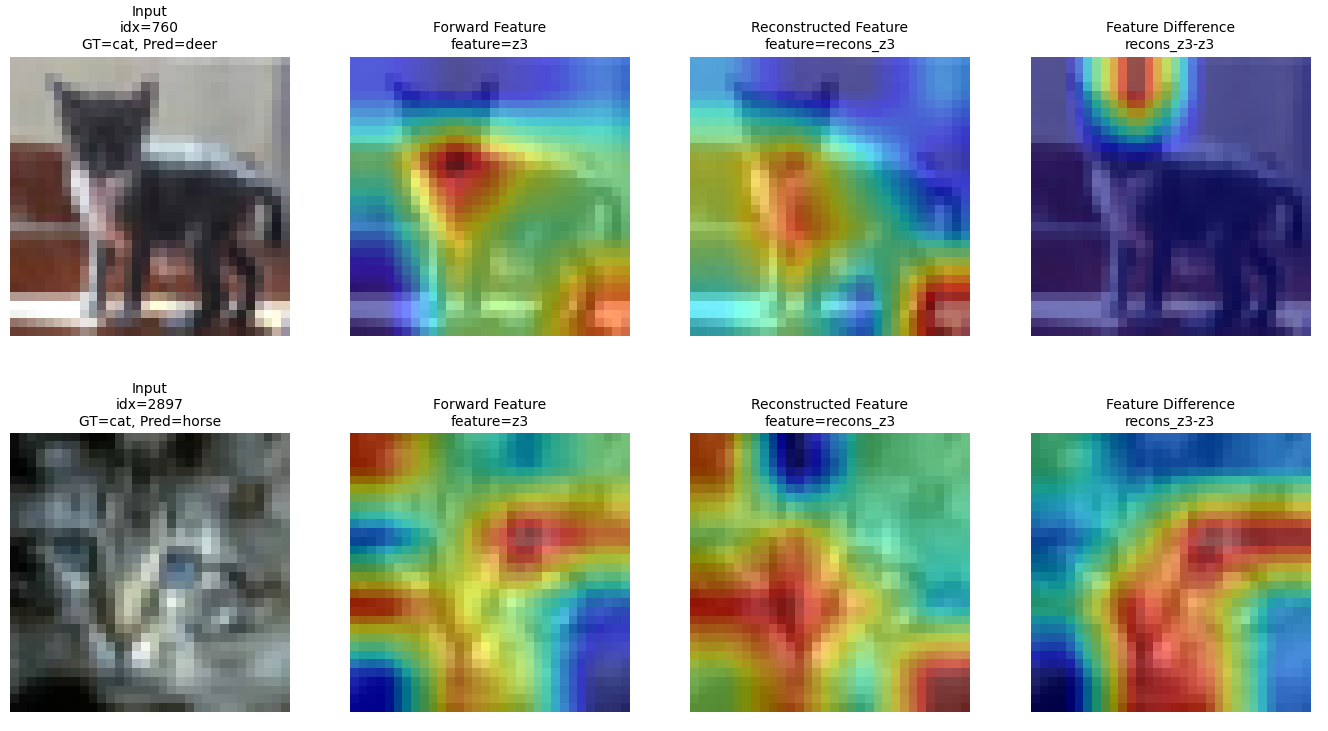}
    \caption{Two examples of Grad-CAM visualization for feature reconstruction comparison at layer $z3$ on CIFAR-10.}
    \label{supp:fig:cifar10_interpretability}
\end{figure}

\subsection{Imagenette}
In the Imagenette example, a tench is misclassified as a gas pump in a visually cluttered scene containing metallic equipment, reflective surfaces, and other high-contrast structures. The forward feature responds strongly to the equipment and surrounding regions above the fish, while the fish itself receives a less coherent response. Such salient man-made structures provide strong edges and textures that are visually prominent but not discriminative for the ground-truth class, indicating that the prediction is affected by both background interference and misplaced emphasis on semantically irrelevant structures.

The reconstructed feature substantially reorganizes this spatial pattern and introduces stronger responses over multiple regions of the fish while changing the representation of the surrounding equipment. Because the reconstructed target is generated toward the ground-truth tench prediction under the frozen downstream mapping, this redistribution provides a concrete reference for which object-related semantics should receive greater representational importance. It can therefore supervise the upstream modules to distinguish features associated with the fish from visually dominant but category-irrelevant structures in the scene.

\begin{figure}[H]
    \centering
    \includegraphics[width=\textwidth]{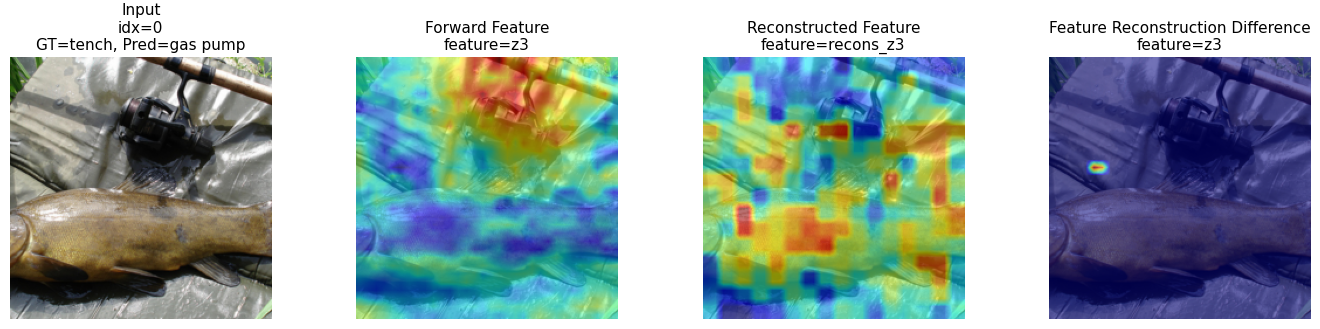}
    \caption{An example of Grad-CAM visualization for feature reconstruction comparison at layer $z3$ on Imagenette.}
    \label{supp:fig:nette_interpretability}
\end{figure}

\subsection{Imagewoof}
Both Imagewoof examples belong to the Shih-Tzu class but are predicted as English foxhound. Since Imagewoof requires fine-grained discrimination among dog breeds, simply localizing a dog is insufficient: successful classification depends on subtle breed-specific cues such as facial proportions, coat appearance, and head morphology. In the first example, the forward response is strongly influenced by the toy and nearby regions in addition to the dog. In the second, the model responds to the dogs but does not clearly prioritize the facial and coat structures that are most useful for distinguishing Shih-Tzu from visually related breeds. The errors therefore arise from both non-target distraction and reliance on features that are insufficiently class-specific.

The reconstructed features redistribute the responses toward the dogs' heads, faces, coat regions, and surrounding body structures, with substantial feature differences appearing in these areas. These changes are particularly informative in a fine-grained task because they emphasize localized appearance cues rather than only the presence of a generic dog. The ground-truth-conditioned reconstructed feature can thus guide post-training toward a representation that gives greater weight to breed-discriminative semantics while reducing the influence of unrelated or weakly discriminative visual cues.

\begin{figure}[H]
    \centering
    \includegraphics[width=\textwidth]{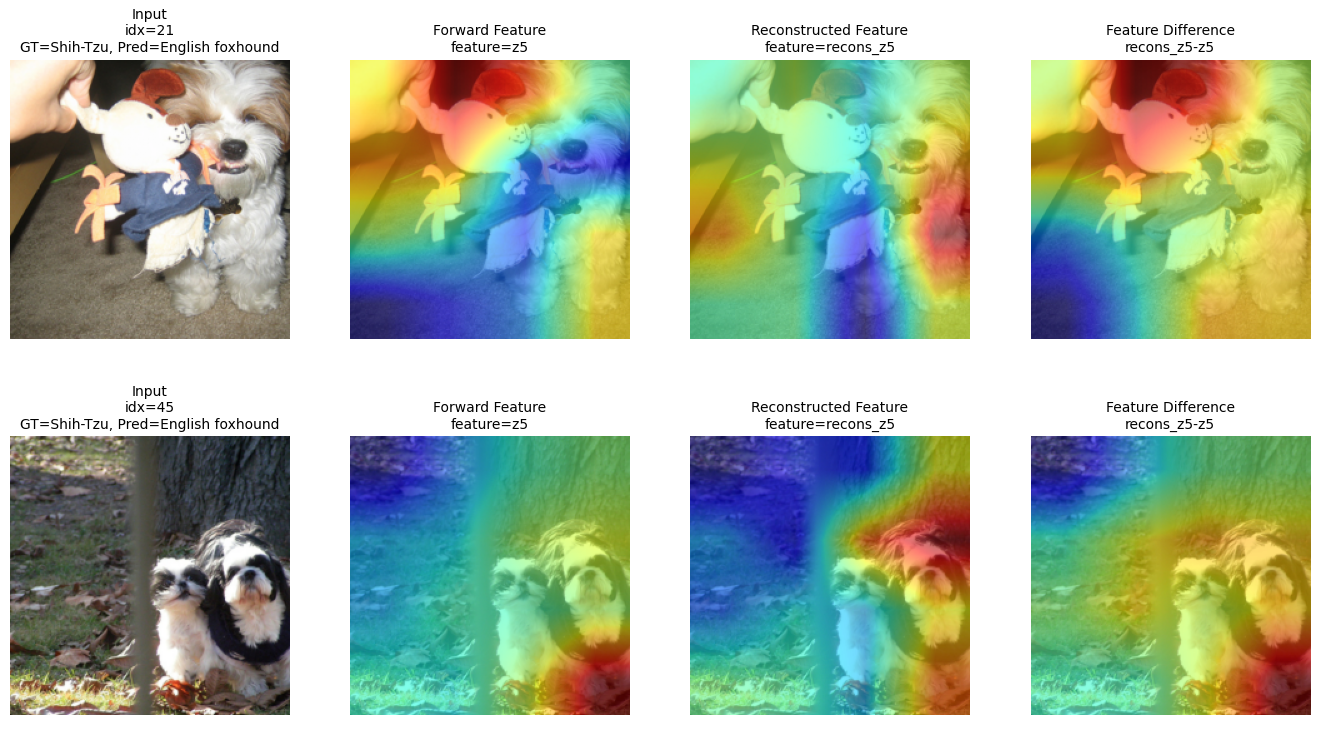}
    \caption{Two examples of Grad-CAM visualization for feature reconstruction comparison at layer $z5$ on Imagewoof.}
    \label{supp:fig:woof_interpretability}
\end{figure}

\subsection{Tiny-ImageNet}
In the Tiny-ImageNet example, a goldfish is misclassified as a snail. The forward response is concentrated mainly on the surrounding stones, whose compact shapes and textures provide visually salient cues that can be more easily matched to the appearance of a snail than to the relatively small goldfish. As a result, the model relies on an incorrect but visually distinctive region instead of adequately representing the target object.

The reconstructed feature shows a substantially different response pattern, with stronger attention shifted toward the goldfish swimming in the water. This label-conditioned redistribution suppresses the misleading stone-related cues and emphasizes the actual object associated with the ground-truth class. The reconstructed target therefore provides a more appropriate feature-level reference for guiding the model from an easily recognized but irrelevant region toward the semantics required for the correct prediction.

\begin{figure}[H]
    \centering
    \includegraphics[width=\textwidth]{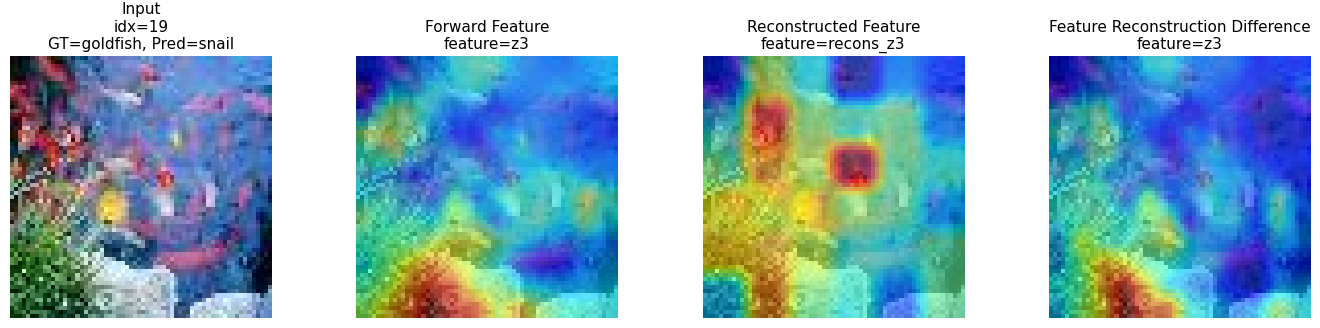}
    \caption{An example of Grad-CAM visualization for feature reconstruction comparison at layer $z3$ on Tiny-ImageNet.}
    \label{supp:fig:tin_interpretability}
\end{figure}

\subsection{Comprehensive Analysis}
Across the five datasets, the visualizations reveal two recurring sources of misclassification. In some cases, such as the Imagenette example and the first Imagewoof example, the forward representation is substantially influenced by background structures or non-target objects. In others, the model already responds to the correct foreground object but emphasizes generic or weakly discriminative regions rather than the structures needed to distinguish the ground-truth category. The latter is particularly evident in MNIST, where stroke connectivity matters, and Imagewoof, where correct localization of a dog alone is insufficient for fine-grained breed recognition. Misclassification therefore cannot be attributed solely to background distraction; it can also arise from misplaced semantic emphasis within the target itself.

The reconstructed features provide a complementary, ground-truth-conditioned reference for diagnosing these representation deficiencies. Across the examples, reconstruction redistributes spatial responses toward local structures that are more relevant to the target category, including discriminative stroke configurations, object contours, foreground structures, and fine-grained appearance cues. The feature-difference maps further localize where the representation is altered most substantially, rather than indicating a uniform amplification of the original feature. Although Grad-CAM visualizations do not establish causal attribution, the repeated alignment between reconstructed responses and class-relevant semantics supports the diagnostic value of reconstructed targets and helps explain why they can serve as meaningful feature-level supervision during localized post-training.

\end{document}